\pdfoutput=1
\documentclass{article} %
\usepackage{iclr2027_conference,times}

\usepackage{amsmath,amsfonts,bm}

\def\eqref#1{equation~\ref{#1}}

\def\1{\bm{1}}

\DeclareMathAlphabet{\mathsfit}{\encodingdefault}{\sfdefault}{m}{sl}
\SetMathAlphabet{\mathsfit}{bold}{\encodingdefault}{\sfdefault}{bx}{n}

\usepackage{hyperref}
\usepackage{url}

\usepackage[utf8]{inputenc}
\usepackage[T1]{fontenc}
\usepackage{hyperref}
\usepackage{url}
\usepackage{booktabs}
\usepackage{amsfonts}
\usepackage{nicefrac}
\usepackage{microtype}
\usepackage{xcolor}
\newif\ifshowrevision
\showrevisionfalse
\definecolor{revisionred}{RGB}{190,0,0}
\newcommand{\rev}[1]{\leavevmode{\ifshowrevision\color{revisionred}\fi #1}}
\newcommand{\revten}[1]{\leavevmode{#1}}
\newcommand{\rebut}[1]{#1}
\usepackage{arydshln}
\usepackage{wrapfig}
\usepackage{needspace}
\usepackage{multirow}
\usepackage{bm}
\usepackage{amsmath}
\usepackage{amssymb}
\usepackage{adjustbox}
\usepackage{makecell}
\usepackage{xspace}
\usepackage{pifont}
\usepackage{animate}
\usepackage{varwidth}
\usepackage{graphicx}
\usepackage{enumitem}

\newcommand{\cmark}{\textcolor{darkgreen}{\ding{51}}}
\newcommand{\xmark}{\textcolor{red}{\ding{55}}}%
\usepackage{mwe}
\usepackage{subcaption,float}
\definecolor{darkgreen}{rgb}{0,0.5,0}
\usepackage{colortbl}
\definecolor{lightgray}{rgb}{0.8, 0.8, 0.8}
\definecolor{best_color}{RGB}{253,155,154}
\definecolor{second_color}{RGB}{254,205,158}
\definecolor{third_color}{RGB}{255,255,163}
\definecolor{gbest}{HTML}{2ECC71}
\definecolor{gsecond}{HTML}{ABEBC6}
\newcommand{\model}{{FIRE3D}\xspace}

\title{\model: Feed-forward Interactive 3D Scene Reconstruction Within A Minute}

\author{%
Hongchi Xia$^{1}$ \quad Tianhang Cheng$^{1}$ \quad Wei-Chiu Ma$^{2}$ \quad Shenlong Wang$^{1}$ \\
\normalfont $^{1}$University of Illinois Urbana-Champaign \quad $^{2}$Cornell University
}
\hypersetup{
  pdftitle={FIRE3D: Feed-forward Interactive 3D Scene Reconstruction Within A Minute},
  pdfauthor={Hongchi Xia, Tianhang Cheng, Wei-Chiu Ma, Shenlong Wang}
}

\iclrfinalcopy %
\begin{document}

\maketitle

\vspace{-6mm}
\begin{figure*}[thbp]
    \centering    
    \vspace{-4mm}
    \includegraphics[width=0.98\textwidth,trim={0 0 0 0.2cm}, clip]{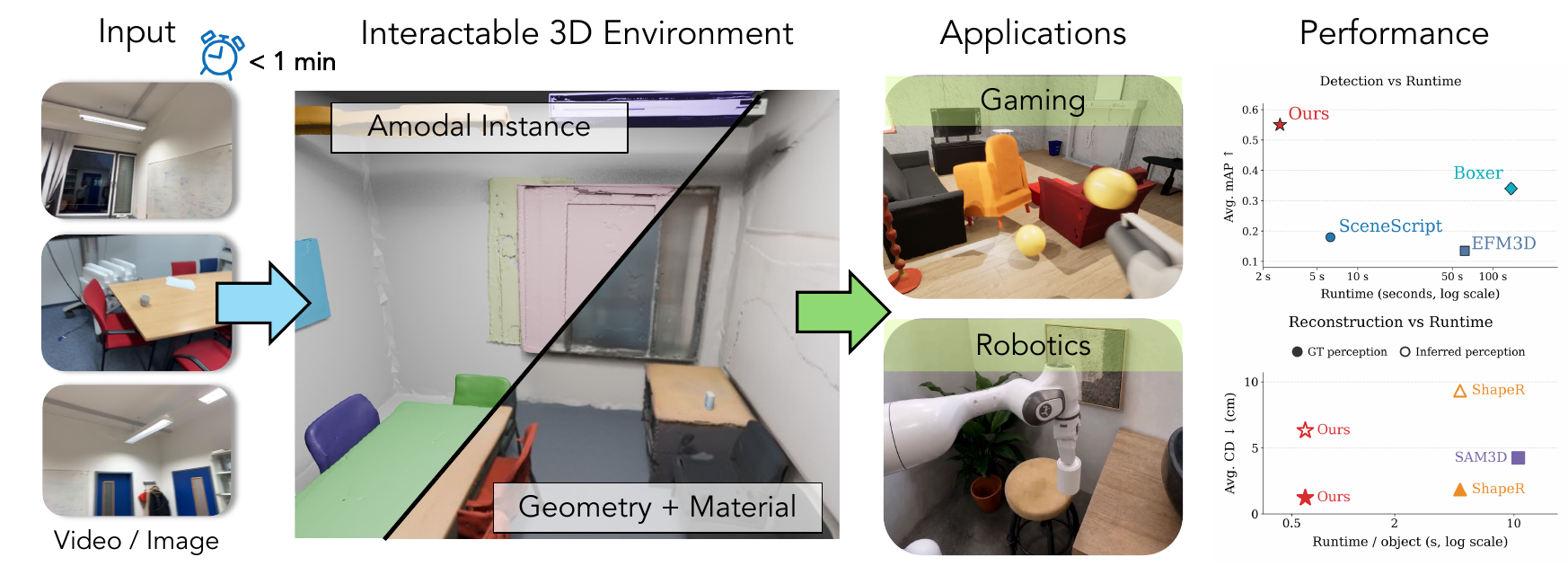}
    \caption{
    \textbf{Overview of \model:} 
    \rev{From an RGB capture,} \model reconstructs \rev{an interactable 3D scene in under a minute for up to twelve textured instances (Sec.~\ref{sec:runtime_protocol}).} It does not require \textit{any} additional object annotation, such as bounding boxes or instance masks, and produces \rev{separate textured object assets together with a background instance.} \model{} \rev{supports downstream applications in} interactive gaming and robotics.
    }
    \label{fig:teaser}
\end{figure*}

\vspace{-10px}
\begin{abstract}
We present \model, a unified framework that \rev{transforms a single RGB image or casual video into interactable 3D scene assets} for games and interactive applications \rev{in \textbf{under a minute} for up to twelve textured instances including preprocessing.} At the core of \model is a \rev{feed-forward inference pipeline} that predicts a compositional scene representation from posed RGB-D observations estimated from the RGB capture, including the 6-DoF pose, bounding box, mesh, and texture \rev{for detected objects}.
By modeling the scene as a collection of discrete entities, \model produces \rev{amodal object assets that can be independently edited and used in interactive applications}. 
Our framework requires no test-time optimization, \rev{runs substantially faster than the compared reconstruction systems}, and provides object-level completeness beyond existing feed-forward 3D approaches. We demonstrate \rev{leading detection accuracy, strong geometry reconstruction, and competitive rendering quality on the evaluated benchmarks, with substantial runtime gains}.

\noindent Project page: \href{https://xiahongchi.github.io/Fire3D}{\textcolor[HTML]{1A5FB4}{https://xiahongchi.github.io/Fire3D}}
\end{abstract}

\section{Introduction}
\label{sec:intro}
Imagine turning a room into an editable 3D world: each chair, table, cabinet, and background surface becomes a complete textured entity that can be moved, rendered, or simulated. Such object-level digital twins are valuable for AR/VR, robotics, gaming, and content creation. Yet real indoor scenes are cluttered, partially observed, and often contain many interacting objects. A practical system must jointly parse object instances, recover complete geometry beyond visible surfaces, synthesize appearance, and \rev{preserve the spatial layout} of the scene.

Existing methods address only parts of this problem. 3D detectors localize objects but do not reconstruct complete geometry or texture~\cite{wang2022detr3d}. Point-cloud segmentation networks parse visible regions but remain perception-only~\cite{qi2017pointnet++}. Object-centric reconstruction methods improve amodal completion, but often assume pre-segmented object inputs or process objects independently. Recent systems move closer to object-level scene reconstruction, but still rely on image prompts, external SLAM and detection, incomplete shape-only reconstruction, or expensive optimization-based refinement~\cite{sam3d,shaper,xia2025holoscene}. Thus, a key gap remains: fast feed-forward reconstruction of complete textured object-level scenes from \textit{unsegmented}, posed RGB-D observations estimated from raw RGB captures.

We introduce \model, a feed-forward framework for object-level textured 3D scene reconstruction from posed RGB-D observations. 
We represent each observation by an RGB image, a camera-frame point map (equivalently, depth with known intrinsics), and a camera-to-world pose. For a single RGB image or casual monocular RGB video, we estimate the required point maps and camera poses with \rev{the Pi3X variant of} Pi$^3$~\cite{wang2026pi} before FIRE3D inference. 
Given these observations without instance masks, \model lifts multi-view features into a 3D feature point cloud and jointly predicts object validity, pose, and 3D instance masks. Each parsed instance is then canonicalized and reconstructed by a point-cloud-conditioned cascaded flow-matching model, which generates structure, shape, and material latents in sequence. A key design of \model is an ultra-compact hierarchical latent space that represents each object with a small set of structure, geometry, and material tokens. This compact representation enables batched flow matching sampling across many instances, substantially accelerating scene-level reconstruction. \rev{This supports fast scene reconstruction without per-scene optimization.}

Training such a feed-forward generative reconstruction model requires large and diverse supervision. To this end, we curate a large-scale training corpus across five scene datasets and four object datasets, totaling \rev{80k scenes, 138,202 rendered sequences, 79,672 additional appearance-augmented sequences, and 500k object assets}. This data provides broad coverage of indoor layouts, object categories, occlusion patterns, and appearance variation, \rev{supporting scene parsing, amodal completion, and textured reconstruction at scale.} 
Together, the feed-forward scene parser, compact latent space, batched object generator, and large-scale training corpus enable fast object-level reconstruction of cluttered indoor scenes. 

Experiments on challenging multi-object indoor scenes \rev{from held-out evaluation splits} \rev{show strong detection and reconstruction results with substantial speedups over the compared systems}, while producing \rev{textured assets that can be independently edited, rendered, and used in interactive applications}. Importantly, 
\rev{from a 60-frame RGB video, \model reconstructs up to twelve textured instances in under 60 seconds, including Pi3X preprocessing, under our timing protocol (Sec.~\ref{sec:runtime_protocol}).}

Our contributions are threefold:
\begin{itemize}[leftmargin=1.2em, itemsep=0.2em, topsep=0.2em]
    \item We introduce \model, a feed-forward framework that reconstructs object-level textured 3D scenes from \textit{unsegmented} posed RGB-D observations estimated from RGB image/video captures, without manual boxes or masks.
    \item We design an ultra-compact latent flow-matching generator that reconstructs up to 16 objects in parallel, \rev{retaining high autoencoding fidelity while enabling fast scene reconstruction}.
    \item We curate a large-scale training corpus with \rev{80k scenes, 218k original/augmented sequences, and 500k object assets} and \rev{demonstrate strong perception and reconstruction performance with substantial runtime gains}.
\end{itemize}

\begin{table}[t]
\centering
\resizebox{\textwidth}{!}{
\begin{tabular}{l|c|ccc|ccccc}
\toprule
\multirow{2}{*}{\textbf{Method}}
  & \multirow{2}{*}{\textbf{Feed-forward}}
  & \multicolumn{3}{c|}{\textbf{Input}}
  & \multicolumn{5}{c}{\textbf{Output}} \\
\cmidrule(lr){3-5}\cmidrule(lr){6-10}
& & \rebut{\textbf{User Input}} & \rebut{\textbf{Network Input}} & \rev{\textbf{No External Instance Parser}}
  & \textbf{BBox \& Mask} & \textbf{Obj. Geo} & \textbf{Texture} & \textbf{BG} & \textbf{Interact-able} \\
\midrule
Pixel-space opt.~\cite{mildenhall2021nerf,kerbl20233d}
  & \xmark & Video & \rebut{Posed RGB} & \cmark & \xmark & \xmark & \cmark & \cmark & \xmark \\
Object-centric opt.~\cite{xia2025holoscene,xia2026simrecon}
  & \xmark & Video & \rebut{Posed RGB} & \xmark & \xmark & \cmark & \cmark & \cmark & \cmark \\
Perception~\cite{scenescript,efm3d}
  & \cmark & Video & \rebut{Posed RGB + points} & \cmark & \cmark & \xmark & \xmark & \xmark & \xmark \\
Image-to-scene~\cite{gen3dsr,midi}
  & \cmark & Image & \rebut{RGB image} & \xmark & \cmark & \cmark & \cmark & $\sim$ & \cmark \\
Video-to-scene~\cite{shaper}
  & \cmark & Video & \rebut{Posed RGB + points} & \xmark & \xmark & \cmark & \rev{\ding{55}} & \xmark & \cmark \\
\midrule
\textbf{Ours}
  & \cmark & Image / Video & \rebut{Posed RGB-D} & \cmark & \cmark & \cmark & \cmark & \cmark & \cmark  \\
\bottomrule
\end{tabular}
}
\caption{\textbf{Comparison with representative scene reconstruction methods}. \rev{FIRE3D estimates instances internally and outputs interactable assets. Instance parsing denotes detection/segmentation, distinct from RGB geometric preprocessing.}
\cmark/\xmark: supported/not supported; $\sim$: partially; BG: background. Obj. Geo: Object-wise geometry. 
\rebut{Network Input: the input to the network, which FIRE3D estimates from the user's RGB capture with Pi3X~\cite{wang2026pi}.}
}
\label{tab:related_cmp}
\vspace{-1em}
\end{table}

\section{Related Works}
\label{sec:related}

\paragraph{Simulation-ready 3D Scene Reconstruction.}
As summarized in Tab.~\ref{tab:related_cmp}, NeRF-~\cite{mildenhall2021nerf,barron2021mip,muller2022instant,yariv2023bakedsdf,guo2022neural,zhou2024neural, ost2021neural} and 3DGS-based methods~\cite{kerbl20233d,huang20242d,yu2024mip, ni2025g4splat} achieve realistic novel-view synthesis, but represent scenes as fields or splats rather than editable object-level assets. Scene-level reconstruction methods~\cite{xia2024video2game,xia2025drawer,xia2025holoscene,xia2026simrecon,ni2025decompositional,niemeyer2021giraffe,cheng2024structure,wu2022object,sautter20253d} can produce simulation-ready environments, yet rely on optimization, search, or iterative refinement, limiting scalability in cluttered scenes.
Recent feed-forward models~\cite{shaper,gen3dsr,midi,sam3d,meng2025scenegen} avoid costly test-time optimization and recover complete shapes or textured objects, but often require pre-segmented inputs, prompts, external perception, or sequential object-wise inference. In contrast, \model performs batched feed-forward inference, \rev{integrating scene perception and textured reconstruction of detected objects}.

\paragraph{Feed-forward 3D Learning.}
Feed-forward 3D learning enables efficient scene understanding and geometric prediction. Existing perception methods predict 3D boxes from multi-view images~\cite{wang2022detr3d, xu20243difftection, boudjoghra2024open}, segment point clouds into semantic or instance regions~\cite{qi2017pointnet,qi2017pointnet++,mask3d,openmask3d}, or infer layouts, global boxes, and egocentric scene representations from images or video~\cite{scenescript,boxer,efm3d, chen2024single}. However, their outputs are typically boxes, masks, layouts, or partial geometry, rather than complete textured assets for simulation.
Recent feed-forward reconstruction models predict dense geometry from images. TRELLIS.2~\rebut{\cite{xiang2026native}} and other image-to-3D methods~\cite{zhao2025hunyuan3d,liu2023one, tochilkin2024triposr, qian2023magic123, long2024wonder3d, voleti2024sv3d,lai2025lattice} recover geometry and texture from a single image, but mainly target individual objects. DUSt3R and successors~\cite{dust3r,mast3r,cut3r,vggt,chen2025ttt3r, yang2025fast3r, lin2025depth, zhang2024monst3r} infer point maps, depth, camera parameters, tracks, or dense scene geometry without per-scene optimization, but do not explicitly produce object-level textured meshes, poses, and editable assets.
\rebut{Amodal reconstruction methods~\cite{zhang2026world,li2025lari,chen2026nova3r,ngo2026volfill,li2024know} also recover geometry beyond the visible surfaces, but do not decompose the scene into object instances.}
Our method unifies perception and reconstruction in a feed-forward framework. From unsegmented posed RGB-D observations constructed from native RGB-D data or estimated from RGB captures, \model directly reconstructs object instances, poses,  complete foreground geometry and texture, plus a static background instance, \rev{producing independently editable scene assets}.

\begin{figure}
    \centering
    \includegraphics[width=\linewidth]{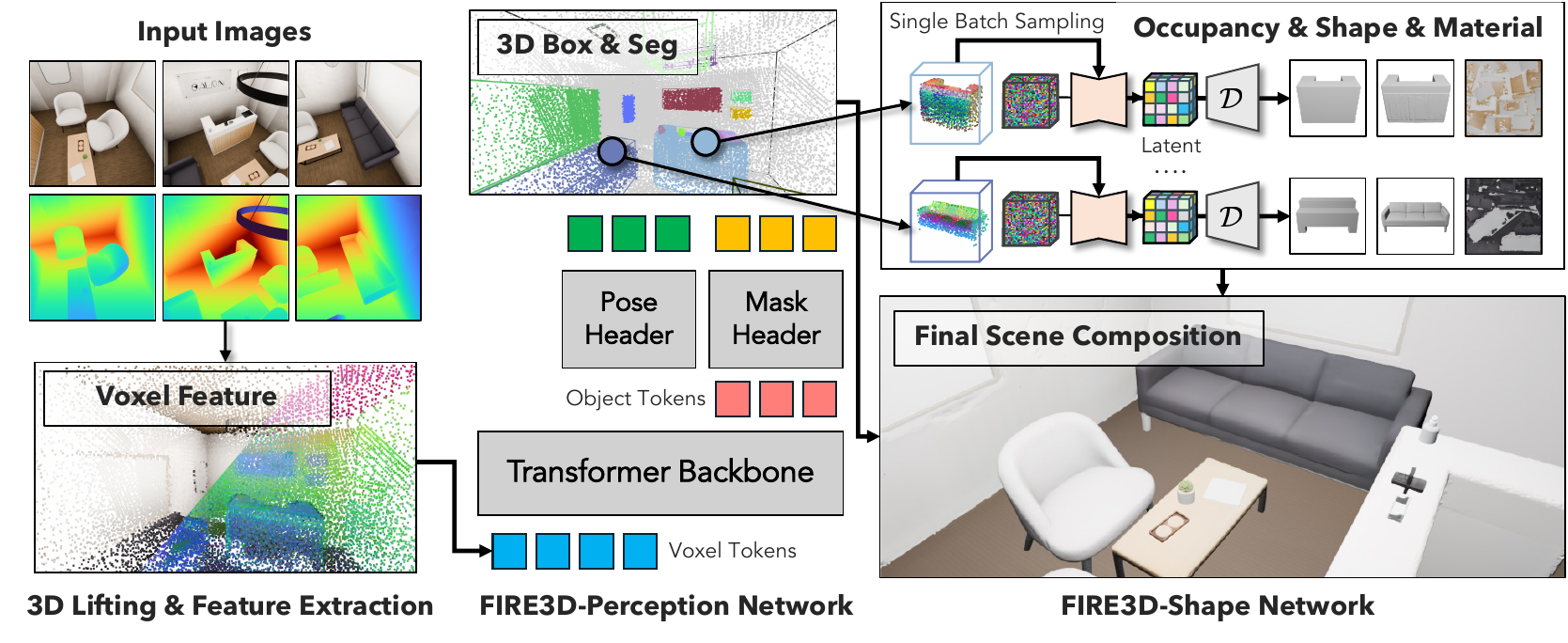}
    \caption{
    {\bf Overview of \model network:} Given the input posed RGB-D observations, \model first performs instance-aware 3D scene perception, and then generates the compact latents of each object conditioned on the predicted instance point clouds. Finally, the complete textured meshes are decoded with the hierarchical VAEs and assembled into a 3D scene with the predicted object 3D poses. 
    }
    \label{fig:overview}
\end{figure}

\begin{figure}
    \centering
    \includegraphics[width=0.8\linewidth]{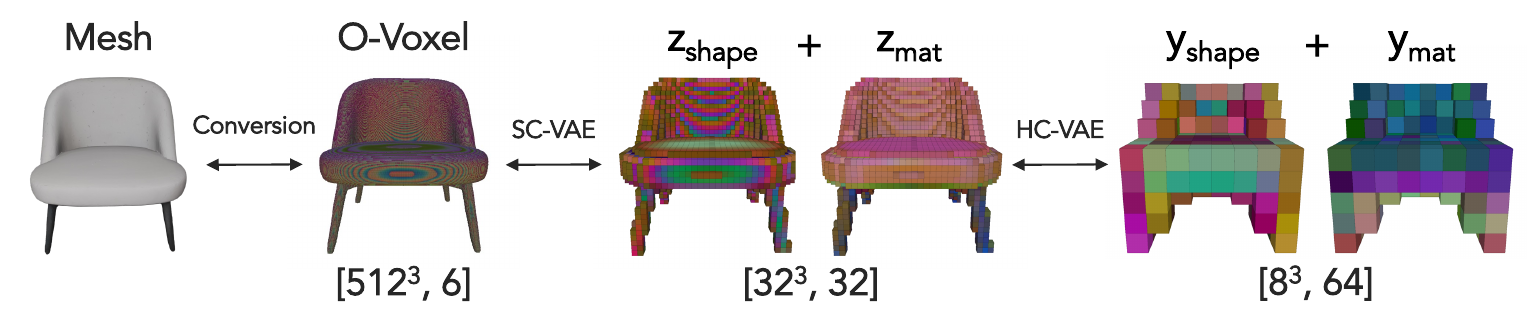}
    \caption{
    {\bf Overview of \model hierarchical compression VAE:}
    To further compress the object latent space and enable efficient \revten{scene-scale object reconstruction}, \model leverages HC-VAE to further compress the SC-VAE latents into an even more compact latent space, \rev{reducing nominal grid feature capacity by $32\times$ relative to SC-VAE.}
    }
    \label{fig:method_vae}
\end{figure}

\section{Method}
\label{sec:method}
In this paper, we propose a model that takes a single RGB image or casually captured monocular RGB video, estimates its posed RGB-D observations, and converts them into \rev{an interactable textured scene under the timing protocol in Sec.~\ref{sec:runtime_protocol}}.  
Based on the observation that existing approaches either 
heavily rely on (multi-stage) optimization \cite{xia2025holoscene}
or employ iterative estimation of scene objects \cite{sam3d}, we propose to develop a feed-forward network that can \rev{reconstruct scene geometry and appearance without per-scene optimization, using a fixed 12-step flow-sampling schedule.} 
At the core of our framework \rev{lie three tightly coupled components}: (i) \rev{a compact}, object-centric shape representation that is 
both memory-efficient and highly expressive; 
(ii) a perception network that extracts object poses and features;
and (iii) a shape generation network that operates on the compressed latent space, enabling parallel batch generation \emph{on a single GPU}. 
Altogether, these components form a unified pipeline \rebut{that works on one shared 3D feature point cloud}, where the efficiency of our scene reconstruction is \rev{supported by compact latents and batched generation and export}. \rebut{Unlike composed pipelines that hand 2D masks from a detector to an image-conditioned generator, \model passes 3D instances directly from perception to generation, so it needs no cross-view mask association or re-alignment of generated objects (Appendix Tab.~\ref{tab:composed_cmp}).}

\rev{We first describe hierarchical compression of object latents.} 
\rev{We then build perception and reconstruction around this representation.} Fig. \ref{fig:overview} summarizes our approach. 

\vspace{-5px}

\subsection{Representing an Object within 256 KBytes}
\label{sec:representation}

\rev{Our goal is fast reconstruction of multiple independently editable scene assets.} 
To efficiently generate these assets simultaneously, we require a representation that is both \emph{expressive} (\emph{i.e.}, capable of encoding diverse geometries and object categories) and \emph{compact} (\emph{i.e.}, fit within the memory constraints of a single GPU).

\vspace{-5px}
\paragraph{Sparse Compression VAE (SC-VAE).} 
One popular 3D object representation is the latent space derived from SC-VAE \rebut{\cite{xiang2026native}}. Specifically, given a textured mesh $\mathcal{M}$, we first convert it into its Occupancy-Voxel (O-Voxel) representation, and then encode it into a shape latent $\mathbf{z}_{\mathrm{shape}}$ and a material latent $\mathbf{z}_{\mathrm{mat}}$ using pretrained SC-VAE. 
While this representation has enjoyed great success in single-object 3D generation methods, its latent resolution (typically $32^3 \times 32$) quickly becomes computationally expensive when scaled to scenes with many objects. \rev{In our configuration, this resolution supports simultaneous generation of two objects on an 80GB A100.} 

\vspace{-5px}
\paragraph{Hierarchical Compression VAE (HC-VAE).} To enable efficient multi-object generation, we propose to further compress the SC-VAE latents. \rev{We learn a compact code on top of the pretrained representation.} We therefore employ an additional sparse 3D CNN to compress the latents into $\mathbf{y}_{\mathrm{shape}}$ and $\mathbf{y}_{\mathrm{mat}} \in 8^3 \times 64$. \rev{Tab.~\ref{tab:vae_recon} shows high autoencoding fidelity after a $32\times$ reduction in nominal grid feature capacity. Two float32 feature grids occupy 256 KiB, excluding sparse metadata.} \revten{At equal flow-training compute, HC-VAE improves generation quality with faster sampling and decoding (Appendix Tab.~\ref{tab:flow_compute_ablation}).} Fig.~\ref{fig:method_vae} illustrates the procedure.

\vspace{-5px}
\subsection{Instance-aware 3D Scene Perception from Posed RGB-D Observations}
\label{sec:perception}

Having established an extremely compact representation for individual objects, we now describe how \model detects and segments objects within a scene to extract features for shape and material reconstruction.

FIRE3D operates on a set of posed RGB-D observations $\mathcal{O}=\{(\mathbf{I}_i,\mathbf{X}_i,\mathbf{T}_i)\}_{i=1}^{N}$, where $\mathbf{I}_i$ is an RGB image, $\mathbf{X}_i$ is its camera-frame 3D point map, and $\mathbf{T}_i$ is the camera-to-world pose. A depth map with known camera intrinsics provides an equivalent representation of $\mathbf{X}_i$. For native RGB-D captures, these quantities are measured directly. For a single RGB image or casual monocular RGB video, we use \rev{Pi3X}~\cite{wang2026pi} to estimate local point maps and camera poses, \rev{with its metric-scale outputs expressed in meters.} \rev{In the single-image case, the camera frame defines the reference frame; output scale follows the input geometry.}

Given the posed RGB-D observations defined above, we use the camera-frame point map and camera pose to transform every pixel $\mathbf{p}_i$ into a 3D point $\mathbf{x}_i$ and augment it with a DINOv3 \cite{dinov3} feature $\mathbf{f}_i$ extracted from the corresponding image frame.  
We then voxelize the resulting feature point cloud $\mathcal{P} = \{(\mathbf{x}_j, \mathbf{f}_j)\}_{j=1}^{N_p}$
 into a sparse 3D feature grid and feed it into a \rebut{class-agnostic} query-based transformer~\cite{detr} to segment the objects and estimate their respective poses:
\begin{equation}
    \left\{
    \hat{v}^{(k)},
    \hat{\boldsymbol{\pi}}^{(k)},
    \hat{\mathbf{m}}^{(k)}
    \right\}_{k=1}^{K_Q}
    =
    f_{\mathrm{percept}}(\mathcal{P}).
\end{equation}
Here, \rebut{$K_Q=384$ is the number of candidate queries, including the background,} $\hat{v}^{(k)} \in [0,1]$ denotes the validity score, $\hat{\mathbf{m}}^{(k)} \in \{0,1\}^{N_p}$ represents the 3D instance mask over $\mathcal{P}$, and $\hat{\boldsymbol{\pi}}^{(k)}$ parameterizes the similarity transformation of the predicted object extent.

Candidates with validity scores below a predefined threshold are discarded, resulting in a total of $K$ predicted objects. 
For each valid candidate, we apply the predicted similarity transformation $\hat{\boldsymbol{\pi}}^{(k)}$ to map the instance point cloud from \rev{world coordinates into} its canonical coordinate frame.  
These canonicalized instance point clouds\rebut{, together with their DINOv3~\cite{dinov3} features,} \rev{serve as} object-centric 3D conditions, which are subsequently used by our shape generation network (Sec.~\ref{sec:reconstruction}) to map each object to the compact HC-VAE latent space, enabling the efficient, parallel reconstruction of the entire scene.

\vspace{-5px}
\subsection{Batched Point Cloud Conditioned Object Reconstruction and Scene Assembly}
\label{sec:reconstruction}

The final module in our network aims to map each detected object to an HC-VAE latent, which is then decoded into complete geometry and material properties:
\begin{equation}
     \{
     {\mathbf{y}}^{(k)}_{\mathrm{shape}},
     {\mathbf{y}}^{(k)}_{\mathrm{mat}}
    \}_{k=1}^K
    =f_{\mathrm{recon}}
    \left(
    \{\hat{\mathcal{P}}^{(k)}\}_{k=1}^K
    \right) 
   .
\end{equation}
We parameterize $f_{\mathrm{recon}}$ as cascaded transformer-based flow-matching models. 
Thanks to our compact HC-VAE latents, we can simultaneously generate \rev{up to 16 objects} on a single 80GB A100 GPU and adopt a smaller flow matching network. 
\rev{The shape latent is generated first}, and then used as the condition for the material latent generation. 
\rev{This supplies explicit geometric context for material prediction.}

To obtain the final textured meshes, the predicted latents are passed through the HC-VAE and SC-VAE decoders to produce canonical textured meshes, followed by an O-Voxel-to-mesh conversion.  The resulting textured meshes $\{\hat{\mathcal{M}}^{(k)}\}_{k=1}^K$ are in their individual canonical object frame. Finally, to assemble the 3D Scene, we transform each reconstructed mesh back into the world coordinate system using the similarity transformations $\{\hat{\boldsymbol{\pi}}^{(k)}\}_{k=1}^K$ predicted by the perception network. 

\rev{We represent the background as one instance and canonicalize it using the room bounding box.}
Structural scene surfaces are grouped into one background instance and follow the same flow-matching, VAE decoding, and scene-assembly pipeline as foreground instances.

\vspace{-5px}
\subsection{Training}

We freeze the DINOv3 backbone and \rev{train the perception model from scratch} on five scene datasets with accurate 3D oriented bounding boxes and per-point instance segmentation annotations. Each scene includes one background instance alongside its foreground instances.
We adopt the SC-VAE from the TRELLIS.2 ~\rebut{\cite{xiang2026native}}, and train the HC-VAE from scratch on latents encoded by the SC-VAE. By compressing each object latent to an $8^3$ grid with $64$ channels, the flow matching models can be trained at the scene level by packing all instances from a scene into a single batch. \rev{We pack up to 48 scene instances per training batch; standalone-object batches contain 64 objects for structure/shape and 48 for material.} The inference time parallelism of 16 objects is still capped by the SC-VAE size.

\subsection{Inference}

\vspace{-5px}
\paragraph{Perception Post-processing.}
Given $K$ candidate object tokens from the transformer decoder, we discard low-confidence proposals by thresholding the predicted validity scores $\hat{v}^{(k)}$ with \rev{$\tau_v=0.2$}, and apply non-maximum suppression (NMS) with IoU threshold \rev{$\tau_{\mathrm{NMS}}=0.4$} to the predicted OBBs, suppressing duplicate detections of the same instance following standard practice~\cite{detr}.

\vspace{-5px}
\paragraph{Batchified Inference for Efficiency.}
At inference time, HC-VAE compression allows all detected instances to be processed in batched forward passes through flow matching models, as each instance is represented by a compact $8^3 \times 64$ latent. The HC-VAE and SC-VAE decoding stages are also batched across instances. The final O-Voxel-to-textured-mesh conversion is batchified using a CUDA C++ implementation of dual contouring for mesh extraction and parallelized UV unwrapping and texture baking. \rev{This design amortizes execution overhead across instances, using multiple chunks when scenes exceed the configured batch capacity.}

\newcommand{\img}[1]{%
  \IfFileExists{#1.png}%
    {\includegraphics[width=0.2\linewidth, trim=8 8 8 8, clip]{#1.png}}%
    {\includegraphics[width=0.2\linewidth, trim=8 8 8 8, clip]{#1.jpg}}%
}

\begin{figure}[ht]
\centering
\resizebox{0.95\linewidth}{!}{%
\setlength{\tabcolsep}{1pt}
\begin{tabular}{ccccccc}
& Input & SceneScript & EFM3D & Boxer & Ours & Ground-truth  \\

\raisebox{30px}{\rotatebox{90}{AEO}} &
\img{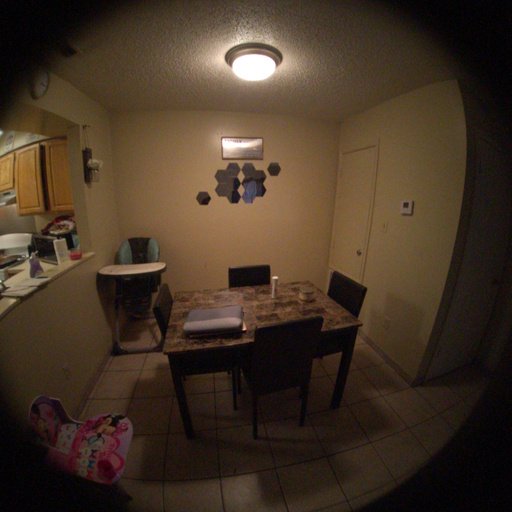} &
\img{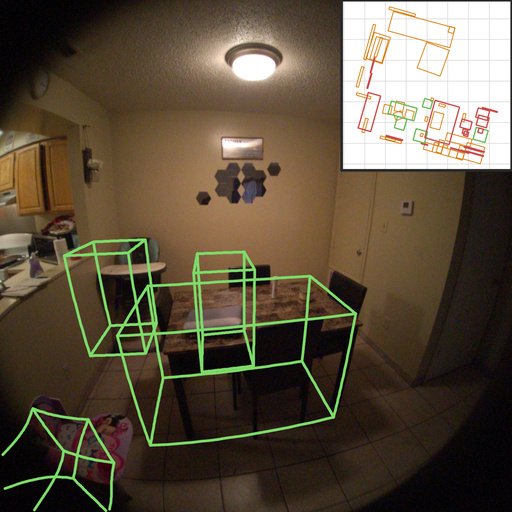} &
\img{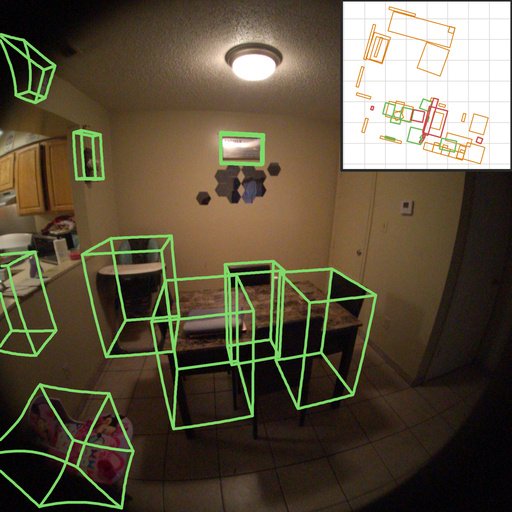} &
\img{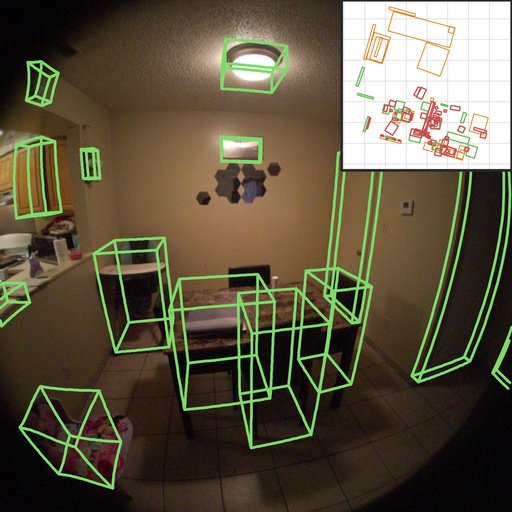} &
\img{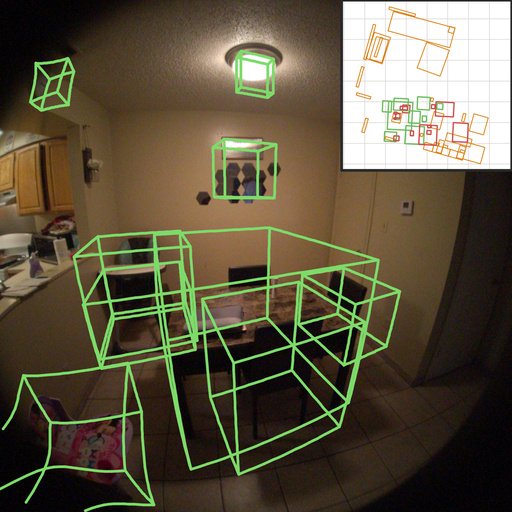} &
\img{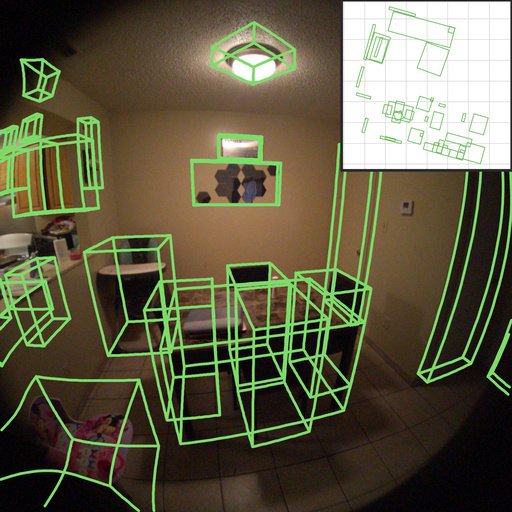} 
\\

\raisebox{20px}{\rotatebox{90}{iTHOR}} &
\img{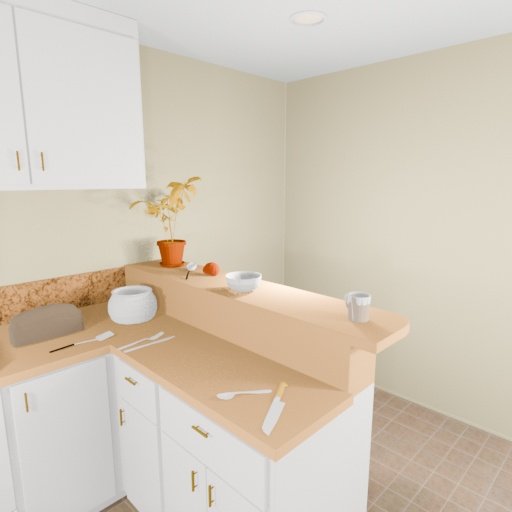} &
\img{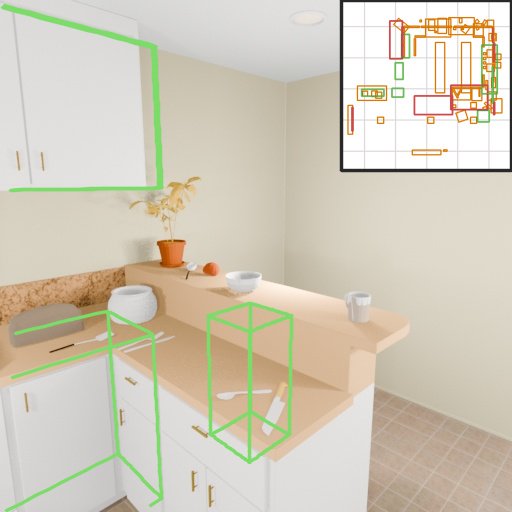} &
\img{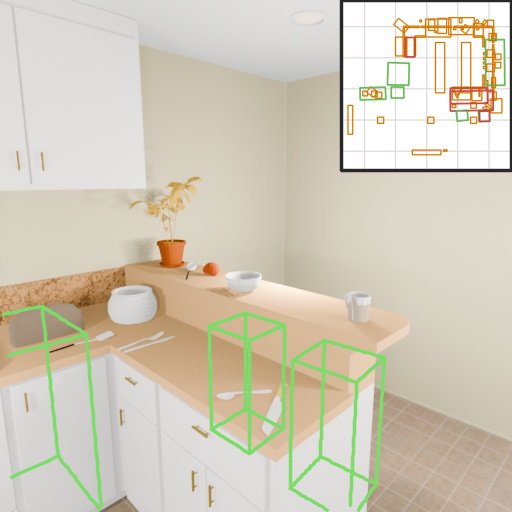} &
\img{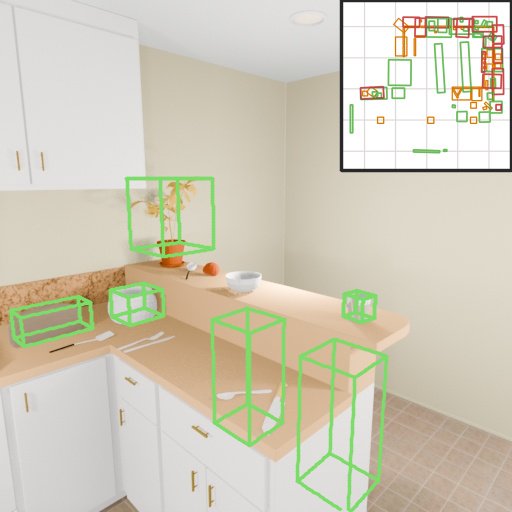} &
\img{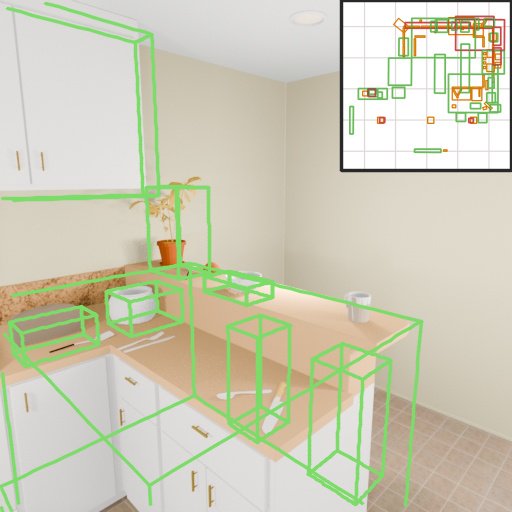} & 
\img{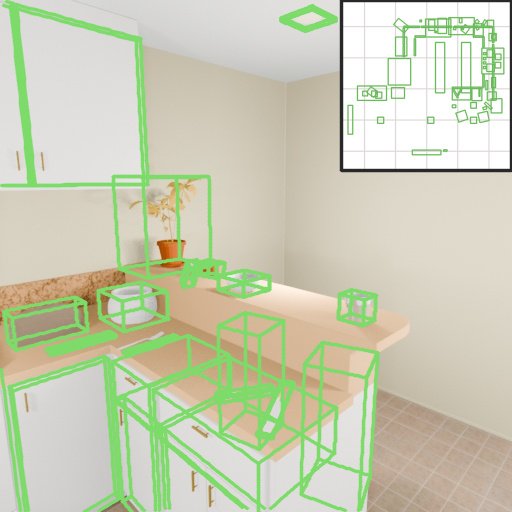}\\

\raisebox{15px}{\rotatebox{90}{Imaginarium}} &
\img{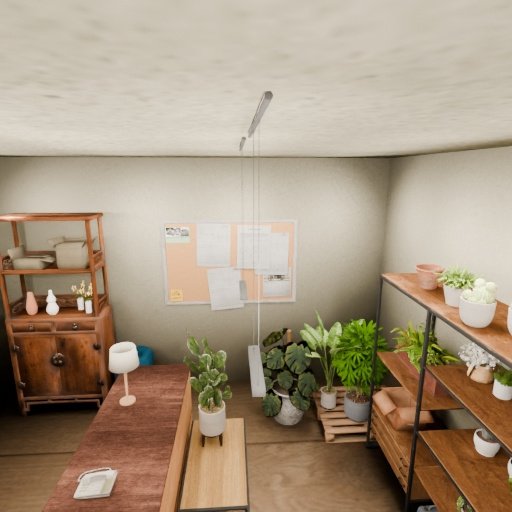} &
\img{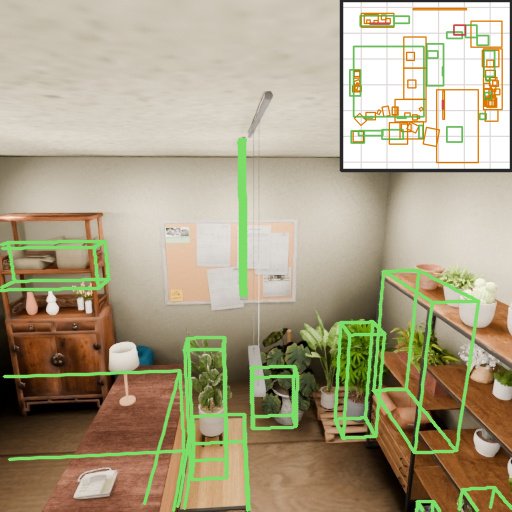} &
\img{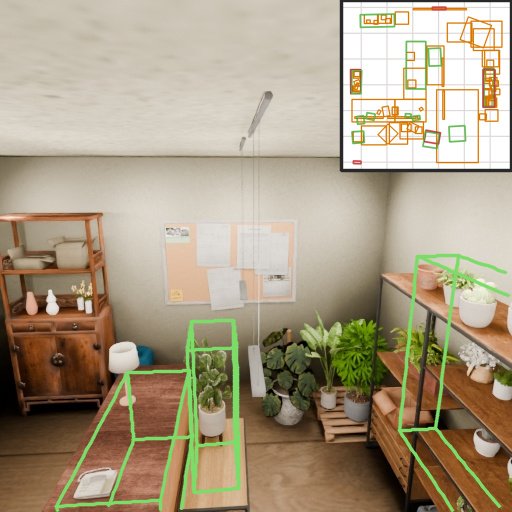} &
\img{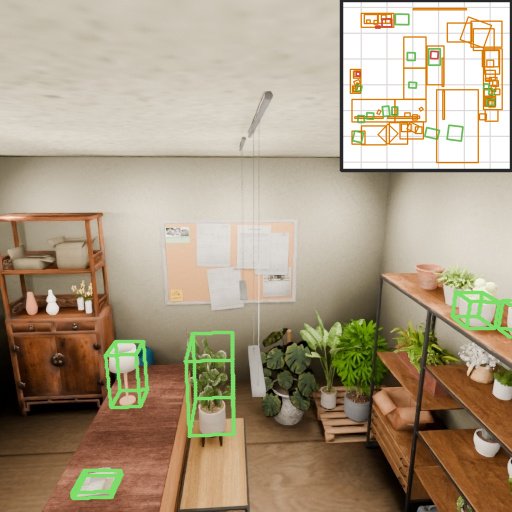} &
\img{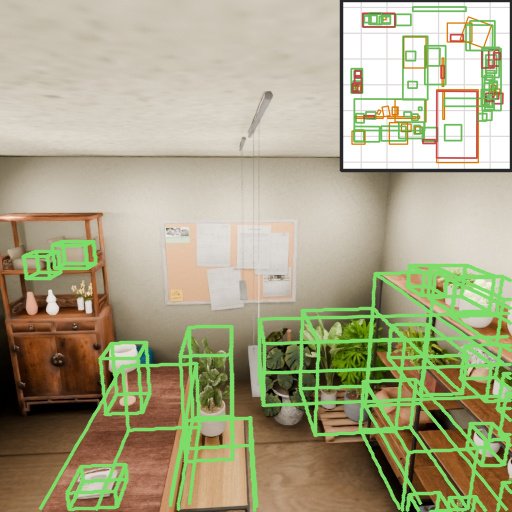} &
\img{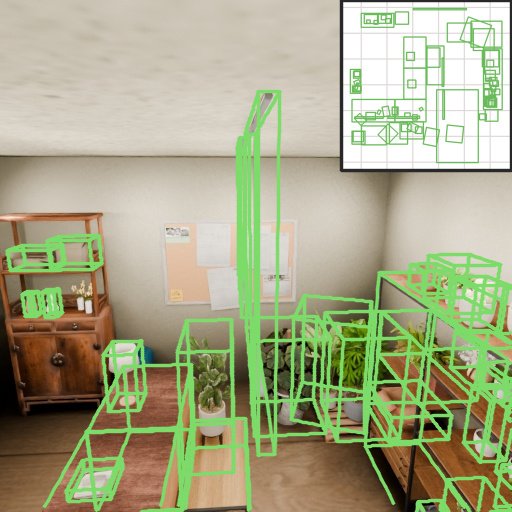} \\

\end{tabular}%
}
\caption{\textbf{Qualitative results of video-based 3D scene perception.}
Correct predictions are visualized in \textcolor{green}{green}. We show the overview perception in the inset figure. Wrong ones are marked in \textcolor{red}{red}, and the undetected are marked in \textcolor{orange}{orange}. \rev{FIRE3D produces coherent instance layouts in these examples.} 
}
\label{fig:perception}
\end{figure}

\begin{table*}[ht]
    \centering
    \setlength{\tabcolsep}{1pt}
    \renewcommand{\arraystretch}{0.5}

    \resizebox{\textwidth}{!}{%
    \begin{tabular}{c c c c c c c c}

        \includegraphics[width=0.245\linewidth]{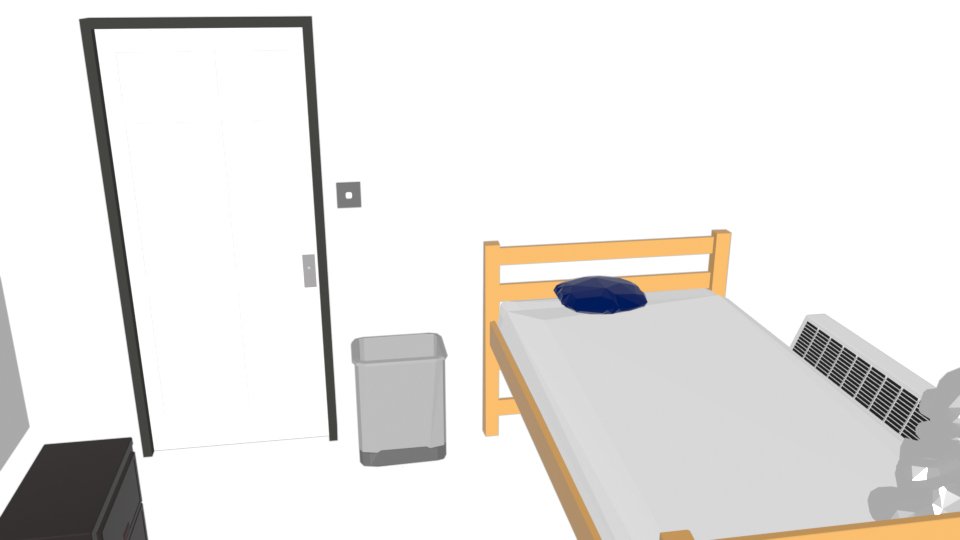} &
        \includegraphics[width=0.245\linewidth]{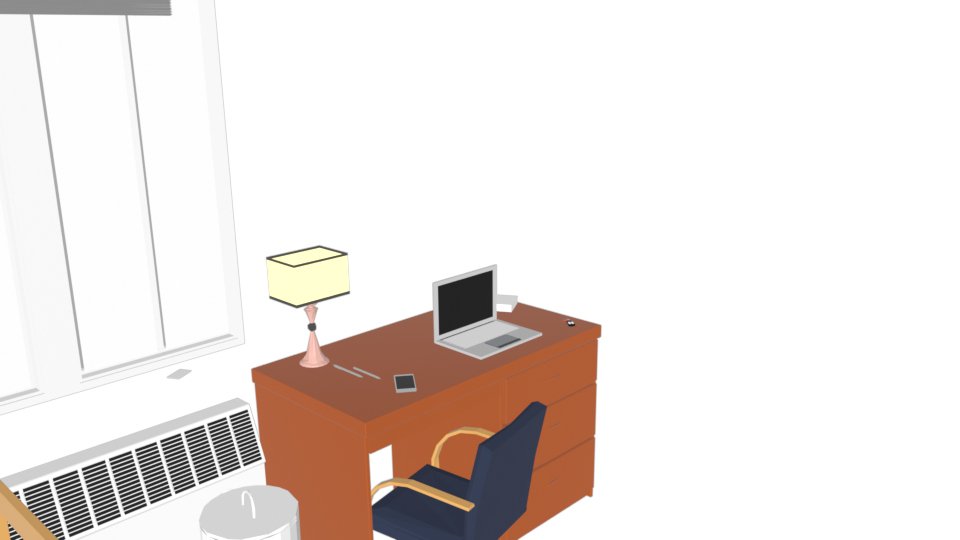} &
        \includegraphics[width=0.245\linewidth]{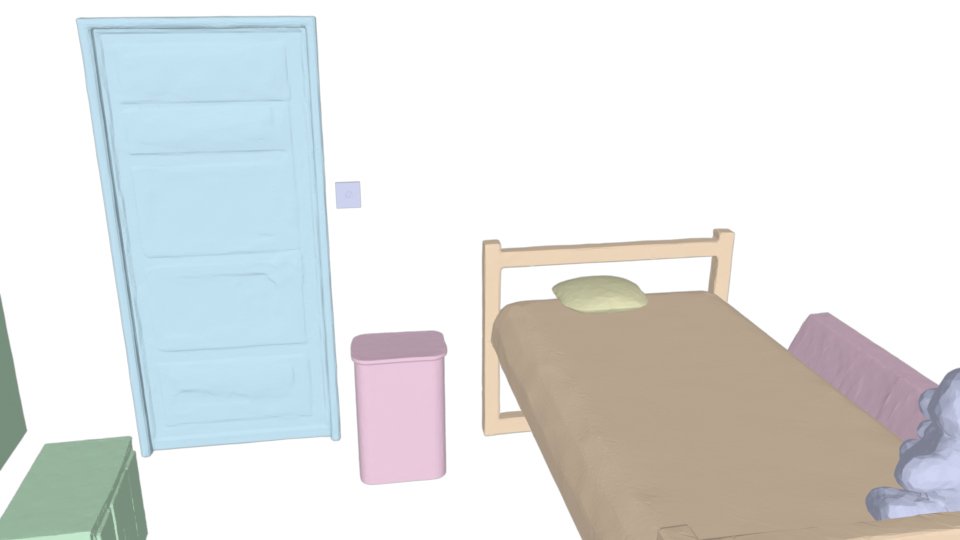} &
        \includegraphics[width=0.245\linewidth]{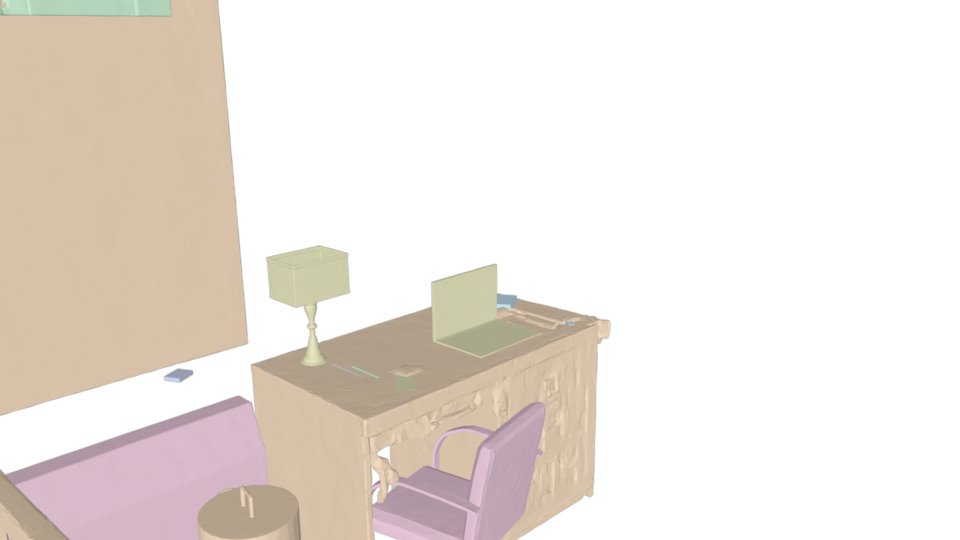} &
        \includegraphics[width=0.245\linewidth]{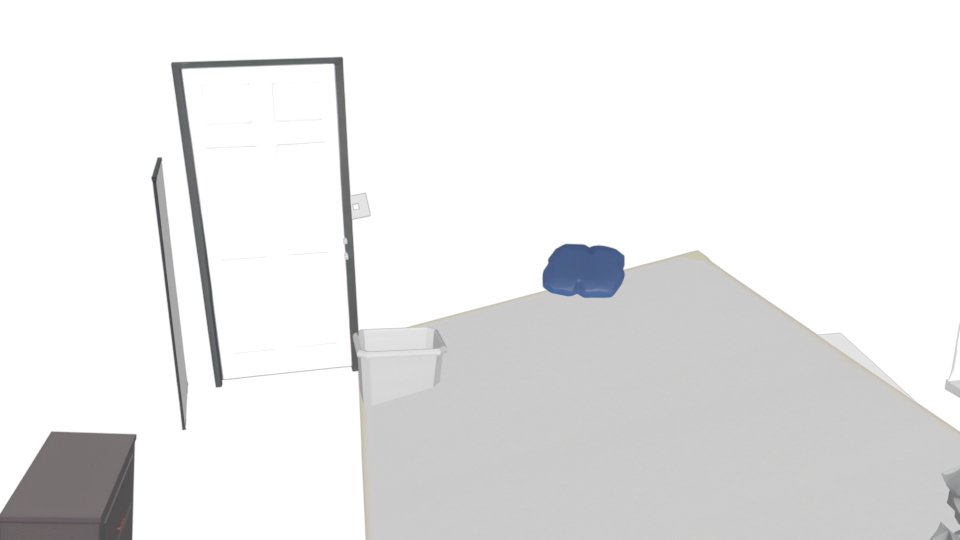} &
        \includegraphics[width=0.245\linewidth]{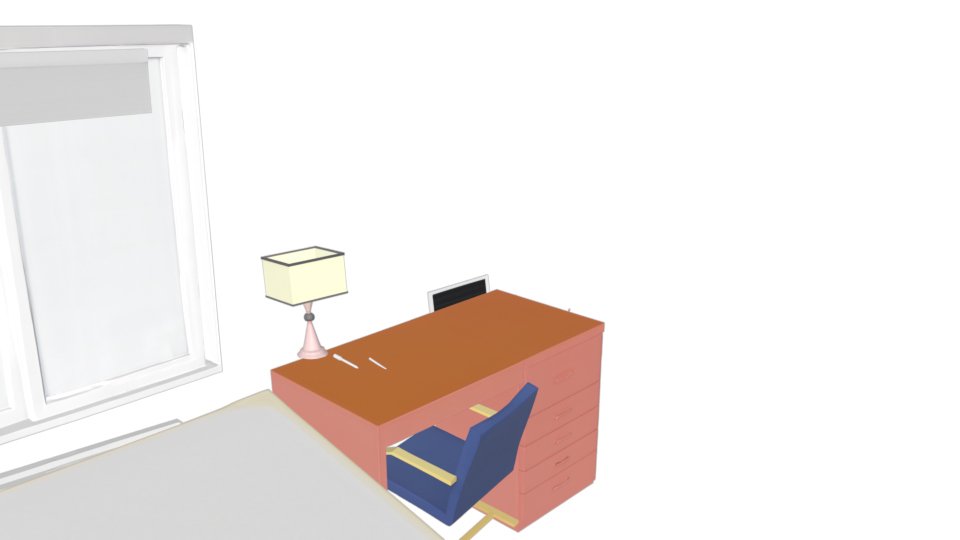} &
        \includegraphics[width=0.245\linewidth]{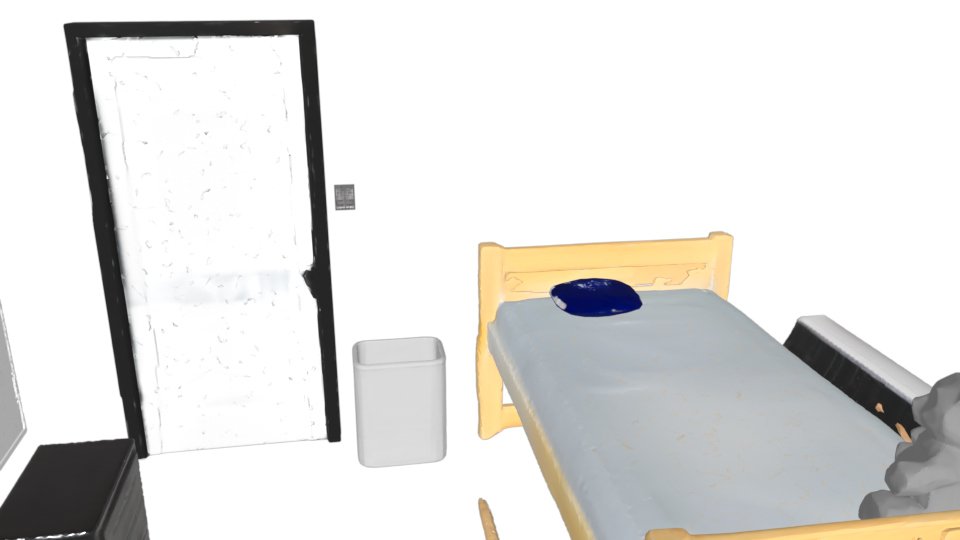} &
        \includegraphics[width=0.245\linewidth]{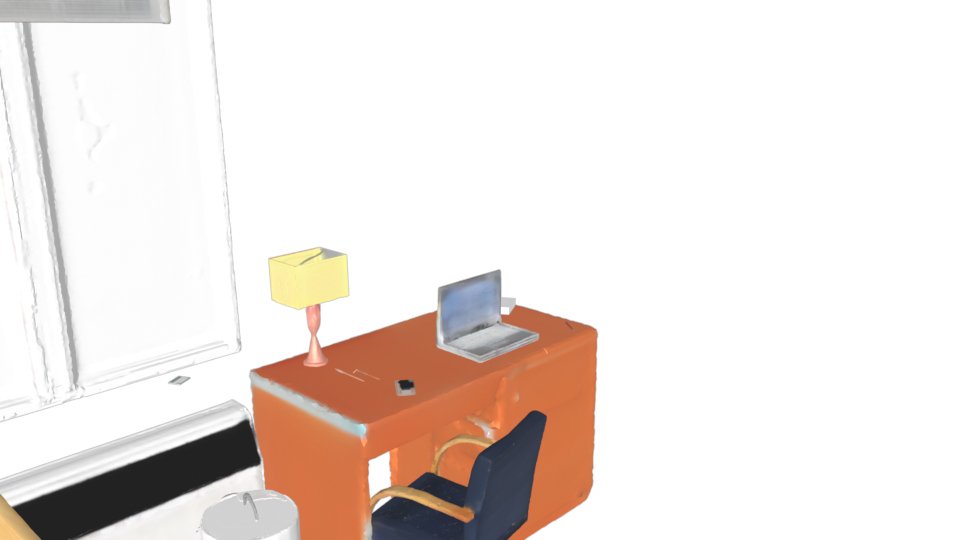} \\[1pt]

        \includegraphics[width=0.245\linewidth]{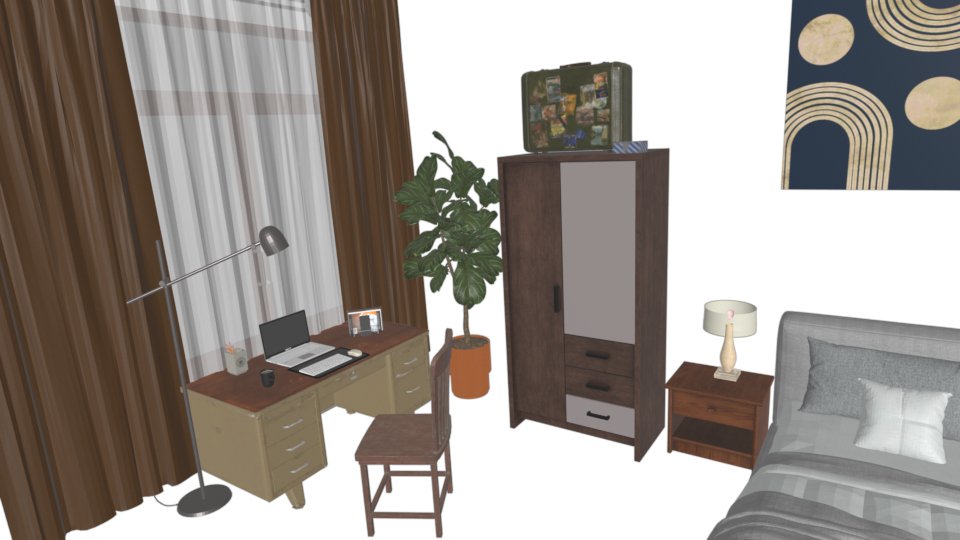} &
        \includegraphics[width=0.245\linewidth]{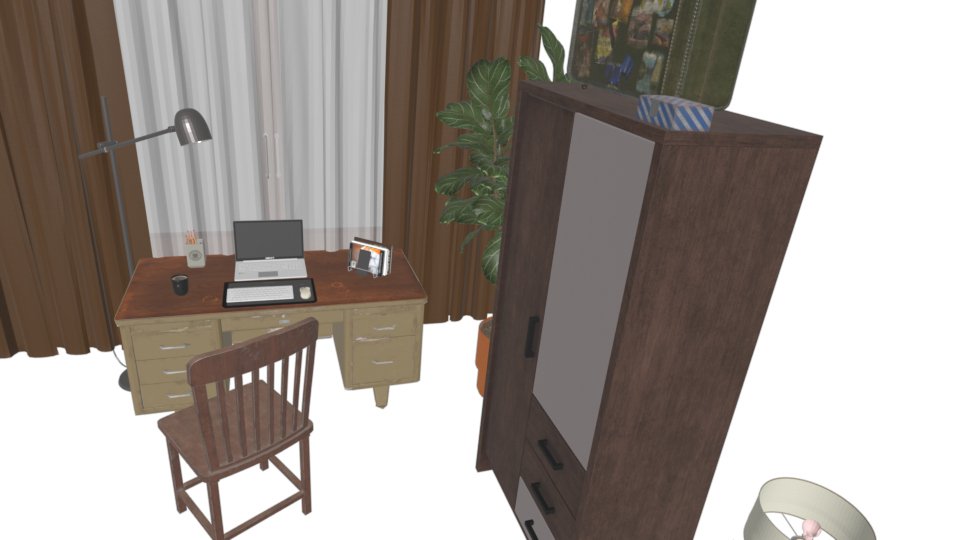} &
        \includegraphics[width=0.245\linewidth]{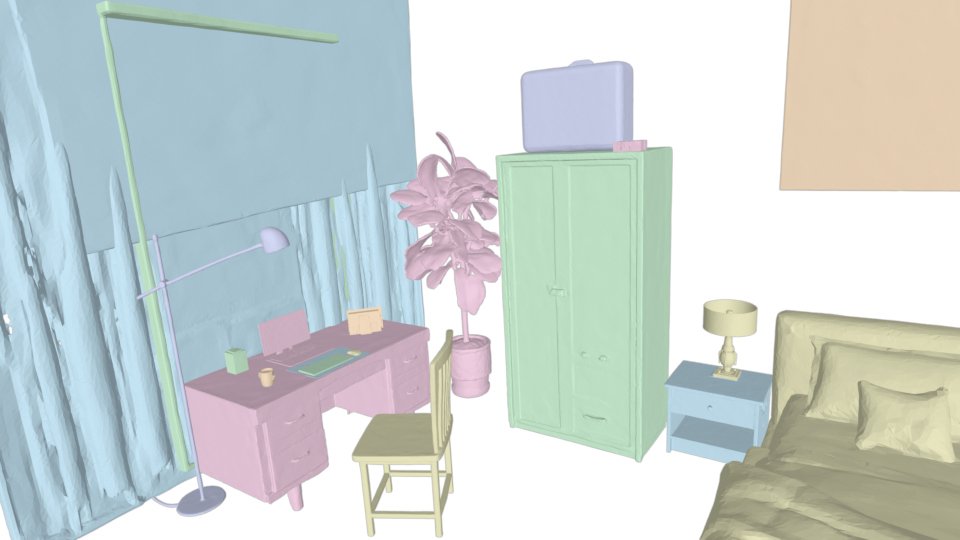} &
        \includegraphics[width=0.245\linewidth]{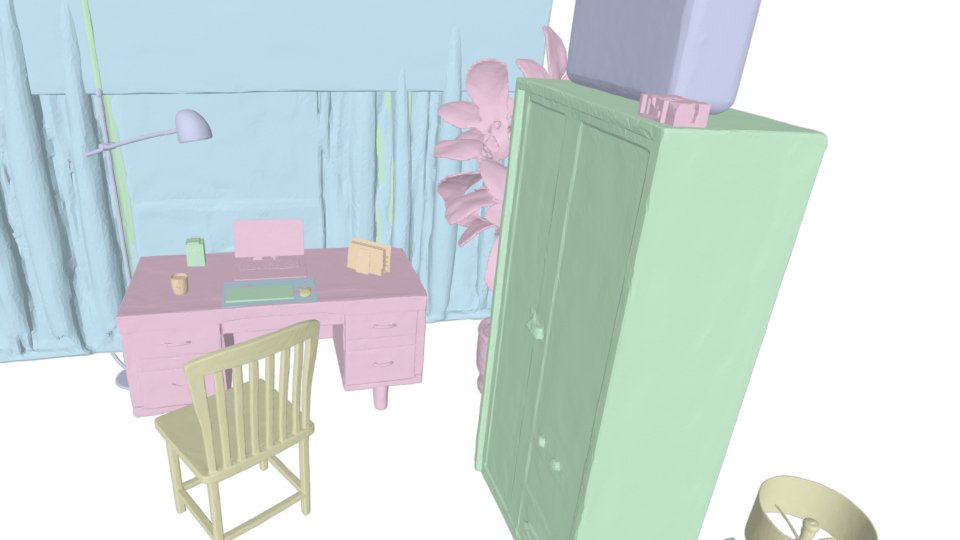} &
        \includegraphics[width=0.245\linewidth]{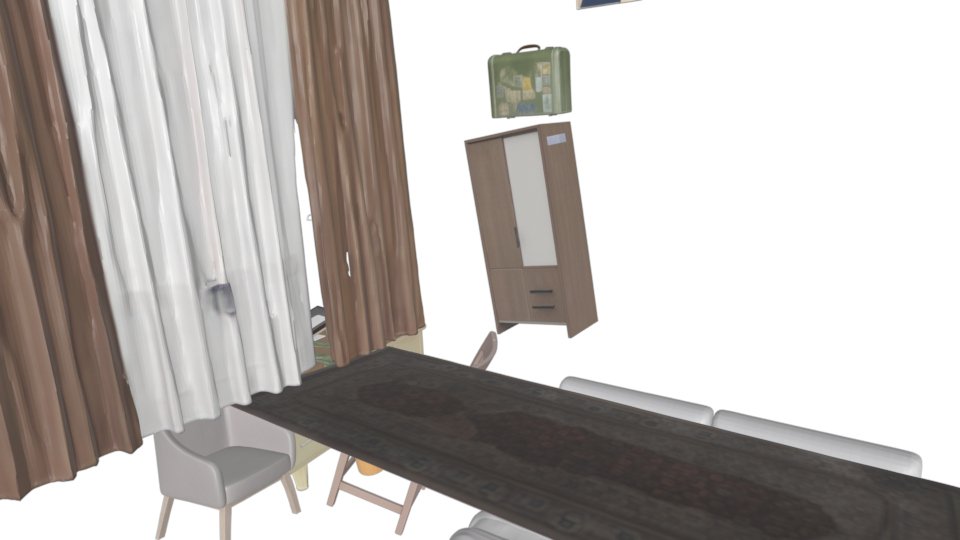} &
        \includegraphics[width=0.245\linewidth]{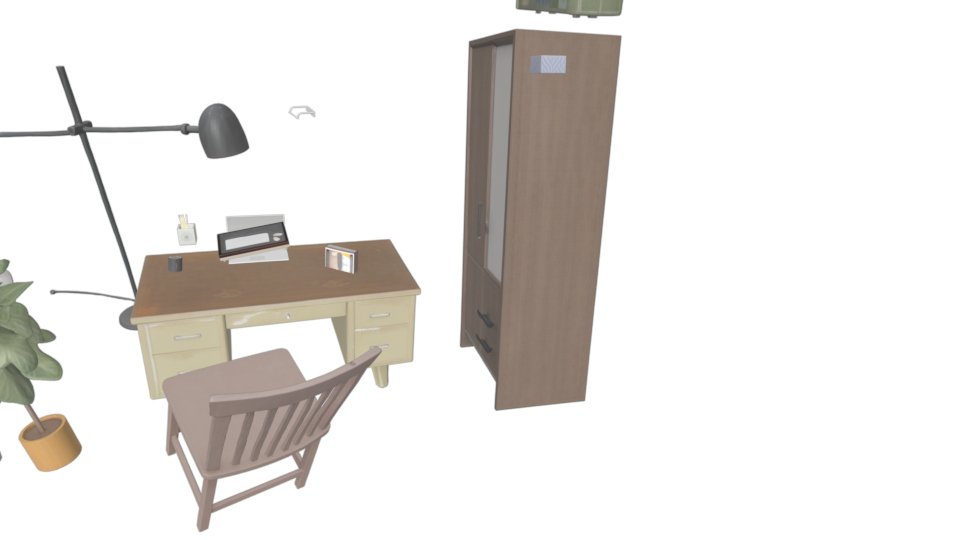} &
        \includegraphics[width=0.245\linewidth]{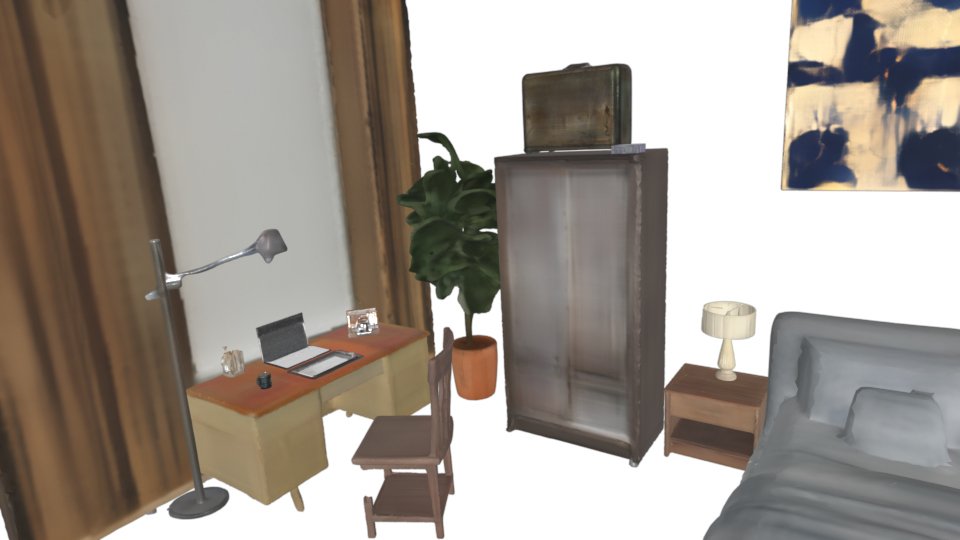} &
        \includegraphics[width=0.245\linewidth]{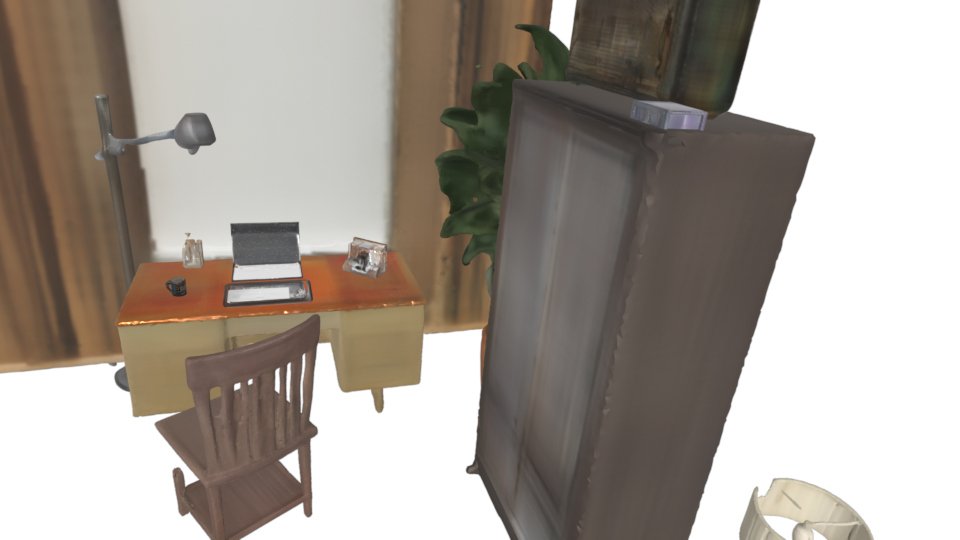} \\[1pt]

        GT View 1 & GT View 2
        & ShapeR View 1 & ShapeR View 2
        & \rev{SAM3D (frame sel.)} View 1 & \rev{SAM3D (frame sel.)} View 2
        & Ours View 1 & Ours View 2
    \end{tabular}%
    }

    \captionof{figure}{\textbf{Qualitative 3D scene reconstruction with ground-truth instance perception.}  \rev{FIRE3D generates textured assets from 3D conditions. ShapeR outputs geometry only; SAM3D uses per-object frame selection.} We exclude BG for matched comparisons.
    }
    \label{fig:recon_cmp_gt_perception}
\end{table*}

\begin{table*}[ht]
    \centering
    \setlength{\tabcolsep}{1pt}
    \renewcommand{\arraystretch}{0.5}

    \resizebox{\textwidth}{!}{%
    \begin{tabular}{c c c c c c}
        \includegraphics[width=0.245\linewidth]{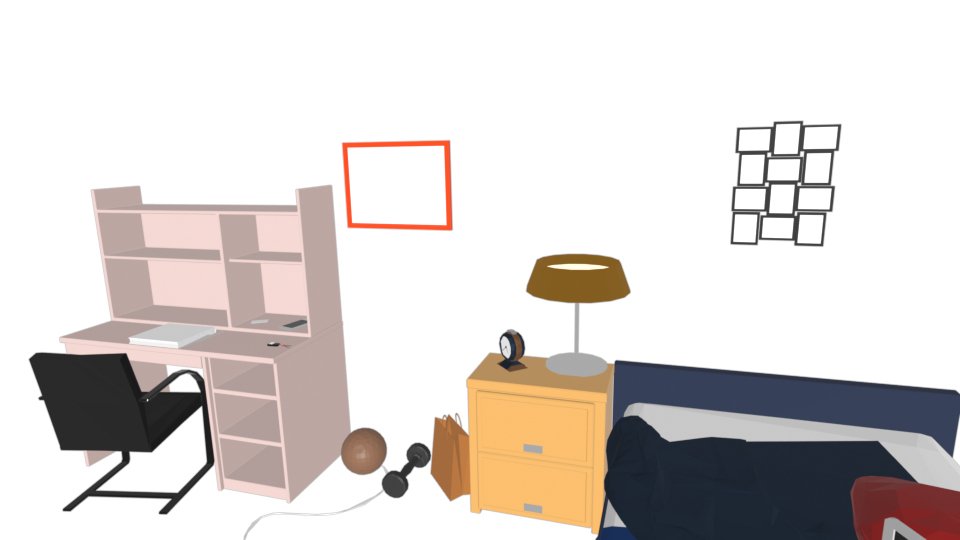} &
        \includegraphics[width=0.245\linewidth]{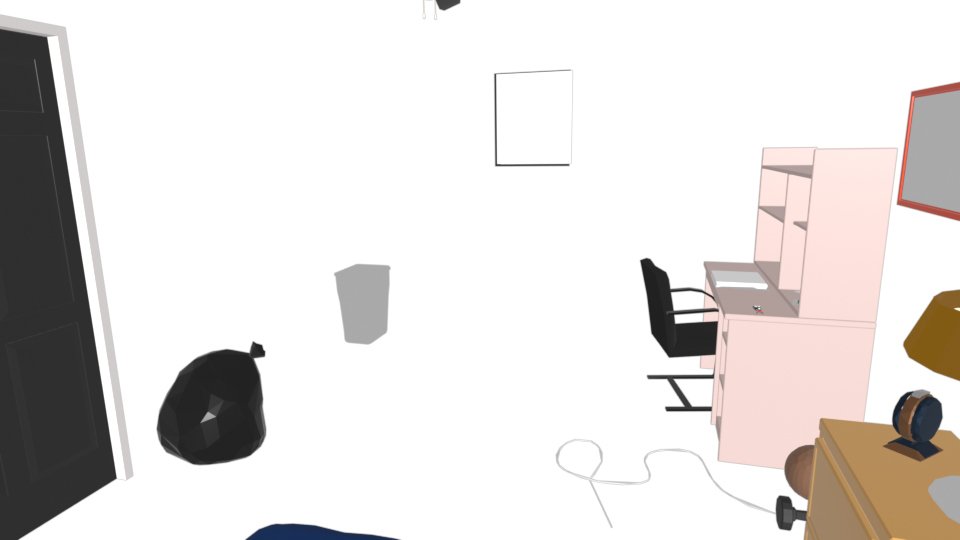} &
        \includegraphics[width=0.245\linewidth]{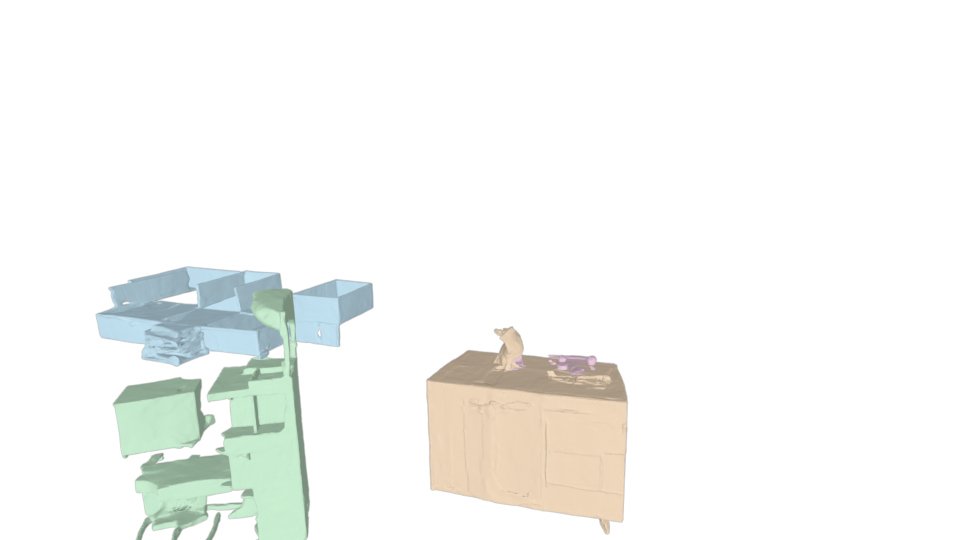} &
        \includegraphics[width=0.245\linewidth]{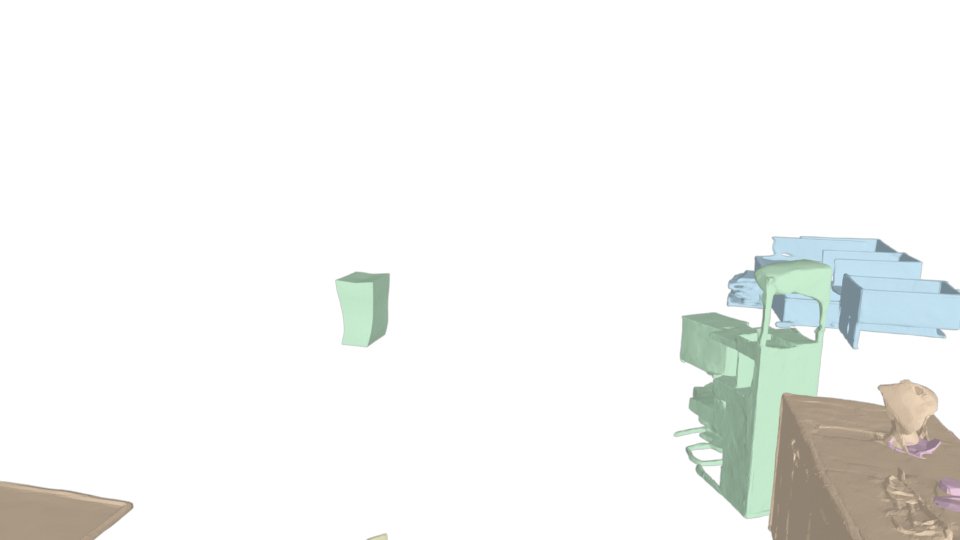} &
        \includegraphics[width=0.245\linewidth]{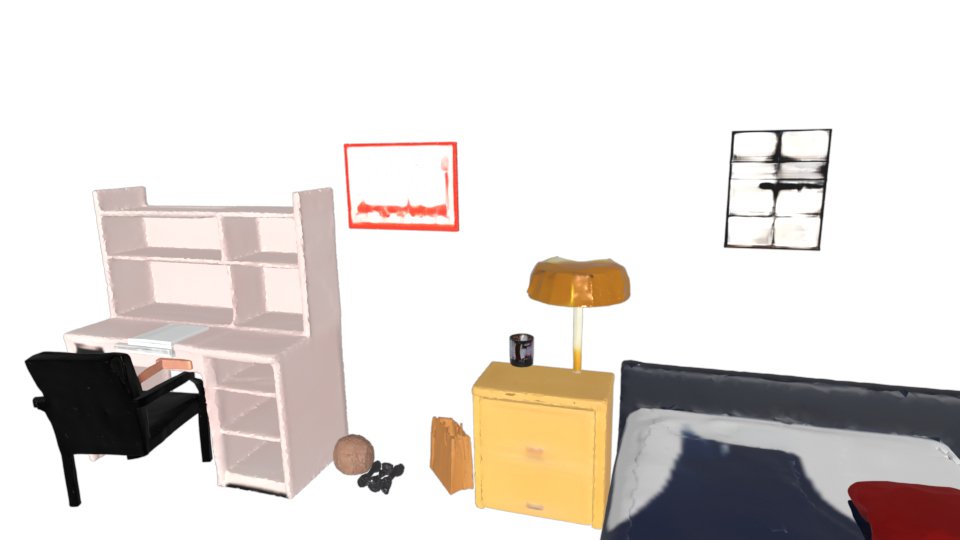} &
        \includegraphics[width=0.245\linewidth]{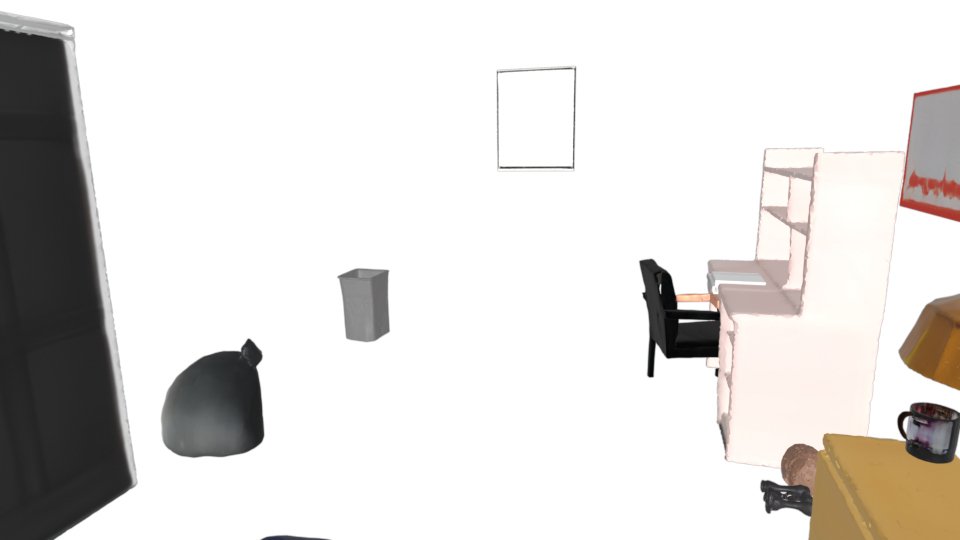} \\[1pt]
        
        \includegraphics[width=0.245\linewidth]{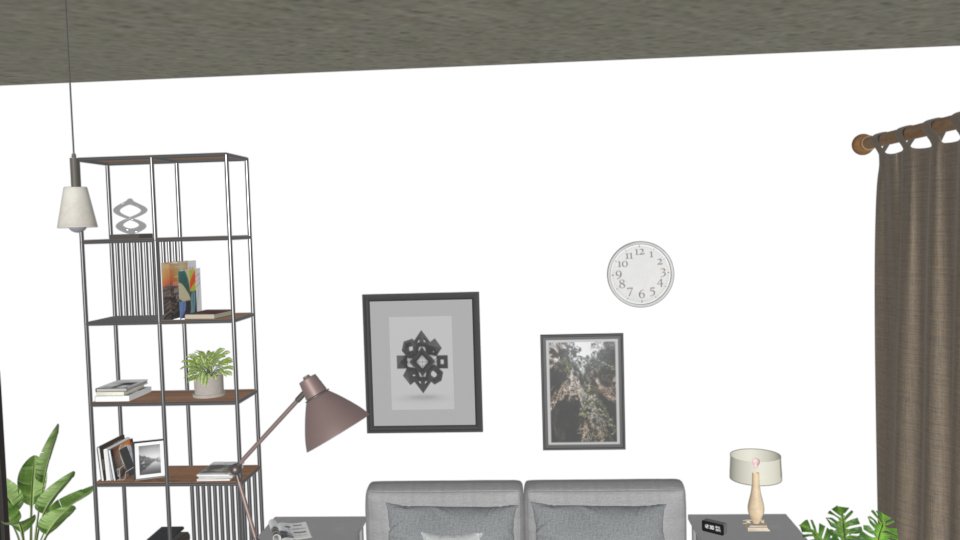} &
        \includegraphics[width=0.245\linewidth]{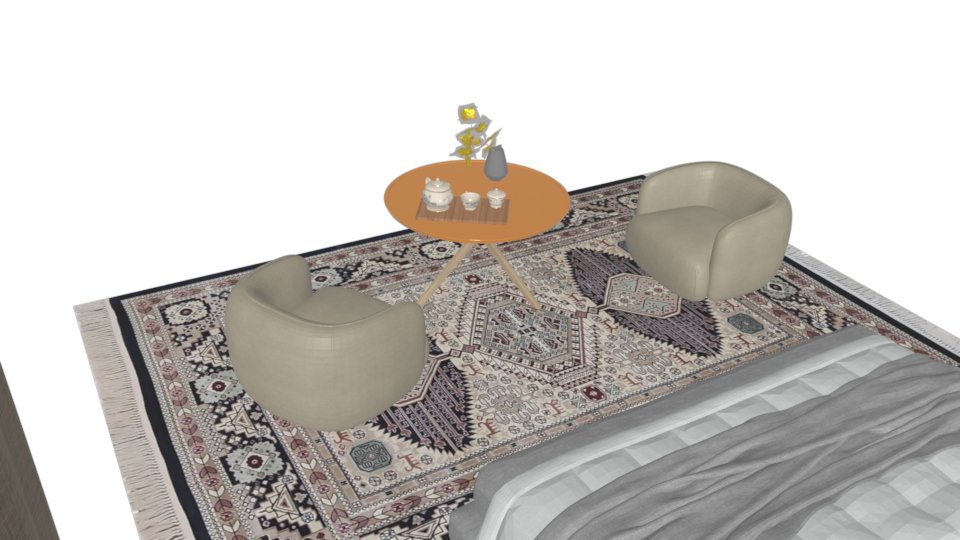} &
        \includegraphics[width=0.245\linewidth]{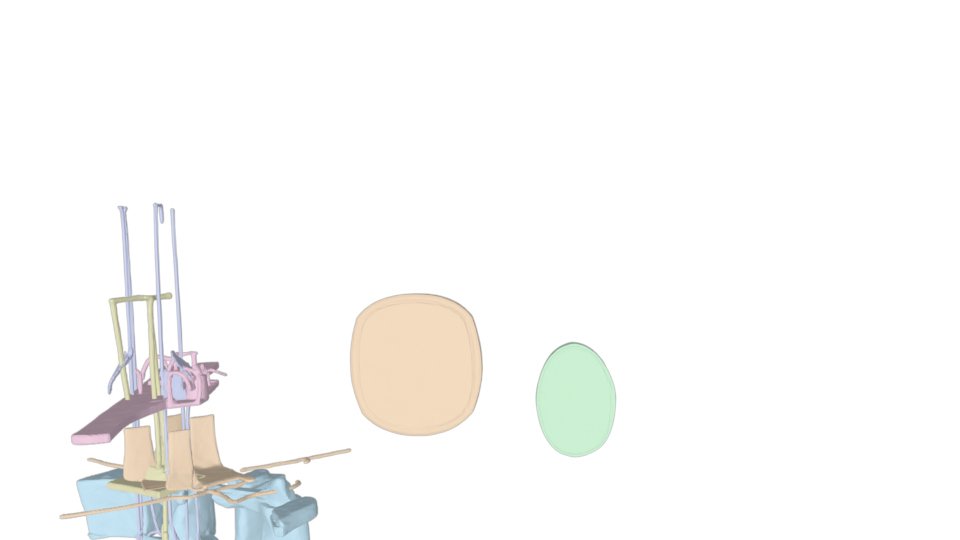} &
        \includegraphics[width=0.245\linewidth]{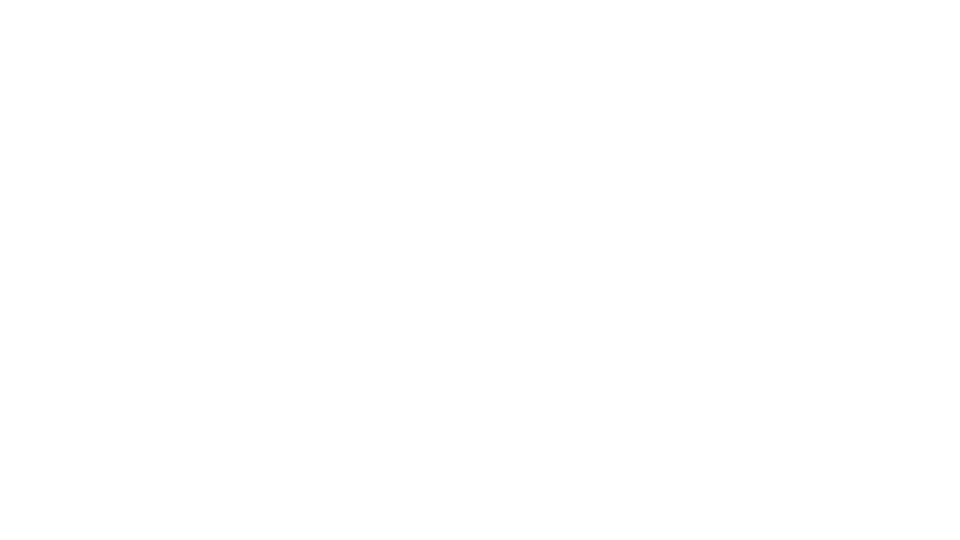} &
        \includegraphics[width=0.245\linewidth]{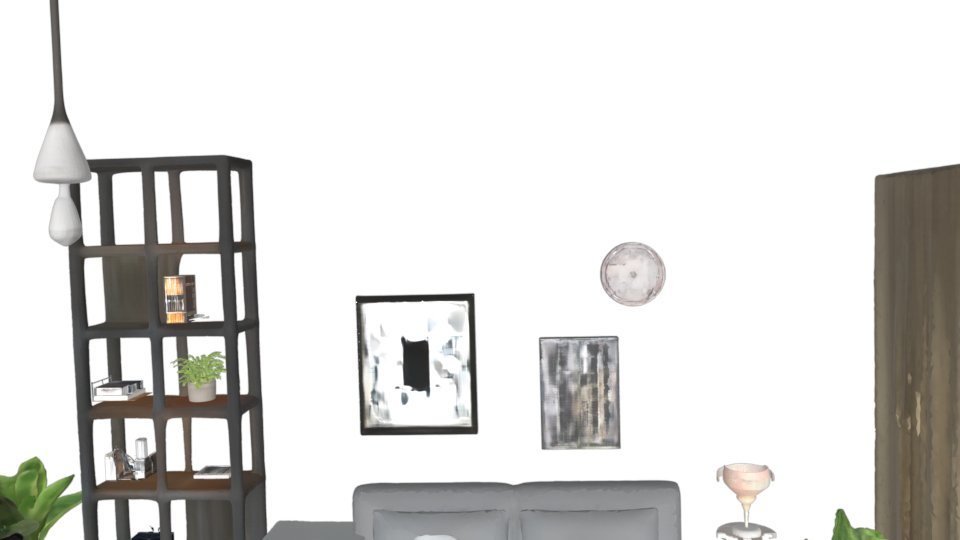} &
        \includegraphics[width=0.245\linewidth]{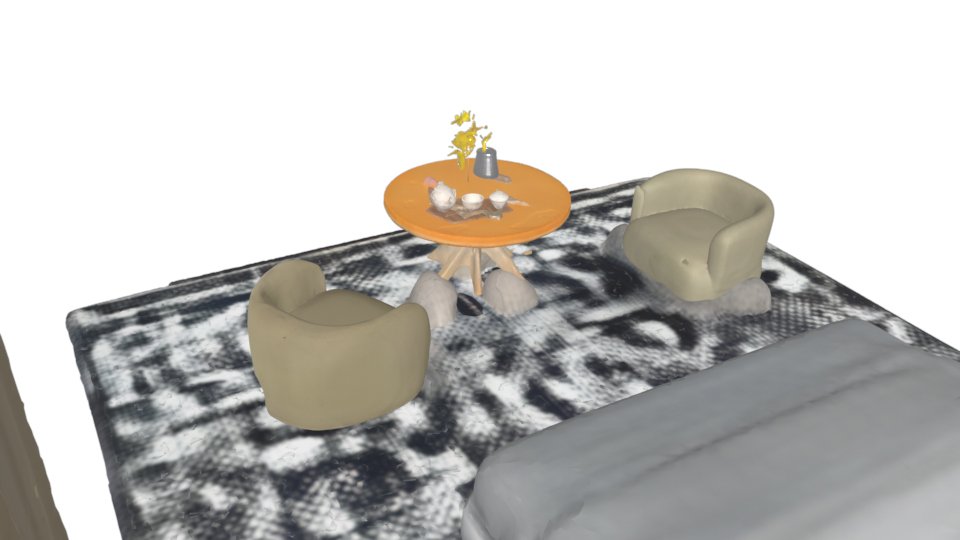} \\[1pt]
        GT View 1 & GT View 2
        & ShapeR View 1 & ShapeR View 2
        & Ours View 1 & Ours View 2
    \end{tabular}%
    }

    \captionof{figure}{\textbf{Qualitative 3D scene reconstruction from inferred perception.} \rev{FIRE3D reconstructs textured objects from estimated instances; the examples also illustrate errors propagated from perception.} We exclude BG for matched comparisons.
}
    \label{fig:recon_cmp_inferred_perception}
\end{table*}

\begin{table*}[ht]
    \centering
    \setlength{\tabcolsep}{1pt}
    \renewcommand{\arraystretch}{0.5}
    \resizebox{0.9\textwidth}{!}{%
    \begin{tabular}{ccccc}
        \includegraphics[width=0.195\linewidth]{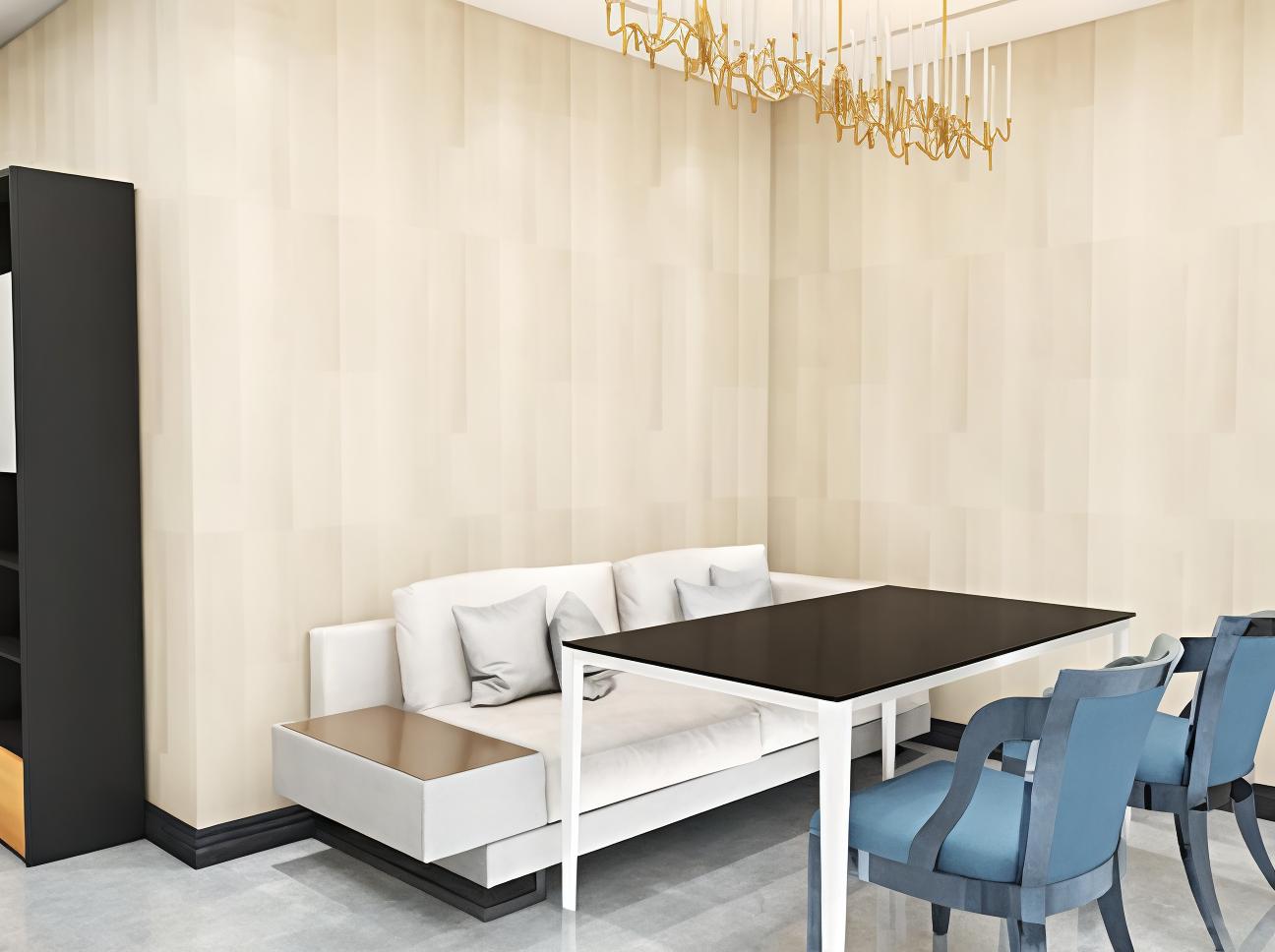} &
        \includegraphics[width=0.195\linewidth]{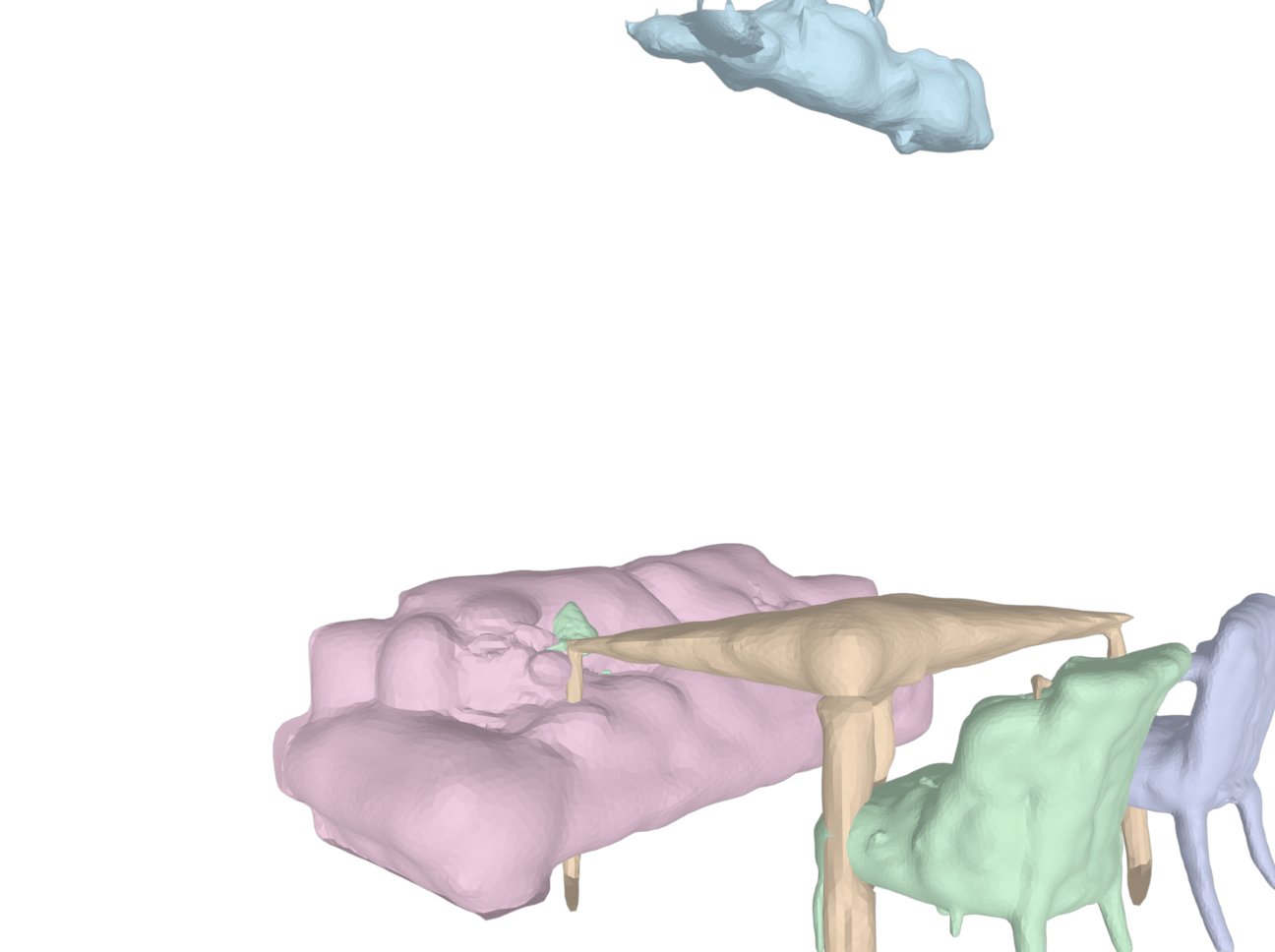} &
        \includegraphics[width=0.195\linewidth]{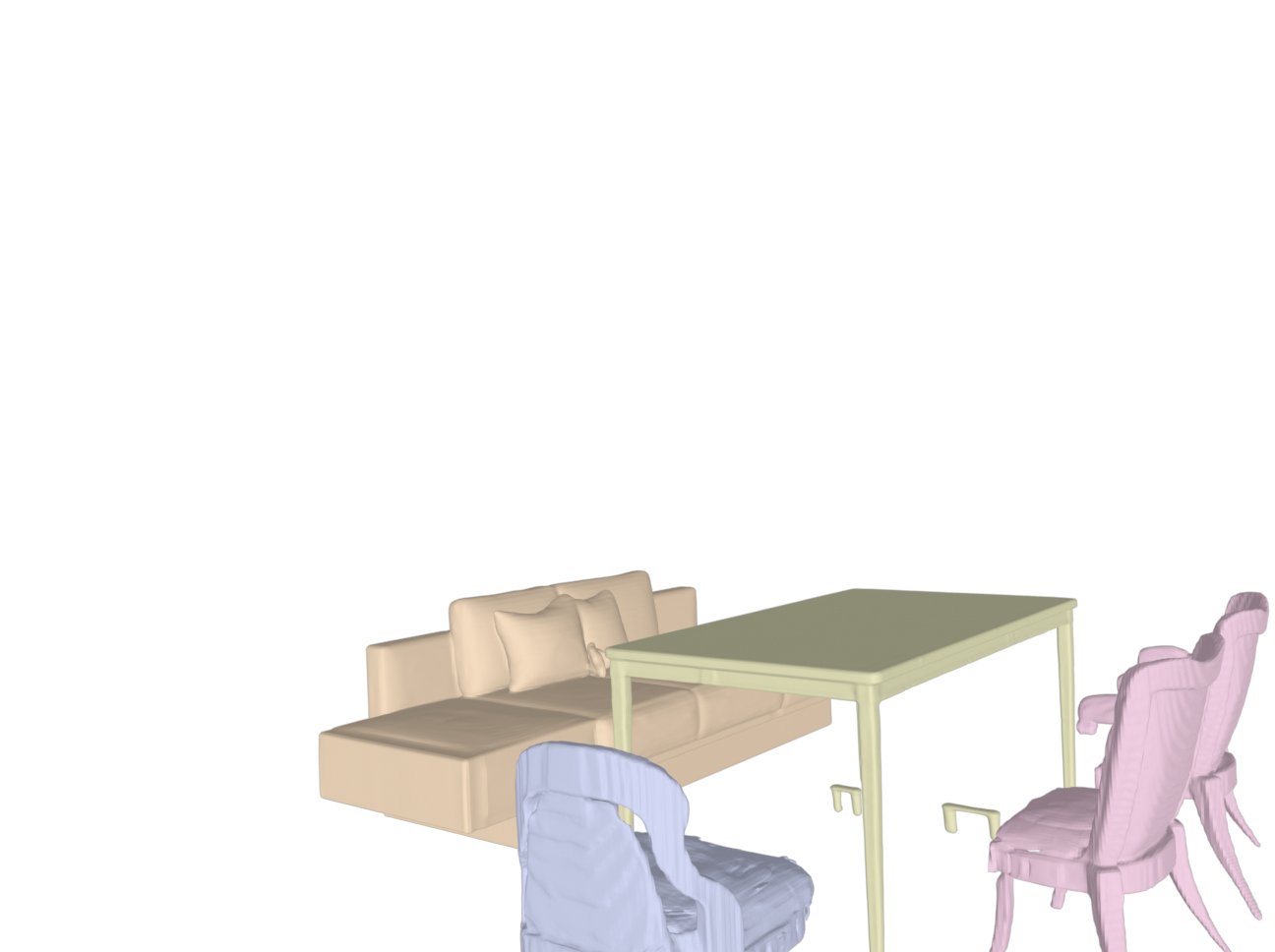} &
        \includegraphics[width=0.195\linewidth]{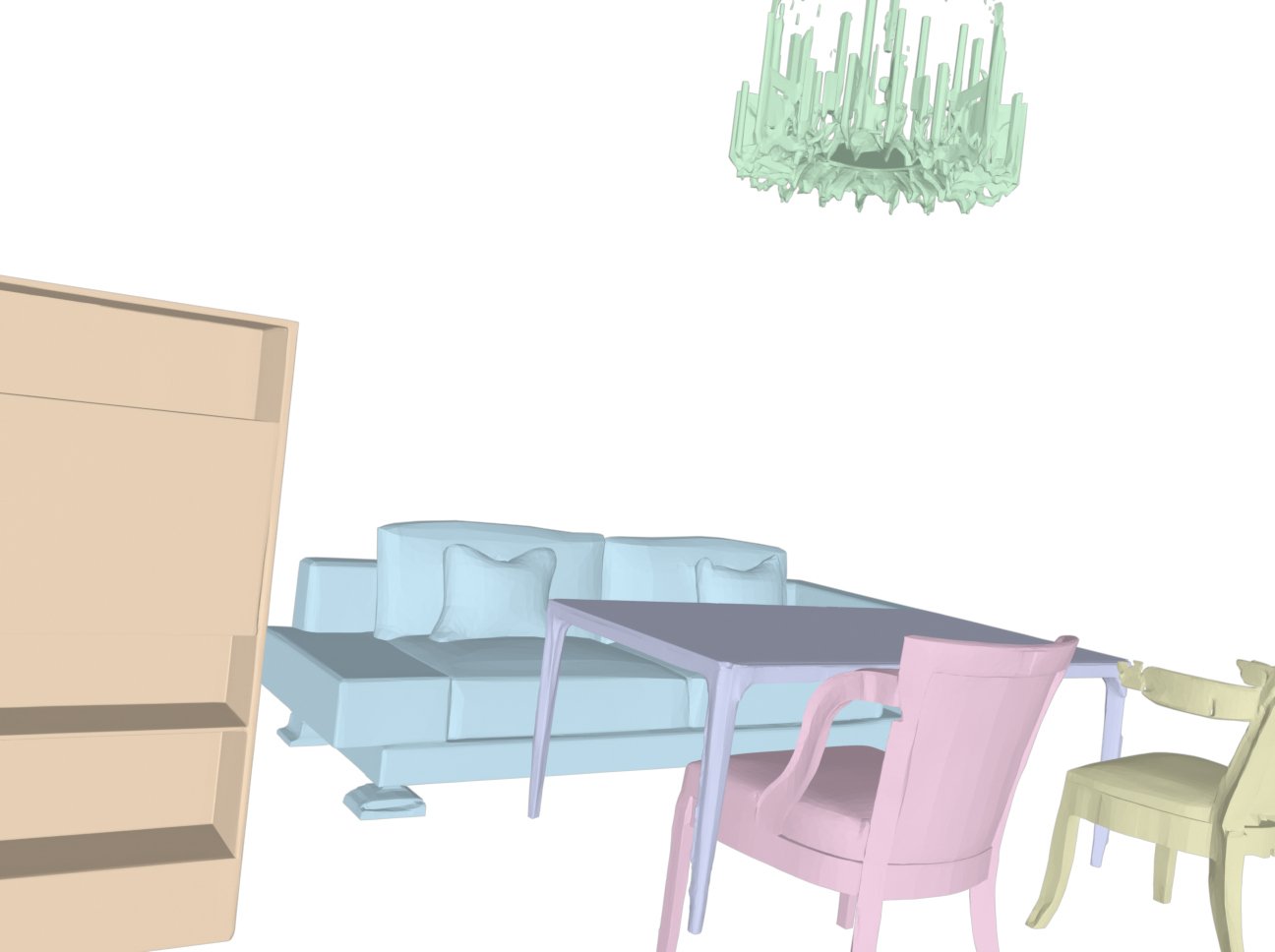} &
        \includegraphics[width=0.195\linewidth]{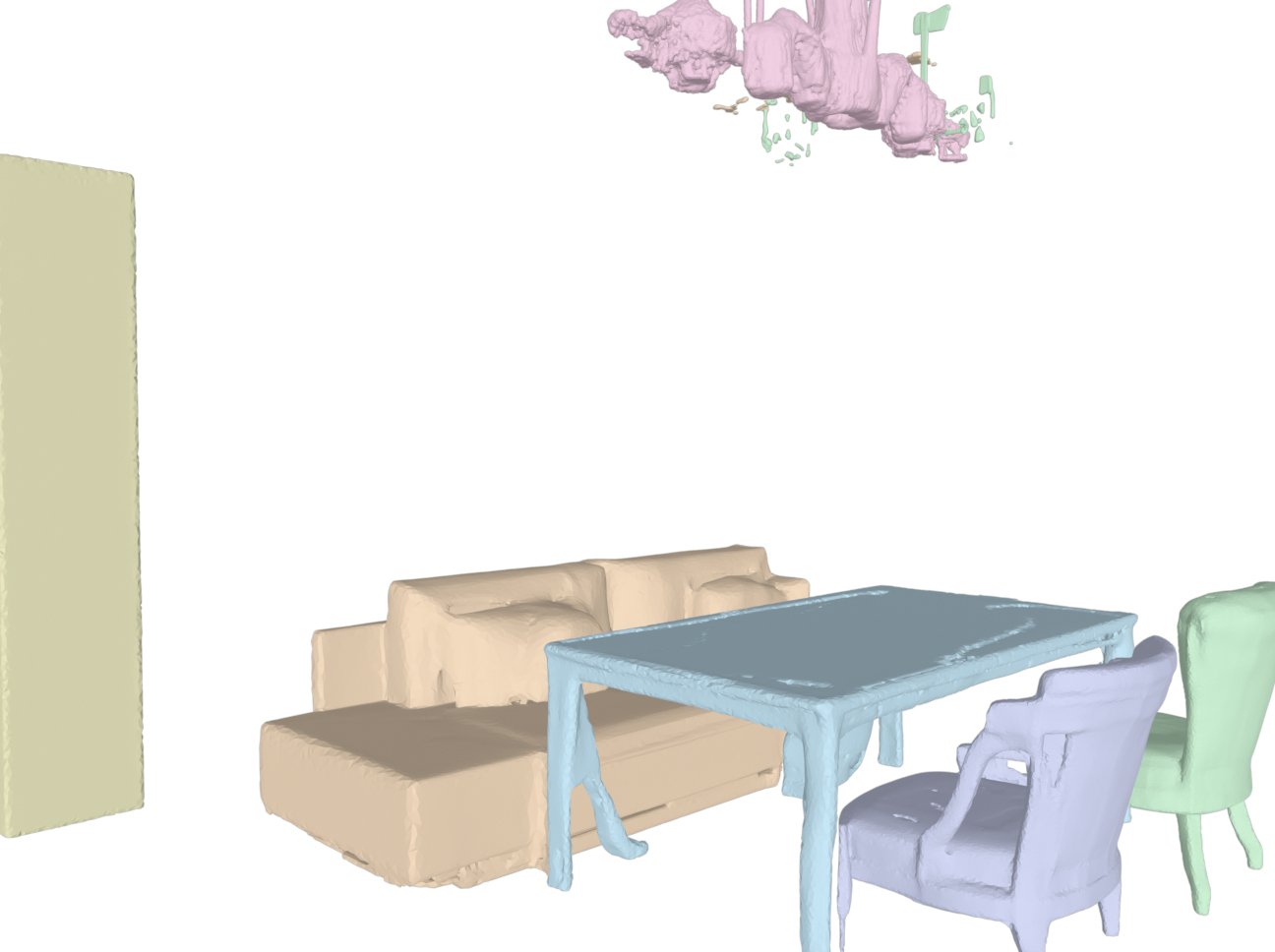} \\[2pt]

        \includegraphics[width=0.195\linewidth]{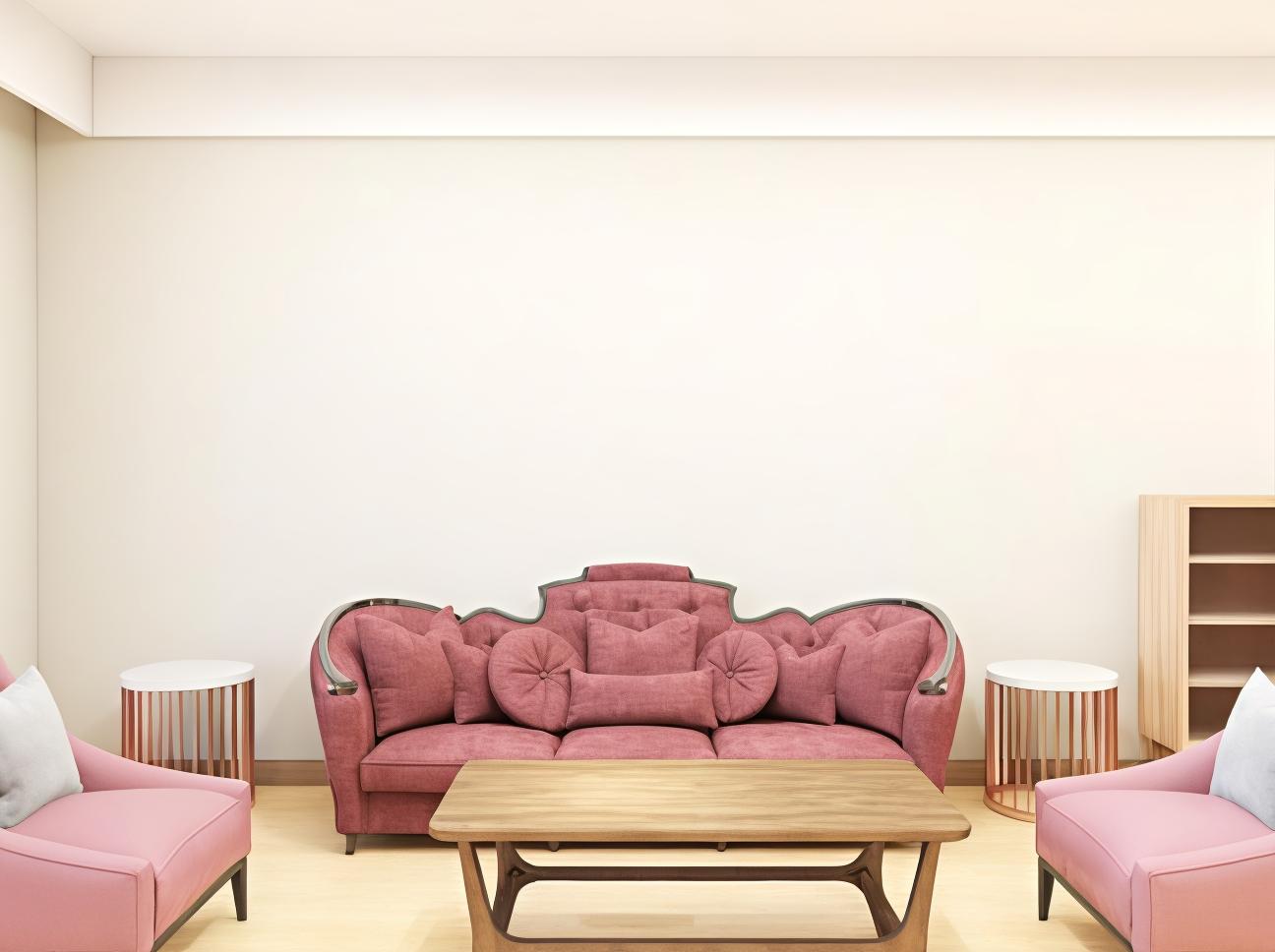} &
        \includegraphics[width=0.195\linewidth]{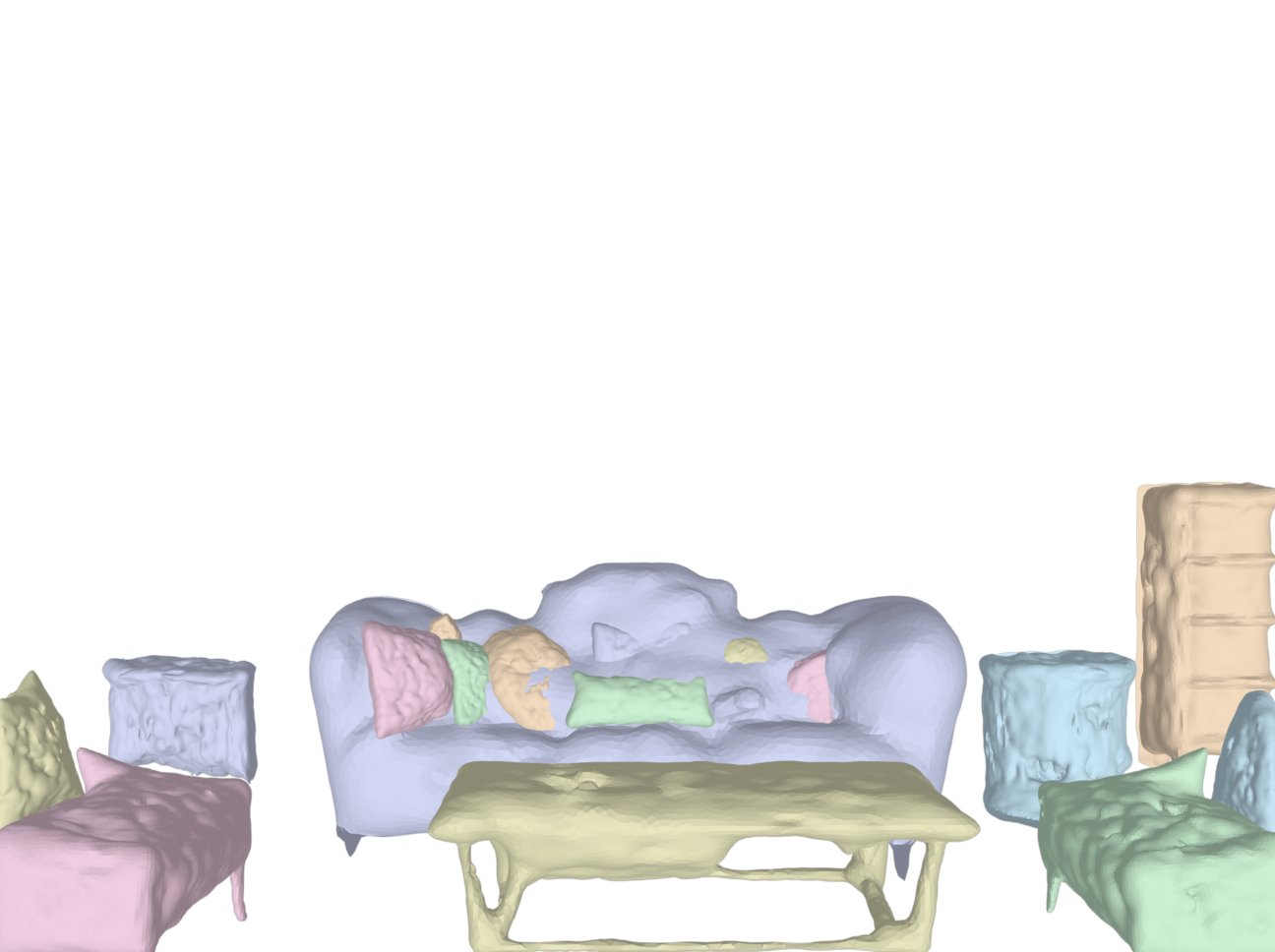} &
        \includegraphics[width=0.195\linewidth]{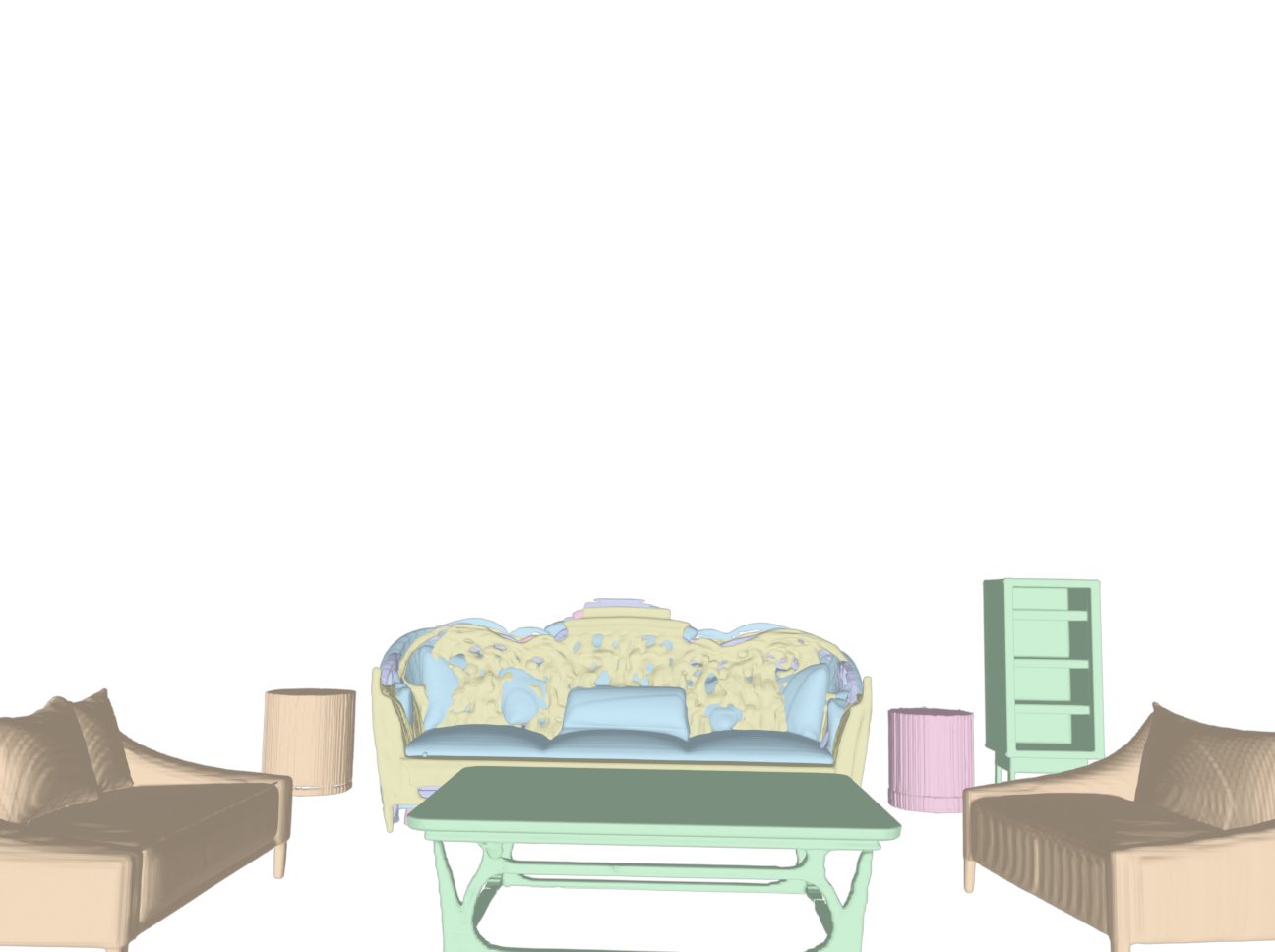} &
        \includegraphics[width=0.195\linewidth]{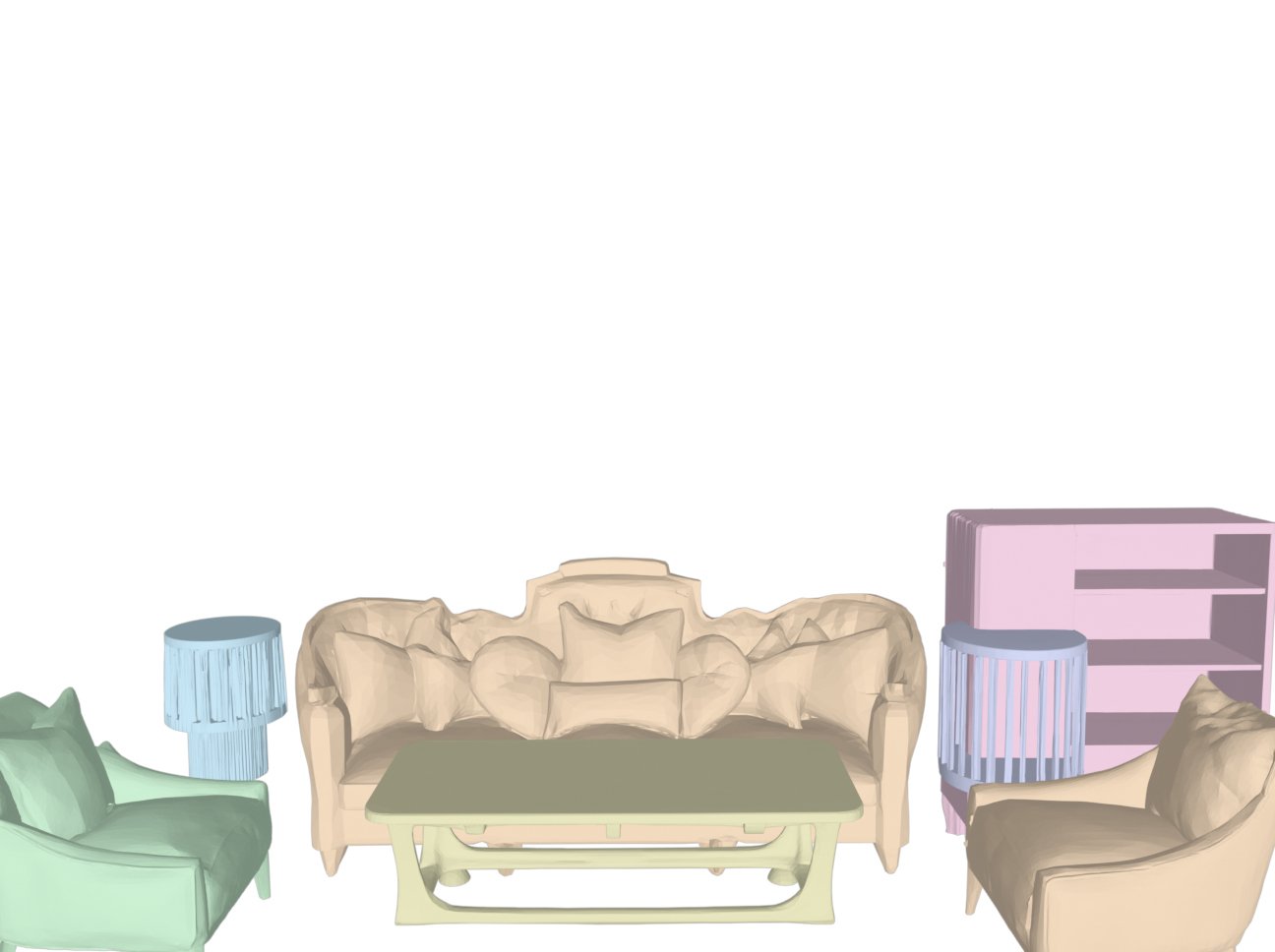} &
        \includegraphics[width=0.195\linewidth]{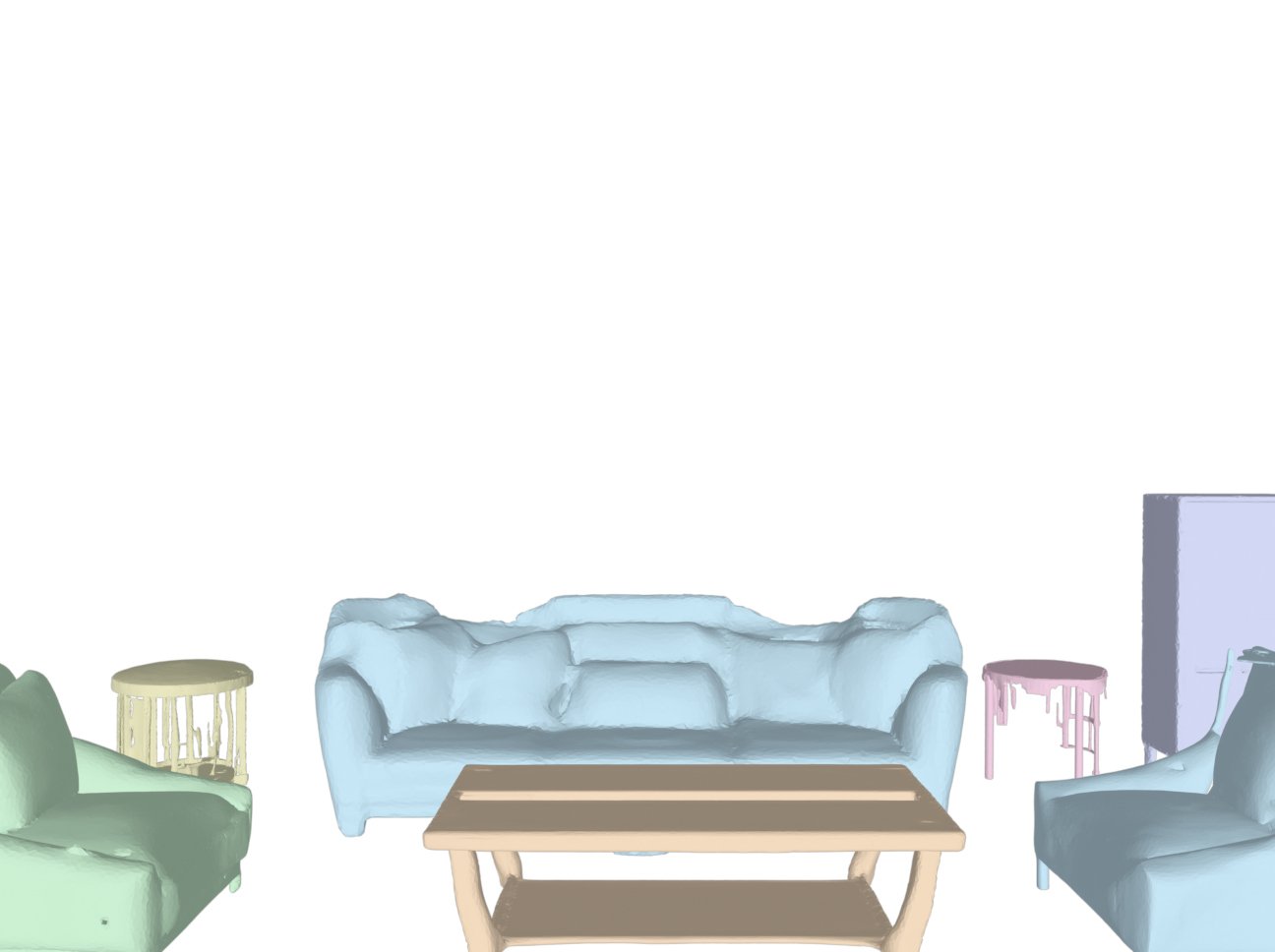} \\[2pt]

        Input & Gen3DSR \cite{gen3dsr} & MIDI \cite{midi} & SceneGen \cite{meng2025scenegen} & Ours
    \end{tabular}
    }
    \captionof{figure}{\textbf{Qualitative results of single-image 3D scene reconstruction.} \rev{\model{} produces coherent object geometry and layouts in these examples.} We exclude BG for matched comparisons.}
    \label{fig:single_image_recon_cmp}
\end{table*}

\begin{table*}[ht]
    \centering

    \setlength{\tabcolsep}{1pt}
    \renewcommand{\arraystretch}{0.5}

    \resizebox{\textwidth}{!}{%
    \begin{tabular}{cccccc}

        \includegraphics[width=0.25\linewidth]{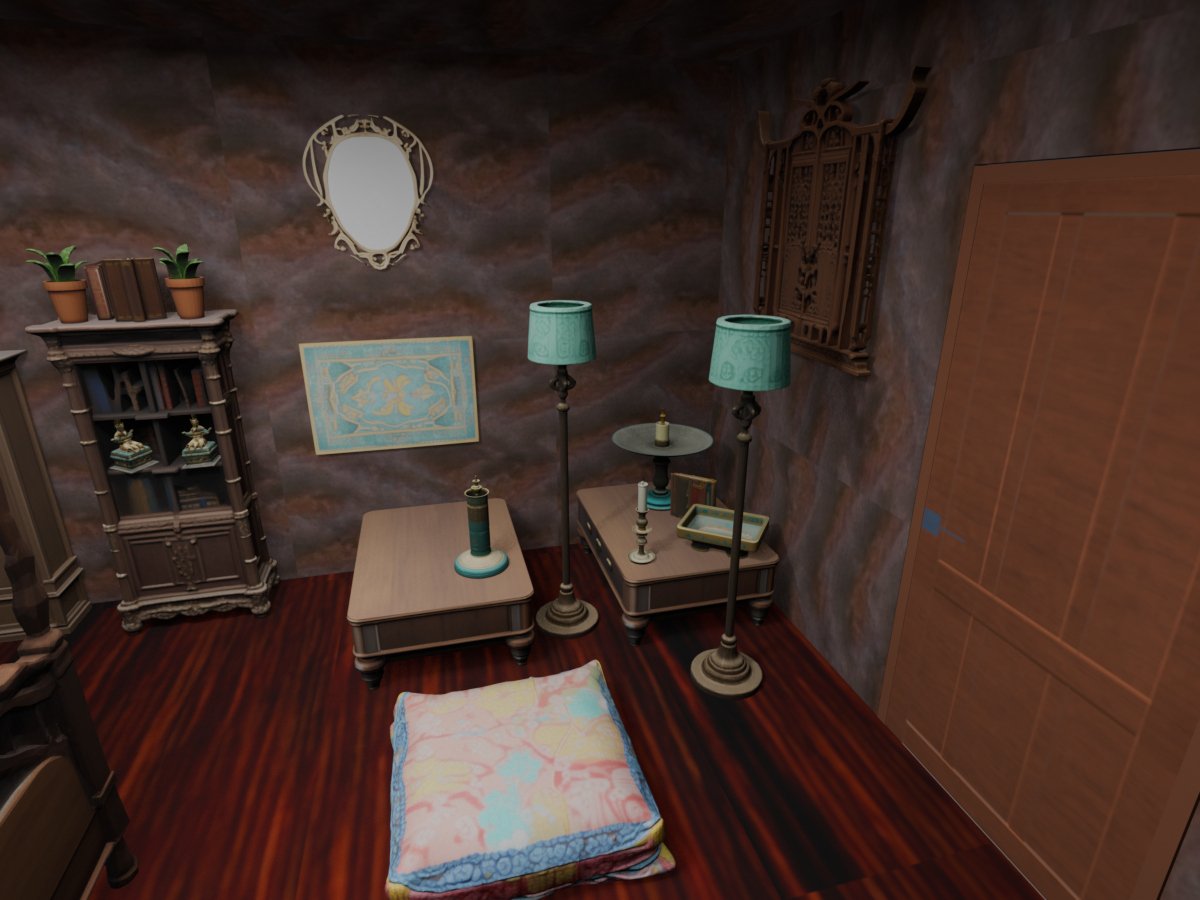} &
        \includegraphics[width=0.25\linewidth]{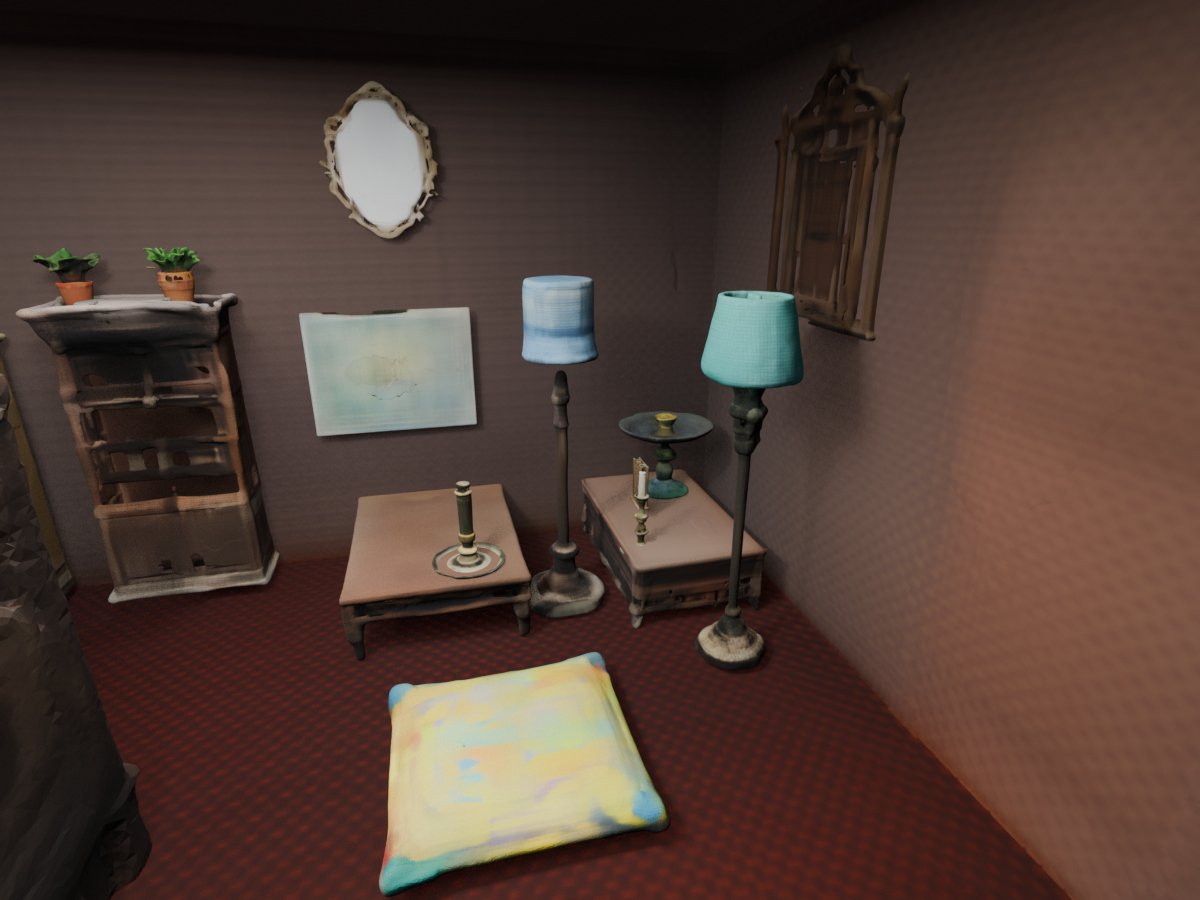} &
        \includegraphics[width=0.25\linewidth]{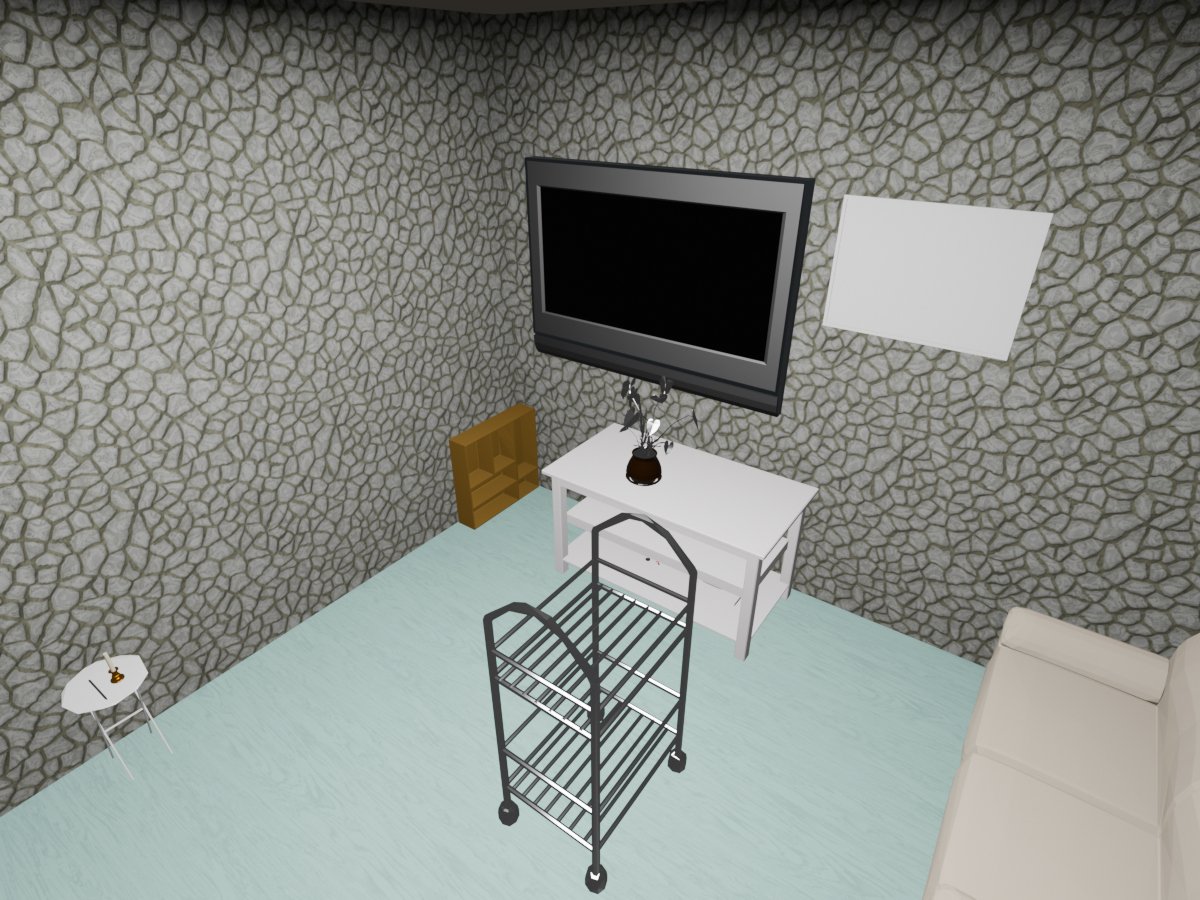} &
        \includegraphics[width=0.25\linewidth]{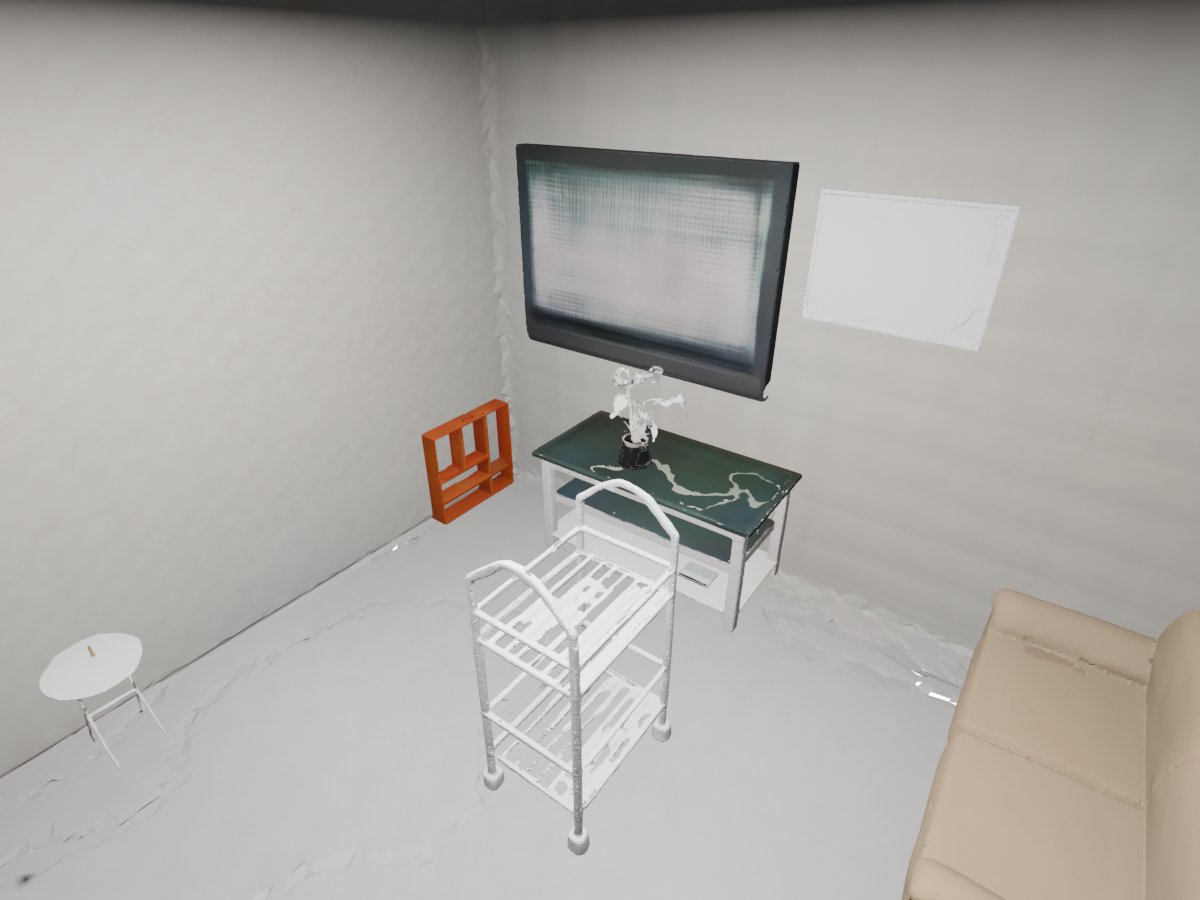} &
        \adjincludegraphics[width=0.25\linewidth, trim={0} {0.25\height} {0} {0}, clip]{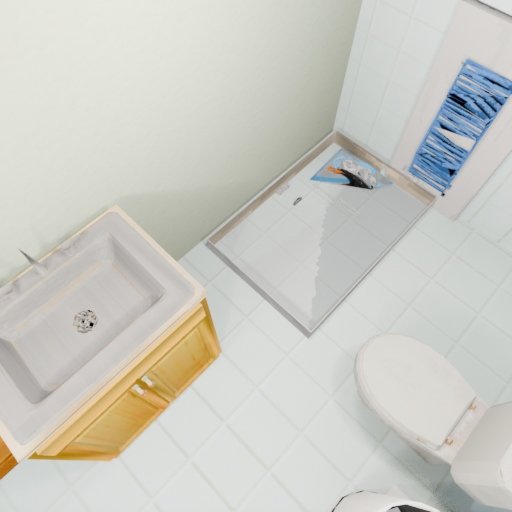} &
        \includegraphics[width=0.25\linewidth]{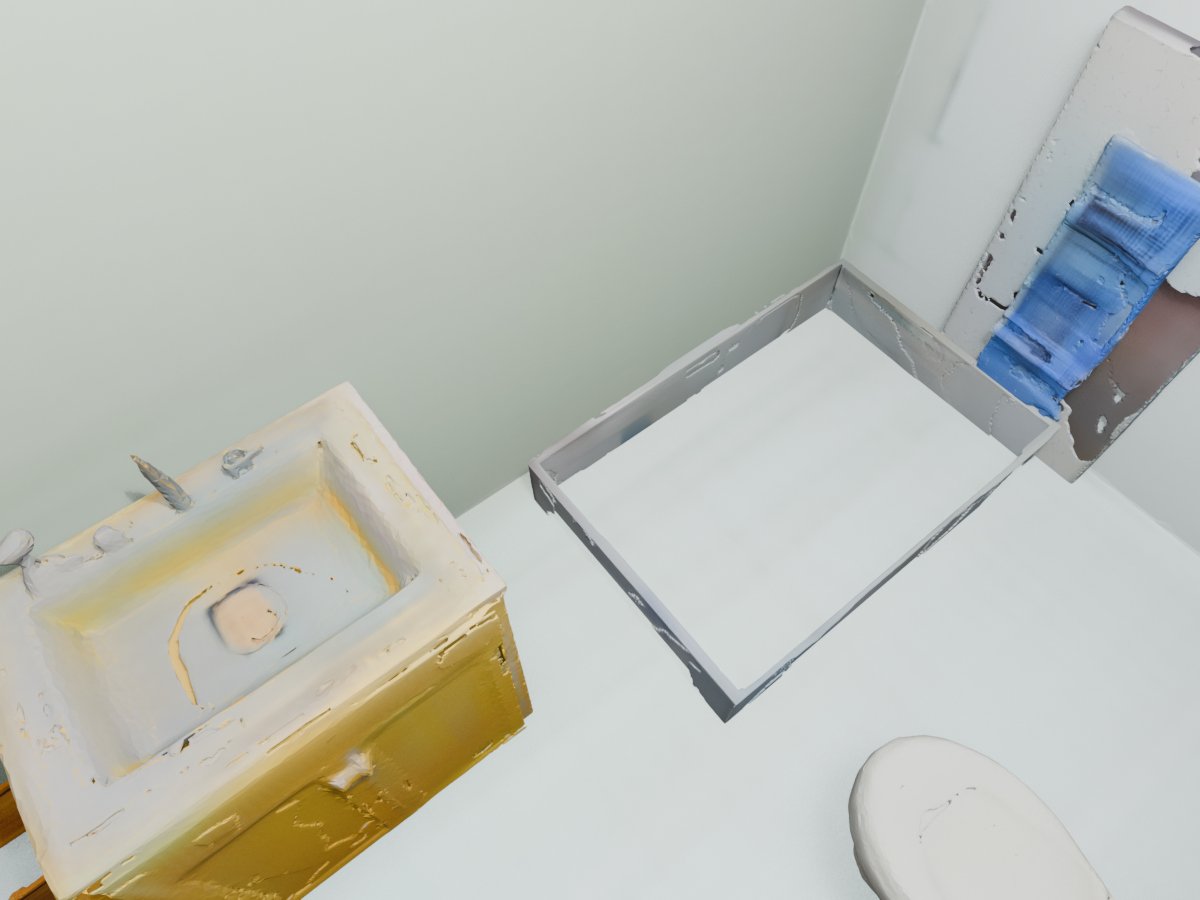} \\
       
        \multicolumn{2}{c}{GT / Ours (SAGE-10k \cite{sage}) } & 
        \multicolumn{2}{c}{GT / Ours (ProcTHOR \cite{ai2thor})} & 
        \multicolumn{2}{c}{GT / Ours (MansionWorld \cite{mansion}) } \\

        \adjincludegraphics[width=0.25\linewidth, trim={0} {0.25\height} {0} {0}, clip]{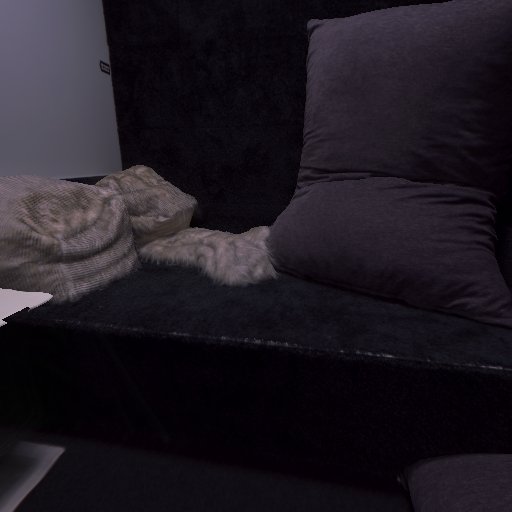} &
        \includegraphics[width=0.25\linewidth]{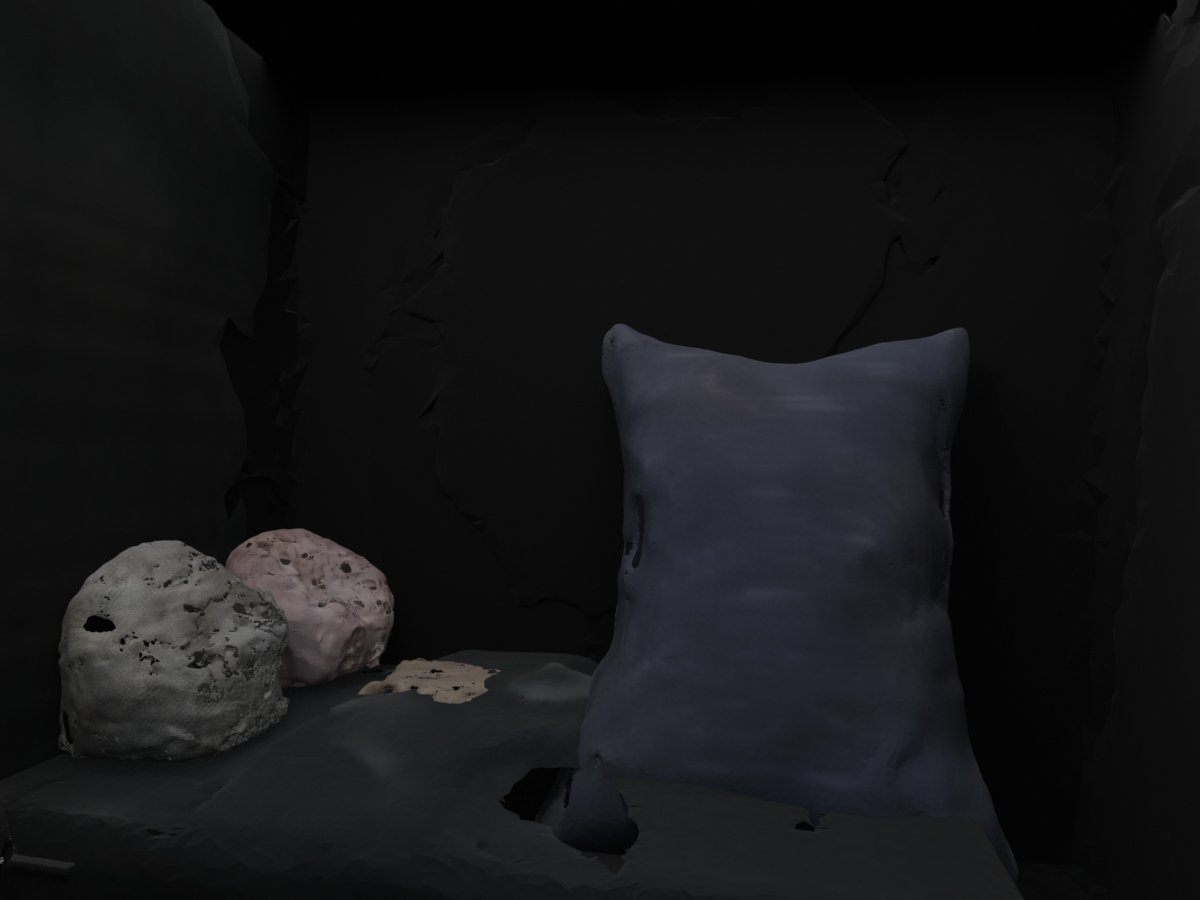} &
        \includegraphics[width=0.25\linewidth]{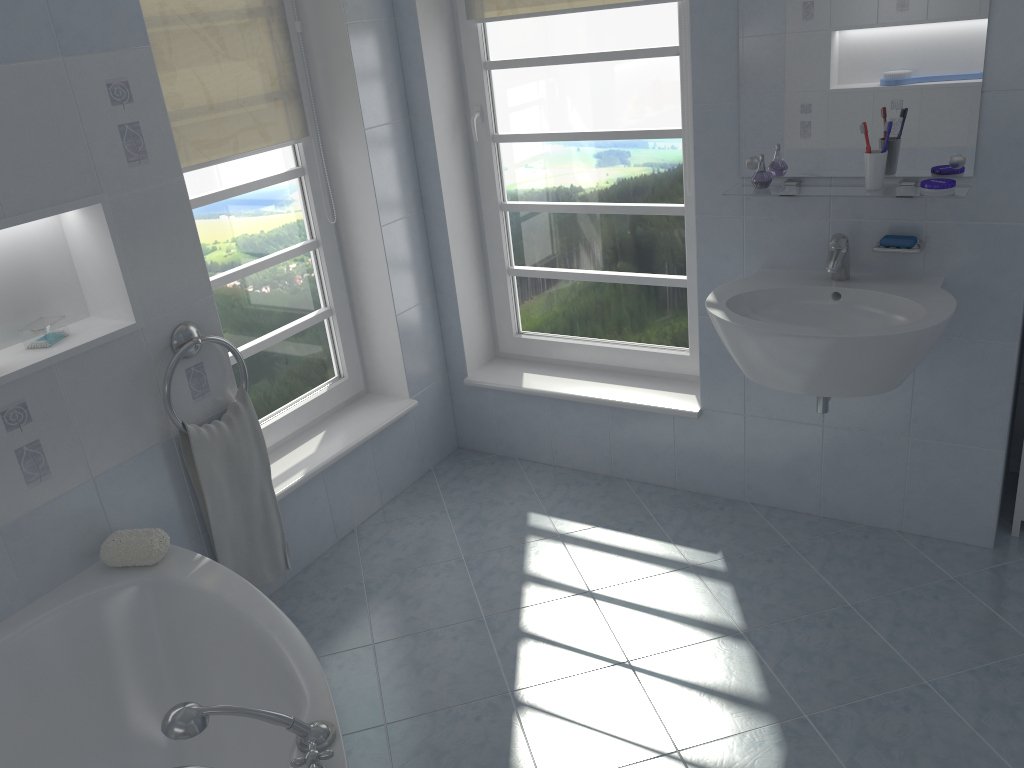} &
        \includegraphics[width=0.25\linewidth]{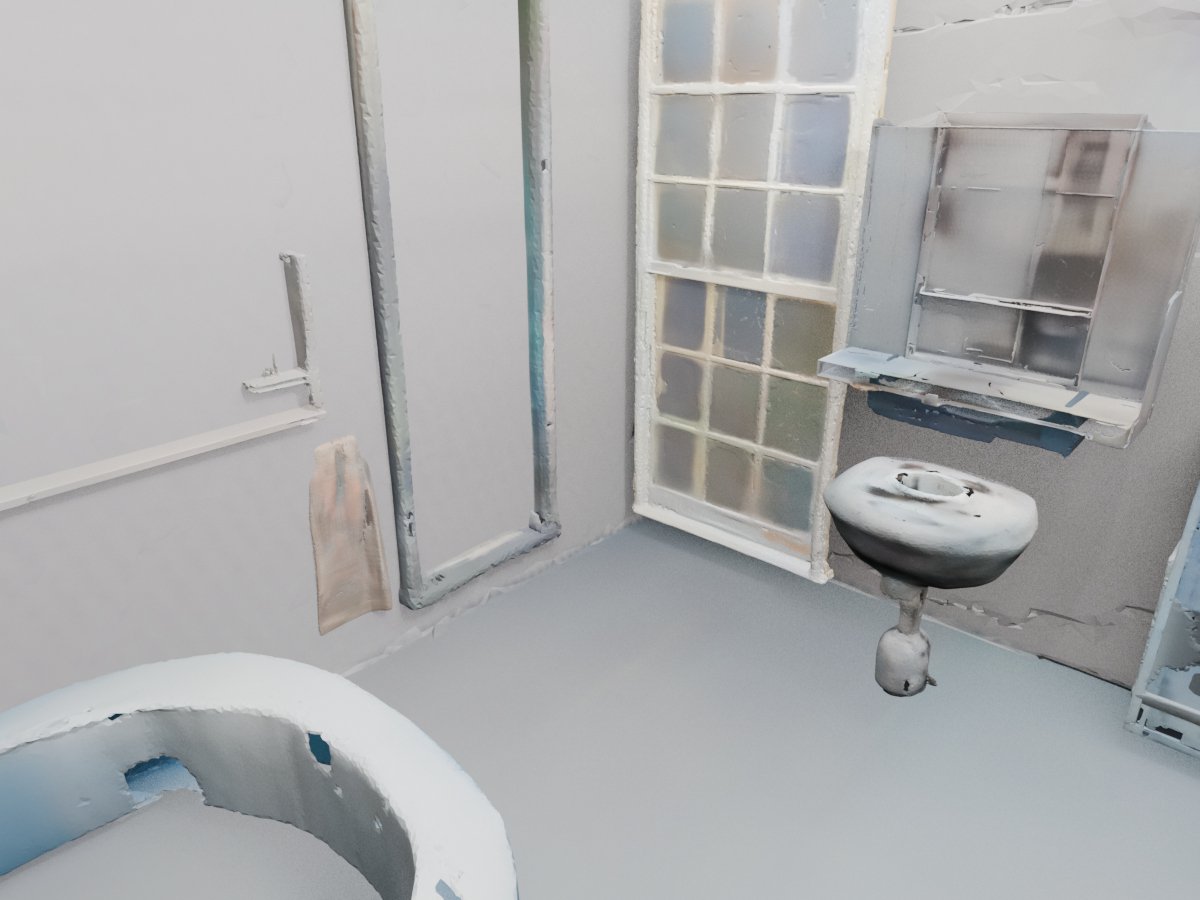} &
        \includegraphics[width=0.25\linewidth]{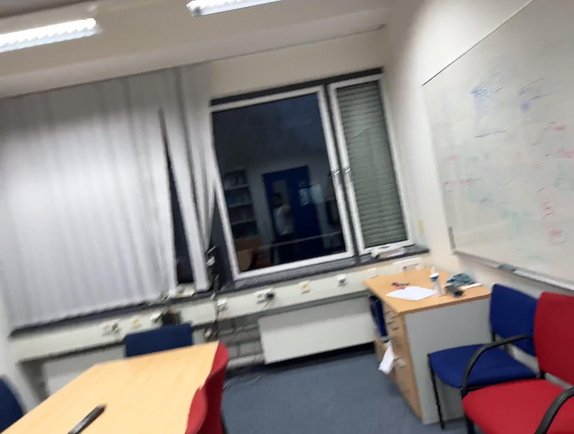} & 
        \includegraphics[width=0.25\linewidth]{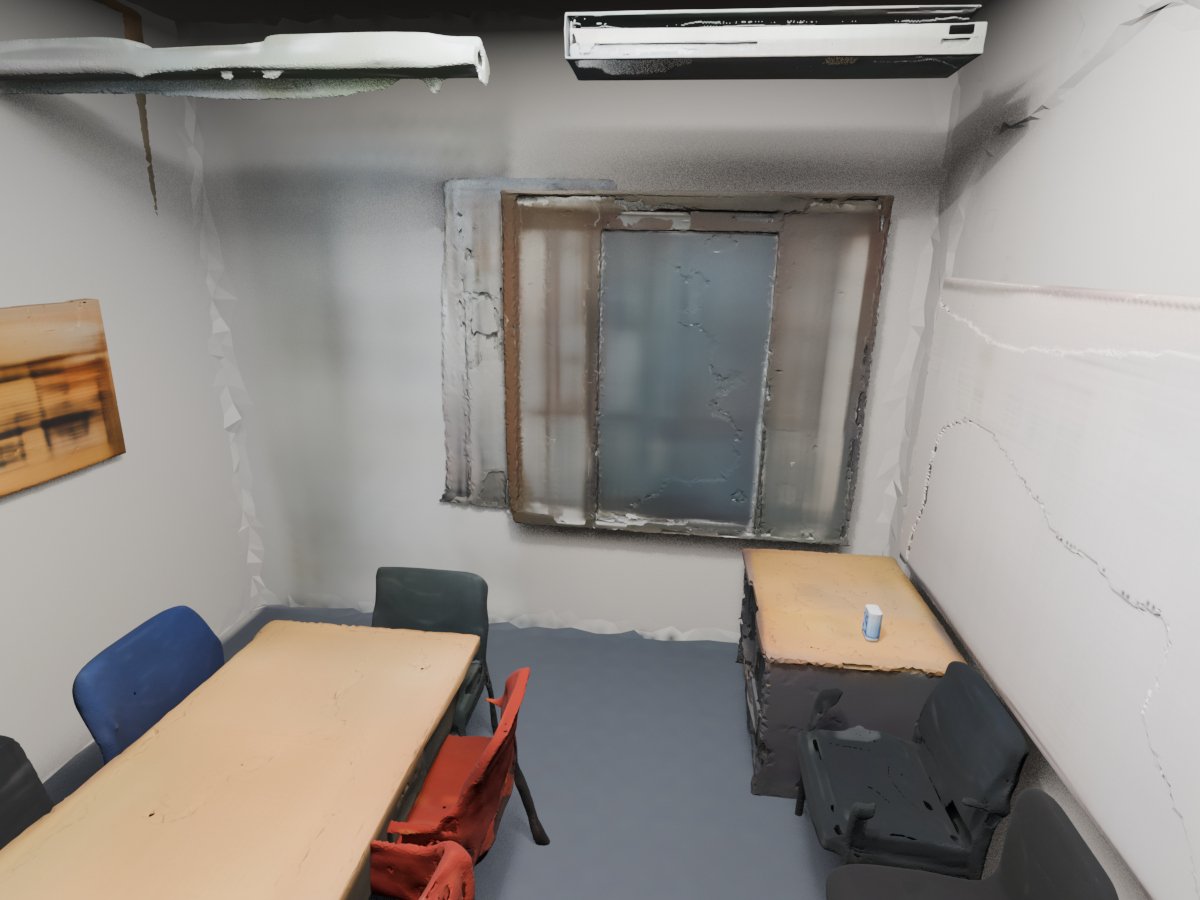} \\
        
        \multicolumn{2}{c}{GT / Ours (Replica \cite{straub2019replica}) } & 
        \multicolumn{2}{c}{GT / Ours (hypersim \cite{roberts2021hypersim}) } &
        \multicolumn{2}{c}{GT / Ours (Scannet++ \cite{scannet++})} \\
    \end{tabular}
    }

    \captionof{figure}{
    \textbf{Qualitative whole-scene reconstruction across diverse datasets.} \rev{All shown scenes and object assets are held out from training, including examples from training-source datasets. Background quality is shown qualitatively.}
    }
    \label{fig:more_examples}
\end{table*}

\begin{table*}[ht]
\centering
\begin{minipage}[t]{0.37\linewidth}
\centering
\resizebox{\linewidth}{!}{
\begin{tabular}{clc|cc}
\toprule
& Method & Runtime (\rev{s/scene}) & mAP $\uparrow$ & mIoU $\uparrow$ \\
\midrule
\multirow{4}{*}{\rotatebox{90}{AEO}} & SceneScript \cite{scenescript} &  \colorbox{second_color}{6.27} & 0.09 & 0.10 \\
& EFM3D \cite{efm3d} & 61.88 & 0.18 & \colorbox{second_color}{0.13} \\
& Boxer \cite{boxer} & 136.24 & \colorbox{second_color}{0.23} & \colorbox{best_color}{0.22} \\
& \textbf{Ours} & \colorbox{best_color}{2.66} & \colorbox{best_color}{0.25} & 0.11 \\
\midrule
\multirow{4}{*}{\rotatebox{90}{iTHOR}} & SceneScript \cite{scenescript} &  \colorbox{second_color}{6.27} & 0.16 & 0.10 \\
 & EFM3D \cite{efm3d} & 61.88 & 0.13 & 0.08 \\
 & Boxer \cite{boxer} & 136.24 & \colorbox{second_color}{0.36}  & \colorbox{second_color}{0.16}  \\
 & \textbf{Ours} & \colorbox{best_color}{2.66} & \colorbox{best_color}{0.52} & \colorbox{best_color}{0.41} \\
\midrule
\multirow{4}{*}[0.5ex]{\rotatebox{90}{Imaginarium}} & SceneScript \cite{scenescript} &  \colorbox{second_color}{6.27} & 0.20 & 0.14 \\
& EFM3D \cite{efm3d}  & 61.88 & 0.14 & 0.09 \\
& Boxer \cite{boxer} & 136.24 & \colorbox{second_color}{0.32} & \colorbox{second_color}{0.17}  \\
& \textbf{Ours} & \colorbox{best_color}{2.66} & \colorbox{best_color}{0.58} & \colorbox{best_color}{0.46} \\
\bottomrule
\end{tabular}%
}
\caption{\textbf{Quantitative results on 3D scene perception.} 
We compare the mAP and mIoU of the perception results for detection and segmentation quality. 
We also measure and report the average model inference time across datasets.
}
\label{tab:perception}
\end{minipage}%
\hfill
\begin{minipage}[t]{0.6\linewidth}
\centering
\resizebox{\linewidth}{!}{%
\begin{tabular}{cll|l|ccc|ccc}
\toprule
& \multirow{2}{*}{Method} & \multirow{2}{*}{\shortstack[l]{\rev{Network}\\Runtime\\(s / Obj.)}}
  & \multirow{2}{*}{\shortstack[l]{\\Perce-\\ption}}
  & \multicolumn{3}{c}{Geometry Quality}
  & \multicolumn{3}{c}{Rendering Quality} \\
\cmidrule(lr){5-7}\cmidrule(lr){8-10}
& & &  & CD \revten{(cm)} $\downarrow$ & F1 $\uparrow$ & NC $\uparrow$ 
 & PSNR \revten{(dB)} $\uparrow$ & SSIM $\uparrow$ & LPIPS $\downarrow$ \\
\midrule
\multirow{3}{*}{\rotatebox{90}{ShapeR}} & ShapeR  \cite{shaper}       & \colorbox{second_color}{4.84} & GT   & \colorbox{best_color}{1.37} & \colorbox{best_color}{0.58} & \colorbox{best_color}{0.81} & - & - & - \\
& SAM3D  \cite{sam3d}   &      10.61      & GT   & 4.07 & 0.26 & 0.71 & - & - & -\\
& \textbf{Ours}            &     \colorbox{best_color}{0.60}             & GT   & \colorbox{second_color}{{1.64}} & \colorbox{second_color}{{0.52}} & \colorbox{second_color}{{0.73}} & - & - & -\\
\midrule
\multirow{5}{*}[-0.5ex]{\rotatebox{90}{iTHOR}} & ShapeR  \cite{shaper}      & \colorbox{second_color}{4.84} & GT   &   \colorbox{second_color}{2.16}   & \colorbox{second_color}{0.68}   & \colorbox{second_color}{0.79}  &- & - & - \\
& SAM3D \cite{sam3d} &     10.61     & GT    &   4.63   &   0.48   &   0.73   &   \colorbox{second_color}{21.35}  &  \colorbox{second_color}{0.90}    &   \colorbox{second_color}{0.19}  \\
& \textbf{Ours}    &      \colorbox{best_color}{0.60}         & GT  & \colorbox{best_color}{{1.38}} & \colorbox{best_color}{{0.71}}  & \colorbox{best_color}{{0.81}} & \colorbox{best_color}{{23.85}} & \colorbox{best_color}{{0.92}} &  \colorbox{best_color}{0.13} \\
\addlinespace[2pt]
\cdashline{4-10}
\addlinespace[2pt]
& ShapeR \cite{shaper} &   \colorbox{second_color}{4.84} & Infer  & \colorbox{second_color}{8.90} & \colorbox{second_color}{0.21} & \colorbox{second_color}{0.68}  &- & - & -\\
& \textbf{Ours} & \colorbox{best_color}{0.60} & Infer  & \colorbox{best_color}{{6.15}} & \colorbox{best_color}{0.29} & \colorbox{best_color}{{0.72}} & {19.04} & {0.86} & {0.25} \\
\midrule
\multirow{5}{*}[-1ex]{\rotatebox{90}{Imaginarium}} & ShapeR  \cite{shaper}      & \colorbox{second_color}{4.84} & GT   &  \colorbox{second_color}{1.54}   &  \colorbox{best_color}{0.72}  &  \colorbox{best_color}{0.83}    & - & - & - \\
& SAM3D \cite{sam3d}   &     10.61     & GT &   3.83    &   {0.45}   &  {0.75}    &  \colorbox{second_color}{18.47}    &  \colorbox{second_color}{0.87}    &   \colorbox{second_color}{0.19}   \\
& \textbf{Ours}        &       \colorbox{best_color}{0.60} & GT  & \colorbox{best_color}{{1.08}} & \colorbox{second_color}{{0.68}} & \colorbox{second_color}{0.82} & \colorbox{best_color}{{20.23}} & \colorbox{best_color}{{0.89}} & \colorbox{best_color}{{0.14}} \\
\addlinespace[2pt]
\cdashline{4-10}
\addlinespace[2pt]
& ShapeR \cite{shaper} &     \colorbox{second_color}{4.84}     & Infer  & \colorbox{second_color}{9.77}& \colorbox{second_color}{0.23} & \colorbox{second_color}{0.67} & - & - & - \\
& \textbf{Ours} & \colorbox{best_color}{0.60} & Infer & \colorbox{best_color}{{6.49}} & \colorbox{best_color}{0.27} & \colorbox{best_color}{{0.70}} & {15.46} & {0.82} & {0.28} \\
\bottomrule
\end{tabular}%
}
\caption{\textbf{Quantitative results on video-based 3D scene reconstruction.}
\rev{GT/Infer denote ground-truth/inferred perception; Infer geometry uses matched objects.}
``-'' indicates metrics not applicable. 
\rev{SAM3D uses frame selection. Runtime is averaged across datasets.} 
We highlight \colorbox{best_color}{best} and \colorbox{second_color}{second best}.
}
\label{tab:main_cmp}
\end{minipage}%
\end{table*}

\vspace{-5px}
\section{Experiments}
\label{sec:exp}
\vspace{-5px}
\subsection{Implementation Details}

\vspace{-5px}
\paragraph{Training Data}
We build training data from diverse indoor scene datasets\rebut{, covering furniture and household objects,} and render RGB-D observations from SAGE-10k~\cite{sage}, InternScenes~\cite{internscenes}, ProcTHOR~\cite{ai2thor}, MansionWorld~\cite{mansion}, and SceneSmith~\cite{scenesmith}, totaling \rev{80k scenes and 138,202 rendered videos} with randomized camera intrinsics. We further augment the diversity and realism with Flux.2~\cite{flux2}, which produces \rev{79,672 additional appearance-augmented videos, for 217,874 sequences sharing the underlying scene corpus}. Additionally, we use an extra 500k objects from object datasets \cite{3dfuture, abo, hssd, objaverse} in the flow matching reconstruction model training \rev{for object reconstruction.} 

\paragraph{Training Details}
We train the perception and generative models separately\rev{ from scratch}, using data augmentations including random frame dropping, scene rotation, and camera-pose/depth noise. \rev{We train the flow models for 200k AdamW steps at $10^{-4}$. Perception training settings are given in the appendix.} We adopt AnyUp \cite{anyup} for higher resolution DINO features. We use 12 sampling steps with a classifier-free guidance scale of 3 for all experiments. 

\vspace{-5px}
\subsection{Experimental Settings}
\paragraph{Tasks Settings}\label{sec:runtime_protocol}
We evaluate our model in three kinds of settings. 
(1) \textbf{Video-based Perception: } Given posed RGB-D observations from evaluation datasets, the model predicts the 3D object-oriented bounding boxes (OBBs) and instance segmentation. 
(2) \textbf{Video-based Reconstruction: }  Given posed RGB-D observations and either ground-truth or inferred instance perception, the model reconstructs each object's complete geometry and texture.
(3) \textbf{Single-image Reconstruction: } Given a single RGB image and its \rev{Pi3X}-estimated point map, \rev{the model automatically estimates instances and reconstructs their geometry and texture.}

\paragraph{Evaluation Datasets} 
(1) \textbf{AEO Dataset} \cite{efm3d}:  a real-world dataset with OBB annotations for \textit{all} 3D objects in the scene, is only used in the video-based perception task.
(2) \textbf{ShapeR Dataset} \cite{shaper}:  a real-world dataset with OBBs, segmented point cloud, and ground-truth object geometry annotations for \textit{selected} 3D objects in the scene, is only used in the video-based reconstruction task.
(3) \textbf{iTHOR Dataset} \cite{ai2thor} and \textbf{Imaginarium \cite{imaginarium} Dataset: } synthetic datasets with OBBs, segmented point cloud, and ground-truth object geometry and textures for \textit{all} 3D objects in the scene, are used in both video-based perception and reconstruction tasks.
(4) \textbf{3D-Front Dataset}~\cite{fu20213d}: a synthetic dataset with rendered images and ground-truth object meshes from the eval split of Gen3DSR \cite{gen3dsr} evaluation benchmark, is used in the single-image reconstruction task. \rev{Quantitative evaluation datasets and their scene/object assets are excluded from training, including across shared asset libraries. Real-world tests cover perception and object geometry; rendering comparisons use synthetic data.}

\paragraph{Metrics}
We evaluate the detection and segmentation quality with mAP and mIoU, geometry quality with Chamfer Distance (CD, \revten{cm}), F-Score (F1\revten{, at 1 cm}), and Normal Consistency (NC), and assess rendering quality using PSNR\revten{ (dB)}, SSIM, and LPIPS. \rev{Metric details are provided in Appendix~\ref{sec:additional_results}.}
\rev{\revten{We report} network inference from posed RGB-D inputs in} \rev{Tab.~\ref{tab:main_cmp}\revten{. We provide a full breakdown} of the one-minute result containing Pi3X preprocessing, feature extraction, and asset export in Tab.~\ref{tab:runtime_breakdown}.}

\paragraph{Baselines}
We evaluate our framework against representative task-specific and simulation-oriented approaches in different evaluation settings.
\textbf{SceneScript} \cite{scenescript} uses an autoregressive model to predict every indoor element pose from the point cloud.
\textbf{EFM3D} \cite{efm3d} predicts 3D OBBs from videos and semi-dense points.
\textbf{Boxer} \cite{boxer} leverages per-image object BBoxes with point clouds to predict every object's OBBs.
\textbf{ShapeR} \cite{shaper} reconstructs only geometry given input object points, images, and text prompts and uses EFM3D \cite{efm3d} to detect objects in the scene.
\textbf{SAM3D} \cite{sam3d} uses user clicks as prompts for instance masks and reconstructs the textured mesh with pose. \rev{For video input, we select each object's largest-mask frame for reconstruction; this adaptation does not fuse views.}
\textbf{Gen3DSR} \cite{gen3dsr}, \textbf{MIDI} \cite{midi}, and \textbf{SceneGen} \emph{automatically} segment and reconstruct every scene object from a single image without the need of user clicks like SAM3D.

\subsection{Experimental Results}
\paragraph{Video-based 3D Scene Perception}
\rev{\model leads mAP on all three datasets and mIoU on iTHOR and Imaginarium, with the fastest reported inference (Tab.~\ref{tab:perception}).} 
\rev{The leading mAP on real-captured AEO further demonstrates transfer to real observations, with room to improve instance overlap.}
Fig.~\ref{fig:perception} shows that our model predicts more structured instance layouts than baselines across datasets.

\paragraph{Video-based 3D Scene Reconstruction}
As shown in Tab.~\ref{tab:main_cmp}, \model \rev{leads all geometry metrics on iTHOR with GT perception and CD on Imaginarium, while ShapeR performs best on its real-world benchmark.}
SAM3D~\cite{sam3d} produces strong single-view textured objects\rev{.}
In contrast, \model{} uses posed RGB-D observations to predict instance point clouds and reconstruct textured foreground objects together with a static background instance without prompts. \rev{FIRE3D improves all rendering metrics over frame-selected SAM3D on both synthetic benchmarks.}
Fig.~\ref{fig:recon_cmp_gt_perception} compares reconstruction under GT perception inputs\rev{.}
Fig.~\ref{fig:recon_cmp_inferred_perception} further shows that, under inferred perception inputs, \model \rev{reconstructs textured scenes and improves all reported geometry metrics over the ShapeR pipeline. The remaining gap to GT perception suggests that more accurate instance estimates could further improve reconstruction.}
Additional results in Fig.~\ref{fig:more_examples} on \rev{diverse scene sources illustrate reconstruction across varied indoor environments.}

\paragraph{Single-image Reconstruction}
\begin{wraptable}{r}{0.45\linewidth}
\vspace{-1em}
\small
\centering
\resizebox{\linewidth}{!}{
\begin{tabular}{lccc}
    \toprule
    \textbf{Method} 
    & \textbf{CD} \revten{(cm)} $\downarrow$ 
    & \textbf{F1} $\uparrow$ 
    & \textbf{NC} $\uparrow$ \\
    \midrule
    Gen3DSR~\cite{gen3dsr} & 20.56 & 0.08 &  0.64 \\
    MIDI~\cite{midi}       & 20.21 & 0.05 & 0.55 \\
    SceneGen~\cite{meng2025scenegen}       & 14.90 & 0.06 & 0.58 \\
    Ours                   & \textbf{11.24} & \textbf{0.10} & \textbf{0.66} \\
    \bottomrule
\end{tabular}
}
\caption{\textbf{\rev{Single-image results}} \rev{on} 3D-Front \cite{fu20213d} \rev{(scene-aligned)}.}
\label{tab:single_image}
\vspace{-1em}
\end{wraptable}
We evaluate \rev{automatic methods from single RGB inputs}, using a \rev{Pi3X}-estimated point map as FIRE3D input. 
\rev{As shown in Fig.~\ref{fig:single_image_recon_cmp} and Tab.~\ref{tab:single_image},}
\rev{\model{} achieves the best CD, F1, and NC among the compared methods on the held-out split, demonstrating effective single-image reconstruction.}

\begin{table*}[ht]
    \centering

    \setlength{\tabcolsep}{1pt}
    \renewcommand{\arraystretch}{0.5}

    \resizebox{\textwidth}{!}{%
    \begin{tabular}{cccccc}

        \includegraphics[width=0.2\linewidth]{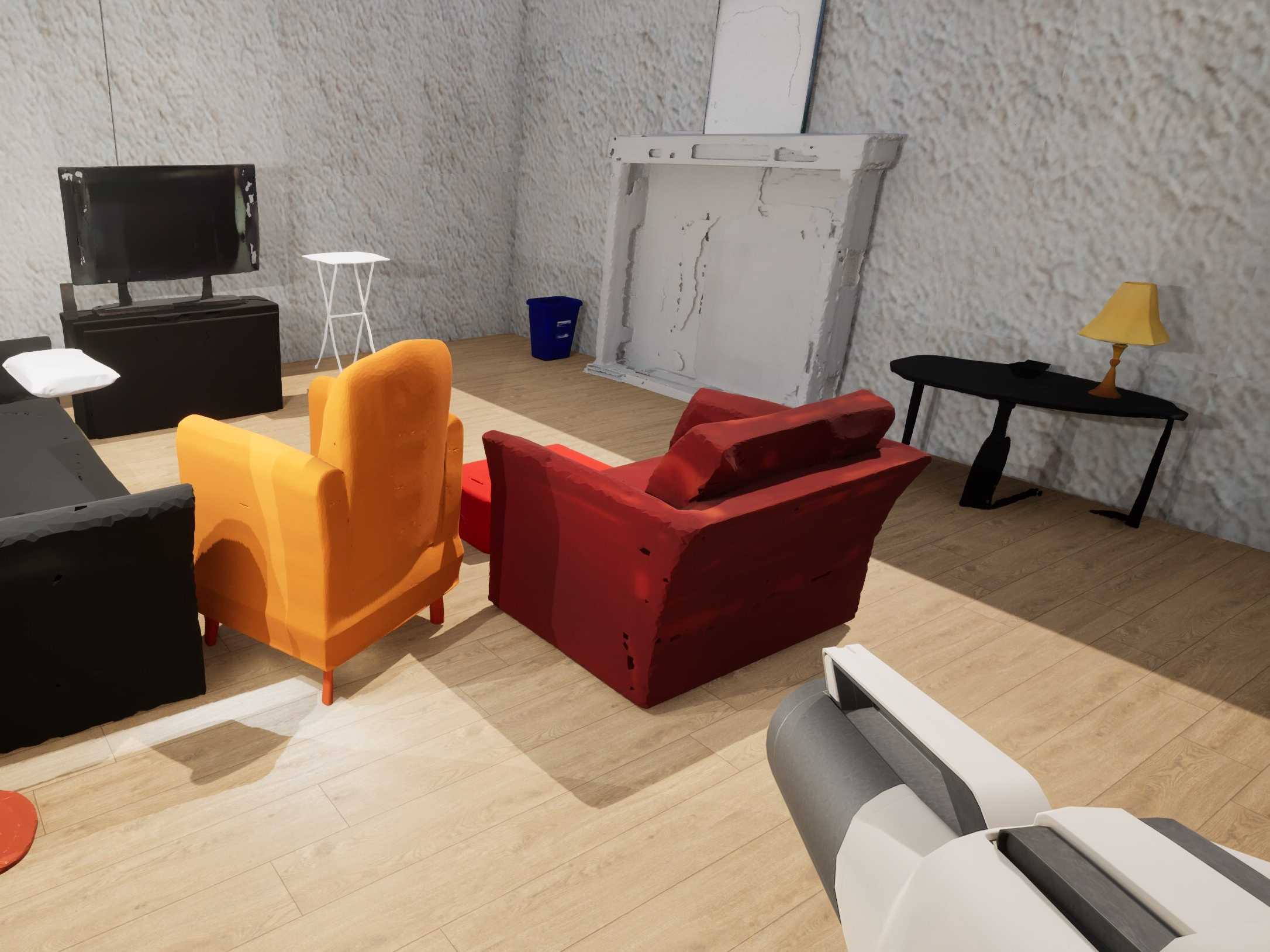} &
        \includegraphics[width=0.2\linewidth]{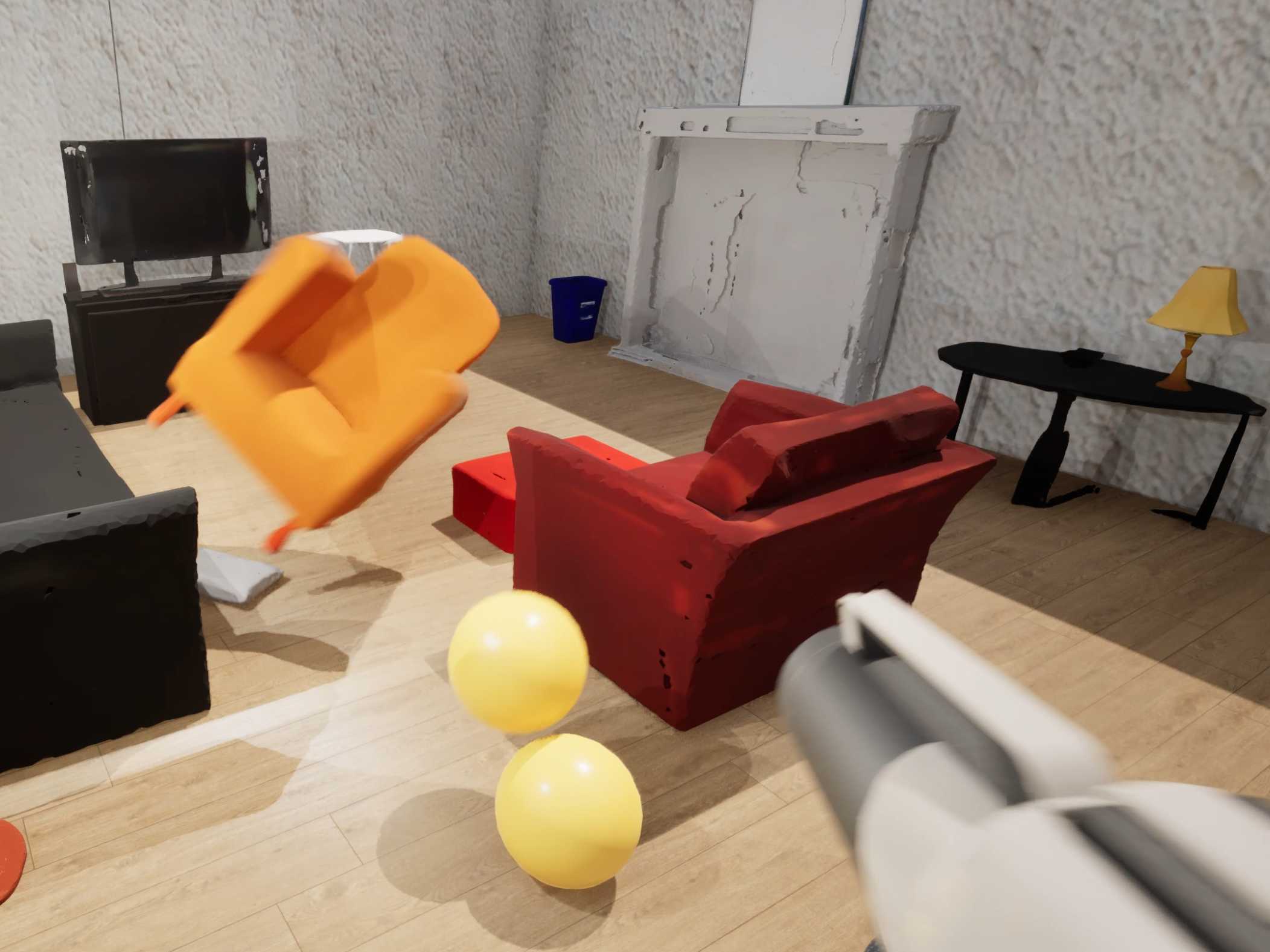} &
        \includegraphics[width=0.2\linewidth]{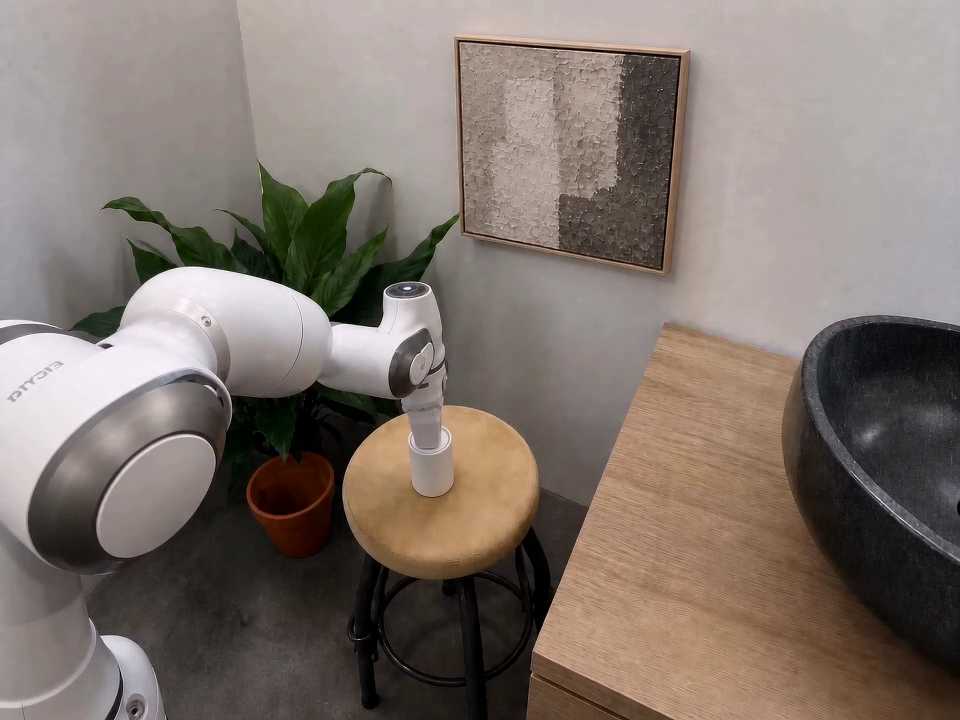} &
        \includegraphics[width=0.2\linewidth]{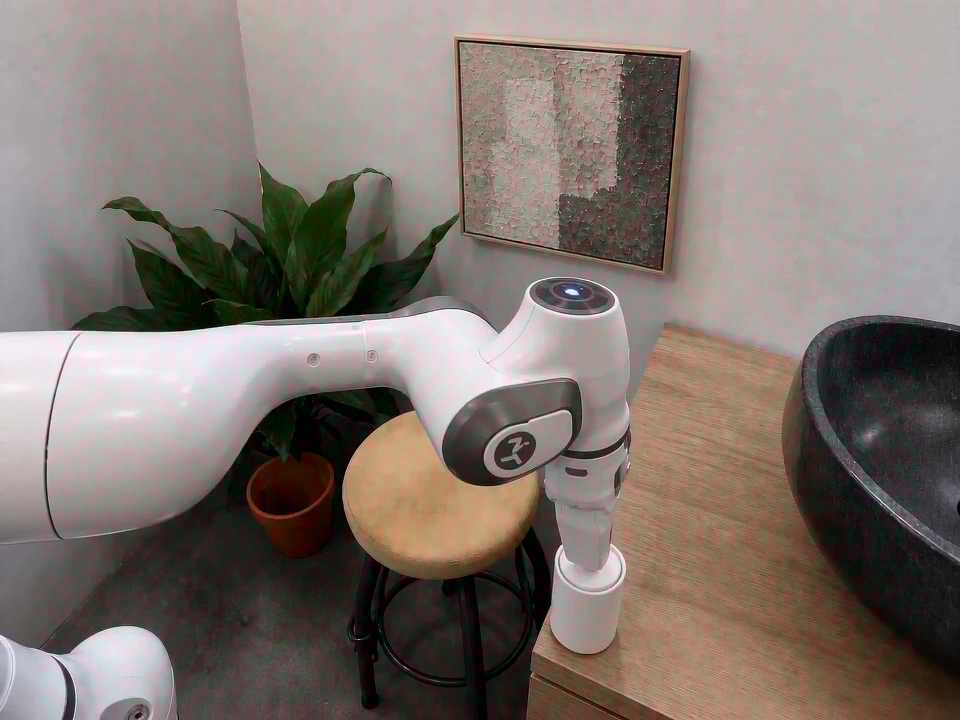} &
        \includegraphics[width=0.2\linewidth]{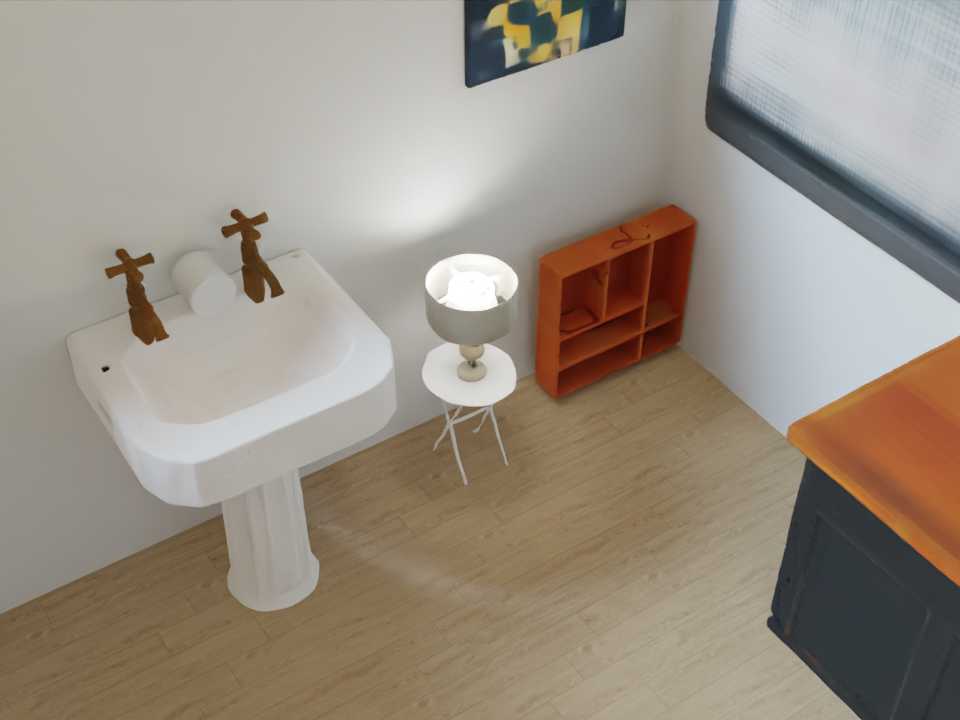} &
        \includegraphics[width=0.2\linewidth]{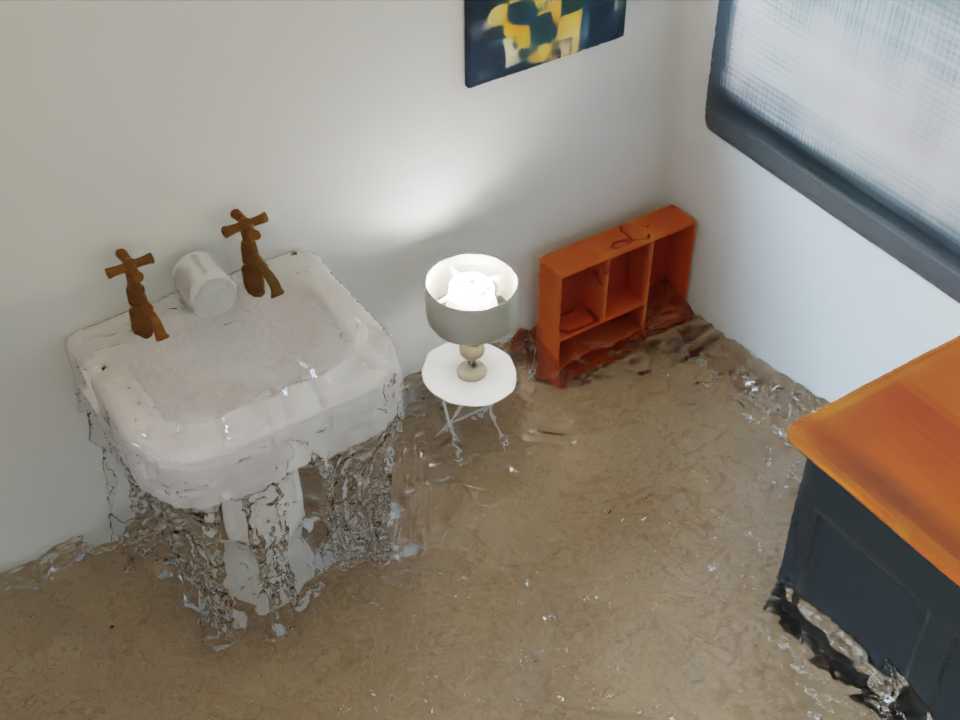} \\
       
        \multicolumn{2}{c}{Interactive Gaming} & 
        \multicolumn{2}{c}{Robotics Simulation} & 
        \multicolumn{2}{c}{Dynamic VFXs} \\

    \end{tabular}
    }

    \captionof{figure}{
    \textbf{Applications of interactive environments created by \model .} 
    Here we showcase diverse downstream applications for \model, including interactive gaming, robotics, and visual effects.
    }
    \label{fig:demo}
\end{table*}

\WFclear
\subsection{Interactive Environment Applications}

\model{} has a wide range of applications across gaming, robotics, and content creation. An overview of the application demonstration can be found in Fig. \ref{fig:demo}.
\paragraph{Gaming \& Dynamic VFXs} 
We create a virtual shooting game with Unreal \cite{ue} using our reconstructed assets. \rev{Independent assets enable rapid scene editing and game integration, extending object-level reconstruction workflows~\cite{xia2024video2game,xia2025drawer}.} It can also be extended to generate imaginary dynamic visual effects with Blender, such as water simulation.
\paragraph{Robotics}
\model{} can also be applied in simulation data generation for Embodied AI. With the help of IsaacSim \cite{isaacsim} and IsaacLab \cite{isaaclab}, we can generate a pick-and-place robot demonstration with Franka Arm, \rev{which grasps an object from the chair and places it on the table.} \rev{This demonstrates compatibility with robotics simulation workflows. We leave systematic evaluation of physical stability and sim-to-real transfer to future work.} 

\section{Conclusion \& Limitation}

We presented \model, a feed-forward framework for object-level textured 3D scene reconstruction from unsegmented, posed RGB-D observations estimated from single-image and casual-video RGB captures. \model leverages compressed object representation with HC-VAE and unifies instance-aware perception and object-centric generation. It enables efficient batched reconstruction of interactable scene elements with \rev{amodal geometry and texture.}
\textbf{Limitations:}
\model{} focuses on static indoor scenes, and requires posed RGB-D observations at the network interface with external geometric preprocessing for RGB-only captures. 
\rev{Its quality depends on depth, camera poses, scale, and instance parsing. Here, interactable means independently editable assets. 
\revten{We leave guarantees} of watertightness, physical stability, relightability, and articulation \revten{to} future work.} \rebut{We discuss further limitations in Appendix~\ref{sec:limitations}.}

\section{AI Disclosure}
\rebut{In this work, we used generative AI tools for the experiments of comparison against Claude Opus 5.5 in scene perception and reconstruction. 
We have not used generative AI tools to 
generate synthetic data sets, help develop theoretical models or conceptual frameworks, formulate mathematical claims,
provide critical ingredients for proving mathematical claims, assist in the writing of proofs, 
propose or refine hypotheses, design or provide feedback on research  methodology or experiments, implement methods, 
assist with translation, clean and reformat dataset, support qualitative and thematic data analysis, and interpret results. 
We have reviewed all AI-assisted work, checking its generation results and conduct evaluations. 
We take responsibility for the final content of this work, including text, claims or artifacts produced with the aid of generative AI.
}

\paragraph{Acknowledgment}
This project is supported by the NSF Awards \#2331878, \#2340254, \#2312102, \#2414227, \#2404385, and \#2525287. We greatly appreciate the NCSA for providing computing resources through Delta and Delta AI program. We thank the support from the NVIDIA Academic Grant Program for Researchers.

\bibliography{iclr2027_conference}
\bibliographystyle{iclr2027_conference}

\appendix
\clearpage

\setcounter{section}{0}
\setcounter{footnote}{0}
\setcounter{figure}{0}
\setcounter{table}{0}
\renewcommand{\thefigure}{A\arabic{figure}}
\renewcommand{\thetable}{A\arabic{table}}
\renewcommand{\theHfigure}{appendix.\arabic{figure}}
\renewcommand{\theHtable}{appendix.\arabic{table}}

\appendix
\setlength{\intextsep}{8pt plus 2pt minus 2pt}
\setlength{\textfloatsep}{10pt plus 2pt minus 2pt}
\setlength{\floatsep}{8pt plus 2pt minus 2pt}
\setlength{\parskip}{3pt plus 1pt minus 1pt}
\setcounter{topnumber}{3}
\setcounter{totalnumber}{4}
\makeatletter
\newcount\wns@tries
\newcommand{\WrapNeedspace}[1]{\par\penalty-100\begingroup
  \setlength{\dimen@}{#1}\wns@tries\z@
  \wns@check
  \endgroup}
\def\wns@check{%
  \ifdim\pagegoal=\maxdimen \dimen@ii\vsize
  \else \dimen@ii\pagegoal \advance\dimen@ii-\pagetotal \fi
  \ifdim\dimen@>\dimen@ii \ifnum\wns@tries<\thr@@
    \advance\wns@tries\@ne
    \ifdim\pagegoal=\maxdimen \null\fi
    \vfil\penalty-\@M
    \expandafter\expandafter\expandafter\wns@check
  \fi\fi}
\makeatother
\section*{Appendix}

\section{More Experiment Results}\label{sec:additional_results}

\begin{figure}[ht]
\centering
\resizebox{0.95\linewidth}{!}{%
\setlength{\tabcolsep}{1pt}
\begin{tabular}{ccccccc}
& Input  & SceneScript & EFM3D & Boxer & Ours & Ground-truth \\

\raisebox{30px}{\rotatebox{90}{AEO}} &
\img{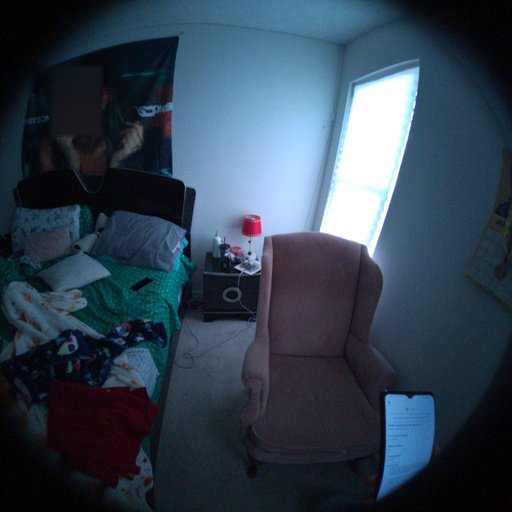} &
\img{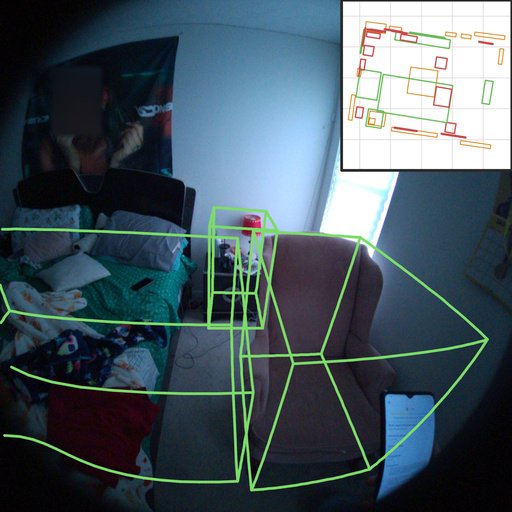} &
\img{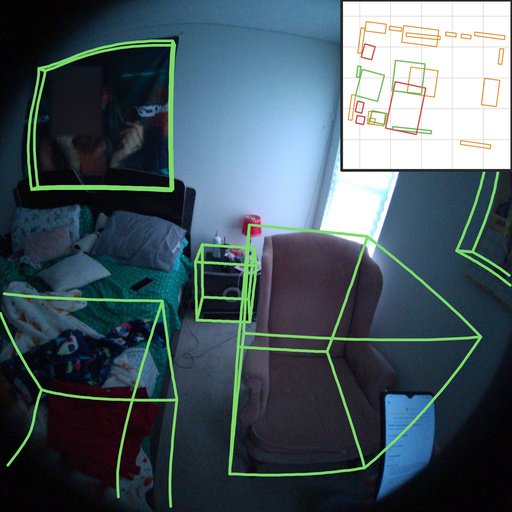} &
\img{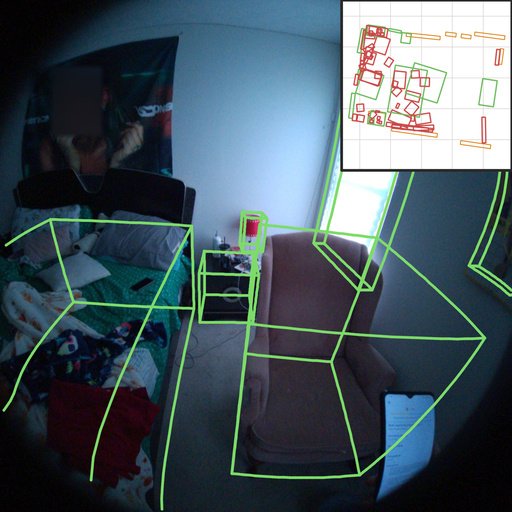} &
\img{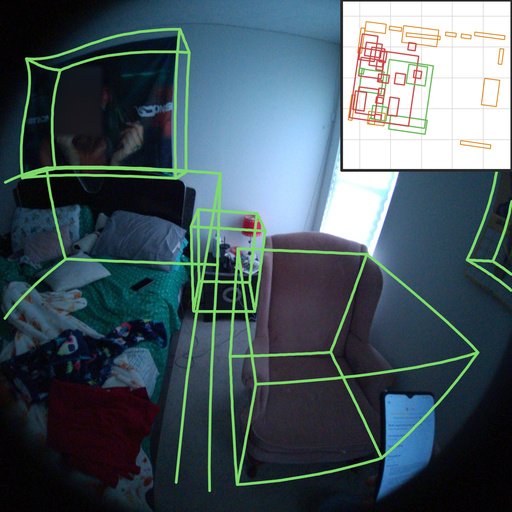} &
\img{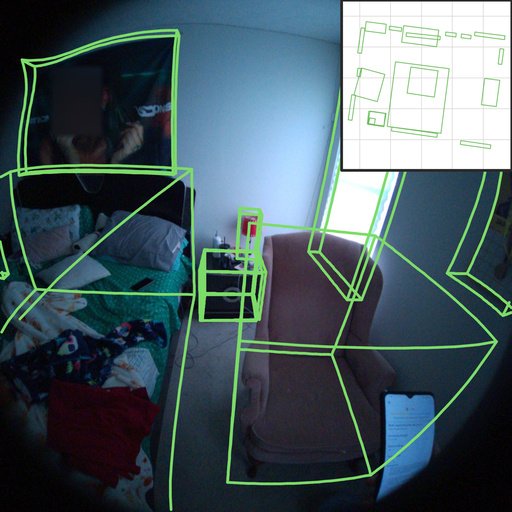} \\

\raisebox{20px}{\rotatebox{90}{iTHOR}} &
\img{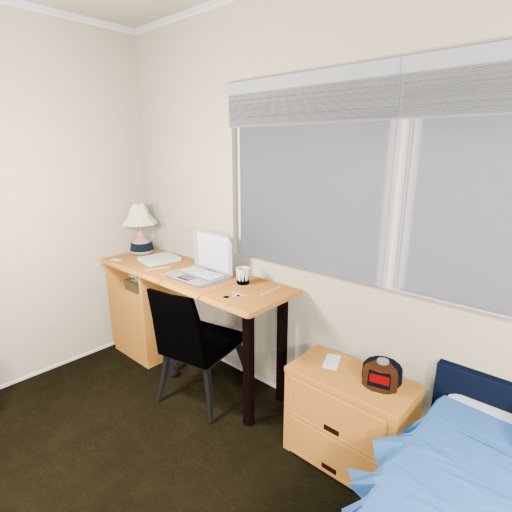} &
\img{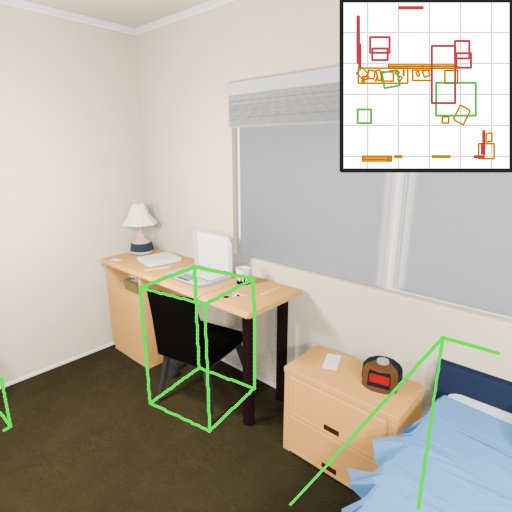} &
\img{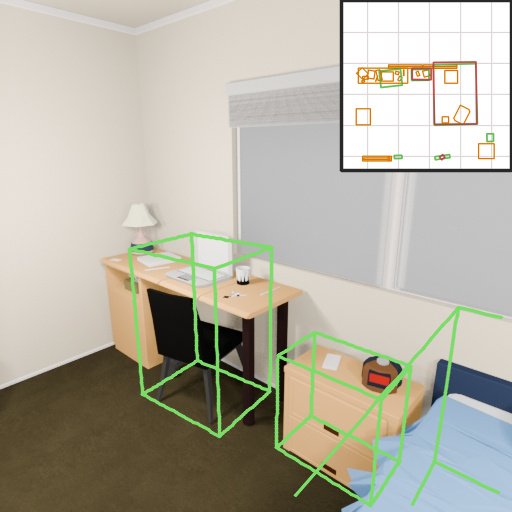} &
\img{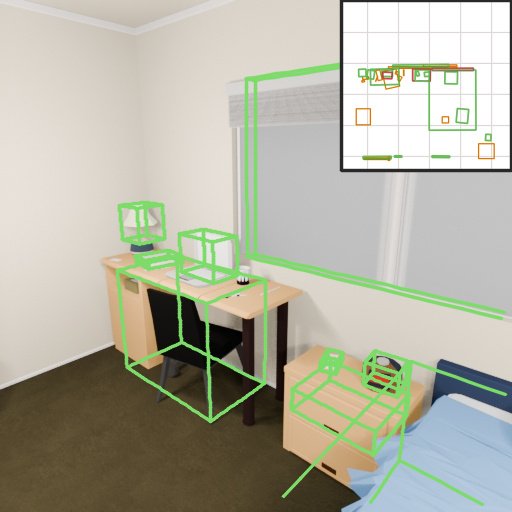} &
\img{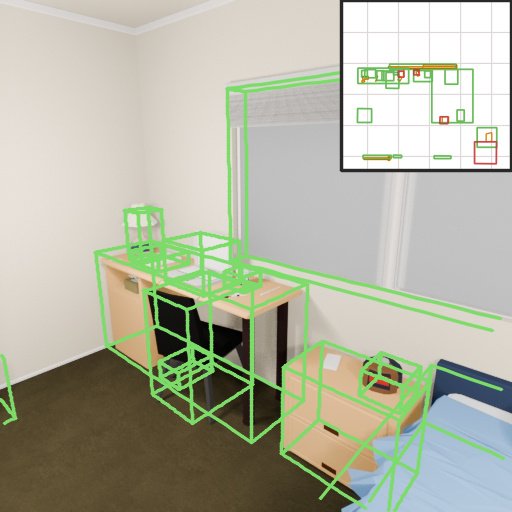} &
\img{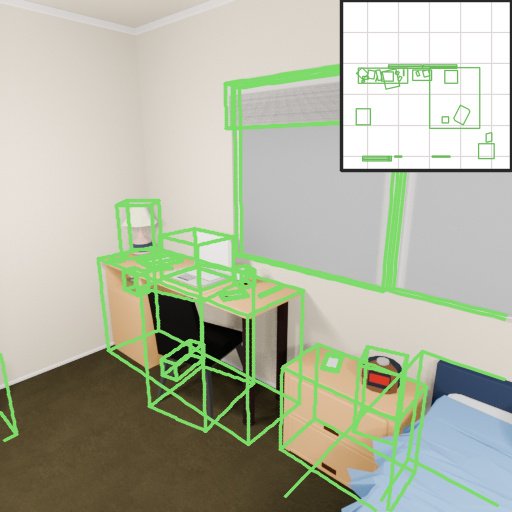} \\

\raisebox{15px}{\rotatebox{90}{Imaginarium}} &
\img{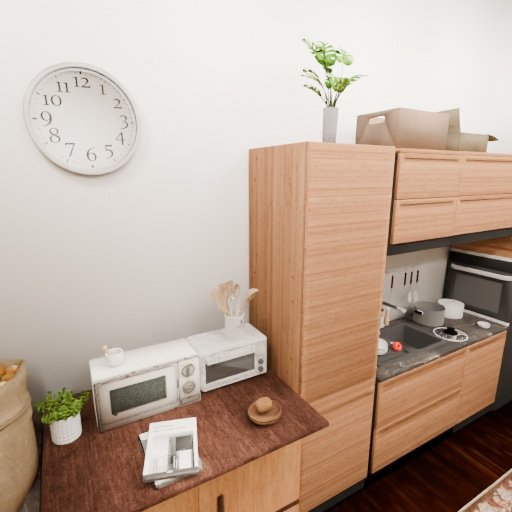} &
\img{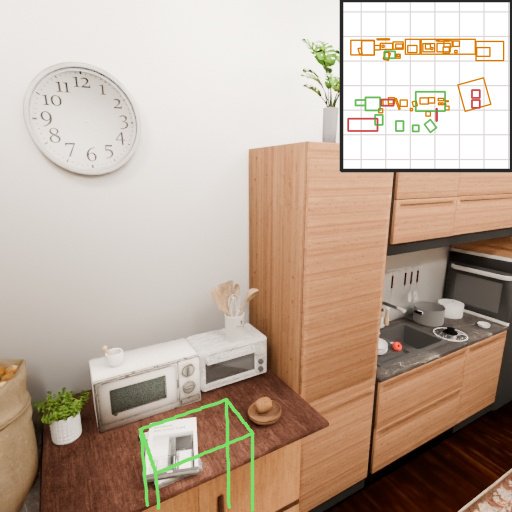} &
\img{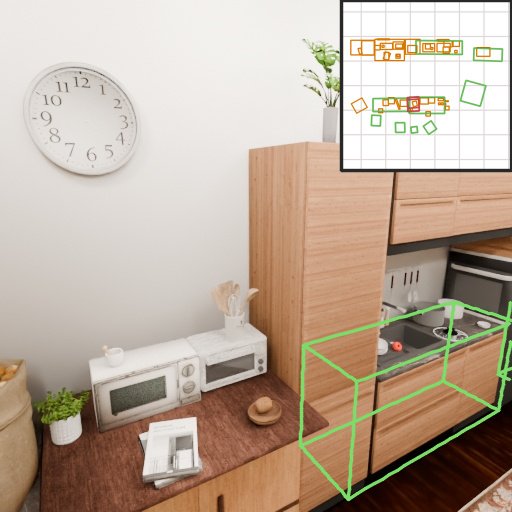} &
\img{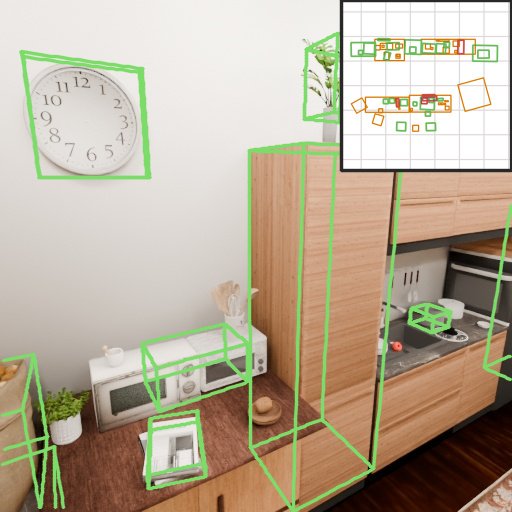} &
\img{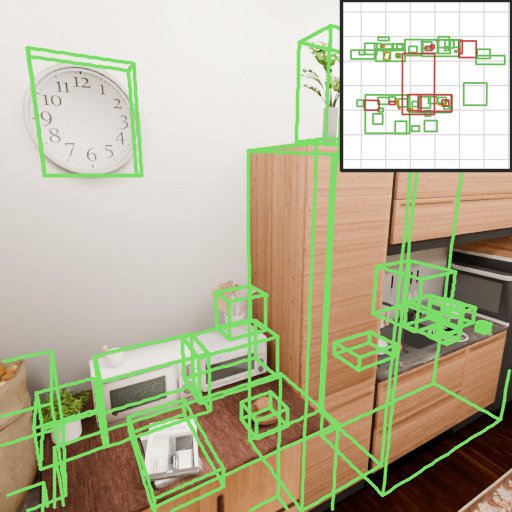} &
\img{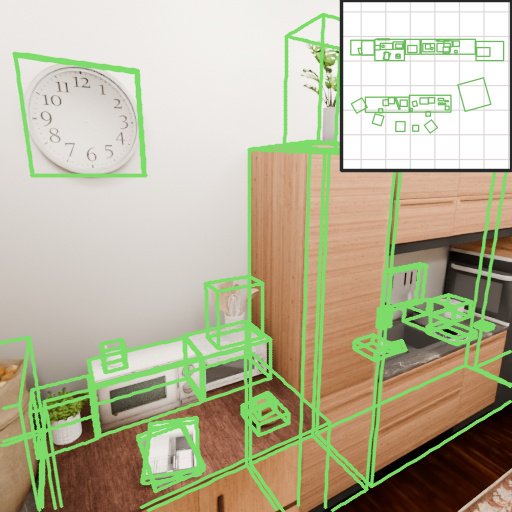} \\

\end{tabular}%
}
\caption{\textbf{More qualitative results of video-based 3D scene perception.}
Correct predictions are visualized in \textcolor{green}{green}. We show the overview perception in the inset figure. Wrong ones are marked in \textcolor{red}{red}, and the undetected are marked in \textcolor{orange}{orange}. \rev{FIRE3D produces coherent instance layouts in these examples.}
}
\label{fig:perception_supp}
\end{figure}

\paragraph{More \rev{Details and} Results on Video-based 3D Scene Perception}
Here in Fig. \ref{fig:perception_supp} we show more visualizations of the results of video-based 3D scene perception against SoTA methods, across multiple diverse datasets. We can observe that \model{} predicts more structured instance layouts than baselines across datasets.
\rev{The AEO evaluation uses 25 scenes. Detection AP uses one-to-one 3D box matching at IoU 0.25, ordered by prediction confidence. mAP is the unweighted mean of per-scene class-agnostic AP. Point-mask IoU is computed after this instance association. mIoU averages foreground GT instance IoUs within each scene, counting missed instances as zero, and then averages scenes equally; background is excluded. All methods use the same metric aggregation. For EFM3D and Boxer, points inside each predicted box form its instance mask. For the main quantitative comparisons, we use all valid samples from each dataset's designated test split. We will release the evaluation splits and sample lists.}

\begin{figure}[ht]
\centering
\resizebox{0.8\linewidth}{!}{%
\setlength{\tabcolsep}{1pt}
\begin{tabular}{ccccc}
& Input & Boxer & Ours & Ground-truth \\

\multirow{2}{*}[69.7pt]{\rotatebox{90}{\makebox[0.4\linewidth][c]{CA-1M}}} &
\img{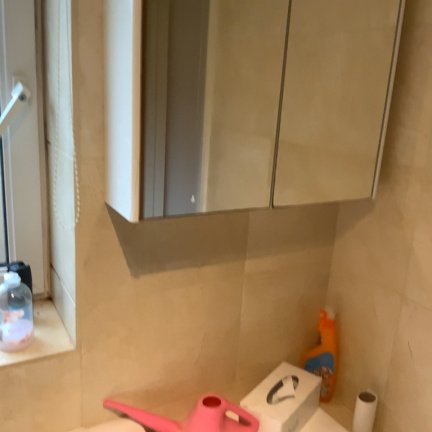} &
\img{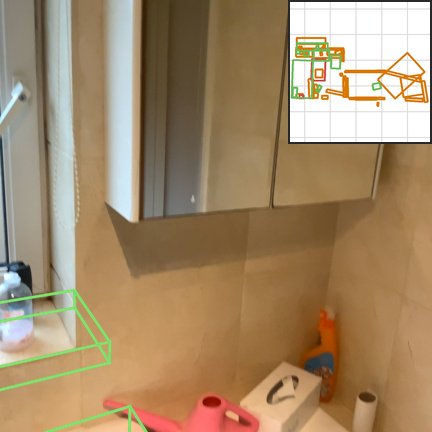} &
\img{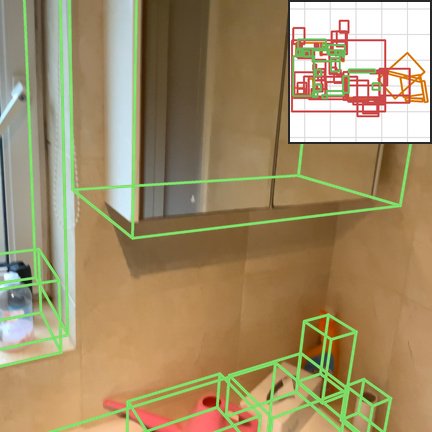} &
\img{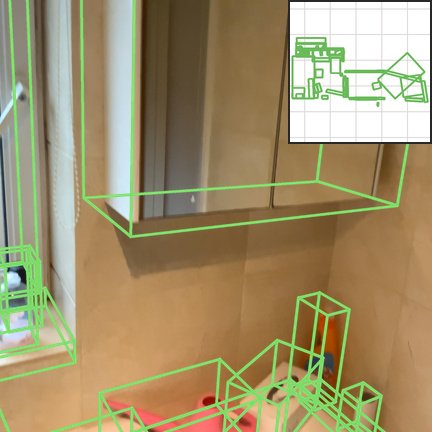} \\

 &
\img{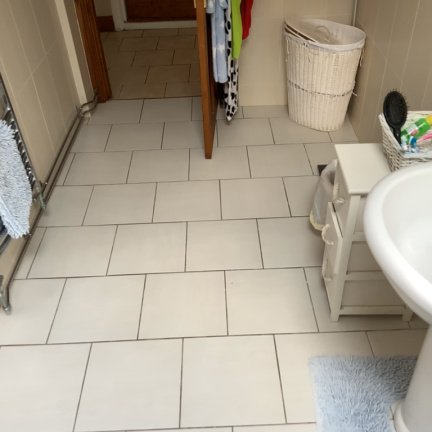} &
\img{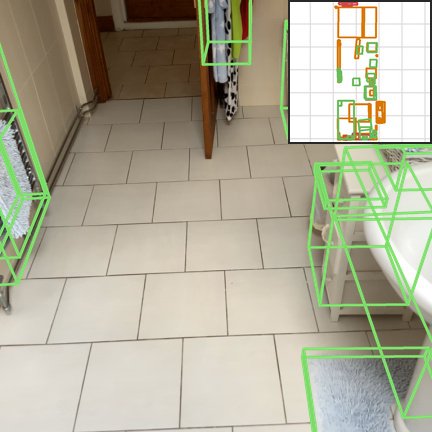} &
\img{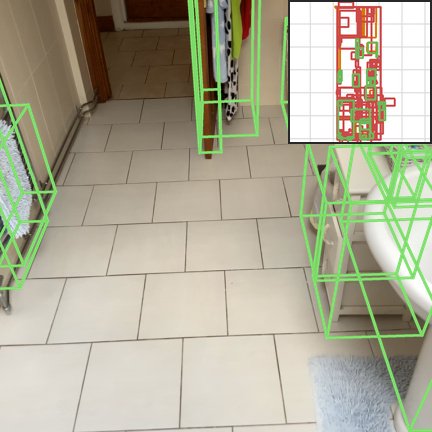} &
\img{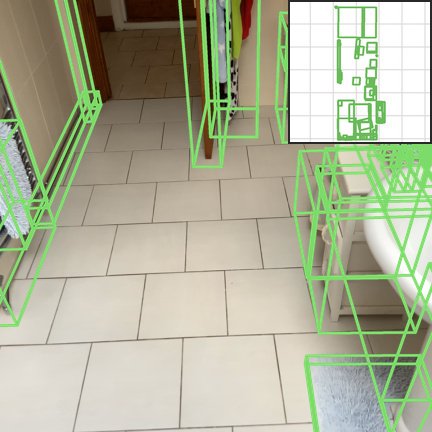} \\

\multirow{2}{*}[69.7pt]{\rotatebox{90}{\makebox[0.4\linewidth][c]{ADT}}} &
\img{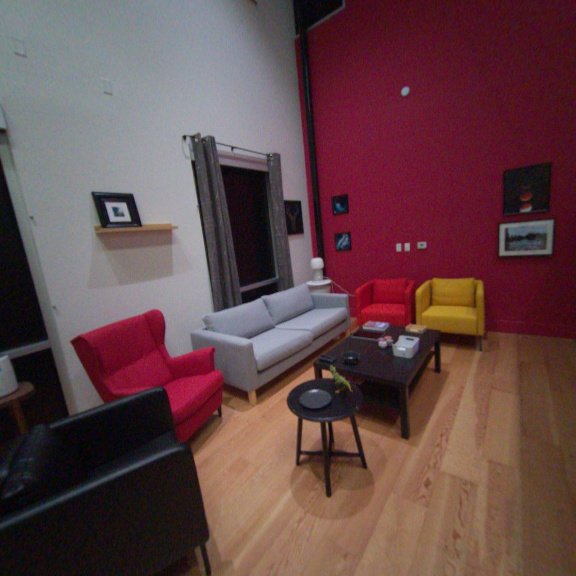} &
\img{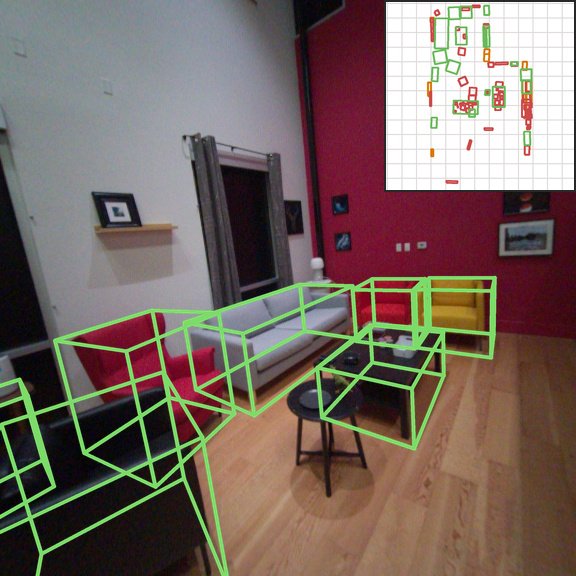} &
\img{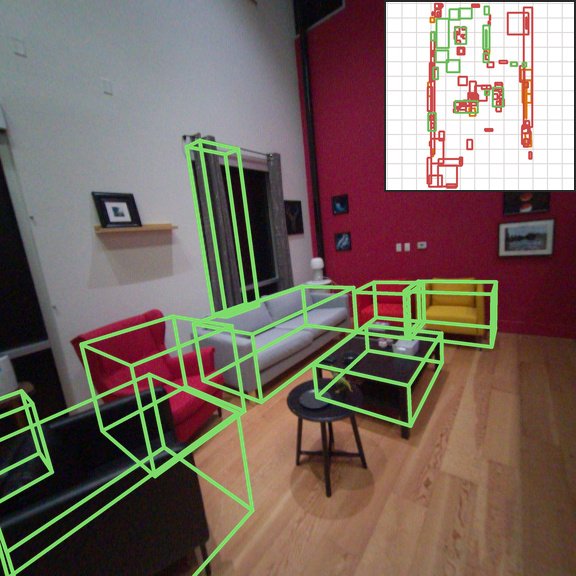} &
\img{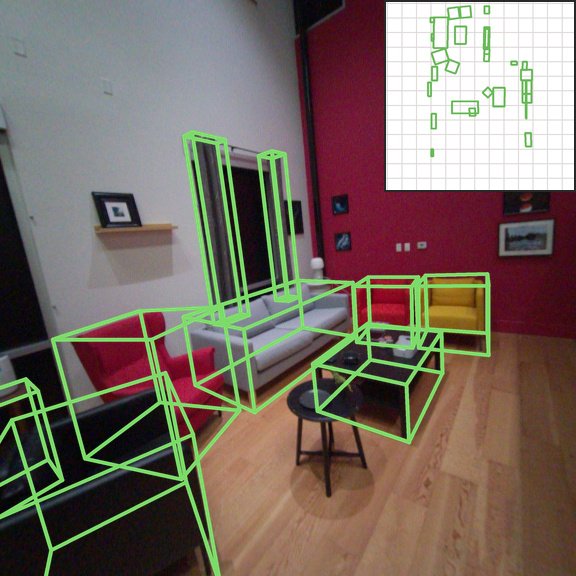} \\

 &
\img{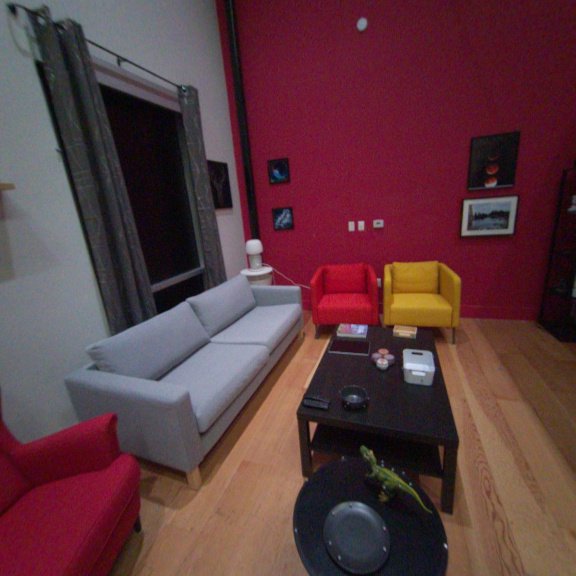} &
\img{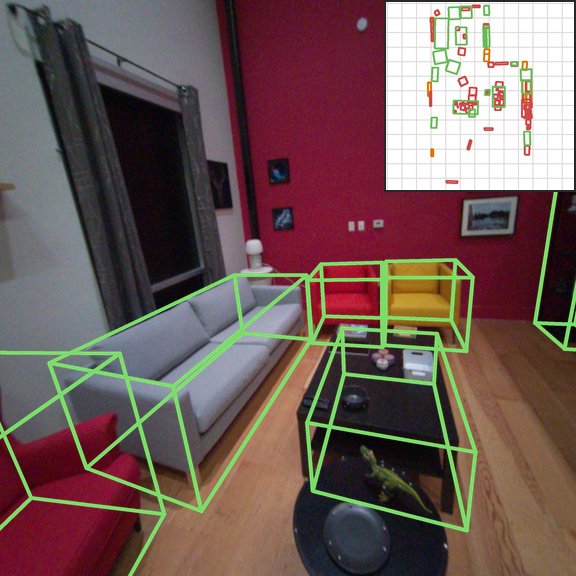} &
\img{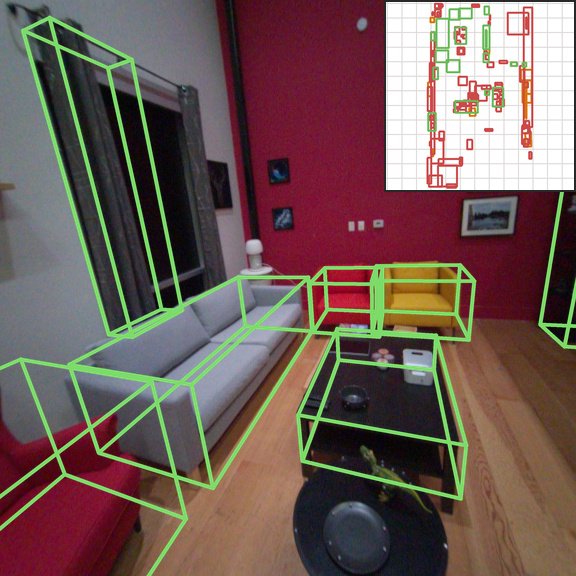} &
\img{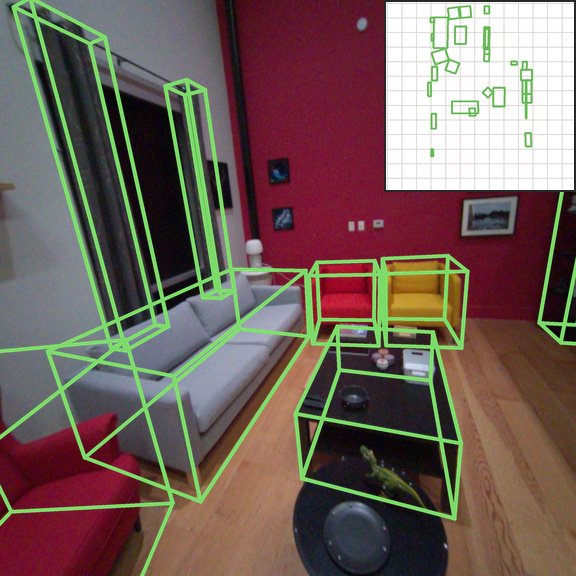} \\

\multirow{2}{*}[69.7pt]{\rotatebox{90}{\makebox[0.4\linewidth][c]{NymeriaPlus}}} &
\img{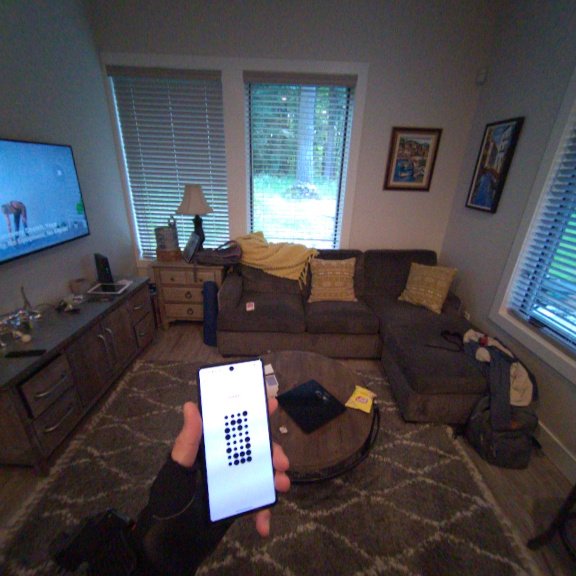} &
\img{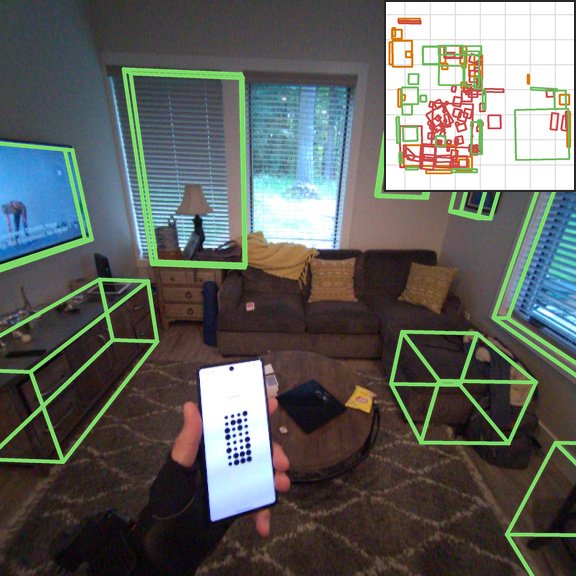} &
\img{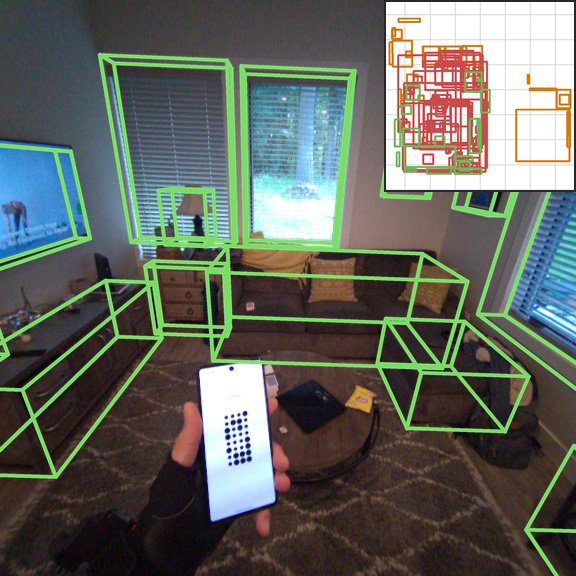} &
\img{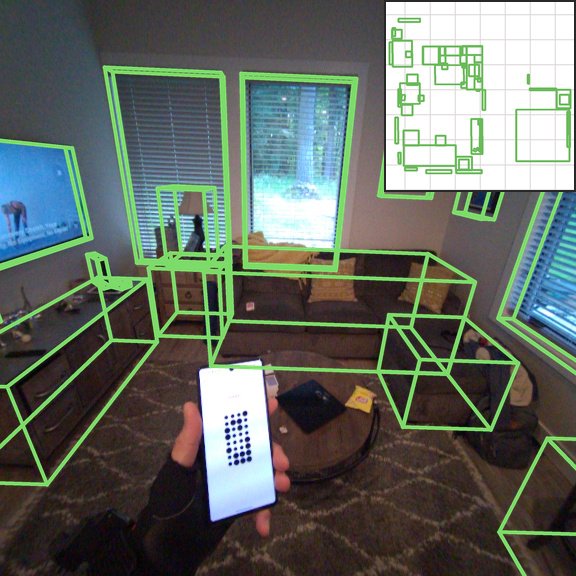} \\

 &
\img{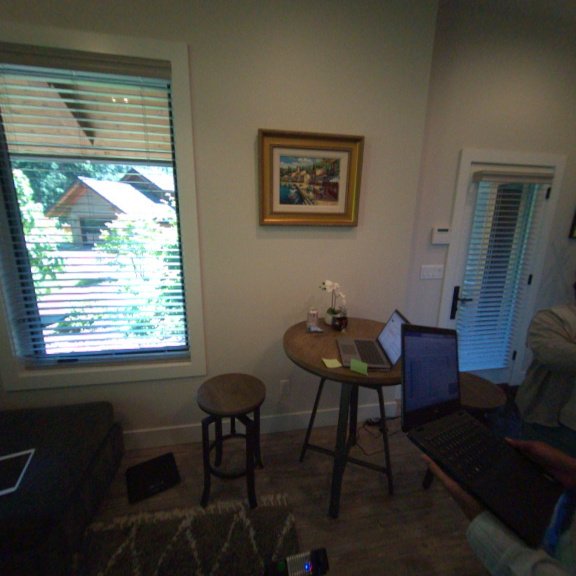} &
\img{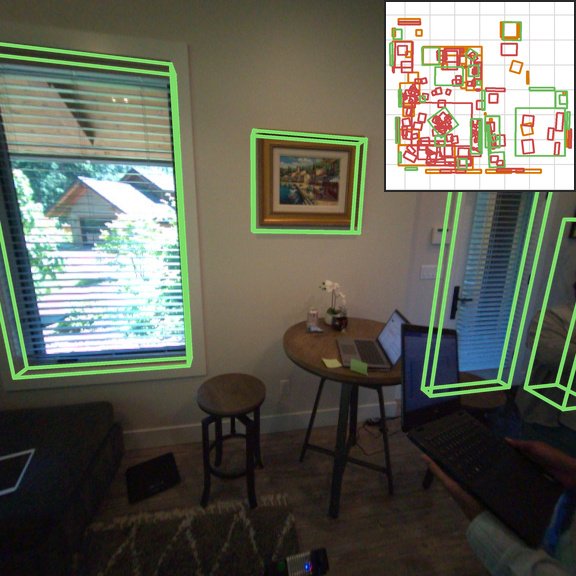} &
\img{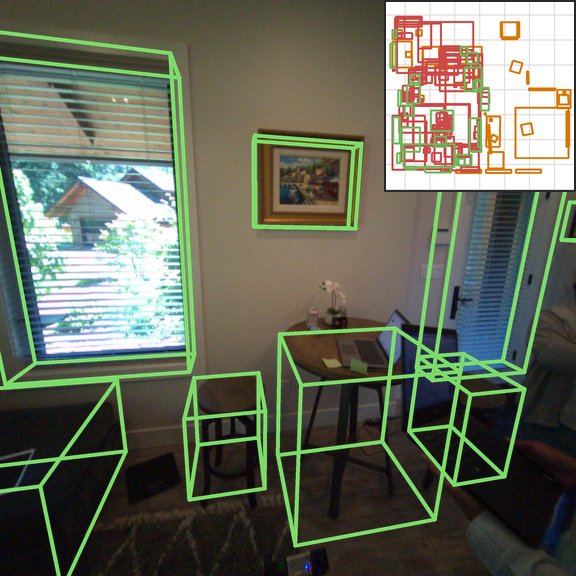} &
\img{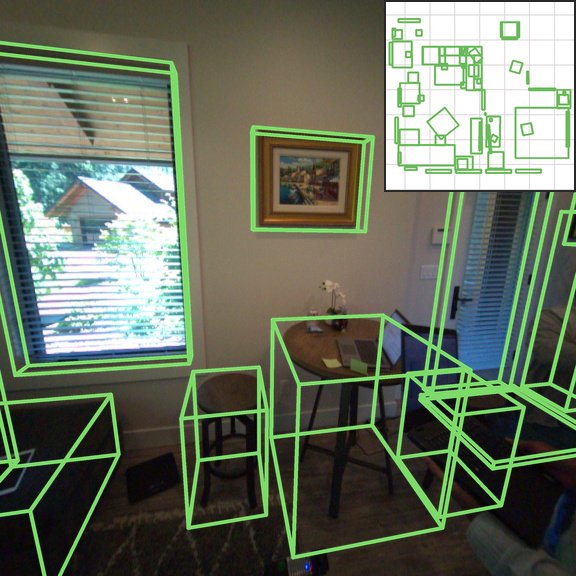} \\

\end{tabular}%
}
\caption{\textbf{Qualitative comparison with Boxer on its real-world benchmarks.} Green denotes GT or a prediction with 3D IoU $\geq 0.25$, red denotes an unmatched prediction, and orange denotes a missed GT object. Every panel includes the bird's-eye-view inset.}
\label{fig:boxer_benchmarks_qual}
\end{figure}

\paragraph{\rebut{More Results on Real-world Perception Benchmarks}}
\rebut{
We evaluate the zero-shot perception transfer capability of \model{} to more real-world captures datasets.
As Tab.~\ref{tab:boxer_benchmarks} shows, we evaluate video-based 3D scene perception on CA-1M~\cite{lazarow2025cubify}, ADT~\cite{pan2023aria}, and NymeriaPlus~\cite{detone2026nymeriaplus}, the three real-world benchmarks in Boxer~\cite{boxer}.
We follow the evaluation protocol of Boxer. 
We adopt the metric of class-agnostic mAP averaged over 3D IoU thresholds from 0.05 to 0.5. 
Since Boxer's evaluation code is not released, we reimplement this protocol and evaluate Boxer's released model with it. 
\textbf{\model{} uses none of these datasets in training, so it is a zero-shot transfer.} 
\model{} exceeds Boxer on CA-1M and ADT datasets.
On NymeriaPlus, Boxer reports results on separately annotated GT that is not released, so we evaluate both methods on the 43 public sequences with the original GT, where the two methods are on par.
Fig.~\ref{fig:boxer_benchmarks_qual} shows qualitative results.
}

\begin{table}[ht]
\centering
\small
\begin{tabular}{lccc}
\toprule
\textbf{Dataset} & \textbf{Boxer (published)} & \textbf{Boxer (our evaluation)} & \textbf{\model{}} \\
\midrule
CA-1M & 0.204 & 0.212 & \textbf{0.250} \\
ADT & 0.215 & 0.215 & \textbf{0.276} \\
NymeriaPlus & 0.278 (unreleased GT) & 0.184 (public GT) & \textbf{0.187} (public GT) \\
\bottomrule
\end{tabular}
\caption{\textbf{Quantitative results on Boxer's real-world benchmarks.} \model{} is evaluated zero-shot with class-agnostic mAP averaged over 3D IoU thresholds from 0.05 to 0.5.}
\label{tab:boxer_benchmarks}
\end{table}

\paragraph{More \rev{Details and} Results on Video-based 3D Scene Reconstruction}
Here we also showcase more qualitative comparison results against SoTA methods \cite{shaper, sam3d}. 
In Fig. \ref{fig:recon_cmp_gt_perception_supp}, we compare reconstruction under GT perception inputs.
In Fig. \ref{fig:recon_cmp_inferred_perception_supp}, we further compare the model performance under inferred perception inputs.
\rev{We sample 100,000 surface points per mesh. CD averages the two directional mean nearest-neighbor distances and is reported in centimeters; F1 uses a 1 cm distance threshold, and NC averages absolute nearest-neighbor normal agreement in both directions.}
\rev{We jointly assess instance recovery using mAP/mIoU, which account for missed objects, and reconstruction quality using matched-object geometry metrics under inferred perception. Rendering metrics in Tab.~\ref{tab:main_cmp} use full images from novel views around each object. The main video comparisons use 67 iTHOR and 120 Imaginarium scenes; the ShapeR evaluation covers all 178 annotated objects across seven scenes.}
\rev{Real-world evaluation follows the available annotations: AEO for perception and ShapeR for amodal object geometry. Since real-world scans generally lack complete per-object geometry and rendering under occlusion~\cite{shaper, scannet++}, synthetic scenes complement these tests with controlled geometry and rendering comparisons.}

\paragraph{\rebut{More Results on Real-world Video-based 3D Scene Reconstruction}}
\rebut{
We evaluate the 3D scene reconstruction capability of \model{} in real-world videos as well. 
We randomly choose 10 real scenes from ScanNet++~\cite{scannet++} dataset.
Since ScanNet++ has no complete per-object GT, we score the predicted foreground geometry of each method against the full scanned 3D surface provided by the dataset. 
We compare against EFM3D + ShapeR and the composed Boxer + SAM2 + TRELLIS.2 pipeline in Tab.~\ref{tab:composed_cmp}. 
We use the same 60 evenly spaced frames for all methods, and \model{} uses no ScanNet++ data in training. 
\model{} reconstructs better scene geometry than both pipelines on every scene in CD, F1 at 5\,cm, and NC (Tab.~\ref{tab:scannetpp}). 
Fig.~\ref{fig:scannetpp_qual} shows qualitative results.
}

\begin{table}[ht]
\centering
\small
\begin{tabular}{lccc}
\toprule
\textbf{Method} & \textbf{CD} (cm) $\downarrow$ & \textbf{F1@5cm} $\uparrow$ & \textbf{NC} $\uparrow$ \\
\midrule
EFM3D + ShapeR & 40.34 & 0.189 & 0.527 \\
Boxer + SAM2 + TRELLIS.2 & 29.02 & 0.221 & 0.557 \\
\model{} & \textbf{16.95} & \textbf{0.458} & \textbf{0.610} \\
\bottomrule
\end{tabular}
\caption{\textbf{Quantitative comparison on ScanNet++.} \model{} achieves better scene geometry than both pipelines on all 10 real scenes.}
\label{tab:scannetpp}
\end{table}

\begin{table*}[ht]
    \centering
    \setlength{\tabcolsep}{1pt}
    \renewcommand{\arraystretch}{0.5}
    \resizebox{\textwidth}{!}{%
    \begin{tabular}{cccc}
        \includegraphics[width=0.245\linewidth]{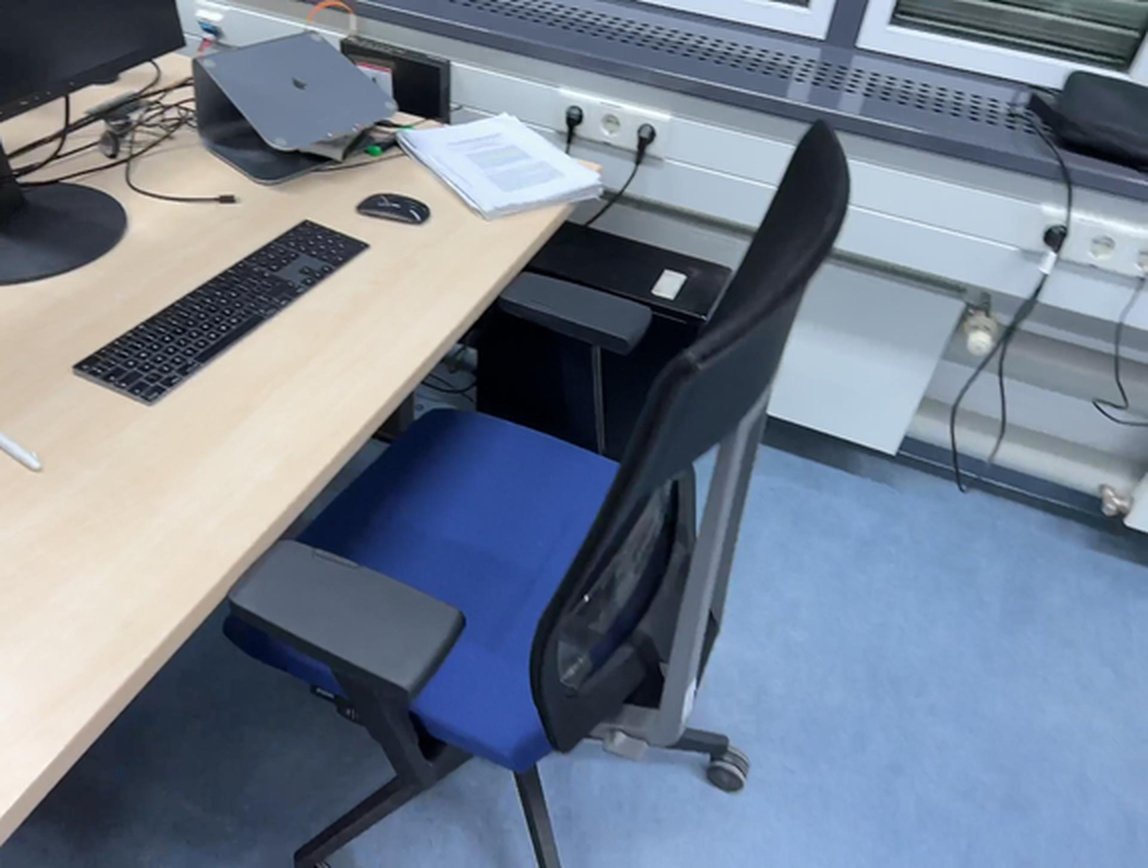} &
        \includegraphics[width=0.245\linewidth]{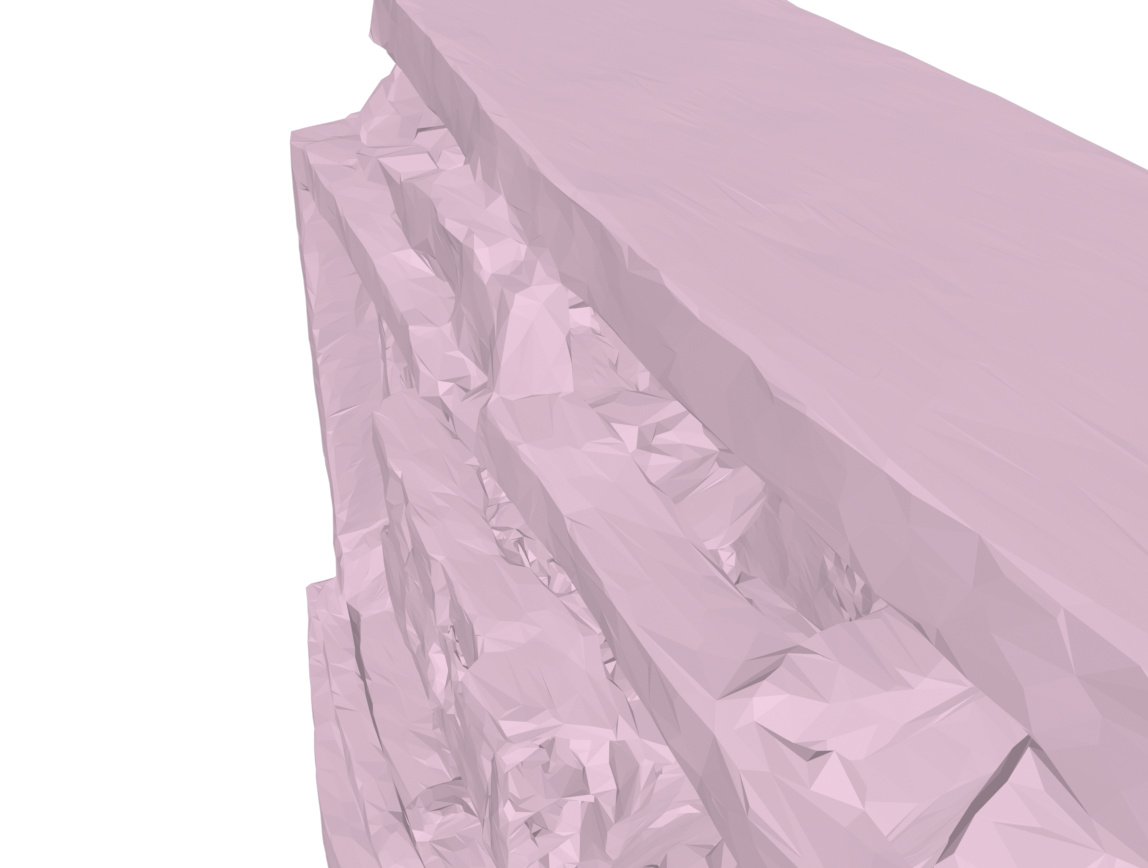} &
        \includegraphics[width=0.245\linewidth]{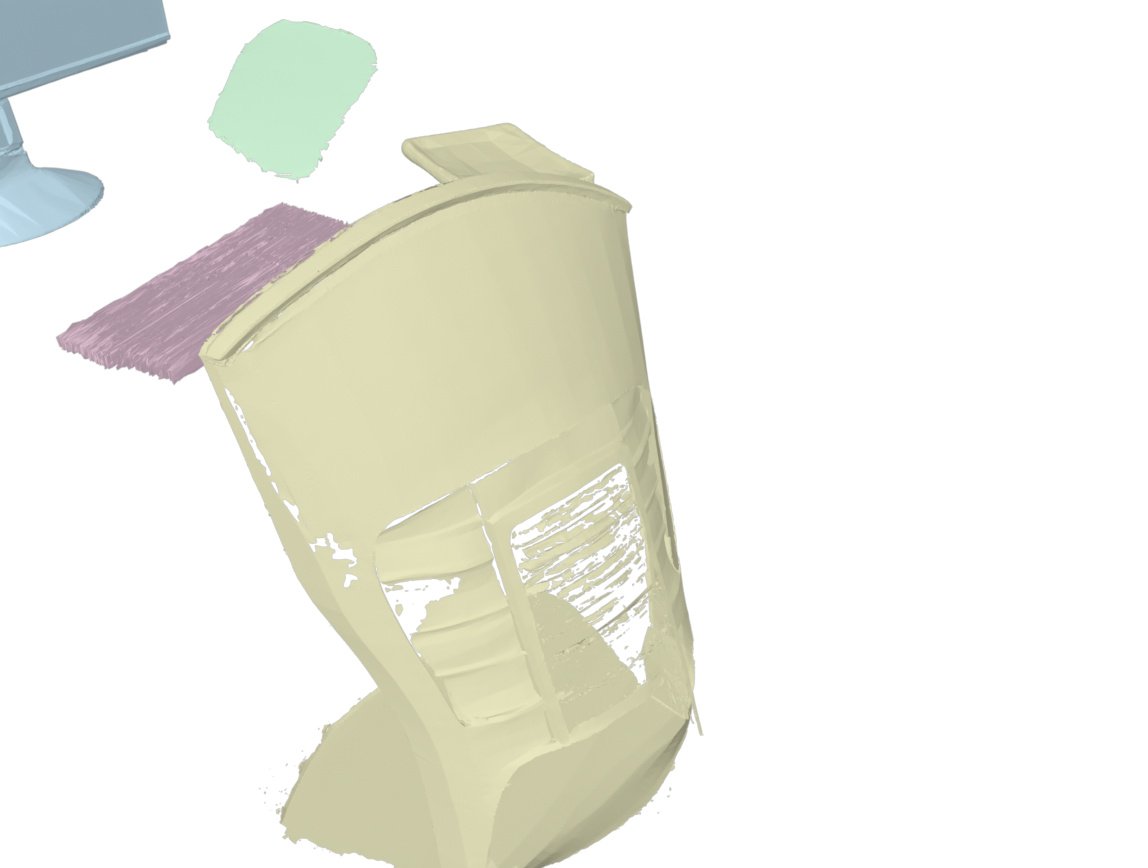} &
        \includegraphics[width=0.245\linewidth]{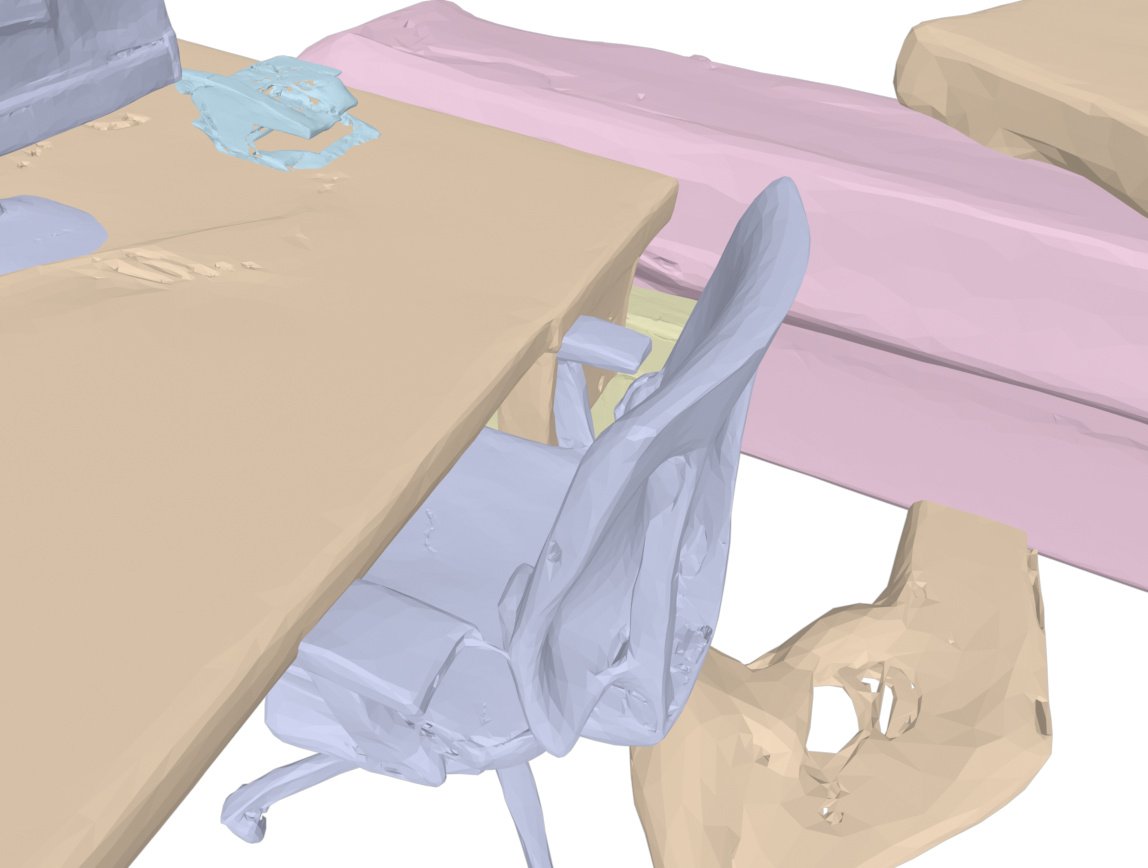} \\[1pt]
        \includegraphics[width=0.245\linewidth]{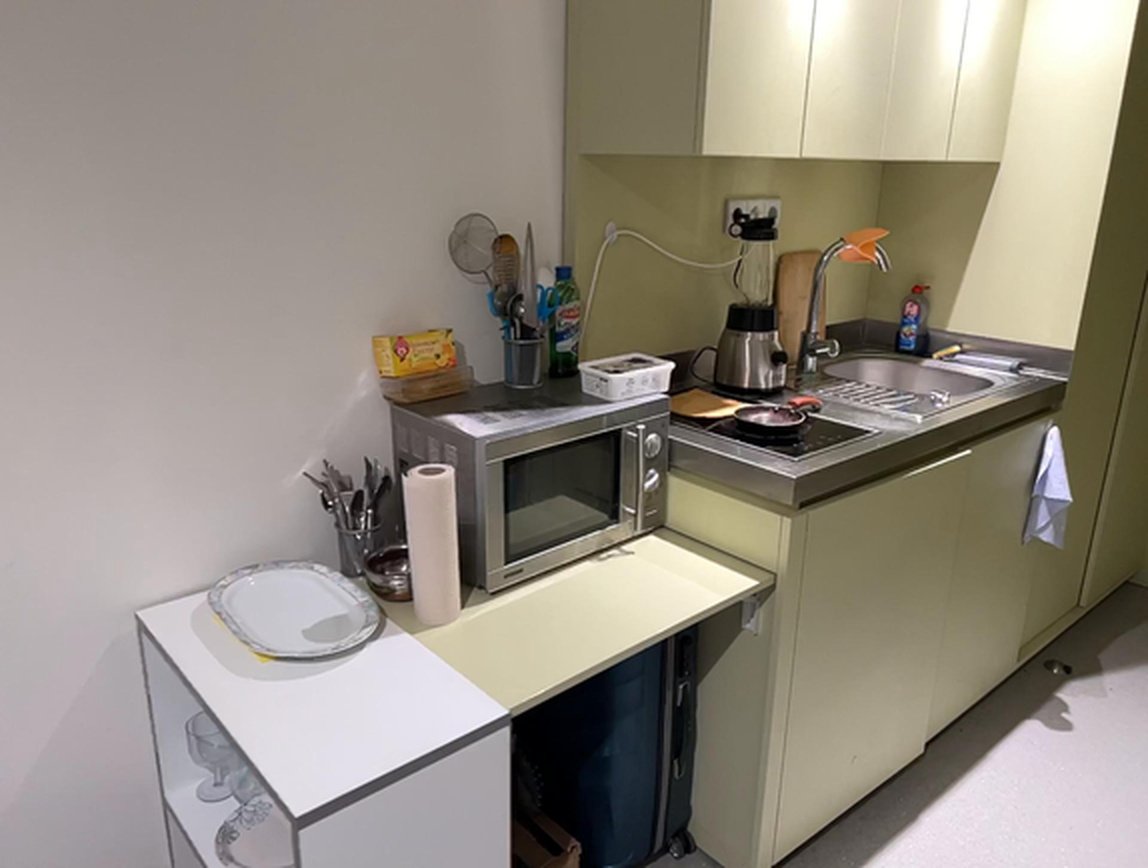} &
        \includegraphics[width=0.245\linewidth]{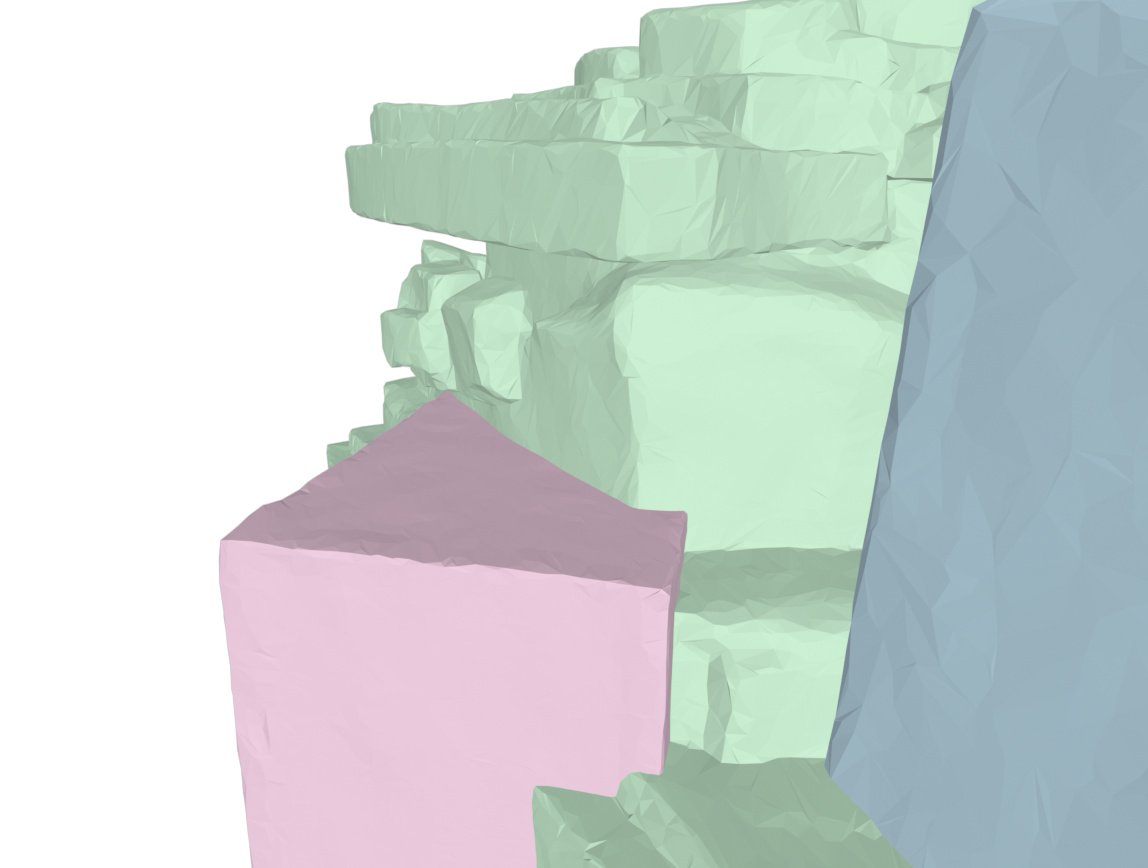} &
        \includegraphics[width=0.245\linewidth]{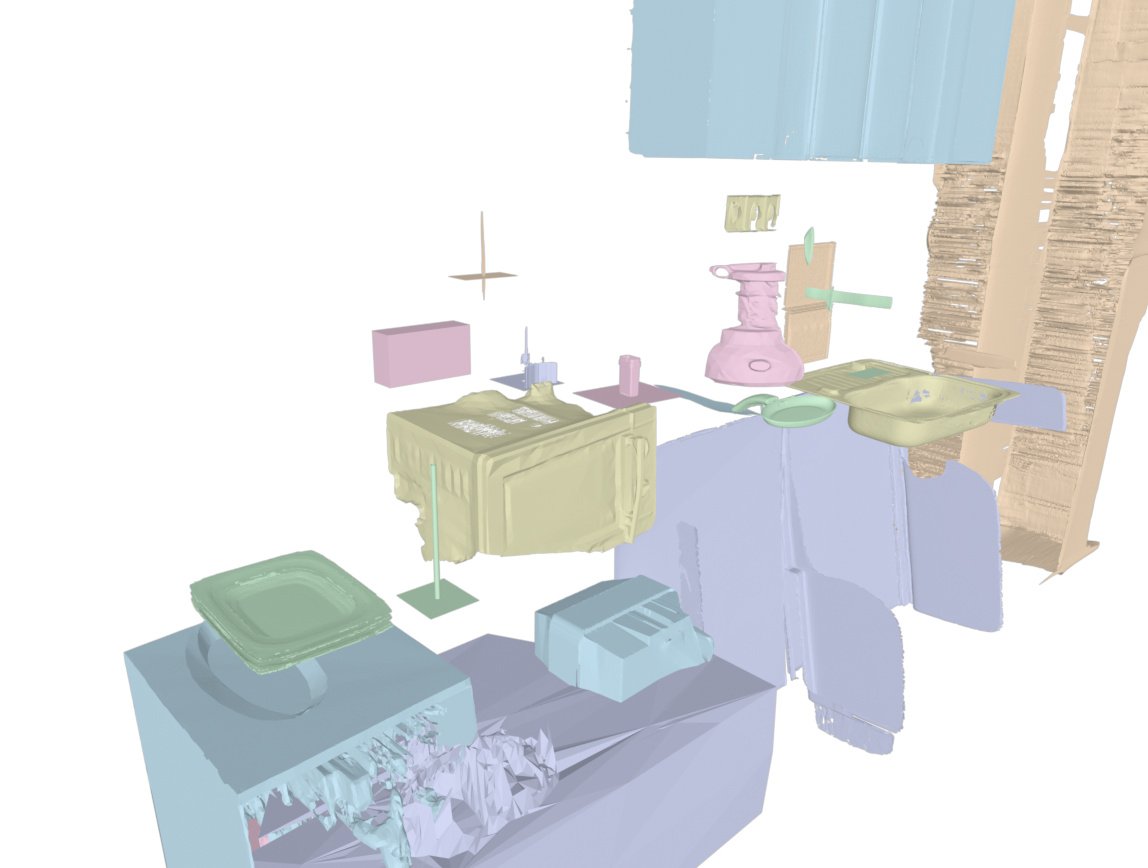} &
        \includegraphics[width=0.245\linewidth]{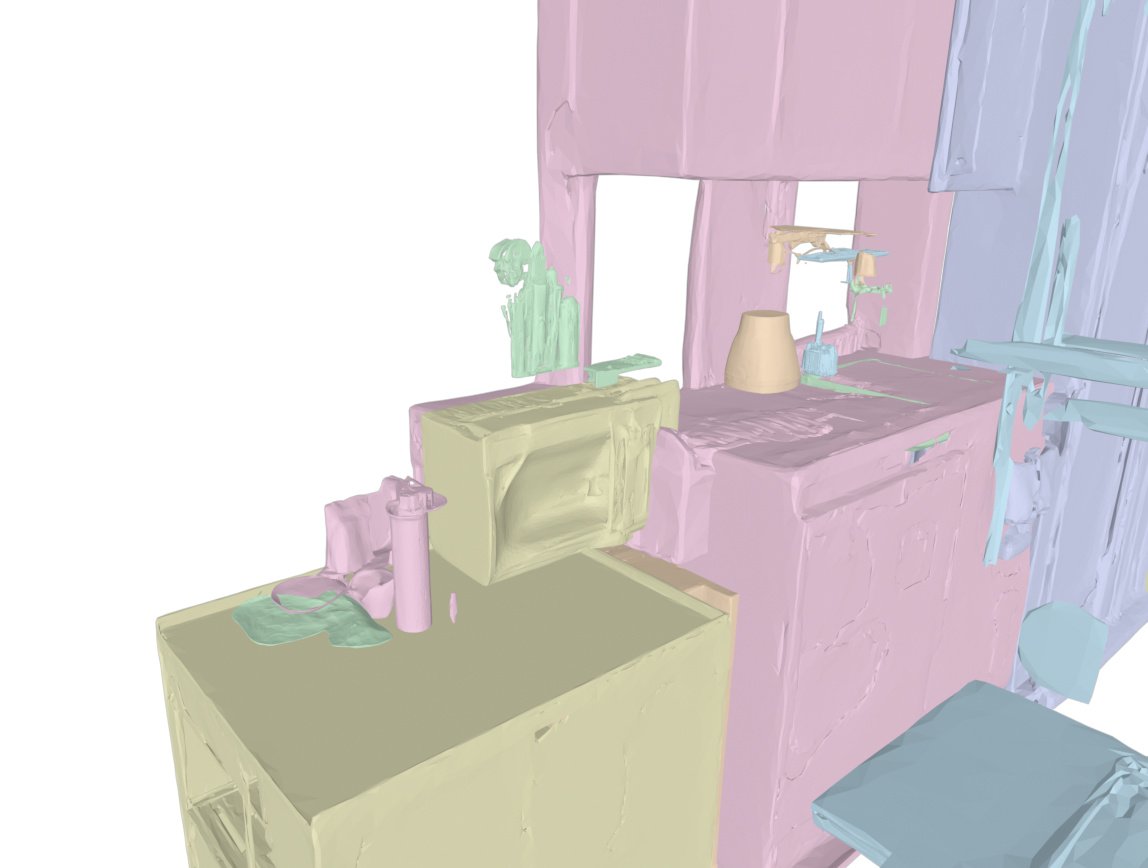} \\[1pt]
        Input & EFM3D + ShapeR & Boxer + SAM2 + TRELLIS.2 & Ours
    \end{tabular}%
    }
    \captionof{figure}{\textbf{Qualitative comparison on ScanNet++.} \model{} reconstructs more complete objects than both pipelines. All methods use the same 60 input frames and the camera of the input view, and background geometry is omitted for all columns.}
    \label{fig:scannetpp_qual}
\end{table*}

\begin{table*}[ht]
    \centering
    \setlength{\tabcolsep}{1pt}
    \renewcommand{\arraystretch}{0.5}

    \resizebox{\textwidth}{!}{%
    \begin{tabular}{c c c c c c c c}

        \includegraphics[width=0.245\linewidth]{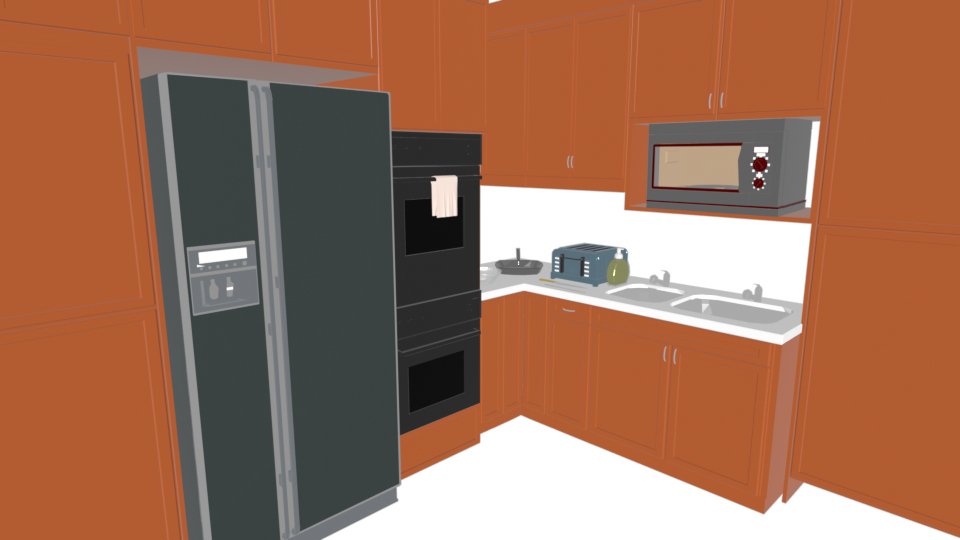} &
        \includegraphics[width=0.245\linewidth]{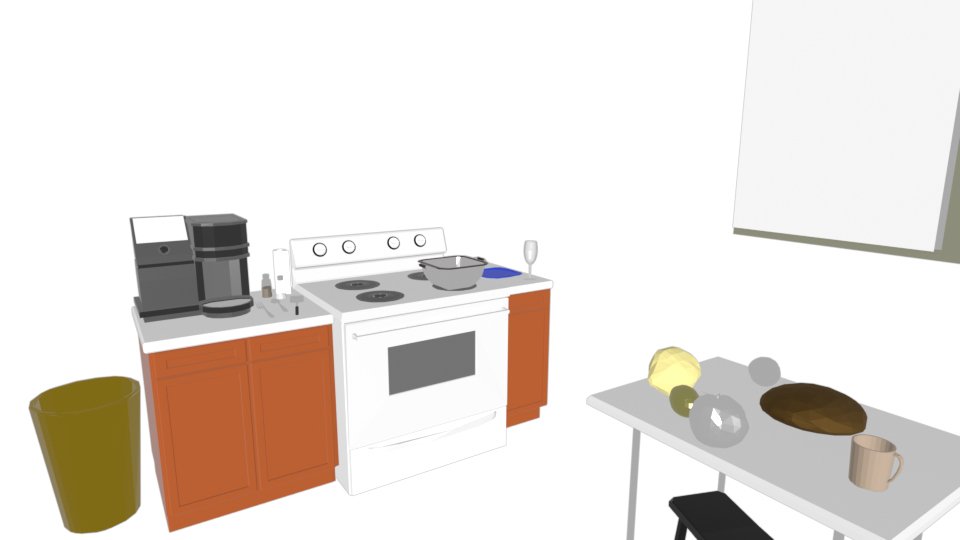} &
        \includegraphics[width=0.245\linewidth]{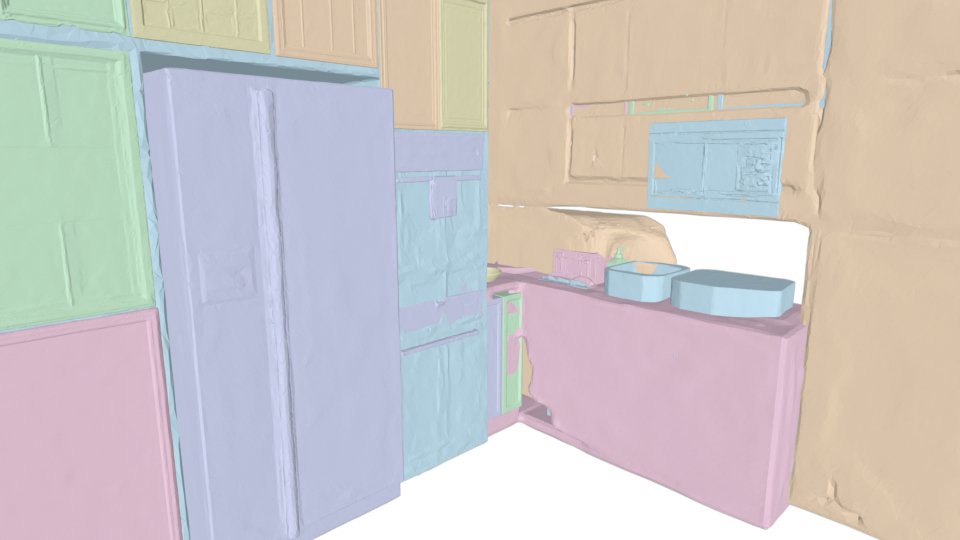} &
        \includegraphics[width=0.245\linewidth]{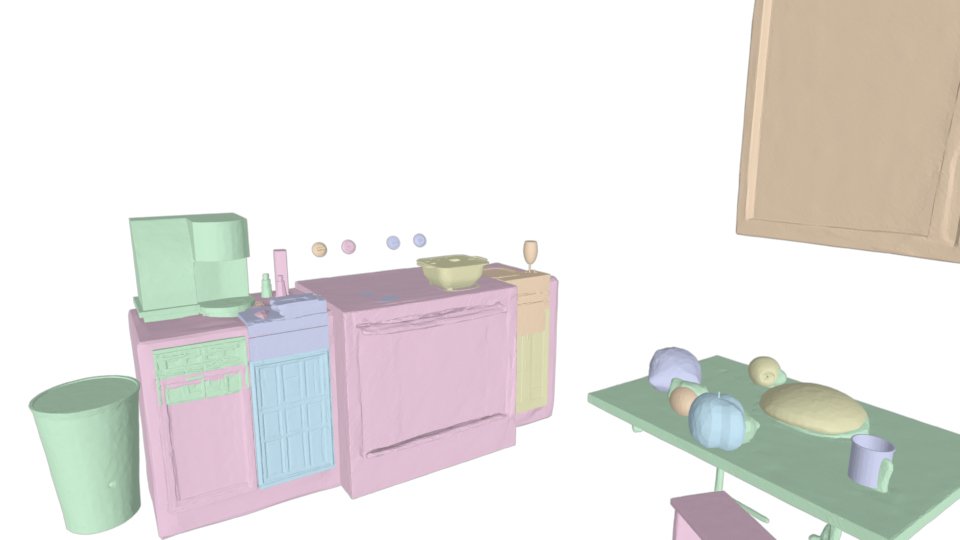} &
        \includegraphics[width=0.245\linewidth]{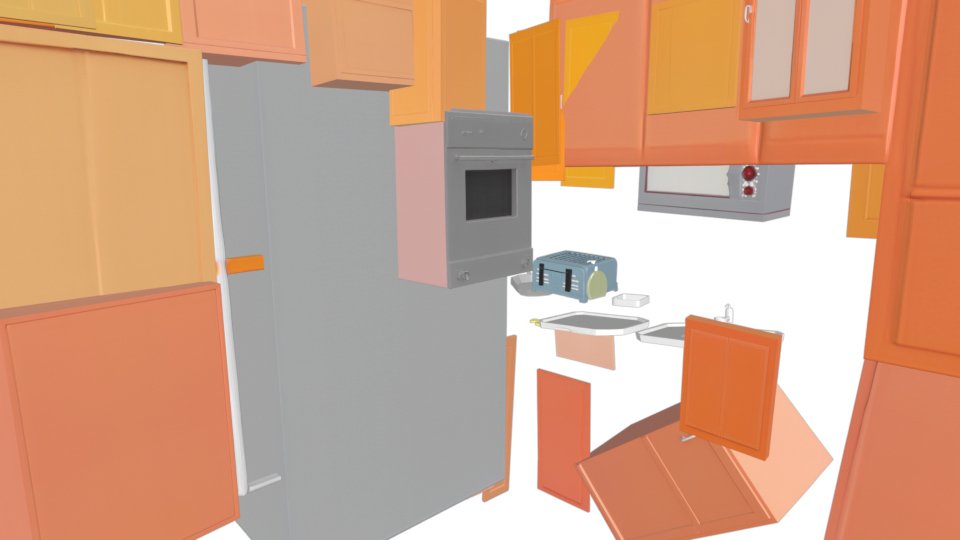} &
        \includegraphics[width=0.245\linewidth]{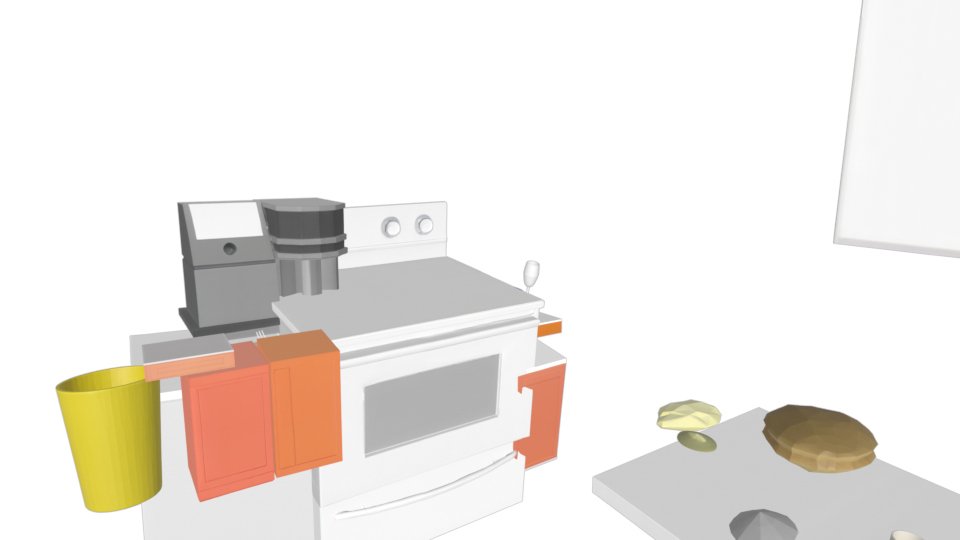} &
        \includegraphics[width=0.245\linewidth]{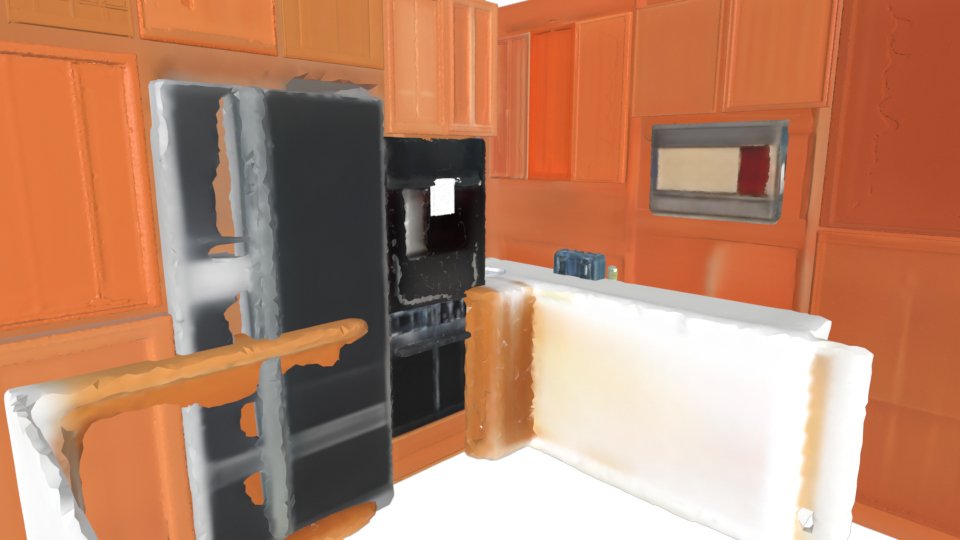} &
        \includegraphics[width=0.245\linewidth]{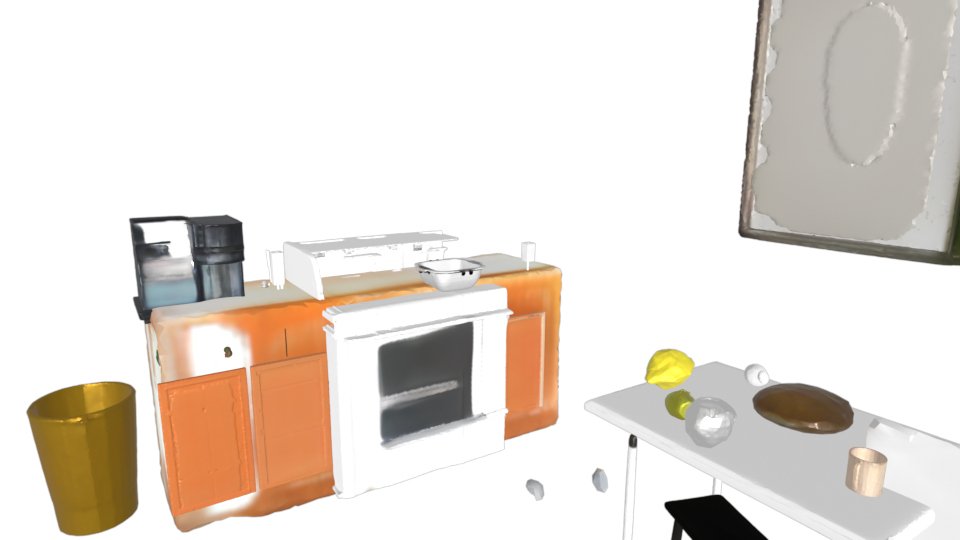} \\[1pt]

        \includegraphics[width=0.245\linewidth]{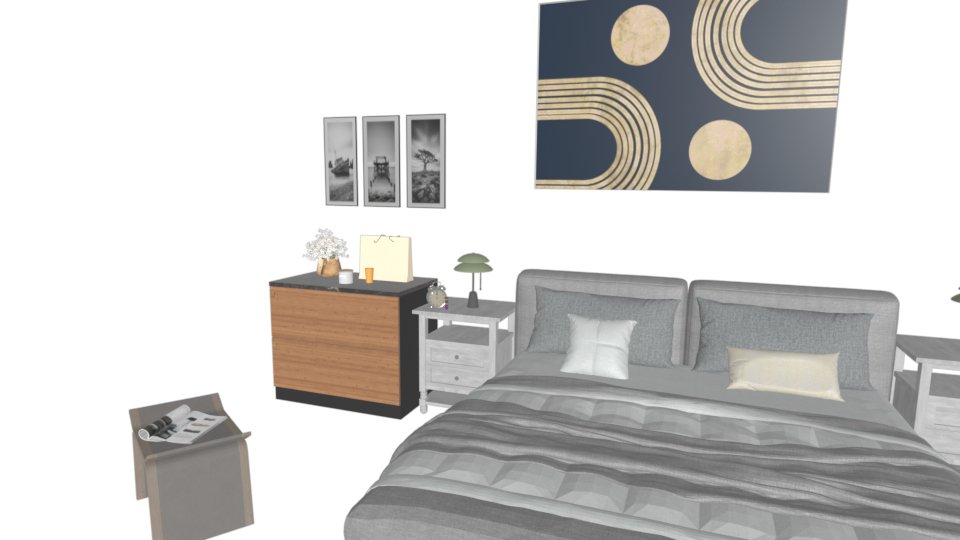} &
        \includegraphics[width=0.245\linewidth]{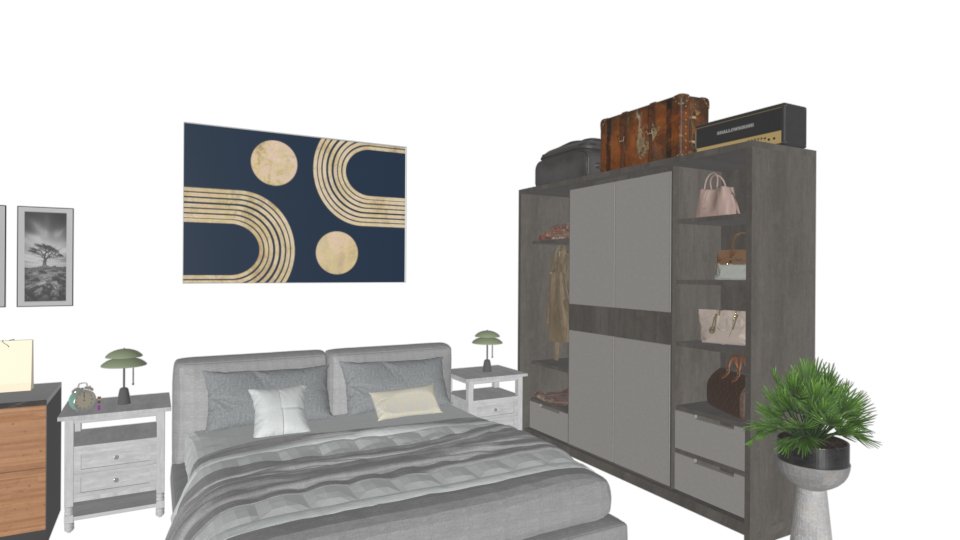} &
        \includegraphics[width=0.245\linewidth]{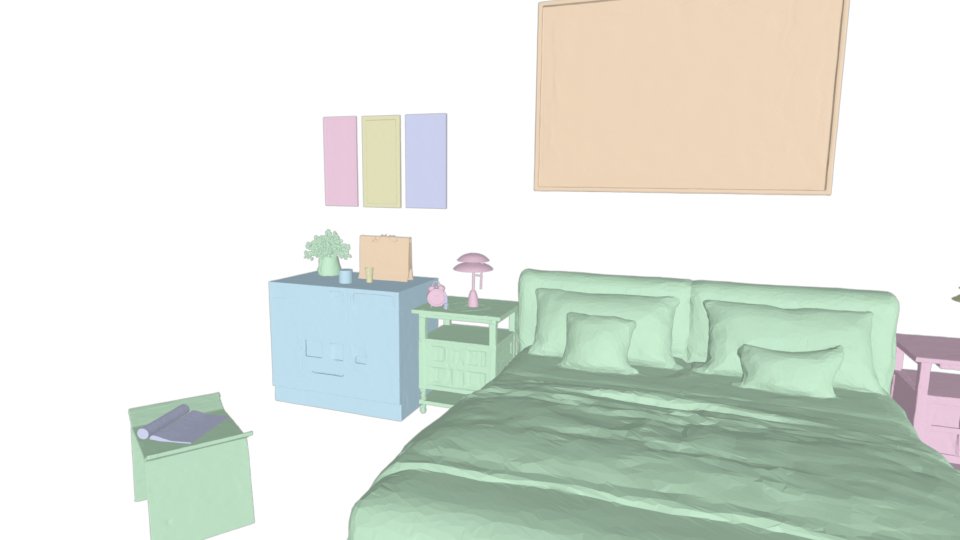} &
        \includegraphics[width=0.245\linewidth]{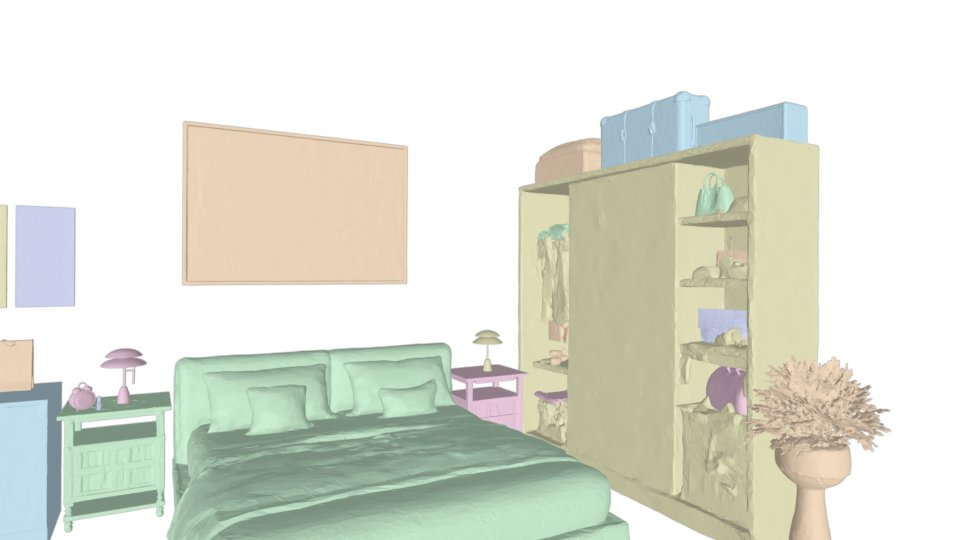} &
        \includegraphics[width=0.245\linewidth]{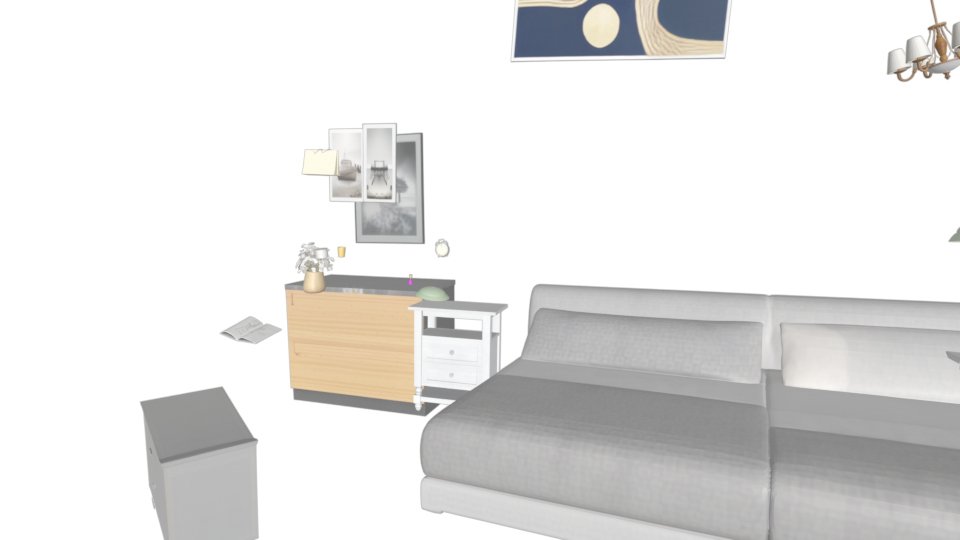} &
        \includegraphics[width=0.245\linewidth]{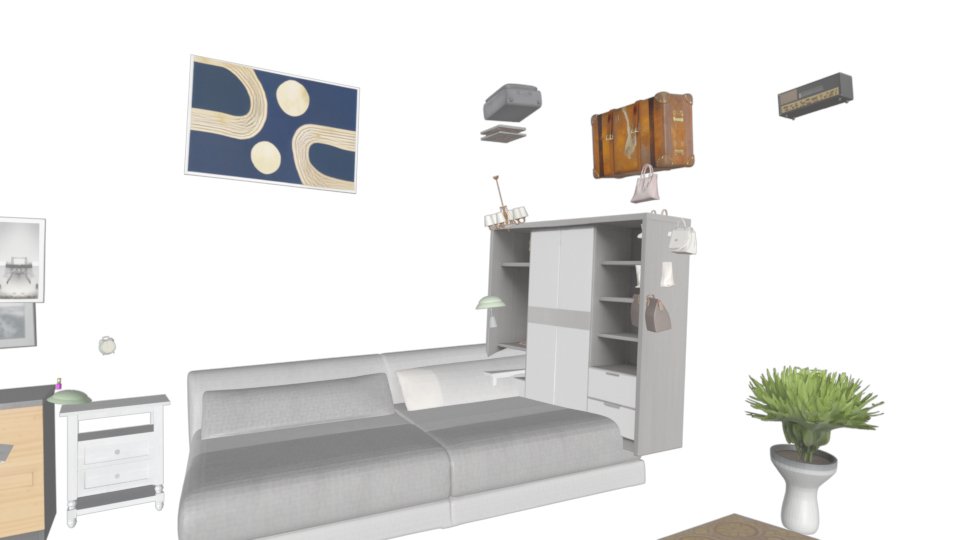} &
        \includegraphics[width=0.245\linewidth]{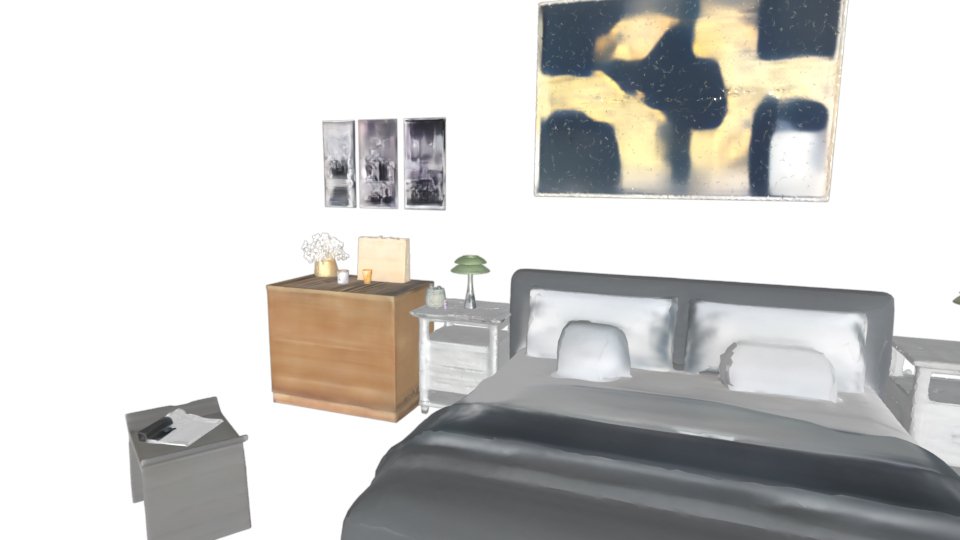} &
        \includegraphics[width=0.245\linewidth]{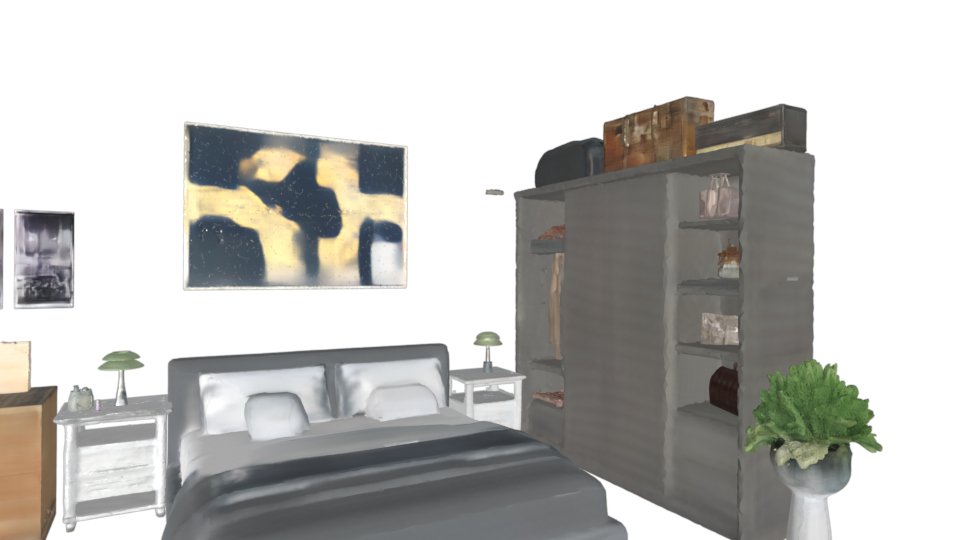} \\[1pt]
        GT View 1 & GT View 2
        & ShapeR View 1 & ShapeR View 2
        & \rev{SAM3D (frame sel.)} View 1 & \rev{SAM3D (frame sel.)} View 2
        & Ours View 1 & Ours View 2
    \end{tabular}%
    }

    \captionof{figure}{\textbf{More qualitative 3D scene reconstruction with ground-truth instance perception.} 
    }
    \label{fig:recon_cmp_gt_perception_supp}
\end{table*}

\begin{table*}[h]
    \centering
    \setlength{\tabcolsep}{1pt}
    \renewcommand{\arraystretch}{0.5}

    \resizebox{\textwidth}{!}{%
    \begin{tabular}{c c c c c c}
        
        \includegraphics[width=0.245\linewidth]{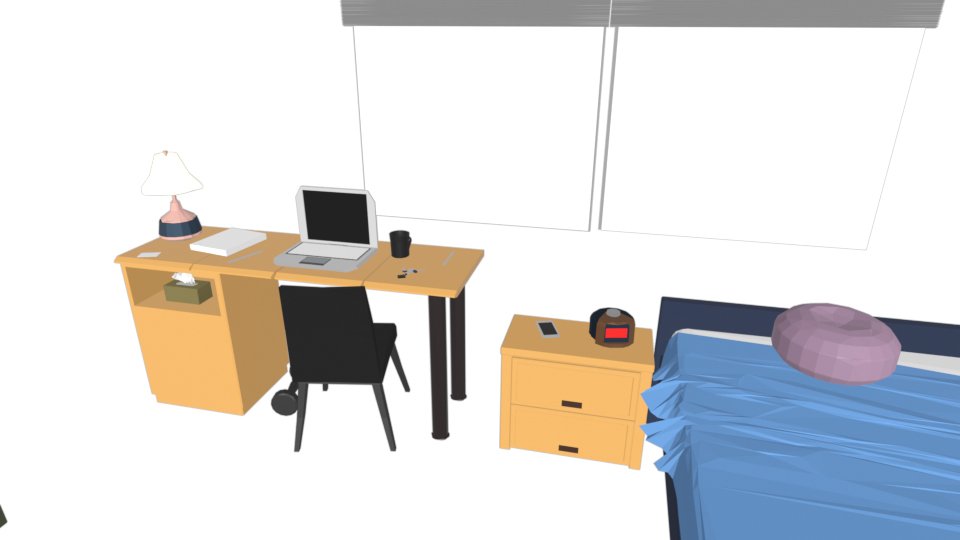} &
        \includegraphics[width=0.245\linewidth]{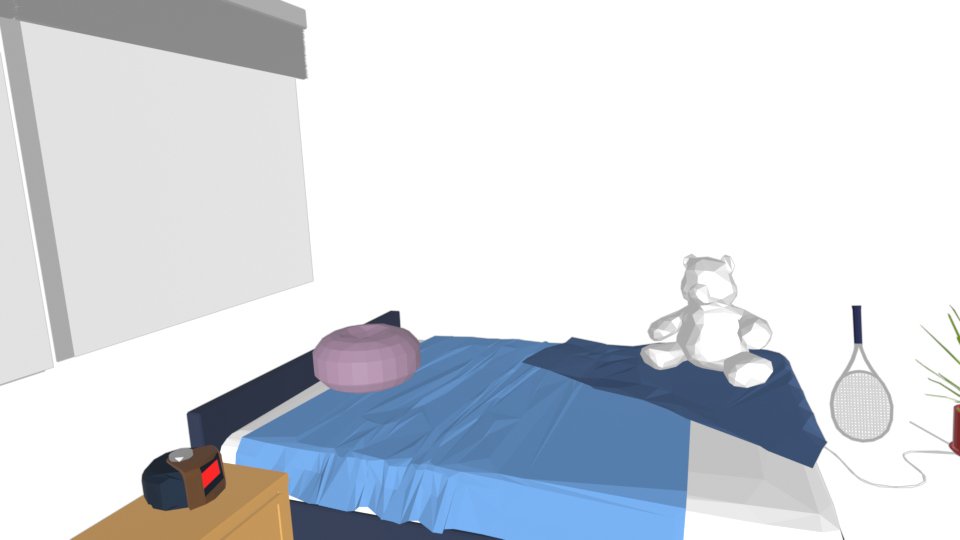} &
        \includegraphics[width=0.245\linewidth]{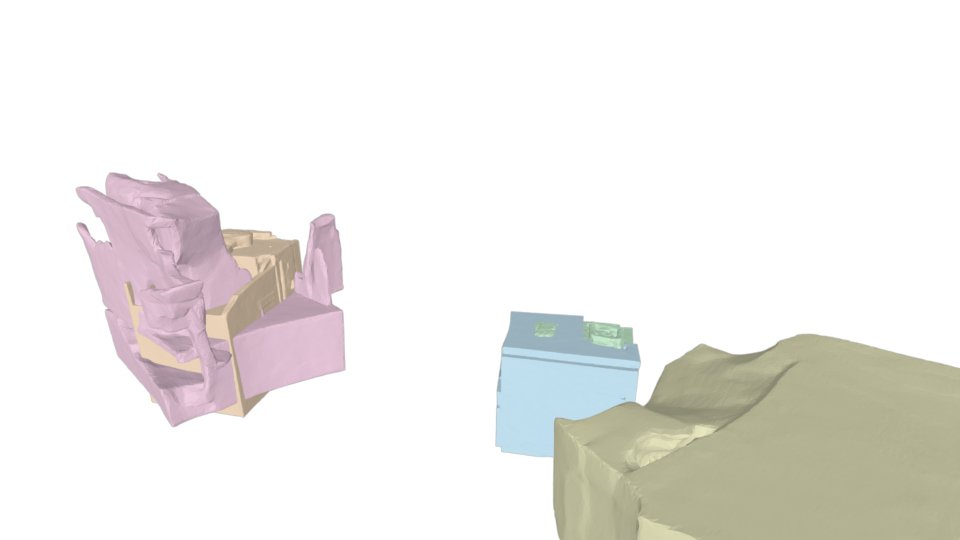} &
        \includegraphics[width=0.245\linewidth]{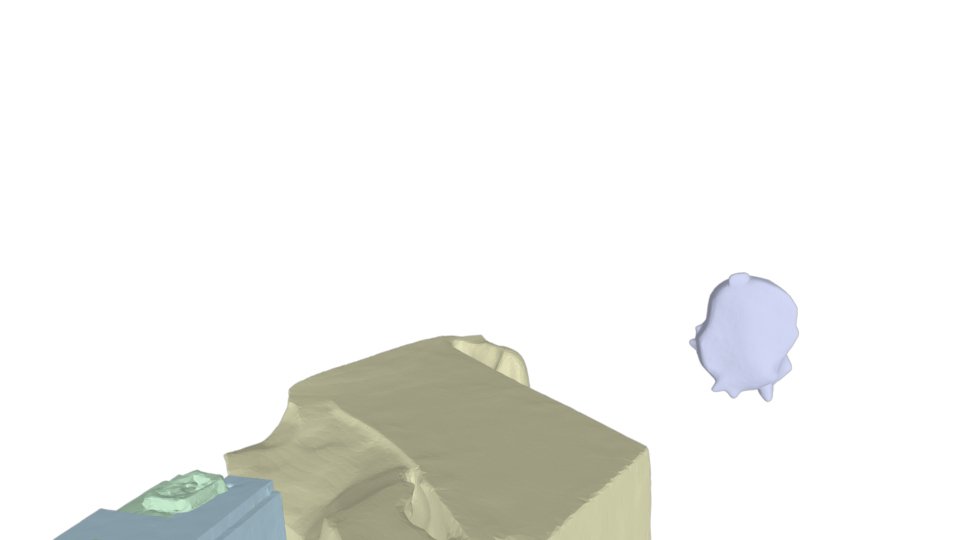} &
        \includegraphics[width=0.245\linewidth]{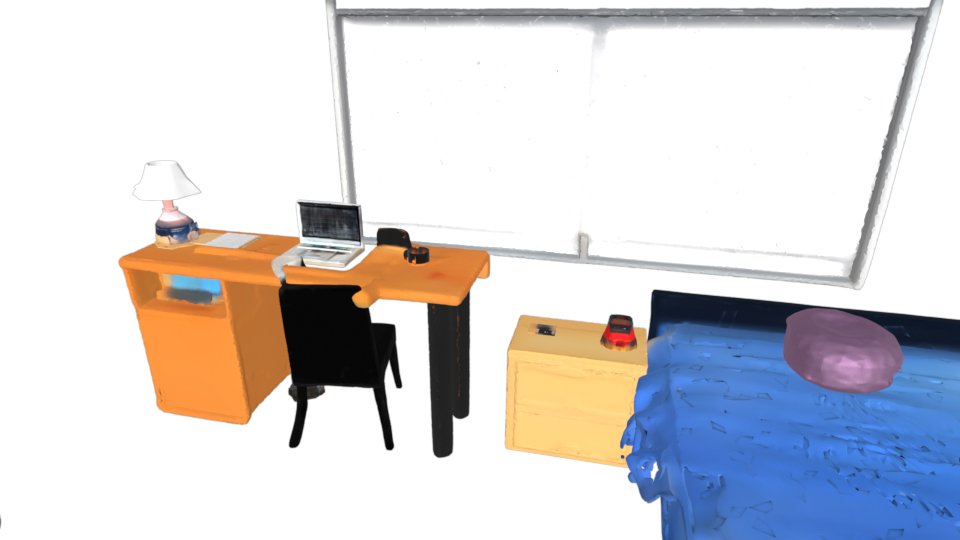} &
        \includegraphics[width=0.245\linewidth]{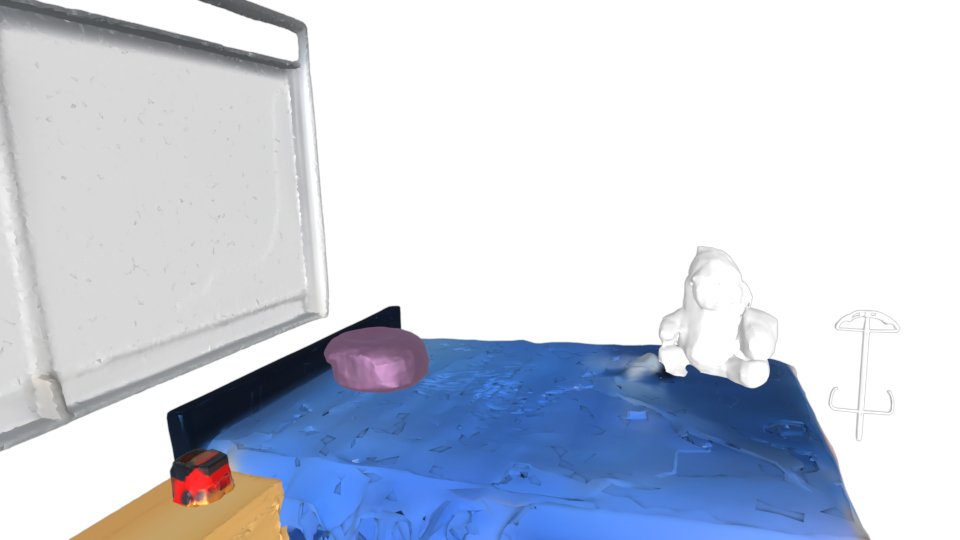} \\[3pt]
        \includegraphics[width=0.245\linewidth]{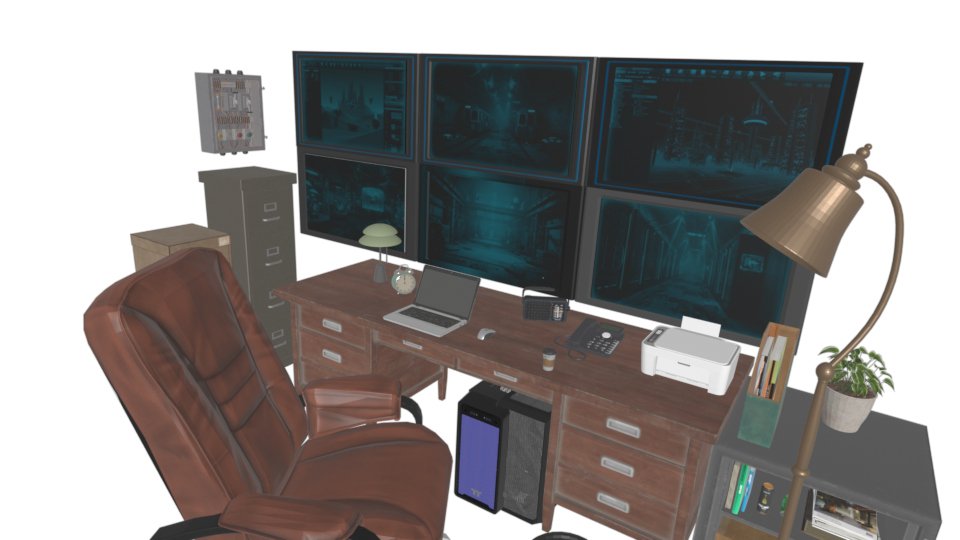} &
        \includegraphics[width=0.245\linewidth]{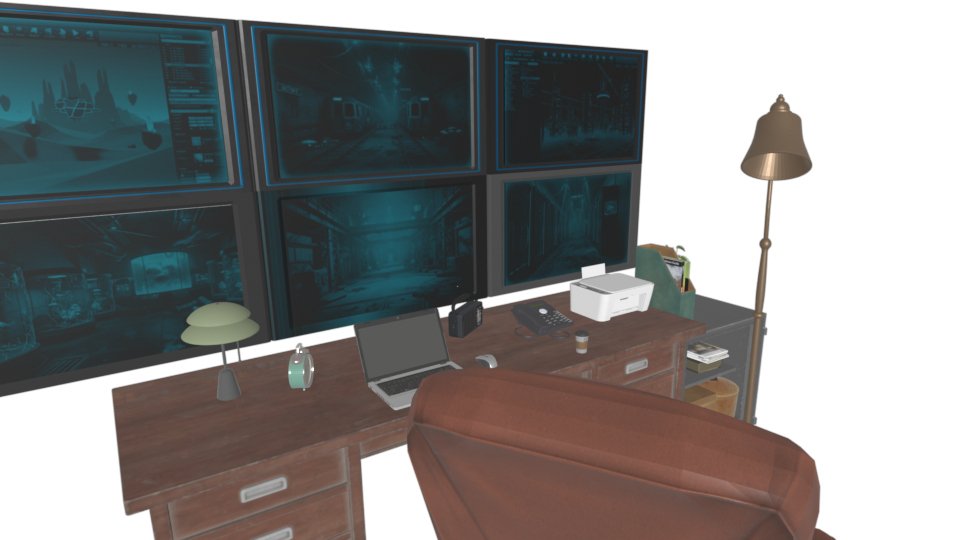} &
        \includegraphics[width=0.245\linewidth]{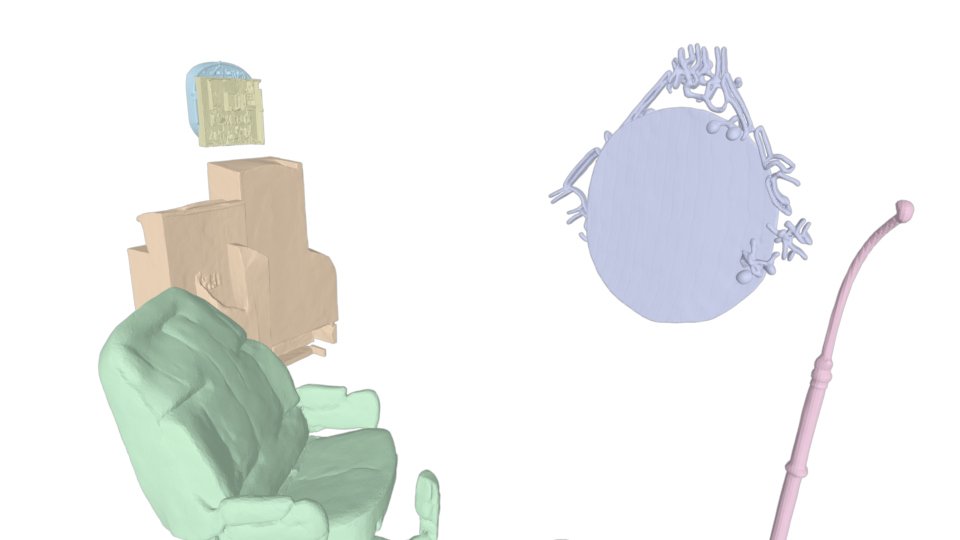} &
        \includegraphics[width=0.245\linewidth]{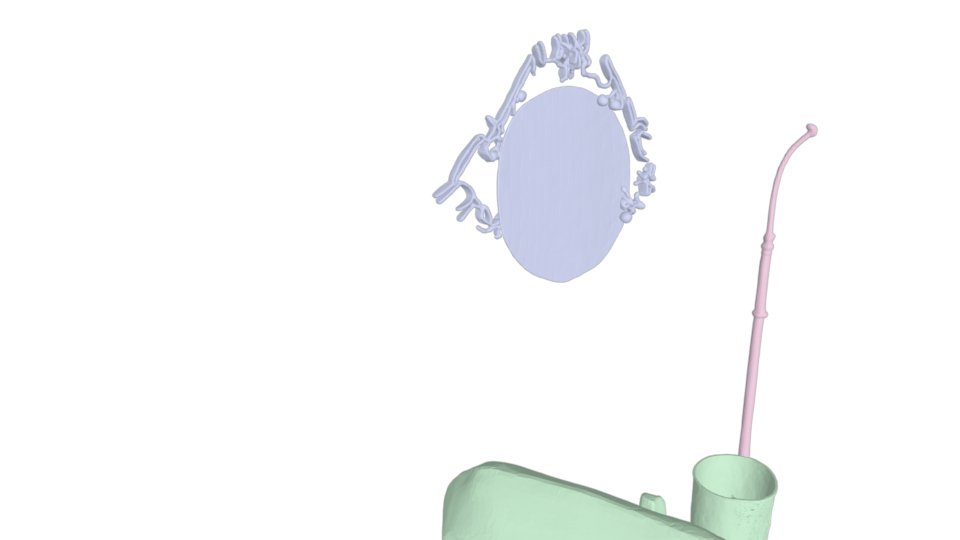} &
        \includegraphics[width=0.245\linewidth]{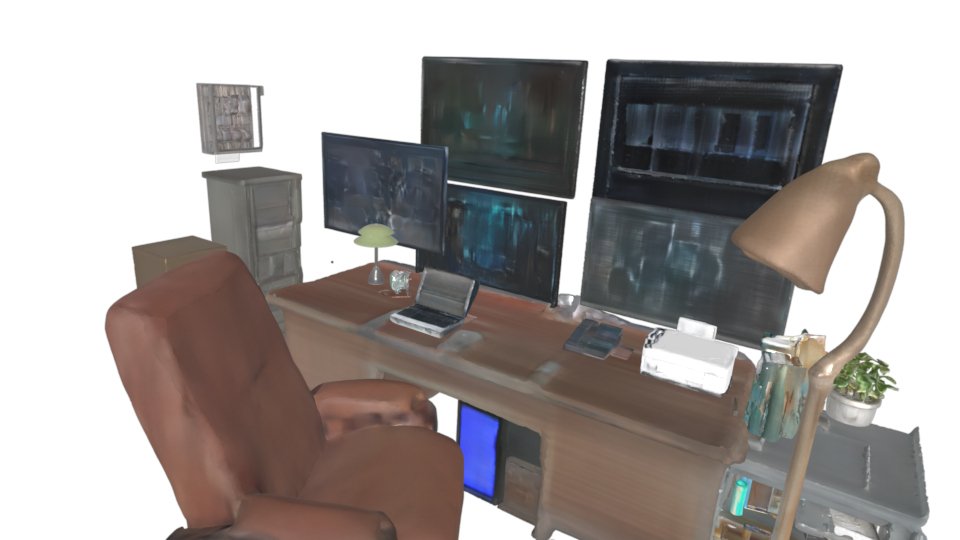} &
        \includegraphics[width=0.245\linewidth]{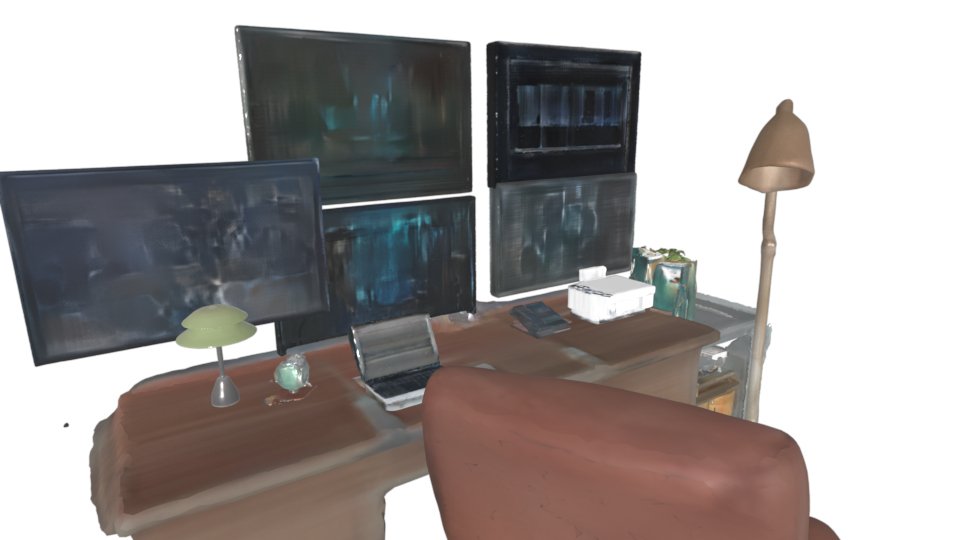} \\[1pt]
        GT View 1 & GT View 2
        & ShapeR View 1 & ShapeR View 2
        & Ours View 1 & Ours View 2
    \end{tabular}%
    }

    \captionof{figure}{\textbf{More qualitative 3D scene reconstruction from inferred perception.} }
    \label{fig:recon_cmp_inferred_perception_supp}
\end{table*}

\pagebreak
\paragraph{Single-image Reconstruction} We showcase more single-image reconstruction visualizations in Fig. \ref{fig:single_image_recon_cmp_more}, which is evaluated in the 3D-Front \cite{fu20213d} dataset. \model{} can achieve better geometry performance against SoTA \textit{automatic} single-image instance scene reconstruction methods, including Gen3DSR \cite{gen3dsr}, MIDI \cite{midi}, and SceneGen \cite{meng2025scenegen}. \rev{Qualitatively, Fig.~\ref{fig:single_image_recon_cmp_more} shows coherent object geometry and scene layouts.} \rev{Single-image evaluation uses 20 test images, with geometry metrics computed after aligning the entire predicted scene to the reference using a single similarity transform (uniform scaling, rotation, and translation).}
\begin{table*}[ht]
    \centering
    \setlength{\tabcolsep}{1pt}
    \renewcommand{\arraystretch}{0.5}
    \resizebox{1\textwidth}{!}{%
    \begin{tabular}{ccccc}

        \includegraphics[width=0.195\linewidth]{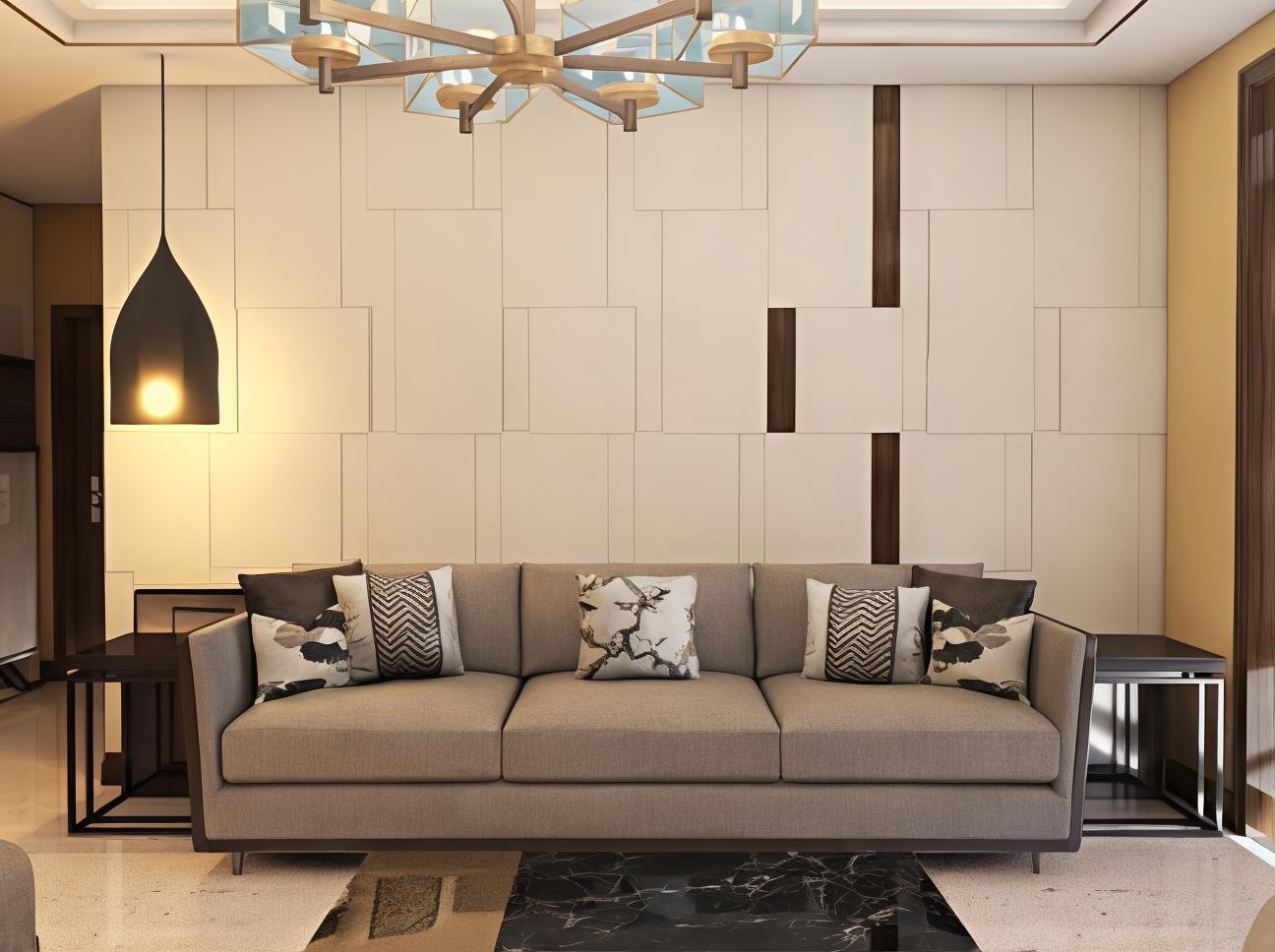} &
        \includegraphics[width=0.195\linewidth]{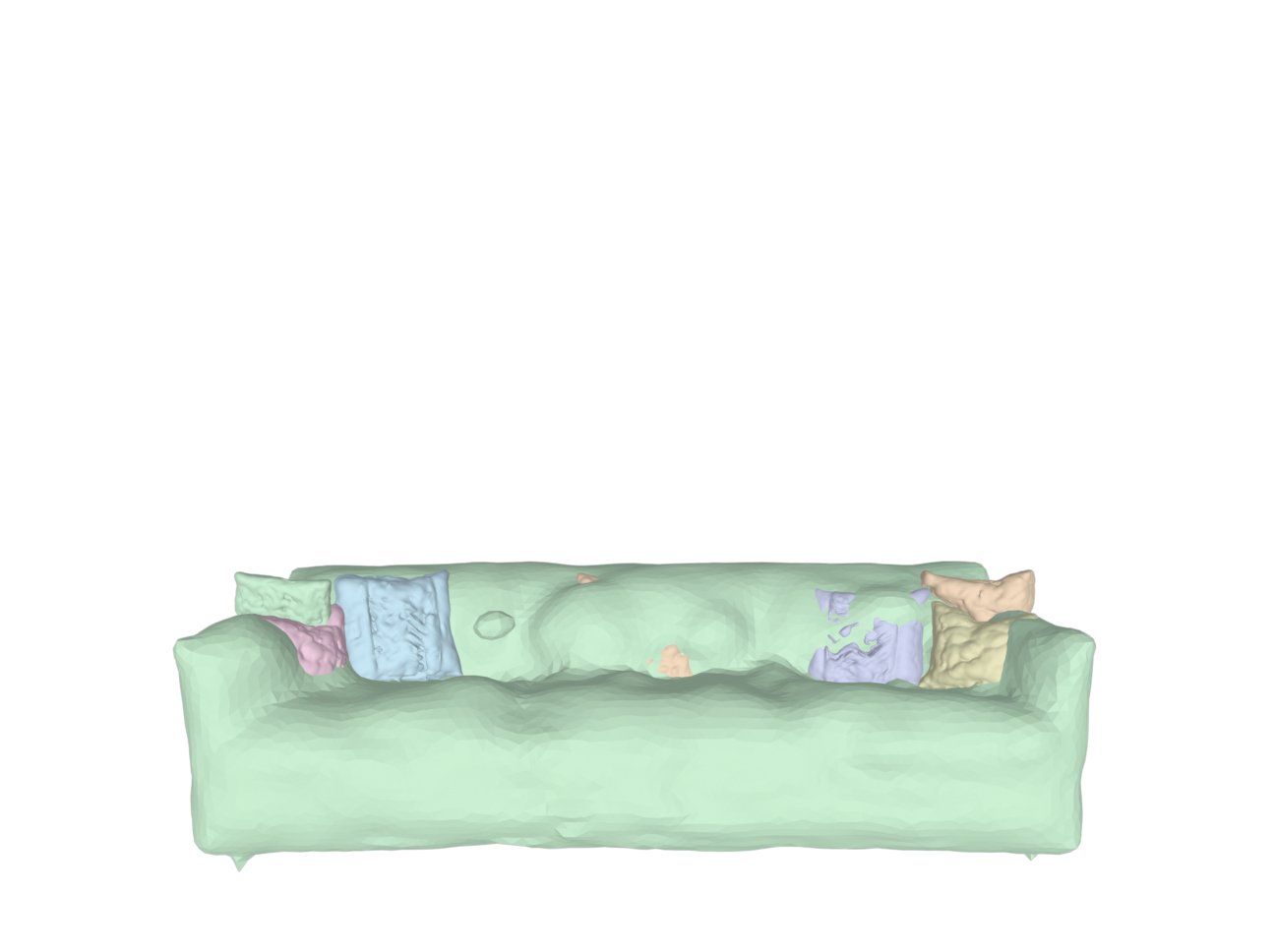} &
        \includegraphics[width=0.195\linewidth]{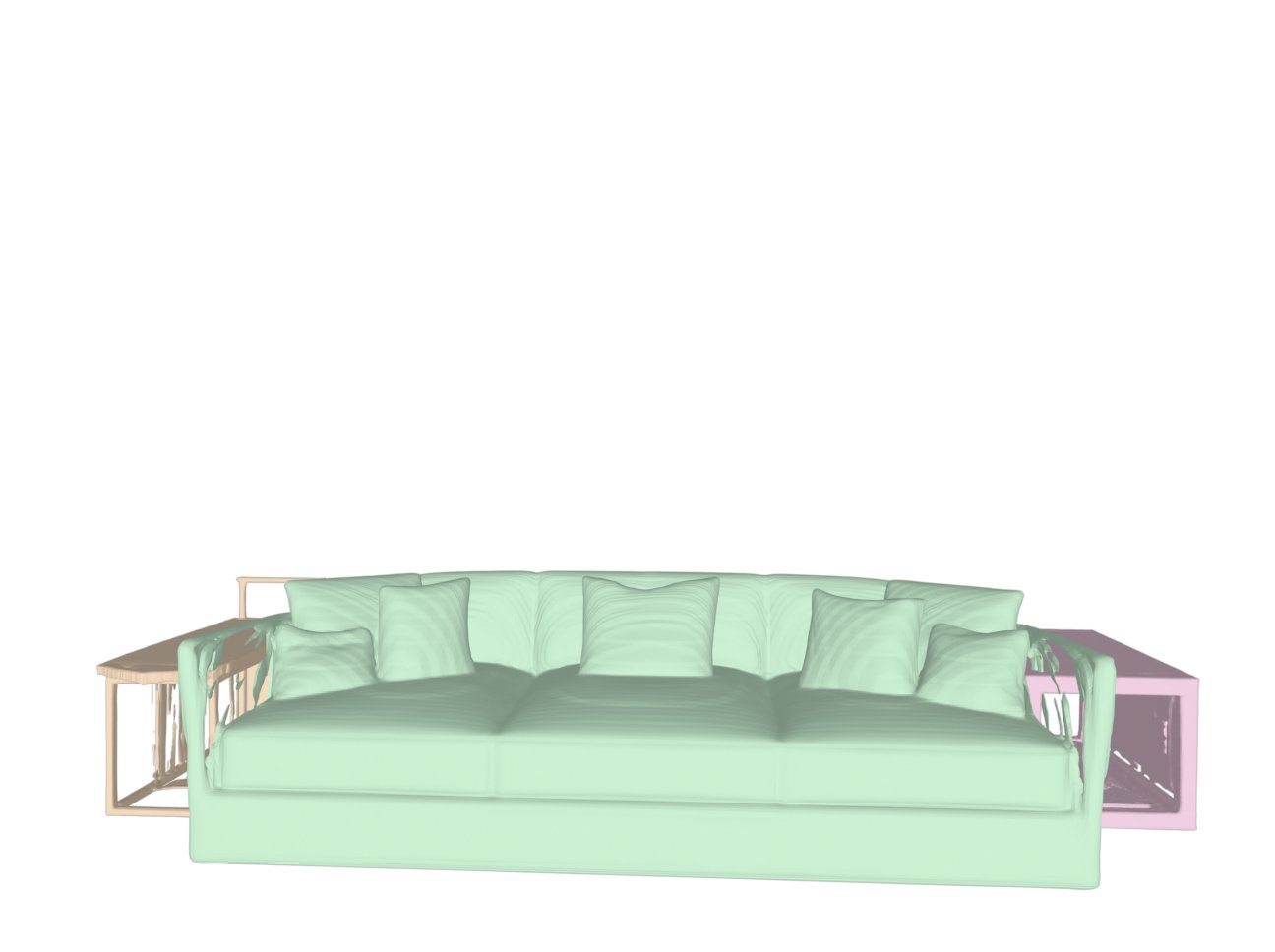} &
        \includegraphics[width=0.195\linewidth]{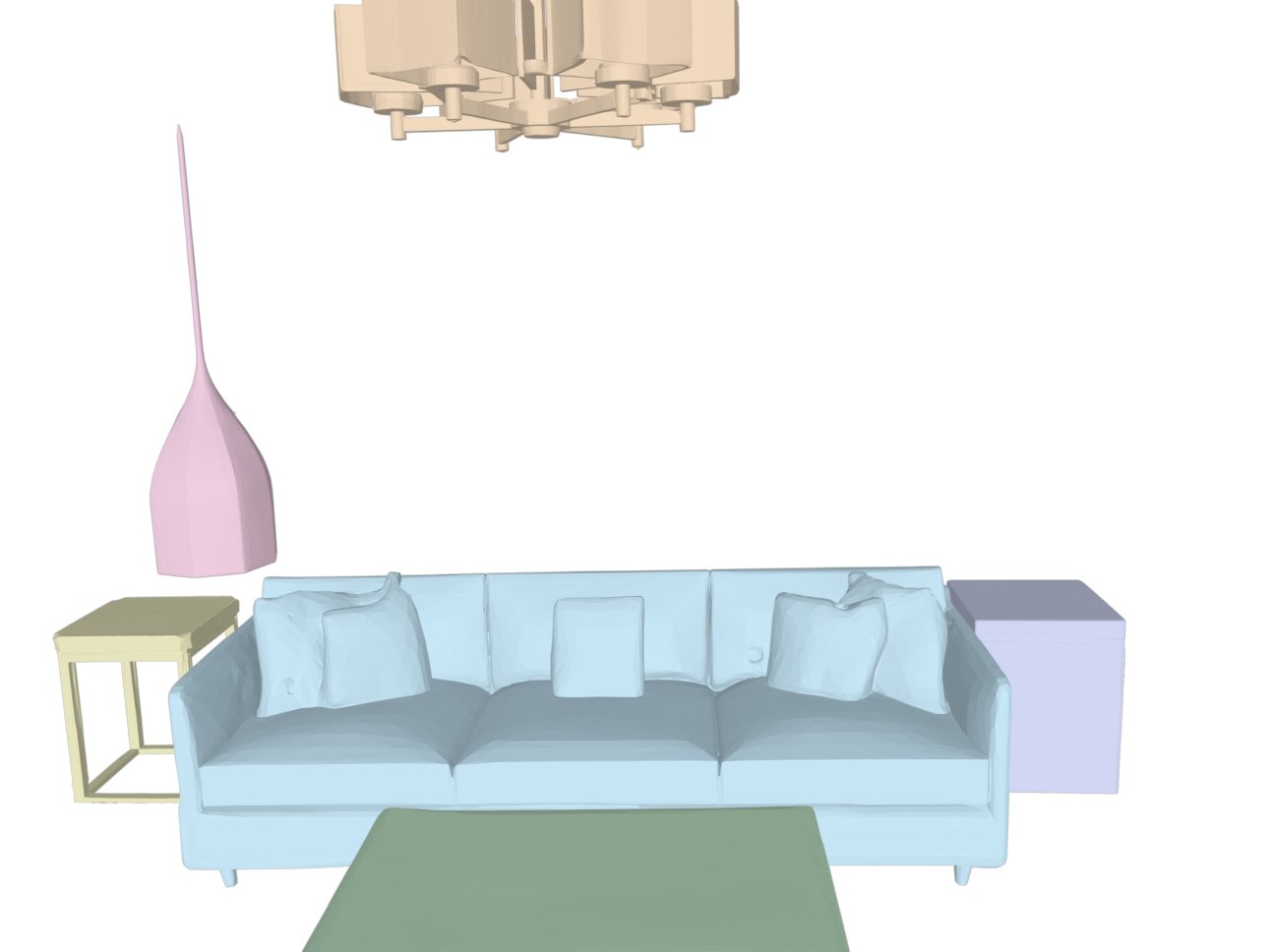} &
        \includegraphics[width=0.195\linewidth]{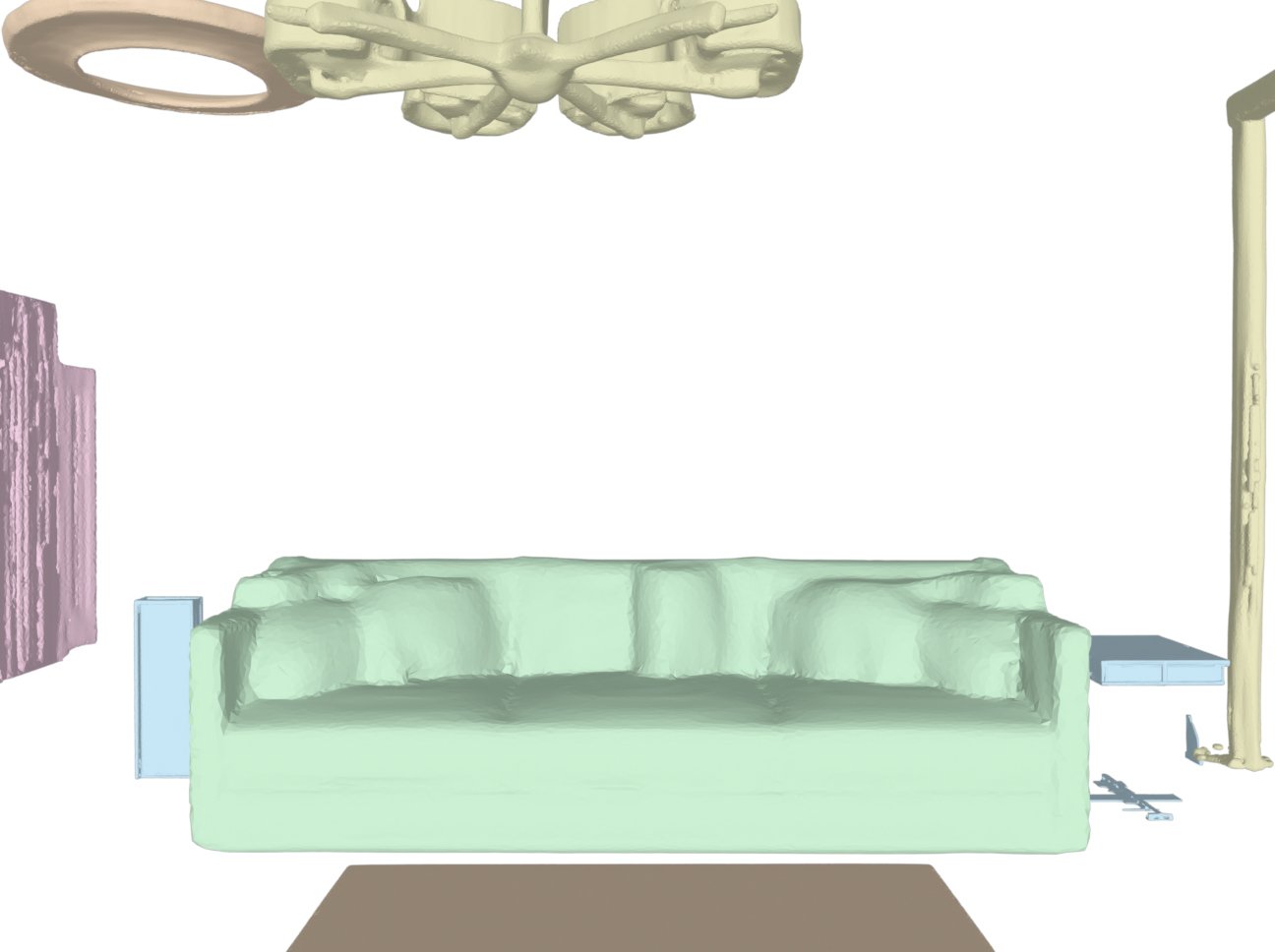} \\[2pt]

        \includegraphics[width=0.195\linewidth]{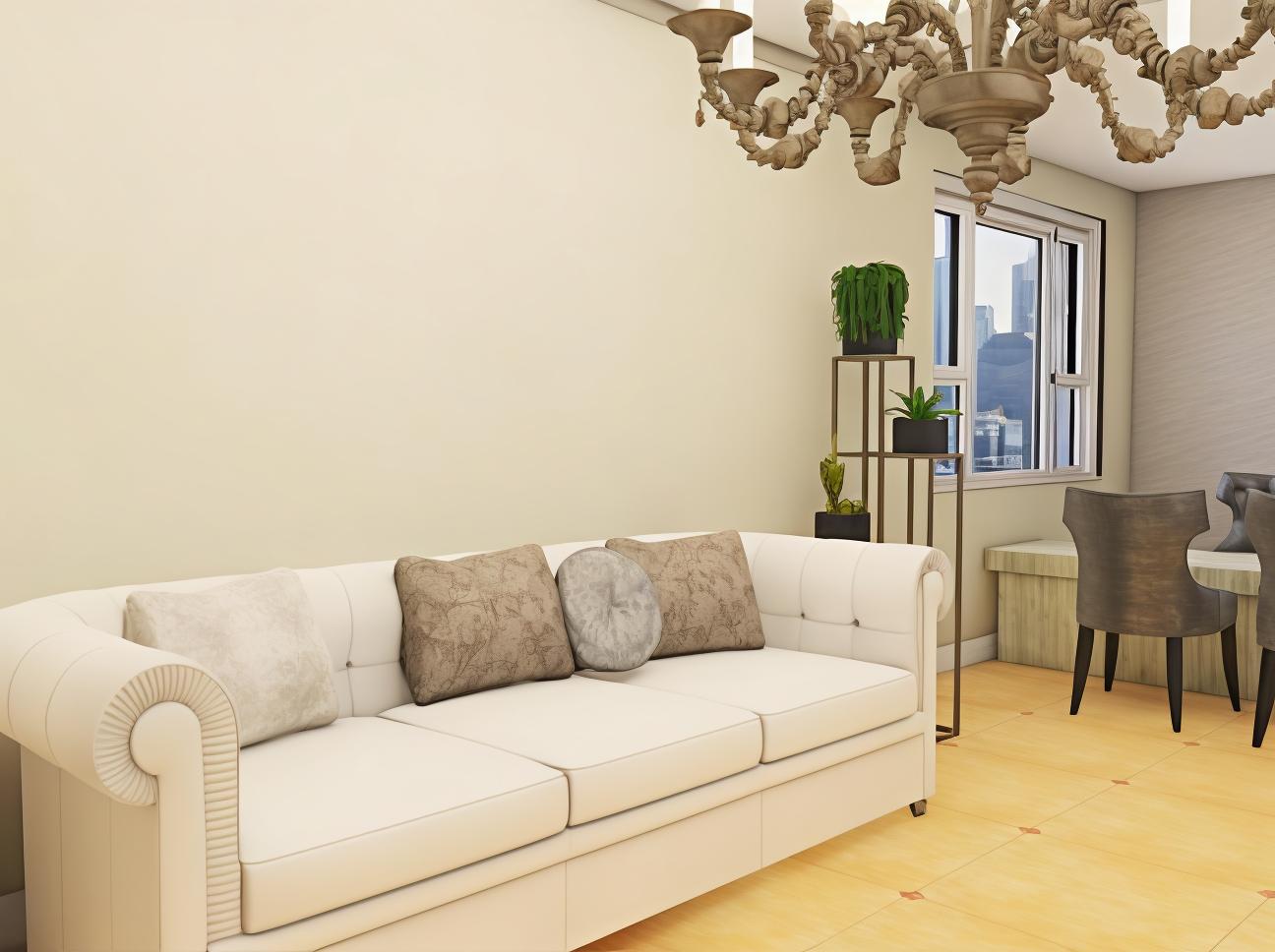} &
        \includegraphics[width=0.195\linewidth]{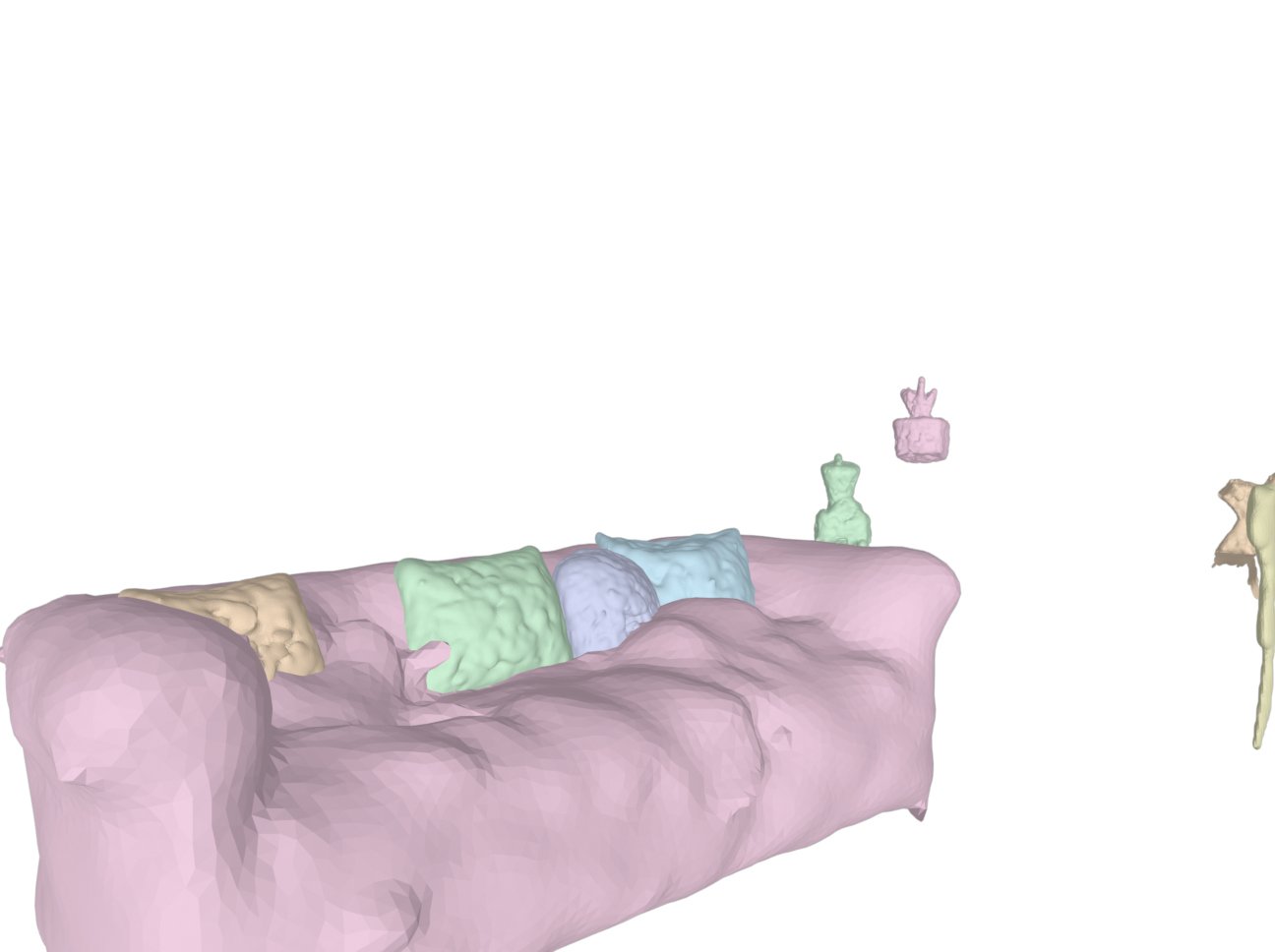} &
        \includegraphics[width=0.195\linewidth]{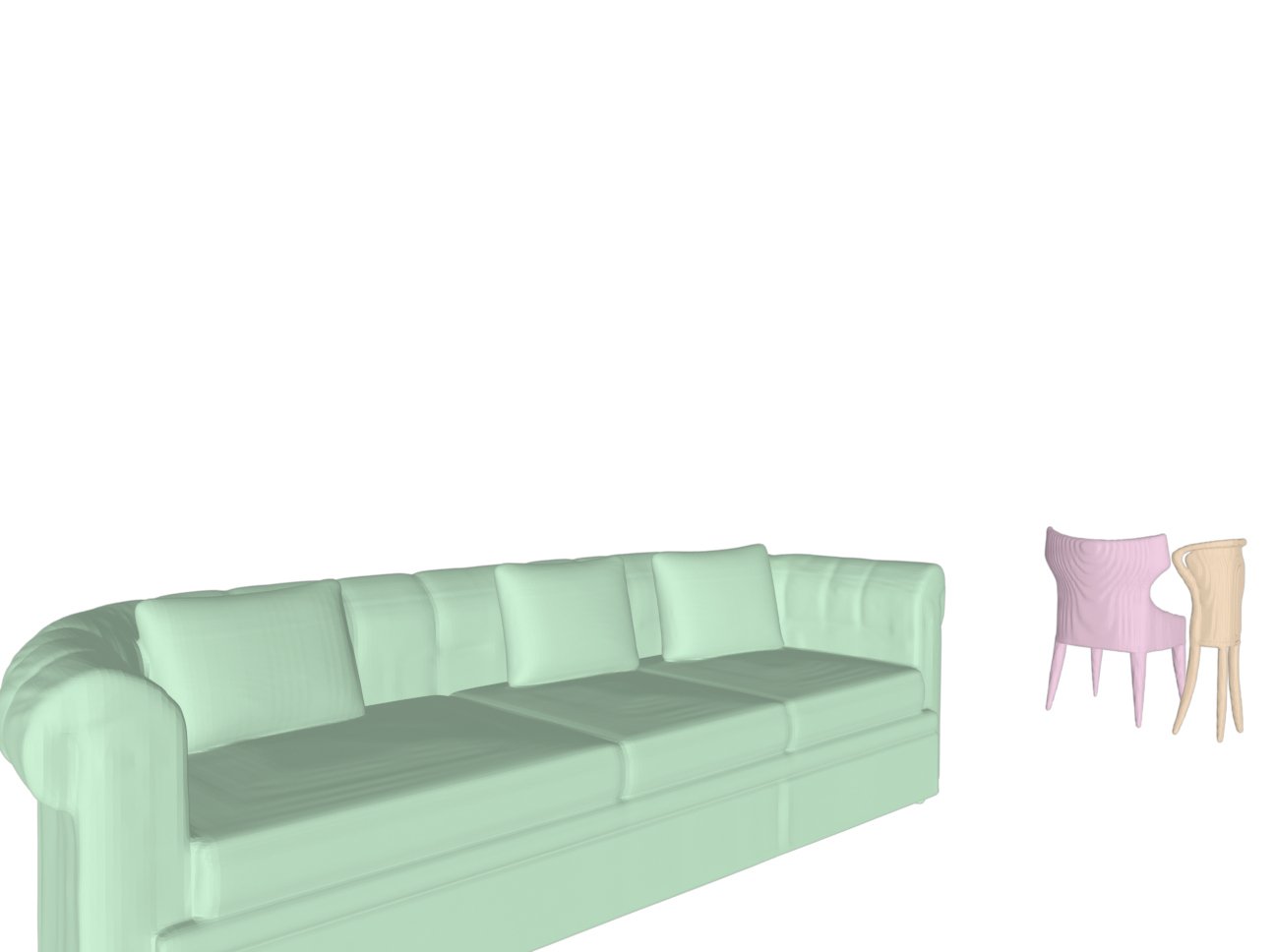} &
        \includegraphics[width=0.195\linewidth]{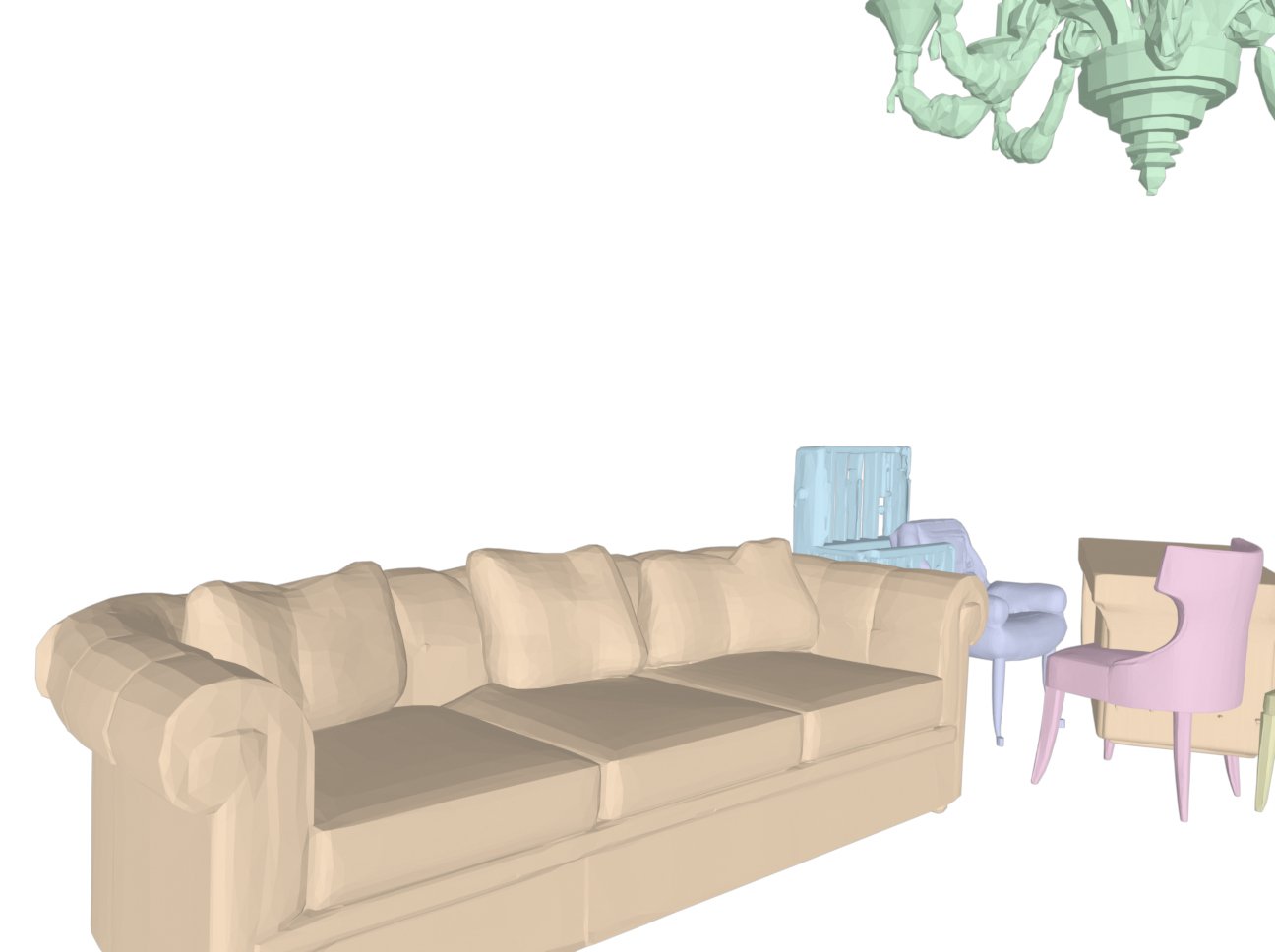} &
        \includegraphics[width=0.195\linewidth]{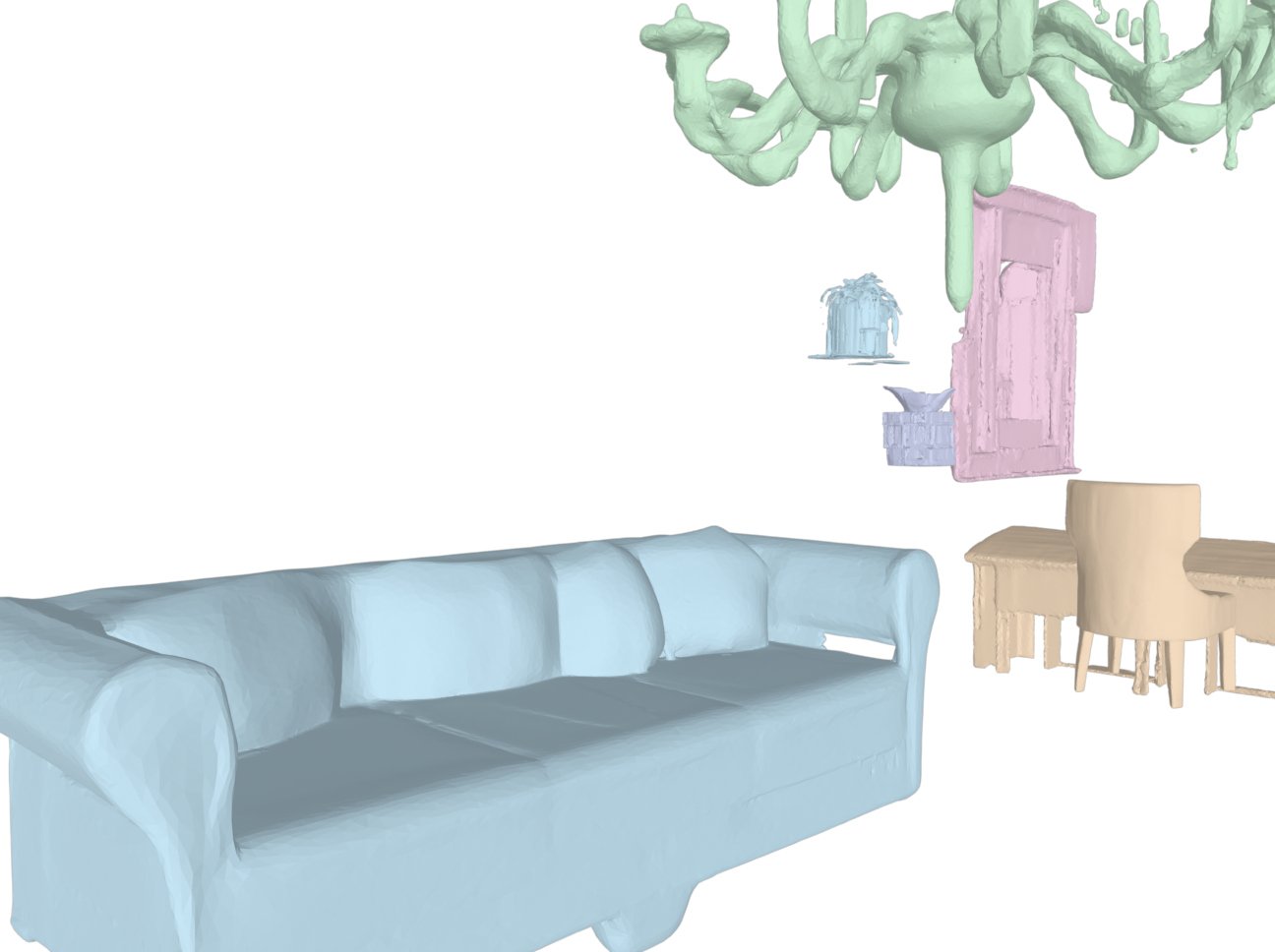} \\[2pt]

        Input & Gen3DSR \cite{gen3dsr} & MIDI \cite{midi} & SceneGen \cite{meng2025scenegen} & Ours
    \end{tabular}
    }
    \captionof{figure}{\textbf{More qualitative results of single-image 3D scene reconstruction.} \rev{\model{} produces coherent geometry in these examples.} We exclude BG for matched comparisons.}
    \label{fig:single_image_recon_cmp_more}
\end{table*}

\WrapNeedspace{93pt}
\paragraph{Ablation Study of \rebut{Input Pose and Depth Sensitivity}}
{
\begin{wraptable}{r}{0.48\linewidth}
\vspace{-1em}
\small
\centering
\resizebox{\linewidth}{!}{
\begin{tabular}{lcccc}
\toprule
\textbf{Setting} & \textbf{mAP}$\uparrow$ & \textbf{mIoU}$\uparrow$ & \textbf{CD} \revten{(cm)} $\downarrow$ & \textbf{PSNR} \revten{(dB)} $\uparrow$ \\
\midrule
GT pose + GT depth & \textbf{0.54} & \textbf{0.44} & 7.98 & 15.93 \\
COLMAP pose + \rev{Pi3X} depth & 0.53 & 0.42 & \textbf{7.23} & \textbf{16.29} \\
\rev{Pi3X} pose + \rev{Pi3X} depth & 0.46 & 0.37 & 8.99 & 15.83 \\
\bottomrule
\end{tabular}
}
\caption{\rev{\textbf{Input pose and depth sensitivity.} Effects vary across metrics and preprocessing configurations.}}
\label{tab:ablation}
\vspace{-1em}
\end{wraptable}
\rev{This ablation uses a total of 20 randomly selected iTHOR and Imaginarium scenes.} To study how preprocessed camera poses and depth will affect the model inference performance, we replace GT posed RGB-D observations with perturbations involving poses from COLMAP \cite{schoenberger2016sfm} and \rev{Pi3X} \cite{wang2026pi}.
\rev{Pi3X poses/depth reduce mAP from 0.54 to 0.46 and increase CD from 7.98 to 8.99 cm, while PSNR remains similar (Tab.~\ref{tab:ablation}).} 
We also validated that batched and sequential execution produce \rev{matching outputs in the tested configuration}, while batching \rev{provides a speedup of} \textbf{more than $10\times$}.\par
}

\paragraph{\rebut{Ablation Study of the Effect of Validity Threshold}}
\rebut{
We want to evaluate how the validity threshold in \model{} affects our perception performance.
On 5 iTHOR and 5 Imaginarium test scenes, we vary only the validity threshold $\tau_v$ and evaluate with the protocol of Tab.~\ref{tab:perception}. 
We found that the validity filter removes very few objects. 
As Tab.~\ref{tab:validity_sweep} shows, removing the filter raises recall by less than 0.01 over the default $\tau_v=0.2$, and results stay flat for $\tau_v$ from 0.05 to 0.4. 
Tab.~\ref{tab:recall_by_size} further shows that recall is high for large objects, at 0.71 on iTHOR and 0.80 on Imaginarium for the largest third, and the remaining misses come mainly from small objects.
}

\begin{table}[ht]
\centering
\small
\begin{minipage}[t]{0.53\linewidth}
\centering
\resizebox{\linewidth}{!}{%
\begin{tabular}{lcccc}
\toprule
$\tau_v$ & iTHOR mAP & iTHOR recall & Imag. mAP & Imag. recall \\
\midrule
None & 0.433 & 0.415 & 0.552 & 0.552 \\
0.05 & 0.425 & 0.408 & 0.545 & 0.540 \\
0.2 (default) & 0.423 & 0.408 & 0.548 & 0.544 \\
0.4 & 0.423 & 0.408 & 0.548 & 0.544 \\
0.7 & 0.400 & 0.382 & 0.483 & 0.468 \\
0.9 & 0.295 & 0.283 & 0.353 & 0.323 \\
\bottomrule
\end{tabular}}
\caption{\textbf{Quantitative results on the validity threshold.} Removing the filter raises recall by less than 0.01 over the default $\tau_v=0.2$.}
\label{tab:validity_sweep}
\end{minipage}\hfill
\begin{minipage}[t]{0.43\linewidth}
\centering
\resizebox{\linewidth}{!}{%
\begin{tabular}{lccc}
\toprule
Dataset & Smallest third & Middle third & Largest third \\
\midrule
iTHOR & 0.125 & 0.376 & 0.714 \\
Imaginarium & 0.241 & 0.598 & 0.795 \\
\bottomrule
\end{tabular}}
\caption{\textbf{Quantitative results on recall by GT object size.} Recall is highest for large objects, and most remaining misses are small objects.}
\label{tab:recall_by_size}
\end{minipage}
\end{table}

\paragraph{\rebut{Ablation Study of the Robustness of Generator}}
\rebut{
Here we quantitatively study the robustness \model{}'s generator.
We evaluate the generator with various intensities of noises and missing points to heavily test the robustness.
To isolate the generator, we use GT poses and instances on 3 iTHOR and 3 Imaginarium scenes and perturb each object's point cloud, either adding Gaussian noise to the point positions or removing a contiguous part of each object with a random plane cut. 
In Tab.~\ref{tab:robustness}, we find that the generator stays robust to moderate noise and missing points. 
With 1\,cm of noise, CD changes by at most 0.1\,cm, and even with 2\,cm of noise or half of the points removed, CD stays within 1.0\,cm of the clean input. 
Fig.~\ref{fig:robust_qual} shows qualitative results.
}

\begin{table}[ht]
\centering
\small
\begin{tabular}{lcccccc}
\toprule
& \multicolumn{3}{c}{\textbf{iTHOR}} & \multicolumn{3}{c}{\textbf{Imaginarium}} \\
\cmidrule(lr){2-4}\cmidrule(lr){5-7}
\textbf{Condition} & CD (cm) $\downarrow$ & F1 $\uparrow$ & NC $\uparrow$ & CD (cm) $\downarrow$ & F1 $\uparrow$ & NC $\uparrow$ \\
\midrule
Clean & 1.18 & 0.690 & 0.790 & 1.26 & 0.666 & 0.790 \\
Noise, 1 cm & 1.26 & 0.621 & 0.749 & 1.36 & 0.609 & 0.745 \\
Noise, 2 cm & 1.74 & 0.438 & 0.674 & 1.49 & 0.463 & 0.676 \\
Noise, 5 cm & 3.17 & 0.248 & 0.562 & 2.65 & 0.279 & 0.569 \\
Keep 75\% of points & 1.64 & 0.665 & 0.764 & 1.40 & 0.630 & 0.767 \\
Keep 50\% of points & 2.16 & 0.615 & 0.747 & 1.74 & 0.561 & 0.757 \\
Keep 25\% of points & 2.88 & 0.536 & 0.711 & 2.34 & 0.478 & 0.714 \\
Keep 10\% of points & 3.42 & 0.453 & 0.677 & 3.27 & 0.377 & 0.665 \\
\bottomrule
\end{tabular}
\caption{\textbf{Quantitative results on generator robustness.} Inputs use GT poses and instances, with noisy or partial instance point clouds.}
\label{tab:robustness}
\end{table}

\begin{table*}[ht]
    \centering
    \setlength{\tabcolsep}{1pt}
    \renewcommand{\arraystretch}{0.5}
    \resizebox{0.92\textwidth}{!}{%
    \begin{tabular}{cccccc}
        & Clean & Noise 1\,cm & Noise 2\,cm & Keep 50\% & Keep 10\% \\
        \raisebox{14px}{\rotatebox{90}{Input}} &
        \includegraphics[width=0.19\linewidth]{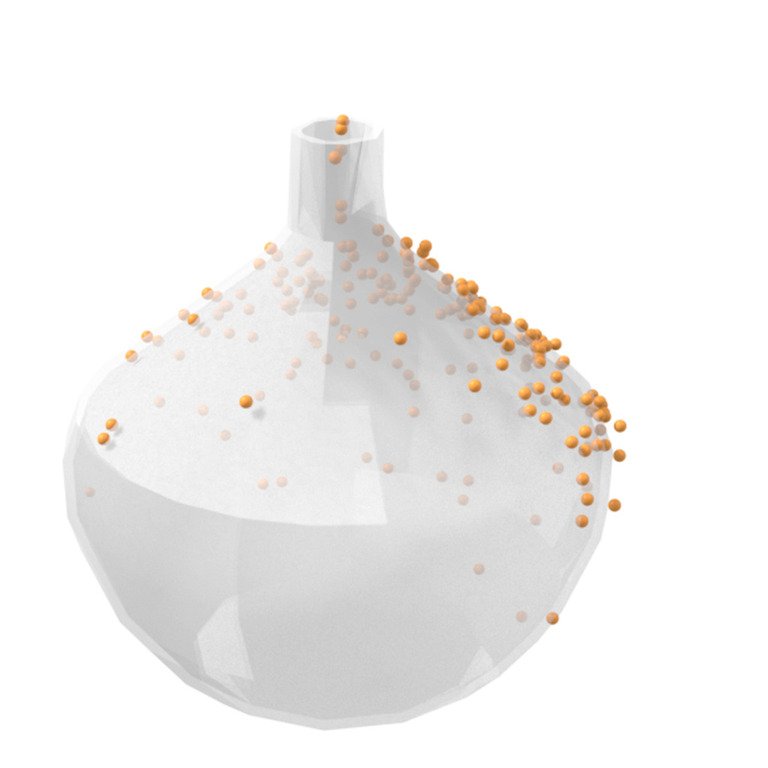} &
        \includegraphics[width=0.19\linewidth]{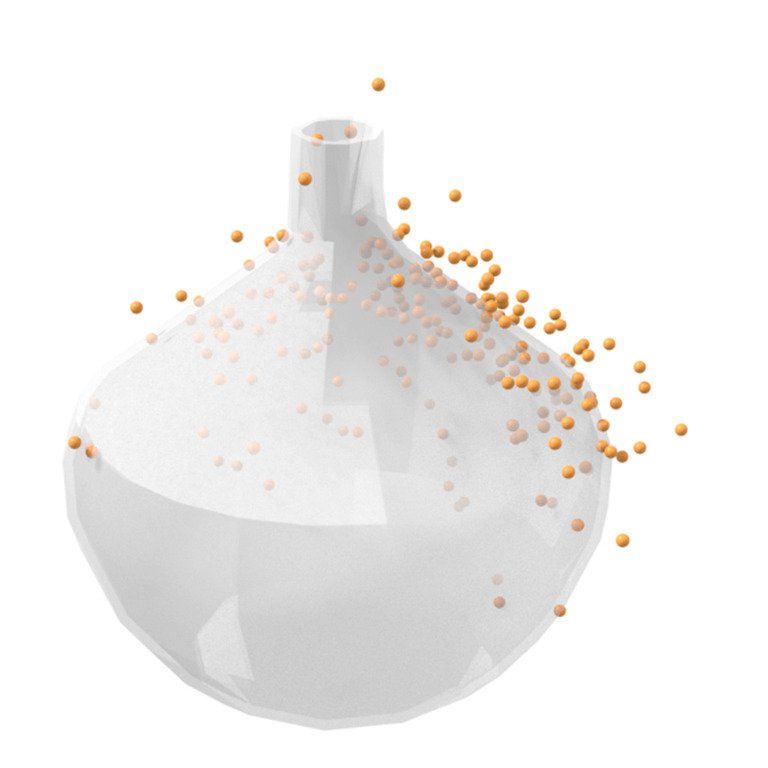} &
        \includegraphics[width=0.19\linewidth]{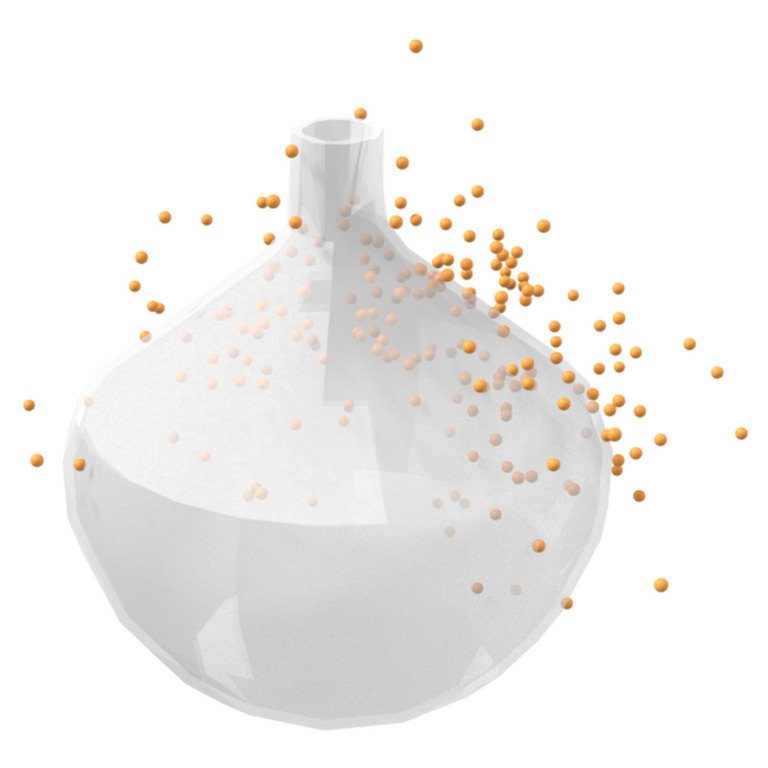} &
        \includegraphics[width=0.19\linewidth]{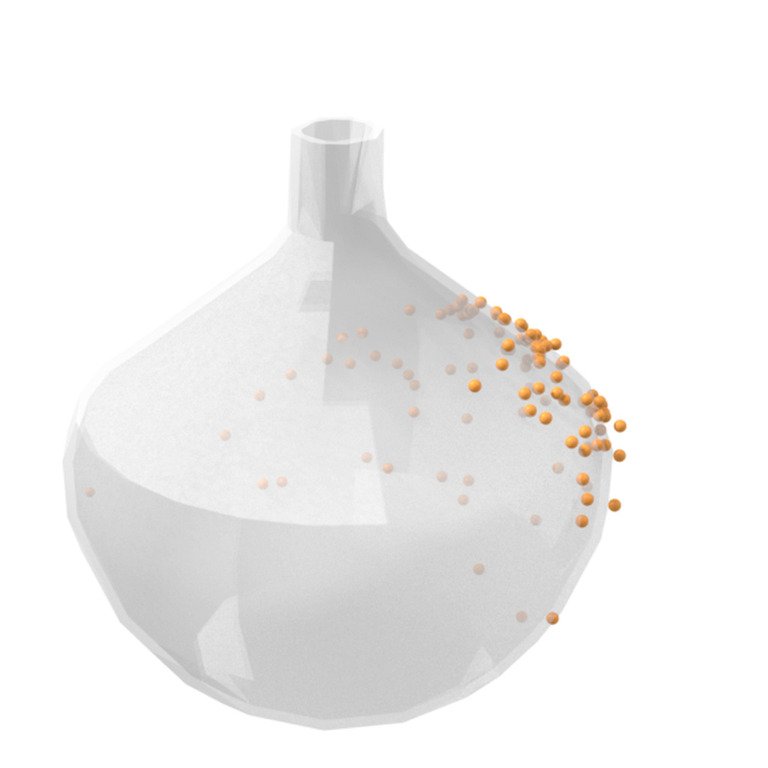} &
        \includegraphics[width=0.19\linewidth]{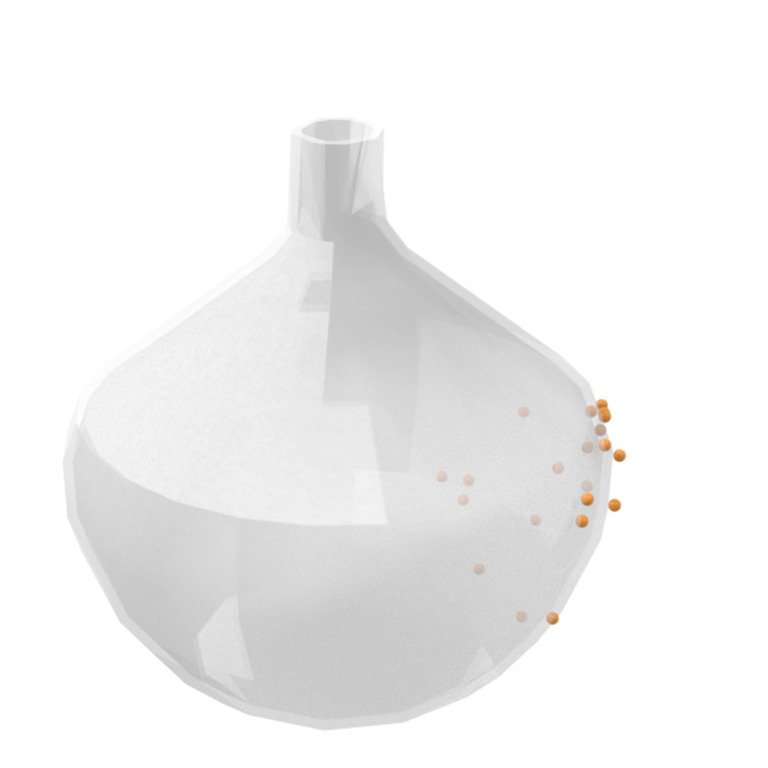} \\[1pt]
        \raisebox{14px}{\rotatebox{90}{Output}} &
        \includegraphics[width=0.19\linewidth]{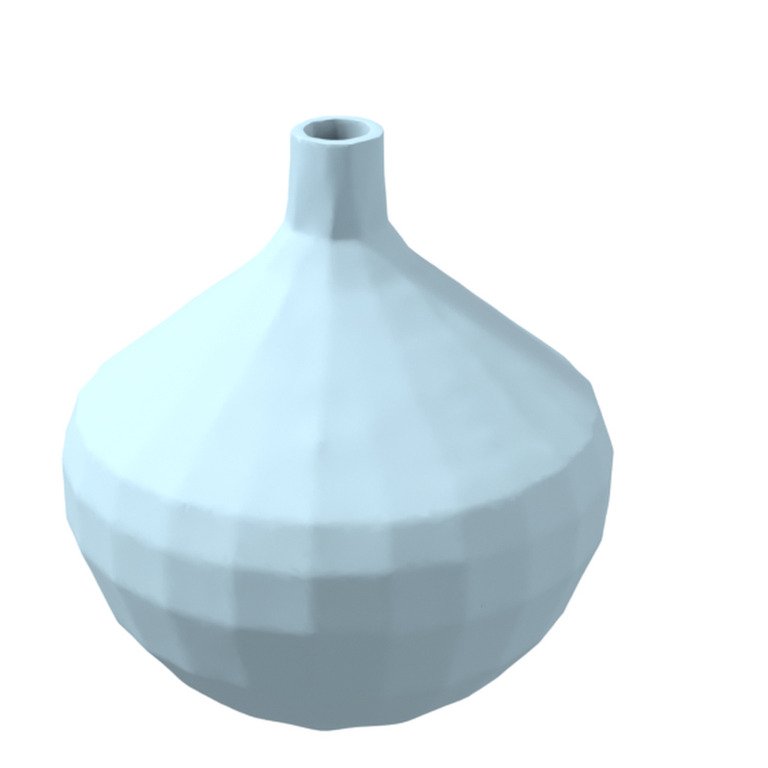} &
        \includegraphics[width=0.19\linewidth]{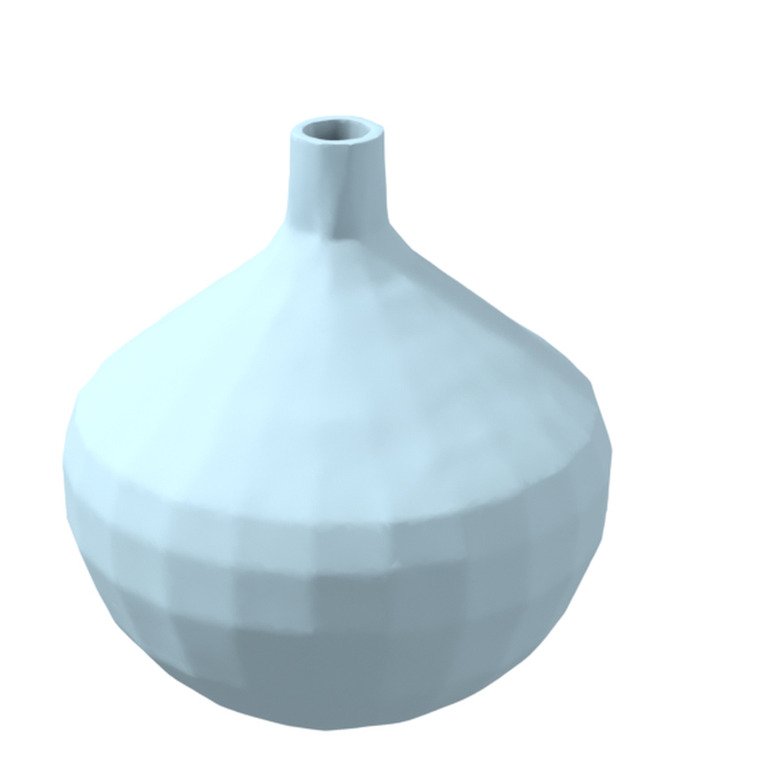} &
        \includegraphics[width=0.19\linewidth]{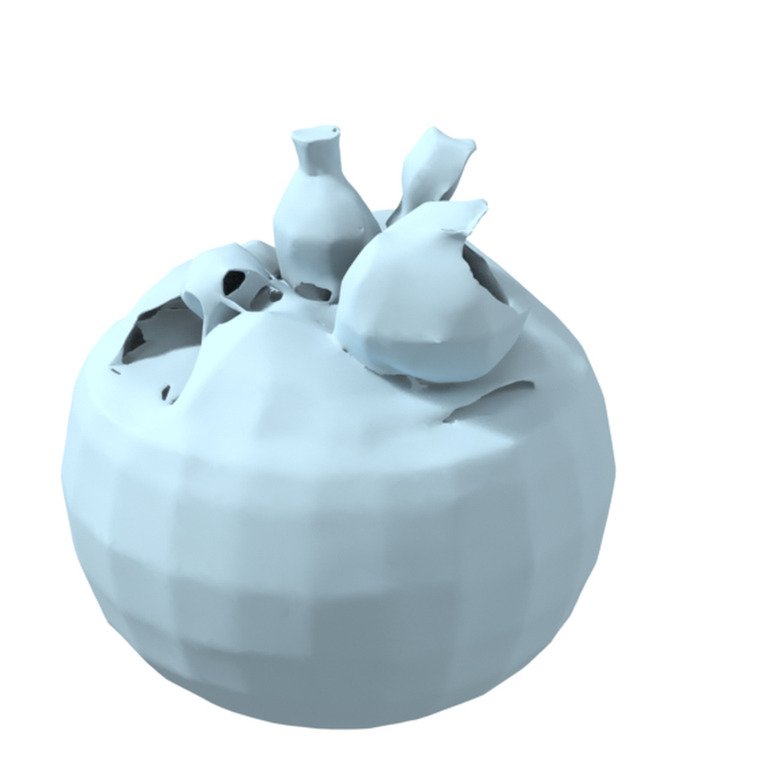} &
        \includegraphics[width=0.19\linewidth]{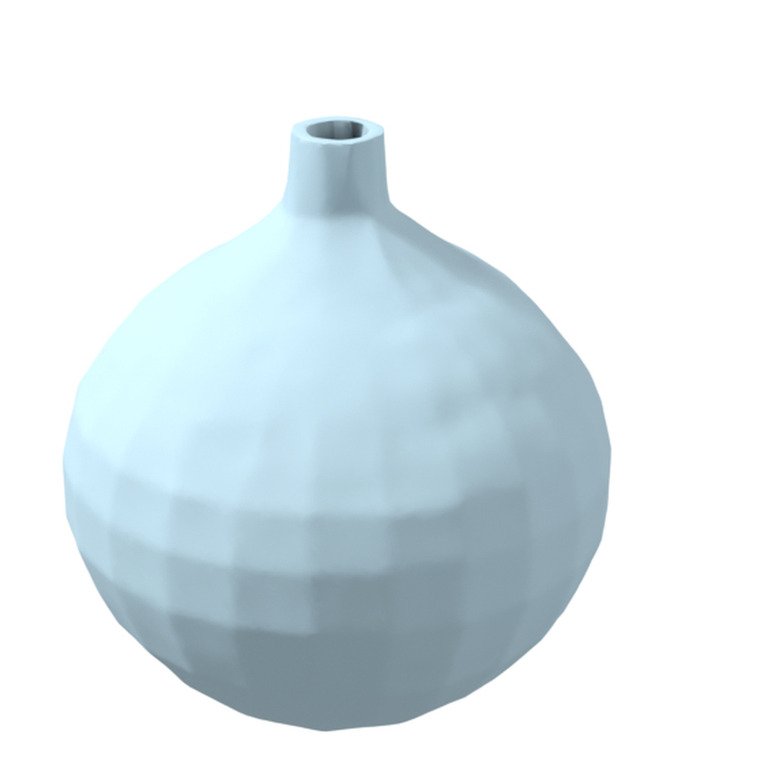} &
        \includegraphics[width=0.19\linewidth]{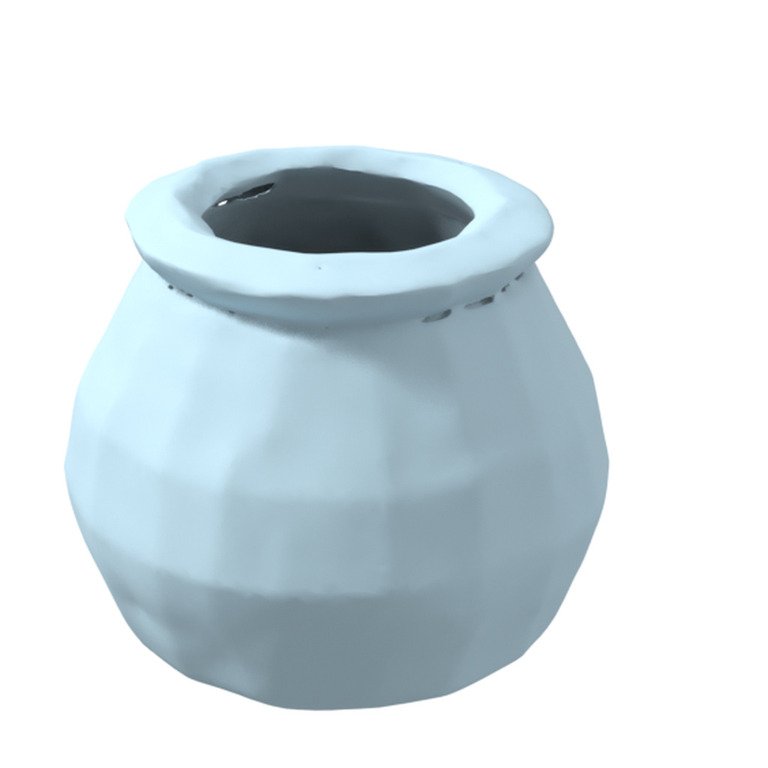} \\[1pt]
        \raisebox{14px}{\rotatebox{90}{Input}} &
        \includegraphics[width=0.19\linewidth]{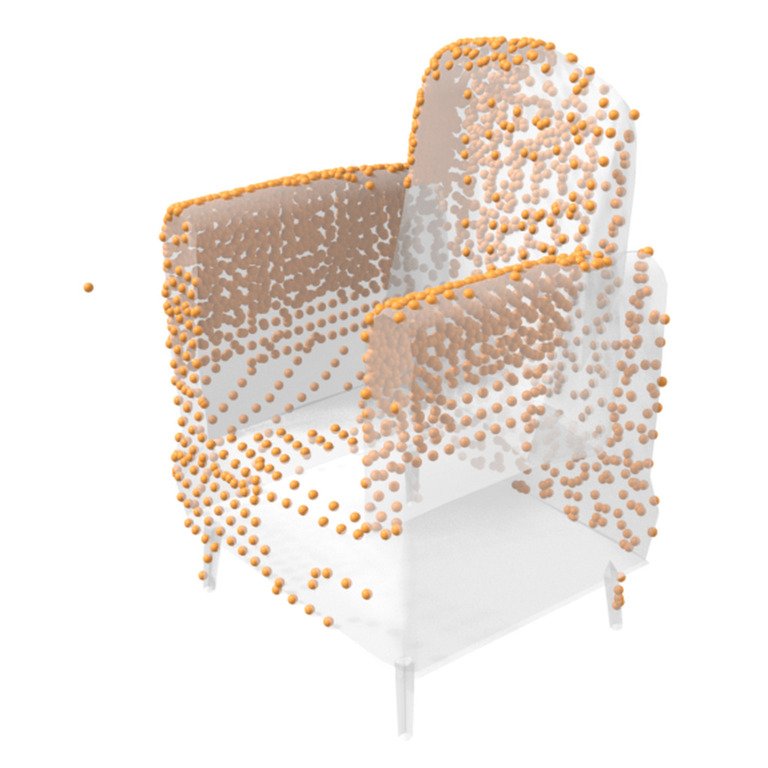} &
        \includegraphics[width=0.19\linewidth]{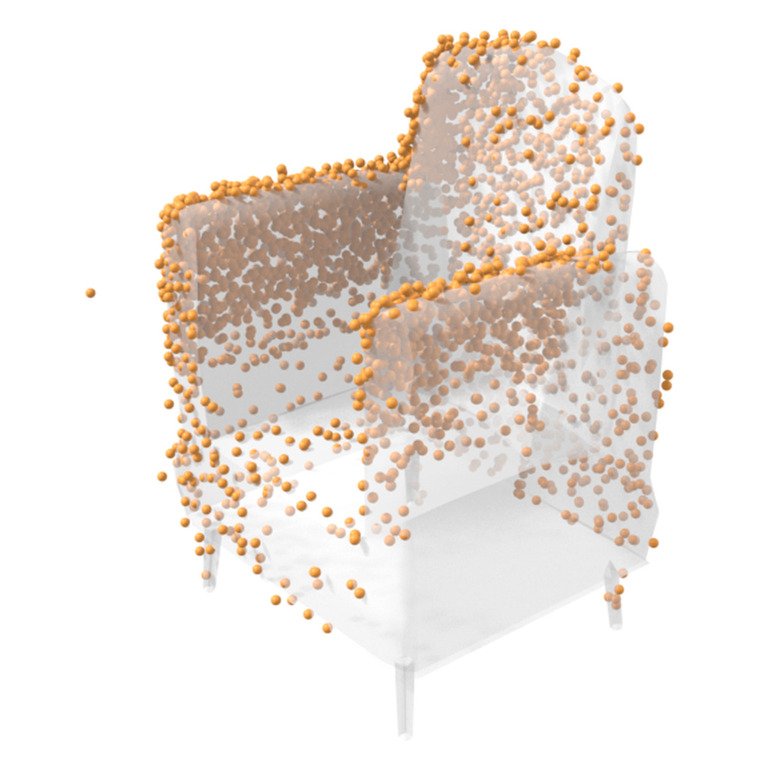} &
        \includegraphics[width=0.19\linewidth]{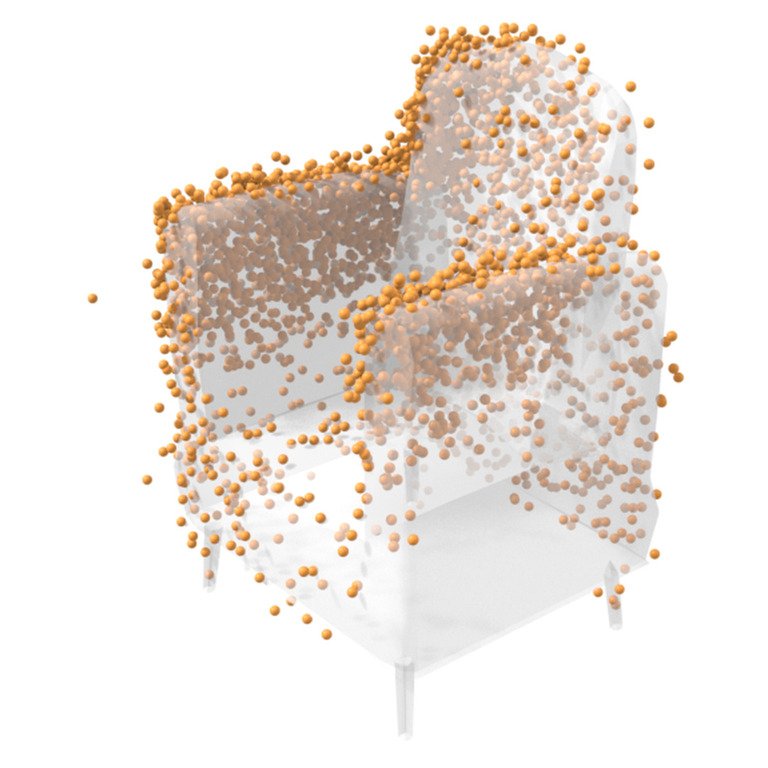} &
        \includegraphics[width=0.19\linewidth]{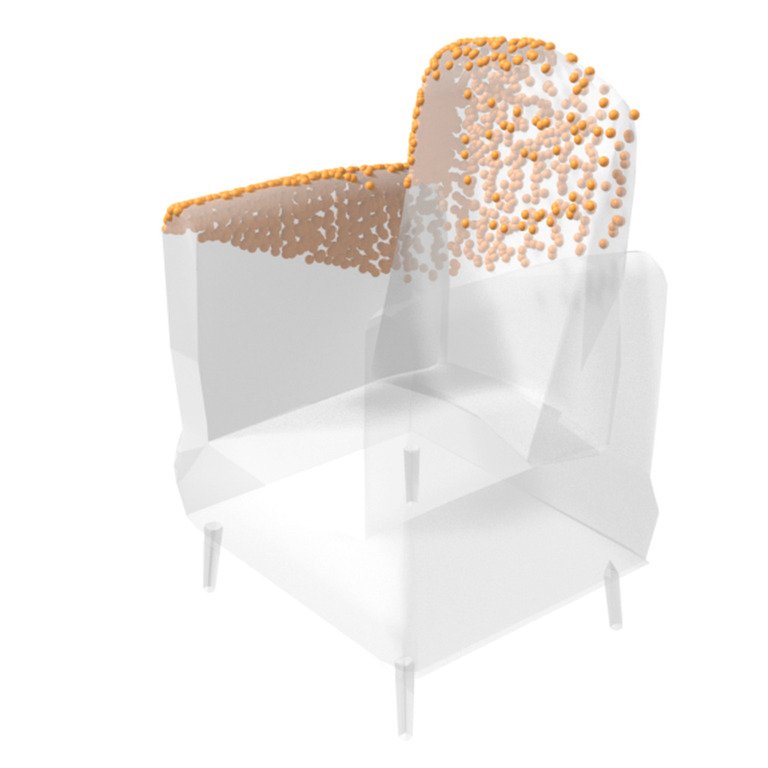} &
        \includegraphics[width=0.19\linewidth]{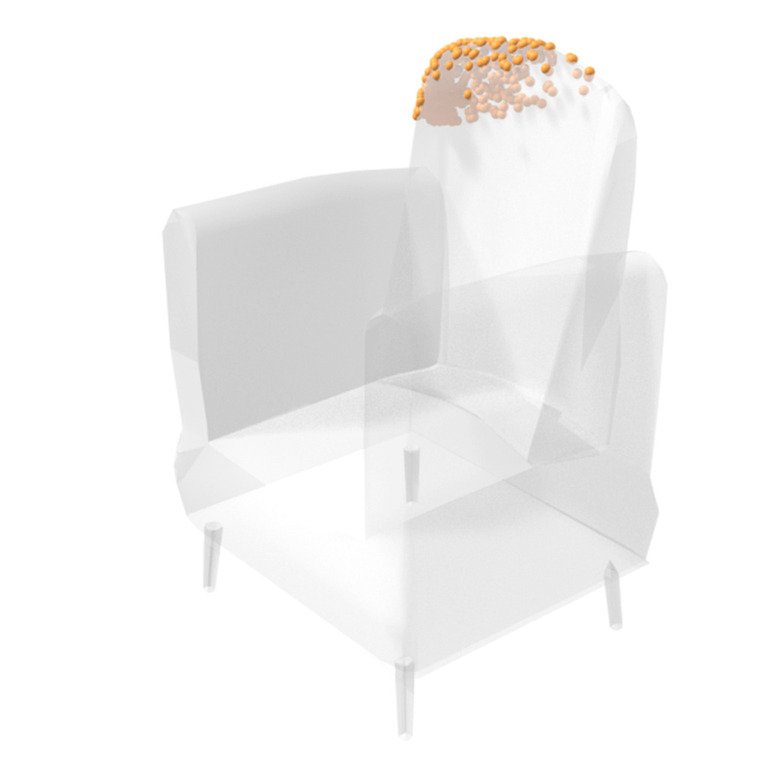} \\[1pt]
        \raisebox{14px}{\rotatebox{90}{Output}} &
        \includegraphics[width=0.19\linewidth]{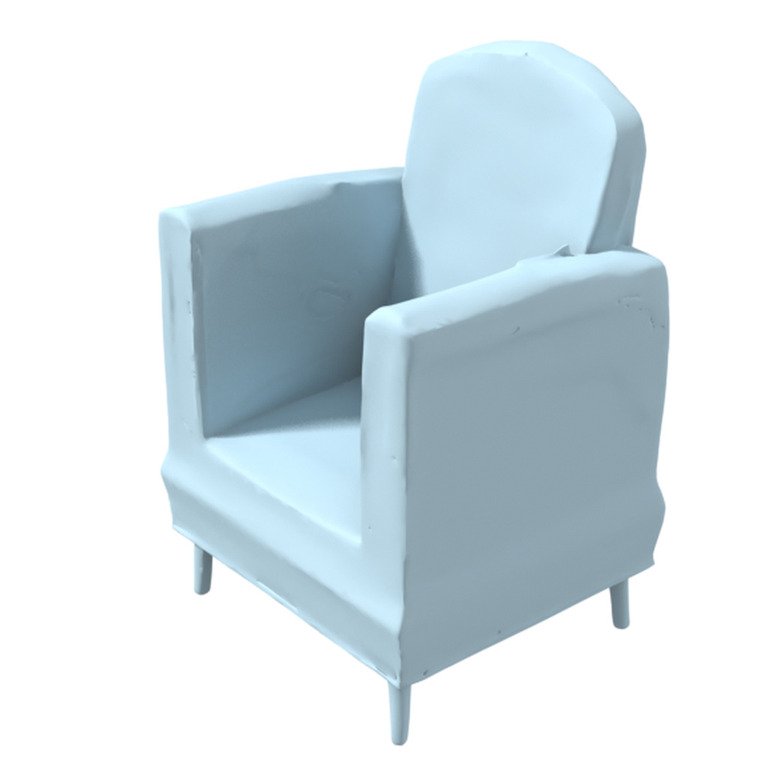} &
        \includegraphics[width=0.19\linewidth]{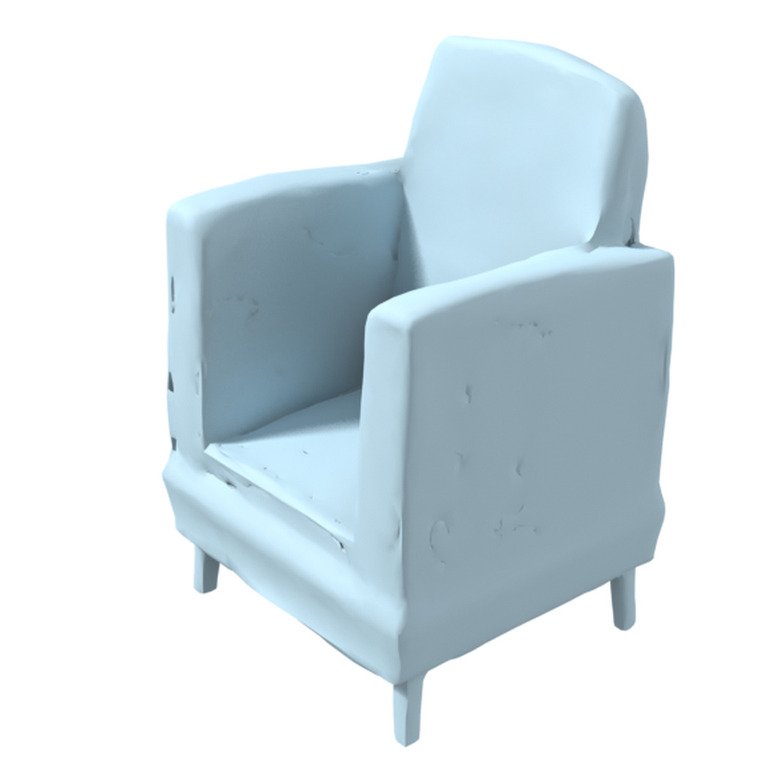} &
        \includegraphics[width=0.19\linewidth]{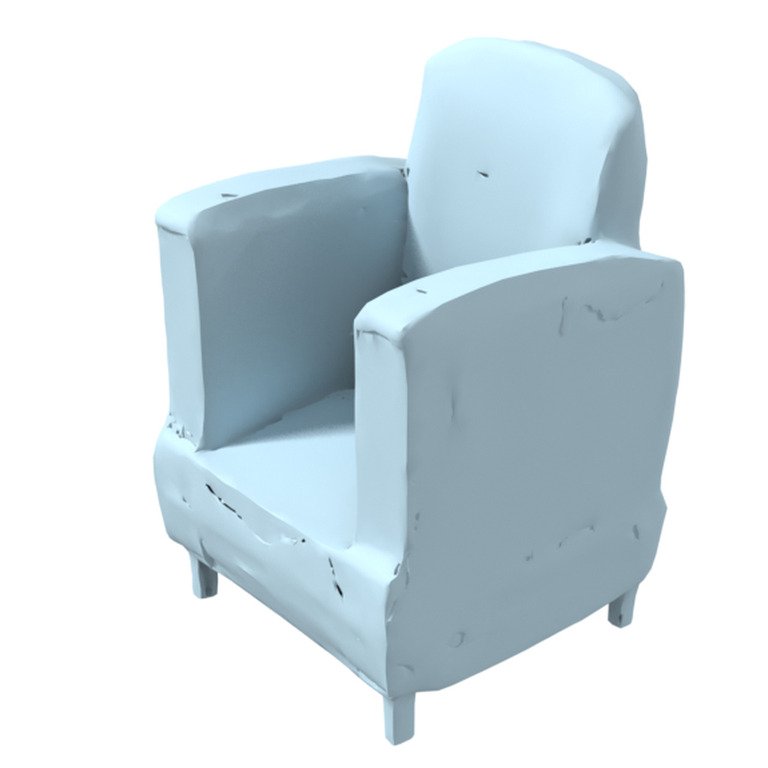} &
        \includegraphics[width=0.19\linewidth]{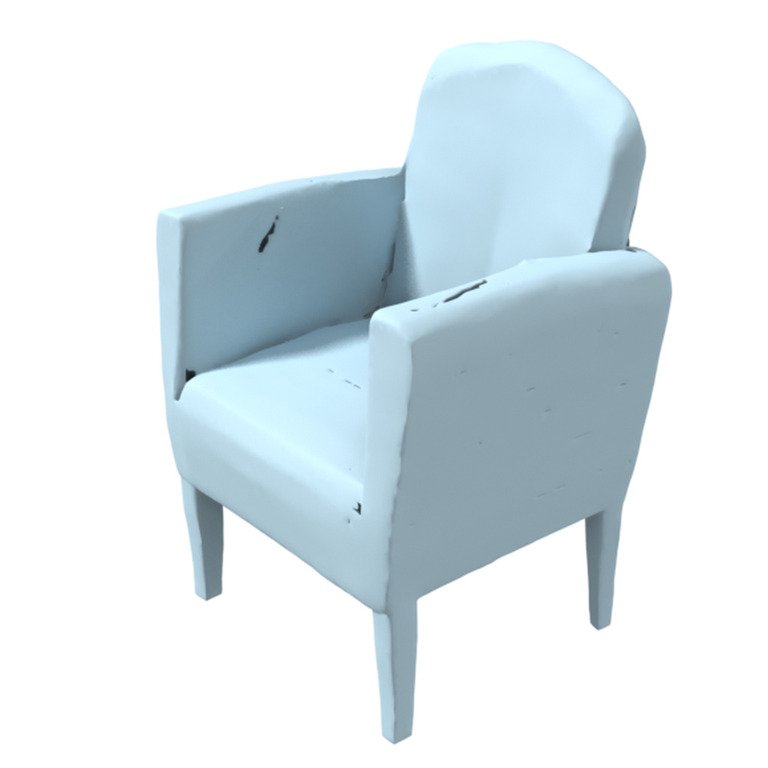} &
        \includegraphics[width=0.19\linewidth]{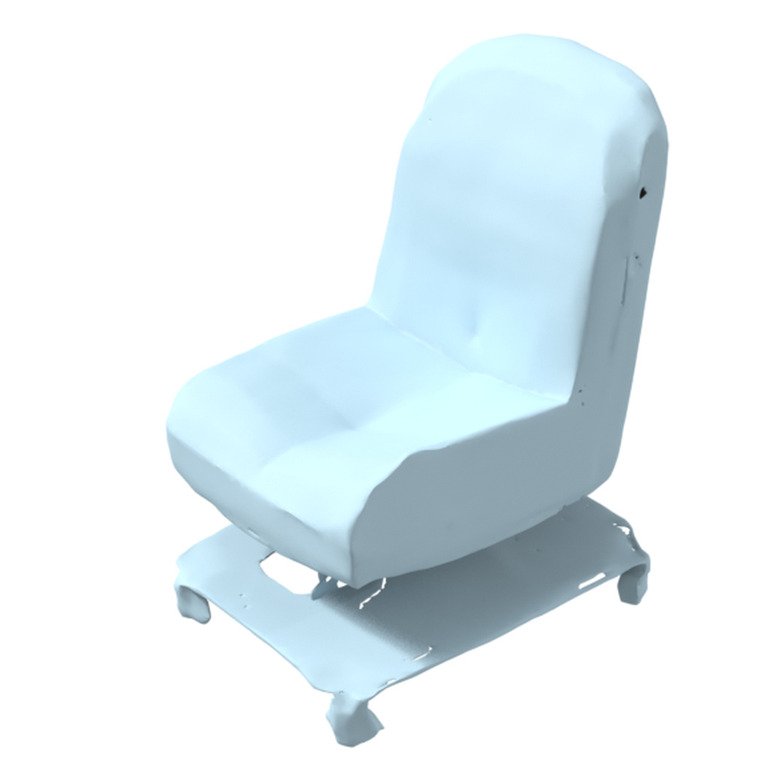} \\[1pt]
        \raisebox{14px}{\rotatebox{90}{Input}} &
        \includegraphics[width=0.19\linewidth]{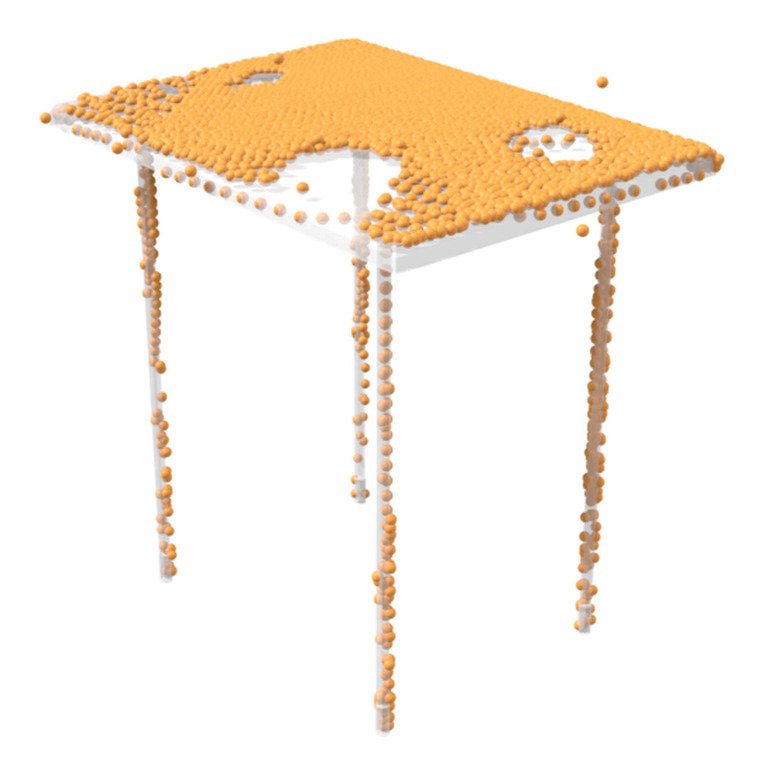} &
        \includegraphics[width=0.19\linewidth]{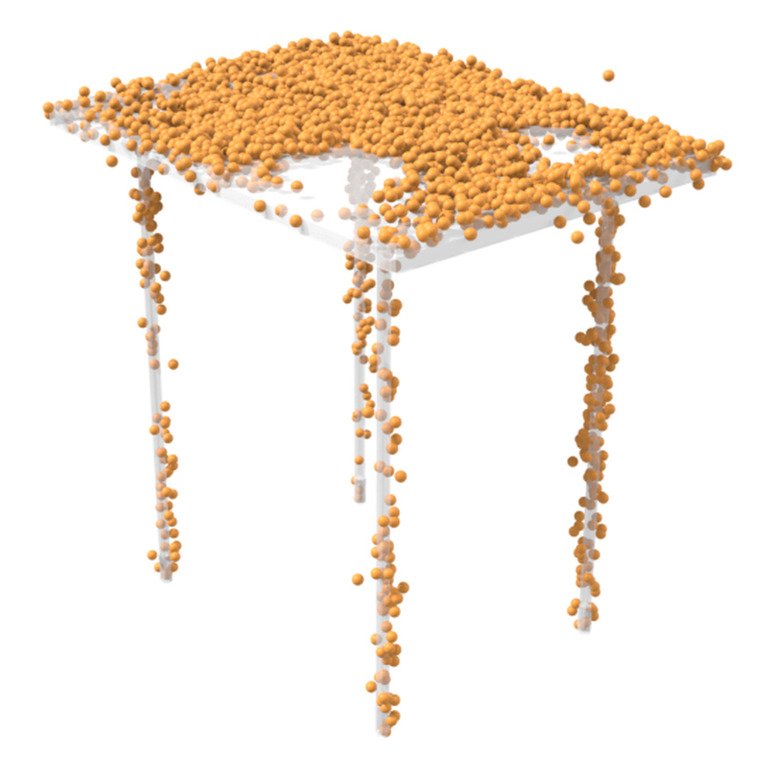} &
        \includegraphics[width=0.19\linewidth]{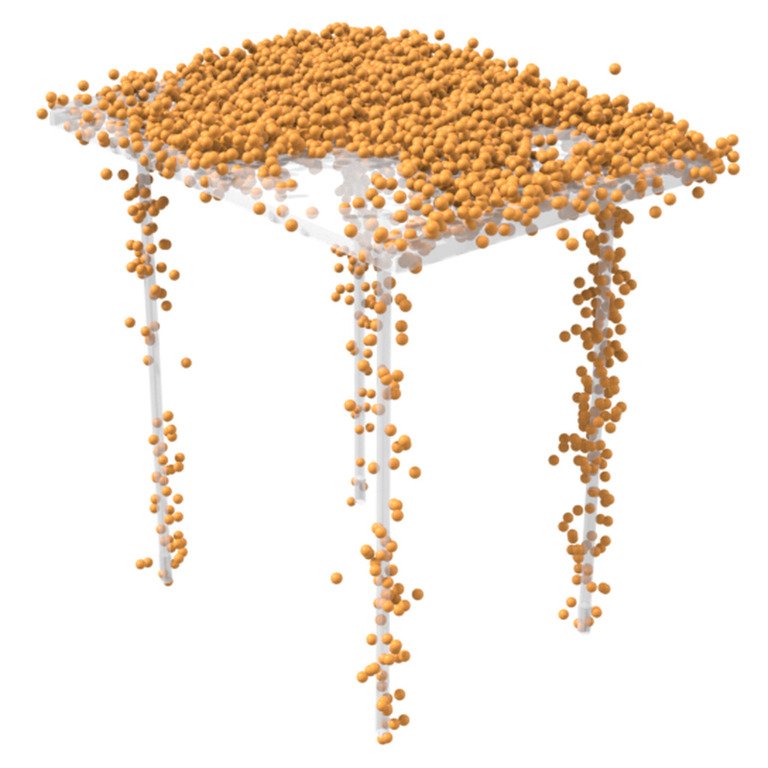} &
        \includegraphics[width=0.19\linewidth]{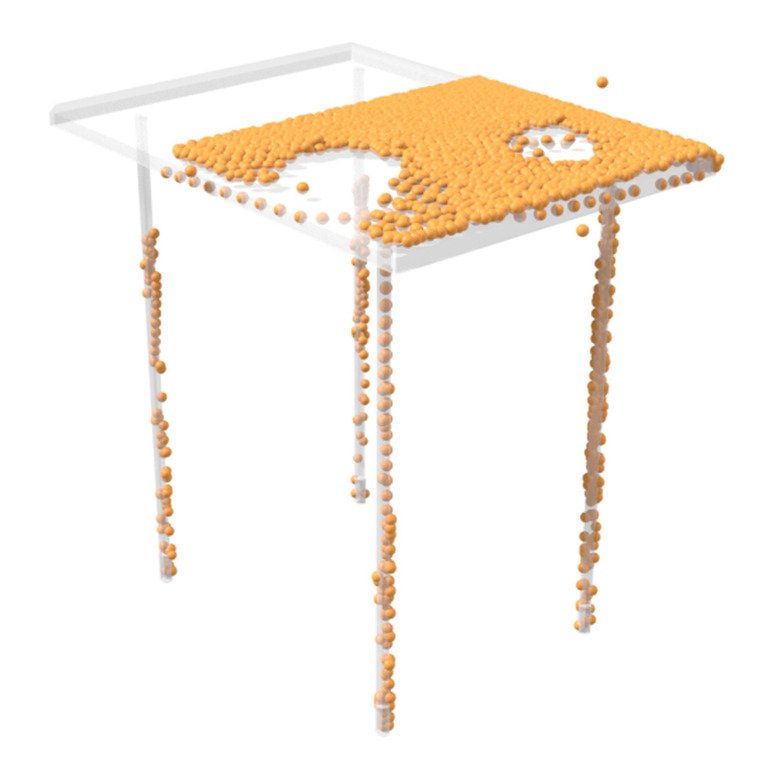} &
        \includegraphics[width=0.19\linewidth]{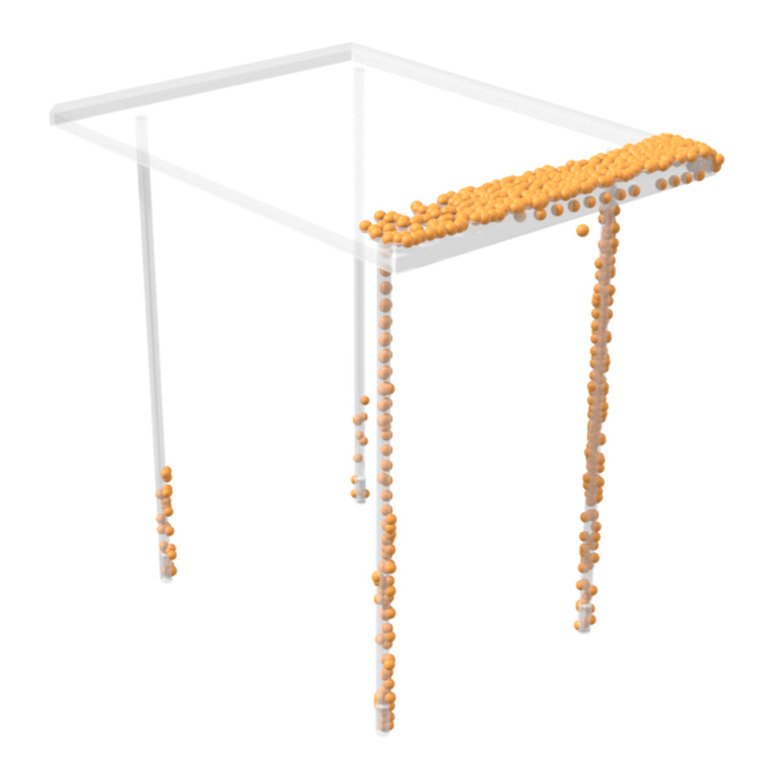} \\[1pt]
        \raisebox{14px}{\rotatebox{90}{Output}} &
        \includegraphics[width=0.19\linewidth]{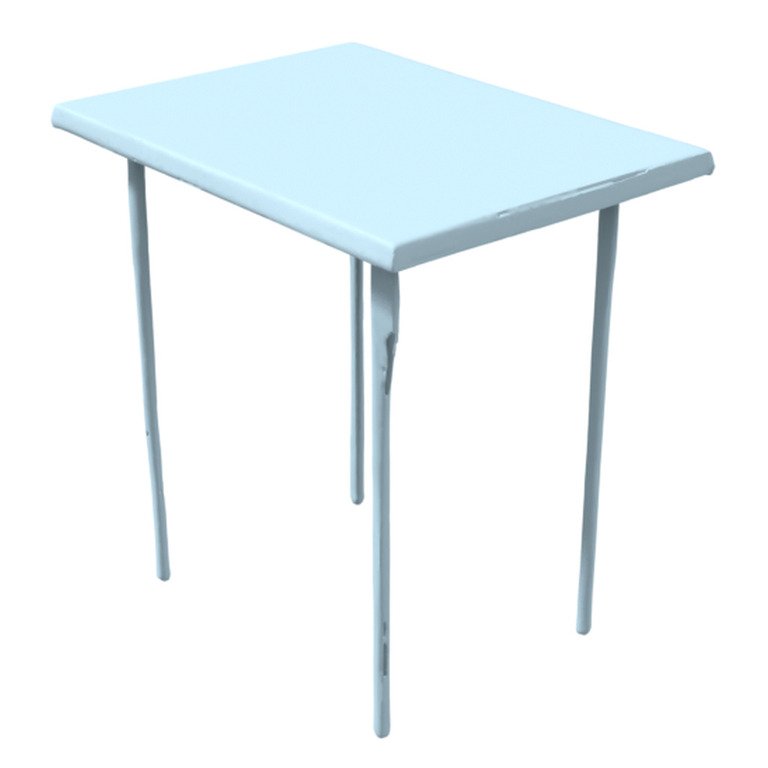} &
        \includegraphics[width=0.19\linewidth]{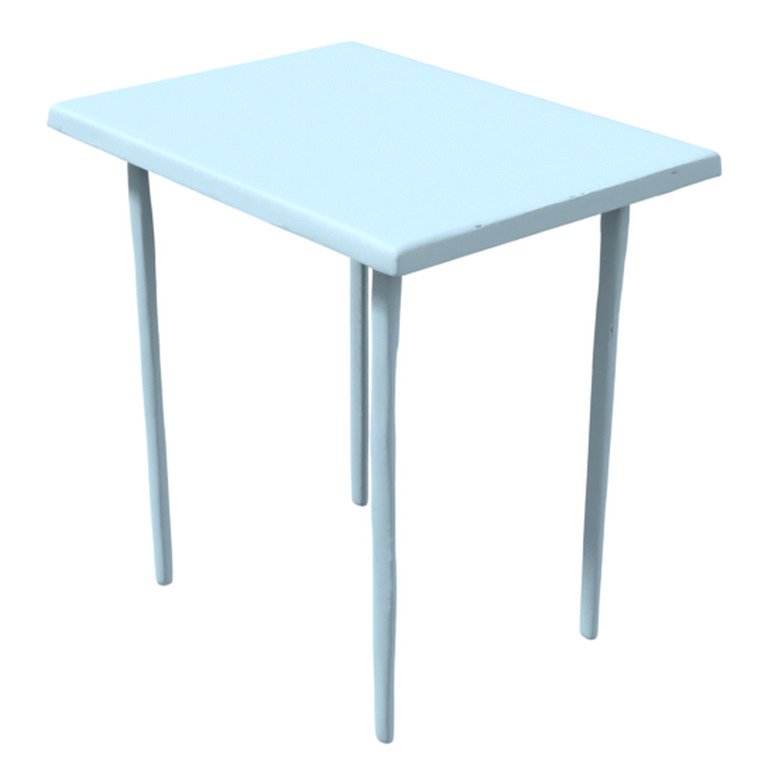} &
        \includegraphics[width=0.19\linewidth]{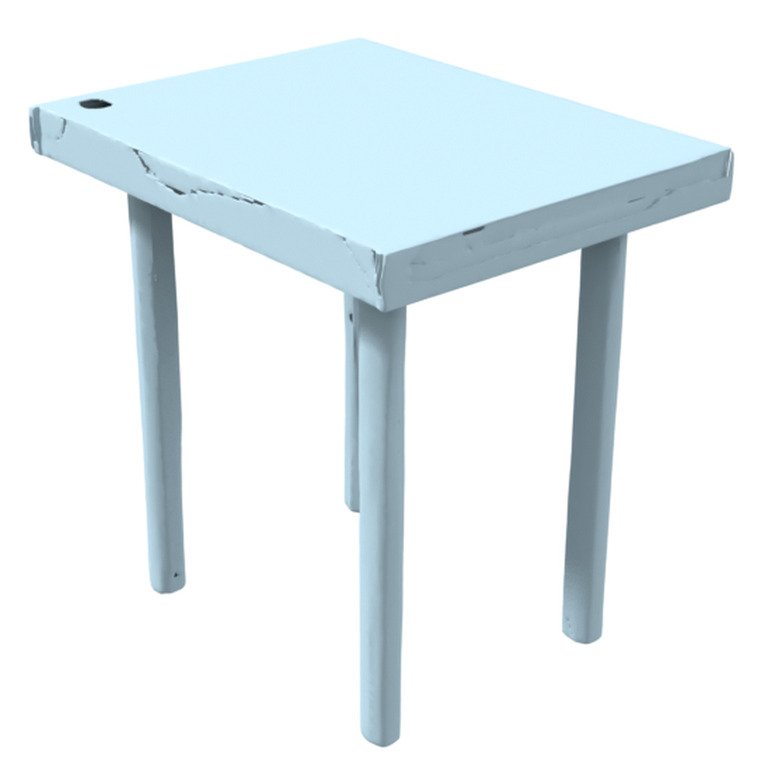} &
        \includegraphics[width=0.19\linewidth]{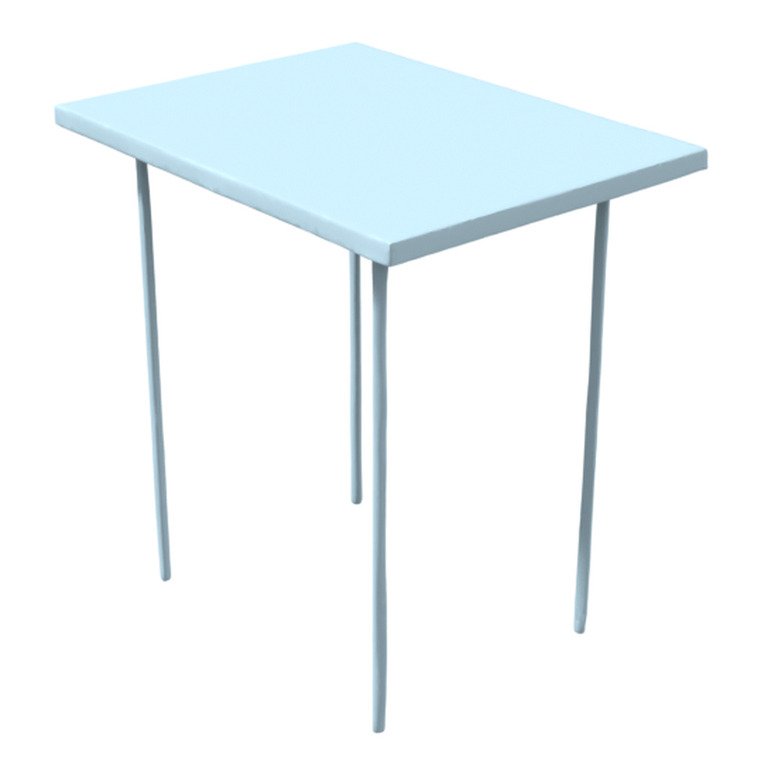} &
        \includegraphics[width=0.19\linewidth]{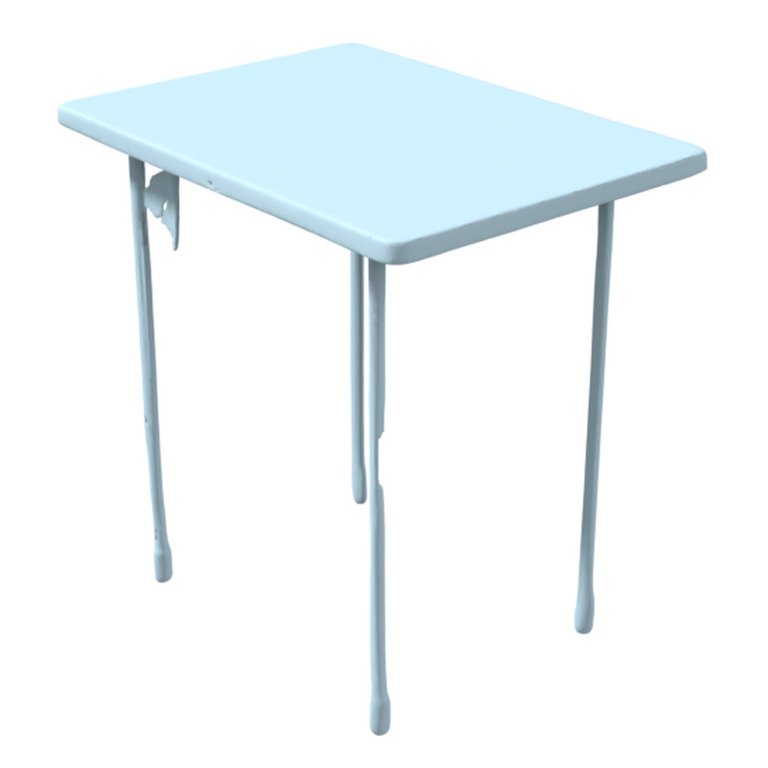} \\[1pt]
    \end{tabular}%
    }
    \captionof{figure}{\textbf{Qualitative results on generator robustness.} \model{} keeps the object shape under moderate noise and missing points. Inputs use GT poses and instances, with noisy or partial instance point clouds (orange) shown over the GT mesh.}
    \label{fig:robust_qual}
\end{table*}

\WrapNeedspace{129pt}
\paragraph{Comparison against HoloScene}
{
\begin{wraptable}{r}{0.45\linewidth}
\vspace{-1em}
\small
\centering
\resizebox{\linewidth}{!}{
\begin{tabular}{llcccc}
\toprule
\textbf{Setting} & \textbf{Method} & \textbf{CD} \revten{(cm)} $\downarrow$ & \textbf{F1}$\uparrow$ & \textbf{NC}$\uparrow$ & \textbf{PSNR} \revten{(dB)} $\uparrow$ \\
\midrule
\multirow{2}{*}{Scene}
 & HoloScene & 2.63 & 0.43 & \textbf{0.86} & \textbf{17.87} \\
 & Ours & \textbf{2.24} & \textbf{0.45} & 0.82 & 13.55 \\
\midrule
\multirow{2}{*}{Object}
 & HoloScene & 2.94 & 0.35 & \textbf{0.81} & \textbf{20.78} \\
 & Ours & \textbf{1.28} & \textbf{0.61} & \textbf{0.81} & 20.64 \\
\bottomrule
\end{tabular}
}
\caption{\textbf{Quantitative comparison against HoloScene.} \model{} achieves comparable scene-level geometry performance and object-level overall performance while reducing runtime by 480$\times$.}
\label{tab:holoscene_cmp}
\vspace{-1em}
\end{wraptable}
HoloScene~\cite{xia2025holoscene} is the closest optimization-based system targeting simulation-oriented reconstruction from video. As shown in Tab.~\ref{tab:holoscene_cmp} and Fig.~\ref{fig:holoscene_qualitative_igibson}, \model improves scene-level CD and F1 and substantially improves object-level CD and F1, while HoloScene obtains higher scene-level NC and PSNR. The two methods are therefore not uniformly ordered by quality. The main difference is efficiency: HoloScene requires {approximately eight hours per scene} \rev{for optimization alone, excluding preprocessing}, whereas \model requires {approximately one minute}, corresponding to a {$480\times$ speedup}.\par
}

\begin{table*}[h]
    \centering
    \setlength{\tabcolsep}{1pt}
    \renewcommand{\arraystretch}{0.5}
    \resizebox{\textwidth}{!}{%
    \begin{tabular}{cccccc}
        \includegraphics[width=0.2\linewidth]{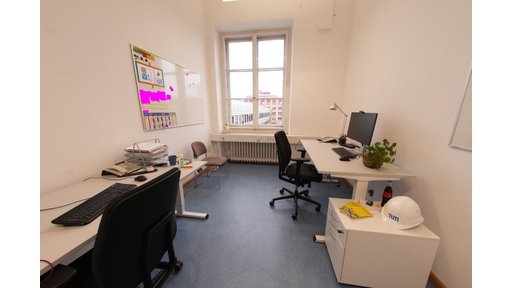} &
        \includegraphics[width=0.2\linewidth]{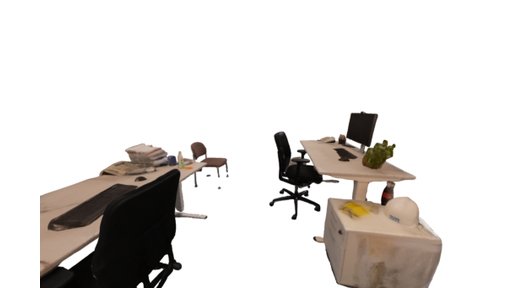} &
        \includegraphics[width=0.2\linewidth]{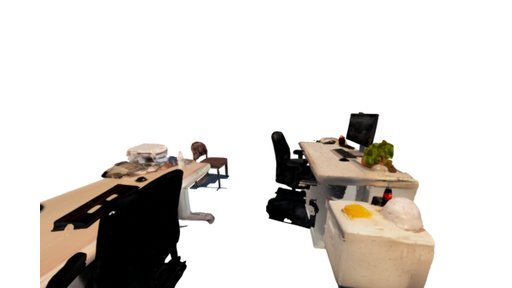} &
        \includegraphics[width=0.2\linewidth]{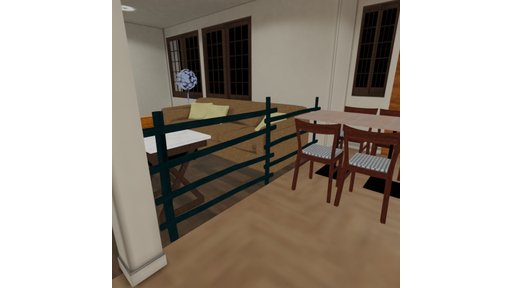} &
        \includegraphics[width=0.2\linewidth]{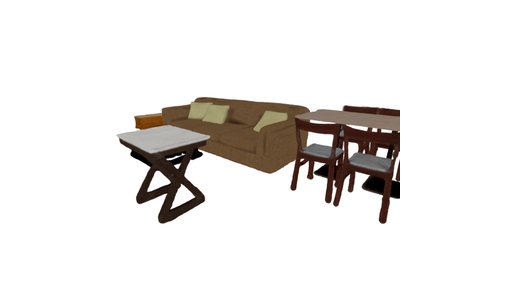} &
        \includegraphics[width=0.2\linewidth]{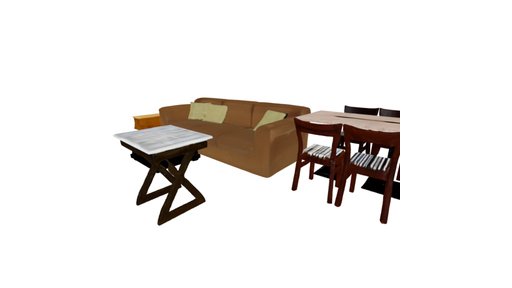} \\[1pt]
        \includegraphics[width=0.2\linewidth]{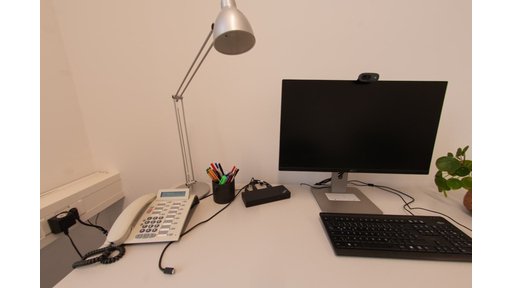} &
        \includegraphics[width=0.2\linewidth]{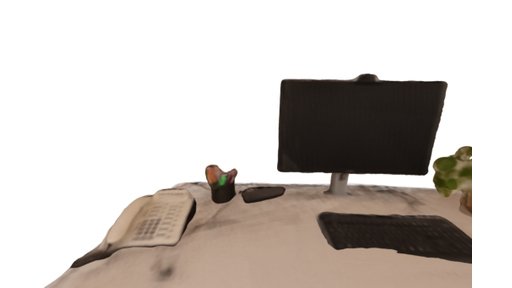} &
        \includegraphics[width=0.2\linewidth]{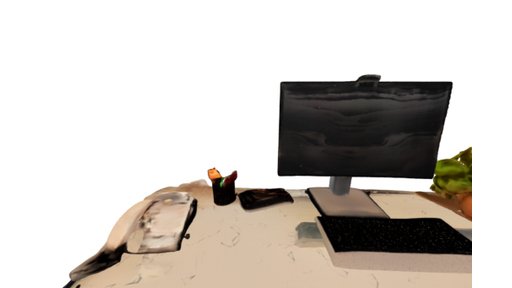}  &
        \includegraphics[width=0.2\linewidth]{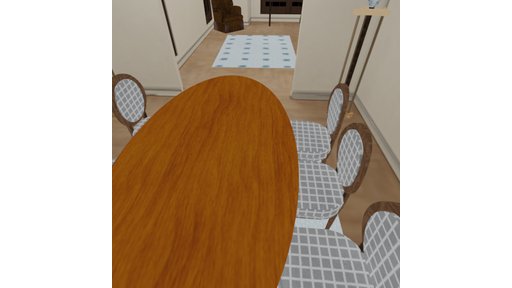} &
        \includegraphics[width=0.2\linewidth]{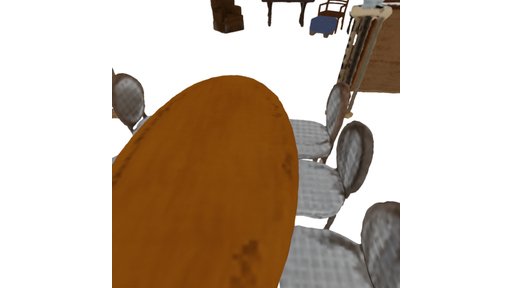} &
        \includegraphics[width=0.2\linewidth]{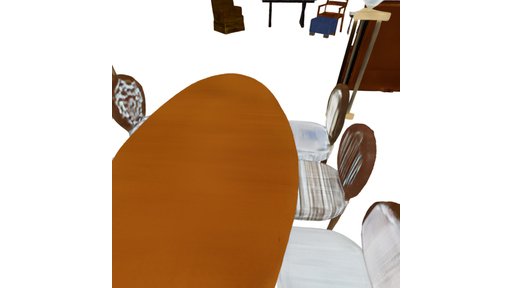} \\[1pt]
        
        Input RGB & HoloScene & Ours & Input RGB & HoloScene & Ours 
    \end{tabular}%
    }
    \captionof{figure}{\textbf{Qualitative comparison with HoloScene.} Input RGB, HoloScene, and \model use identical \rev{camera parameters}. Background surfaces are excluded. Appearance is evaluated on GT foreground-object pixels.}
    \label{fig:holoscene_qualitative_igibson}
\end{table*}

\WrapNeedspace{159pt}
\paragraph{Comparison against SimRecon}
{
\begin{wraptable}{r}{0.4\linewidth}
\vspace{-1em}
\small
\centering
\resizebox{\linewidth}{!}{
\begin{tabular}{llcc}
\toprule
\textbf{Dataset} & \textbf{Method} & \textbf{mAP}$\uparrow$ & \textbf{mIoU}$\uparrow$ \\
\midrule
\multirow{2}{*}{iTHOR}
 & SimRecon & 0.36 & 0.31 \\
 & Ours & \textbf{0.49} & \textbf{0.37} \\
\midrule
\multirow{2}{*}{Imag.}
 & SimRecon & 0.61 & \textbf{0.56} \\
 & Ours & \textbf{0.67} & 0.52 \\
\midrule
\multirow{2}{*}{Overall}
 & SimRecon & 0.48 & 0.44 \\
 & Ours & \textbf{0.58} & \textbf{0.45} \\
\bottomrule
\end{tabular}
}
\caption{\textbf{Quantitative comparison against SimRecon.} \model{} achieves superior overall perception performance while enjoying a 31$\times$ speedup.}
\label{tab:simrecon_cmp}
\vspace{-1em}
\end{wraptable}
We compare with SimRecon~\cite{xia2026simrecon} on matched 10-scene subsets of iTHOR \cite{ai2thor} and Imaginarium \cite{imaginarium}. Tab.~\ref{tab:simrecon_cmp} shows that \model improves the mAP detection evaluation metric on both datasets and obtains a higher overall mIoU on instance segmentation, \rev{while trailing SimRecon in Imaginarium mIoU (0.52 versus 0.56).} In terms of runtime analysis, \model{} reduces average runtime from 262.74 to 8.23 seconds per scene, yielding a {$31.93\times$ speedup}. \rev{FIRE3D thus improves overall mAP/mIoU at substantially lower runtime.} See visualizations of perception results in Fig. \ref{fig:simrecon_rgb_obb_bev_selected}.\par
}

\begin{table*}[h]
    \centering
    \setlength{\tabcolsep}{1pt}
    \renewcommand{\arraystretch}{0.5}
    \resizebox{\textwidth}{!}{%
    \begin{tabular}{cccccc}
        \includegraphics[width=0.2\linewidth]{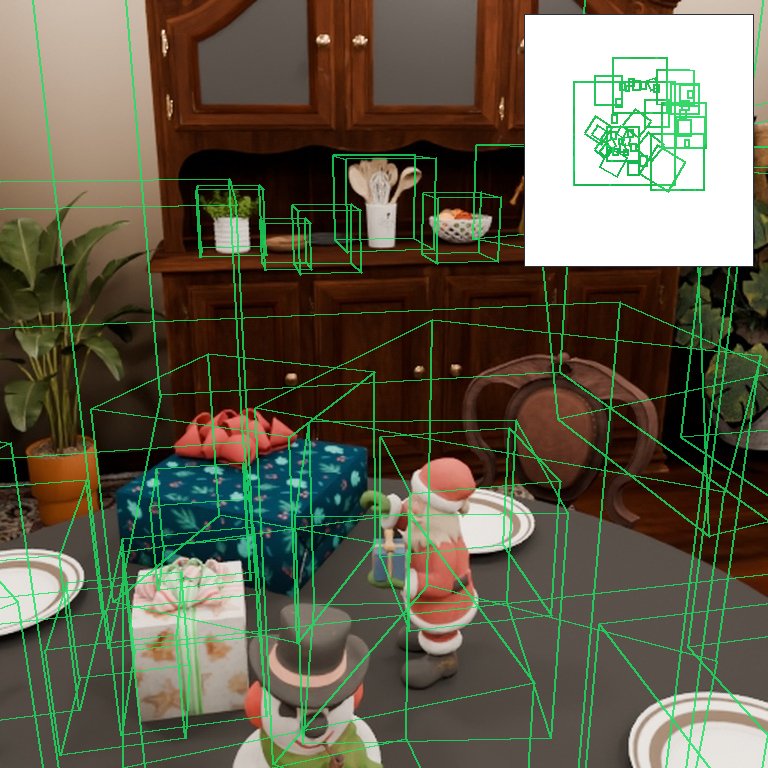} &
        \includegraphics[width=0.2\linewidth]{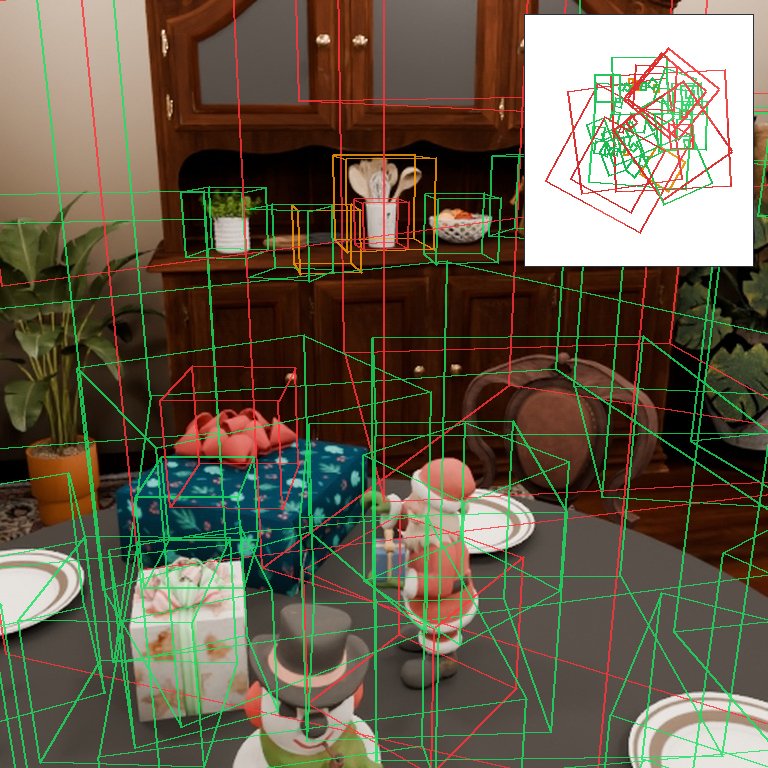} &
        \includegraphics[width=0.2\linewidth]{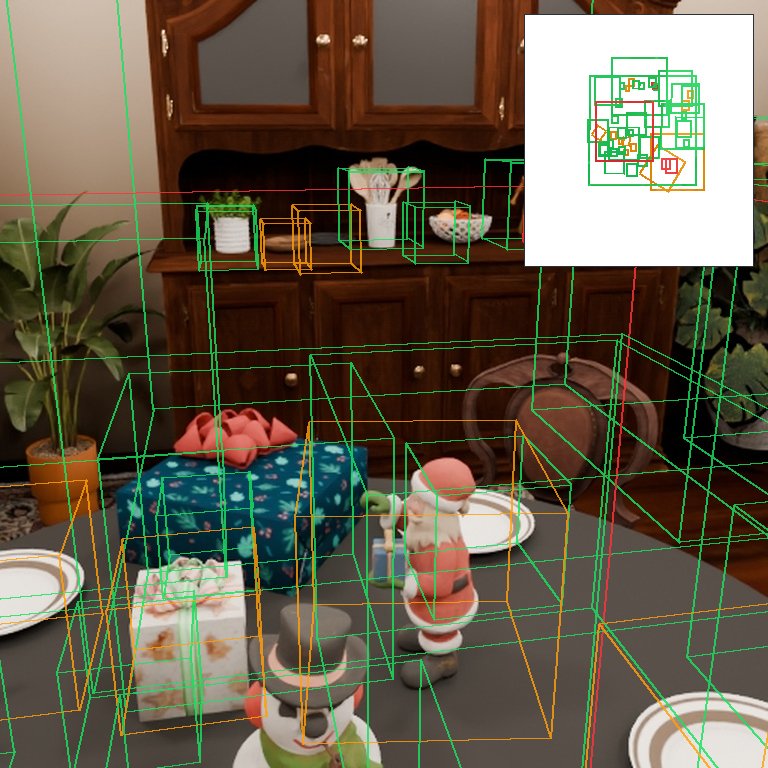} &
        \includegraphics[width=0.2\linewidth]{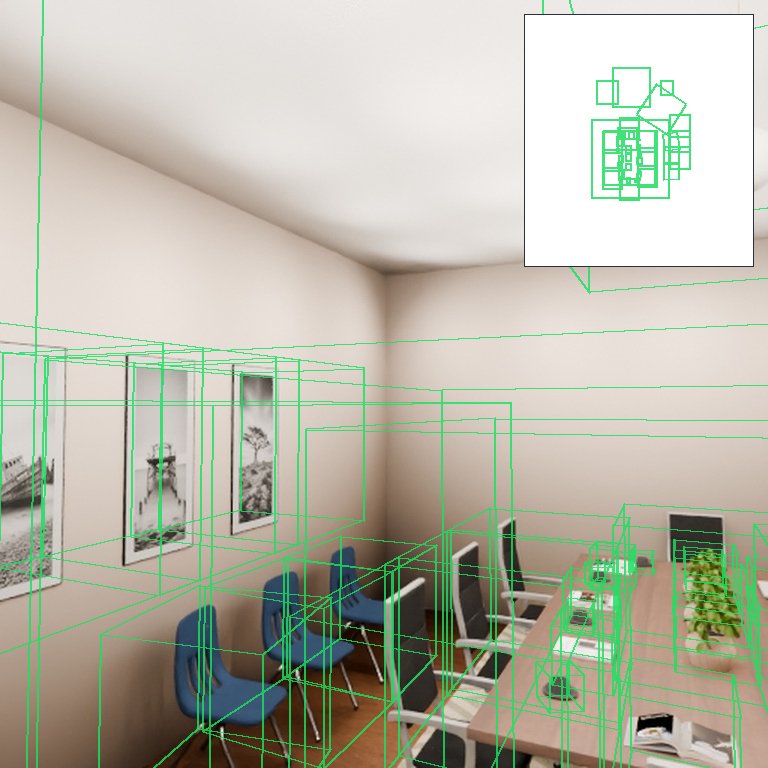} &
        \includegraphics[width=0.2\linewidth]{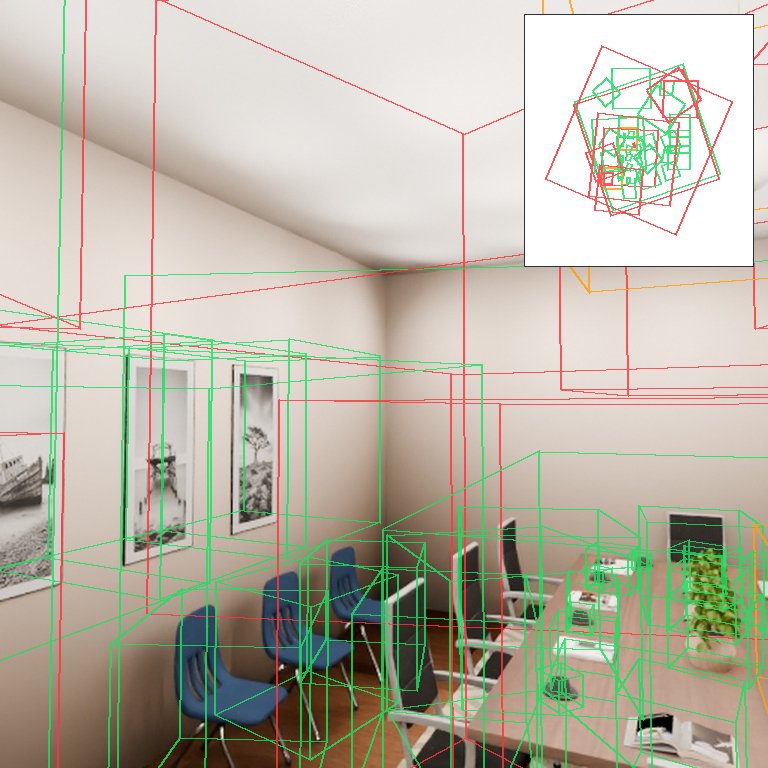} &
        \includegraphics[width=0.2\linewidth]{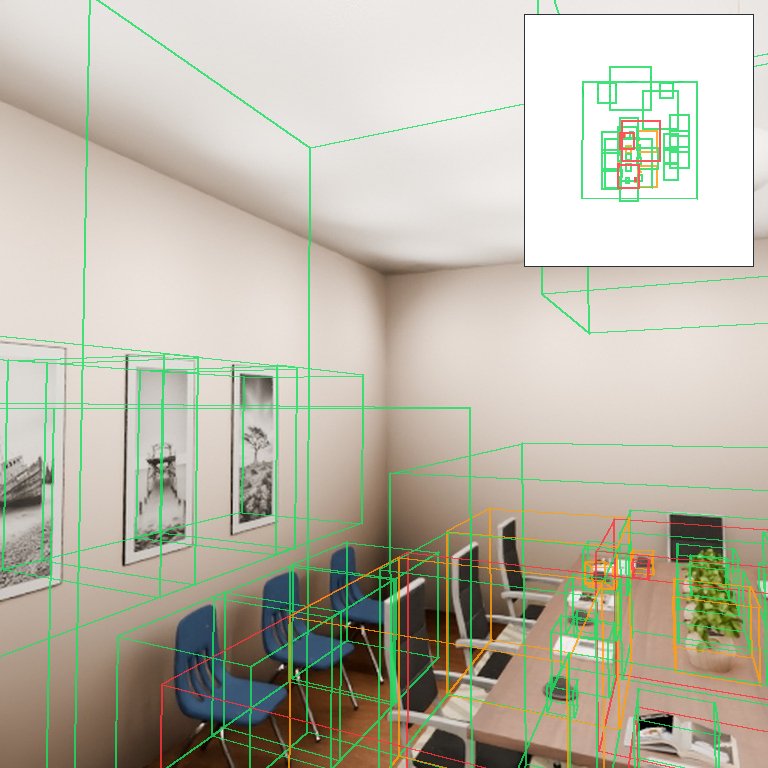} \\[1pt]
        Ground-truth & SimRecon & Ours & Ground-truth & SimRecon & Ours
    \end{tabular}%
    }
    \captionof{figure}{\textbf{Qualitative 3D-detection comparison with SimRecon.} Each row uses the same original RGB and camera. Green denotes GT or a prediction with 3D IoU $>0.25$, red denotes an unmatched prediction, and orange denotes a missed GT object. Every panel includes the same-scale bird's-eye-view inset. \model{} has fewer missed objects and higher accuracy than SimRecon \cite{xia2026simrecon}.}
    \label{fig:simrecon_rgb_obb_bev_selected}
\end{table*}

\WrapNeedspace{111pt}
\paragraph{Comparison against LiteReality}
{
\begin{wraptable}{r}{0.38\linewidth}
\vspace{-1em}
\small
\centering
\resizebox{\linewidth}{!}{
\begin{tabular}{lccc}
\toprule
\textbf{Method} & \textbf{CD} \revten{(cm)} $\downarrow$ & \textbf{F1}$\uparrow$ & \textbf{NC}$\uparrow$ \\
\midrule
LiteReality & 10.60 & 0.18 & 0.46 \\
Ours & \textbf{1.47} & \textbf{0.70} & \textbf{0.81} \\
\bottomrule
\end{tabular}
}
\caption{\textbf{Comparison against LiteReality.} \model{} achieves superior geometry performance over LiteReality while enjoying a $7.46\times$ faster runtime.}
\label{tab:litereality_cmp}
\vspace{-1em}
\end{wraptable}
LiteReality \cite{huang2026litereality} follows a lift-then-instance pipeline based on external scanning and asset retrieval. To isolate object reconstruction, we provide both methods with GT detection and segmentation on the same randomly selected iTHOR \cite{ai2thor} and Imaginarium \cite{imaginarium} scenes, and include inference and mesh post-processing in the runtime. Tab.~\ref{tab:litereality_cmp} shows that \model improves all three geometry metrics and reduces the geometry runtime from 18.21 to 2.44 seconds per object ({$7.46\times$ faster}). See visualizations comparisons in Fig. \ref{fig:litereality_floorplan12_411} \par
}

\begin{table*}[h]
    \centering
    \setlength{\tabcolsep}{1pt}
    \renewcommand{\arraystretch}{0.5}
    \resizebox{\textwidth}{!}{%
    \begin{tabular}{cccccc}
        \includegraphics[width=0.2\linewidth]{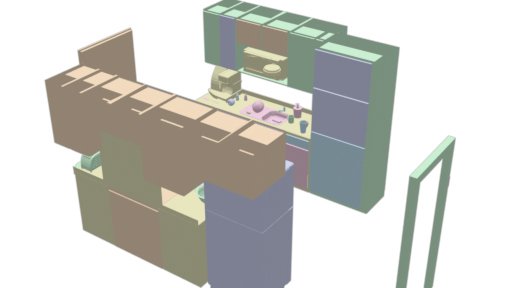} &
        \includegraphics[width=0.2\linewidth]{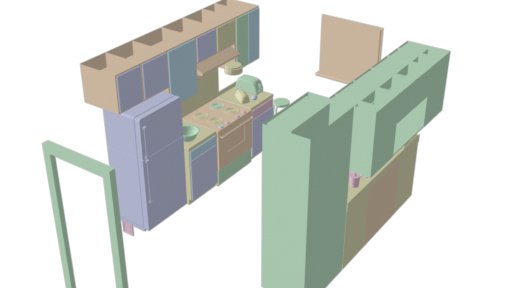} &
        \includegraphics[width=0.2\linewidth]{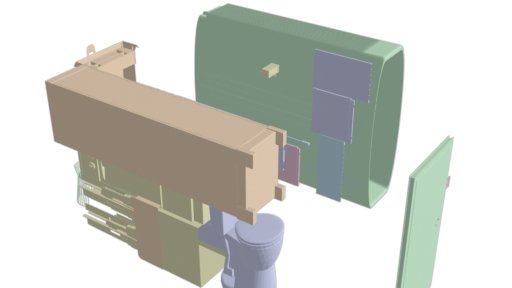} &
        \includegraphics[width=0.2\linewidth]{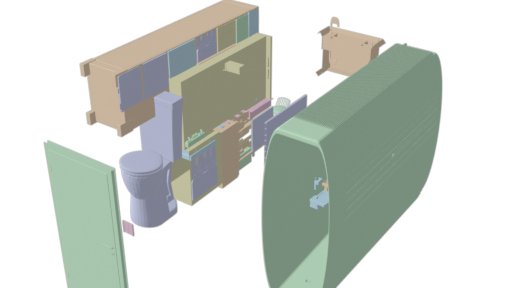} &
        \includegraphics[width=0.2\linewidth]{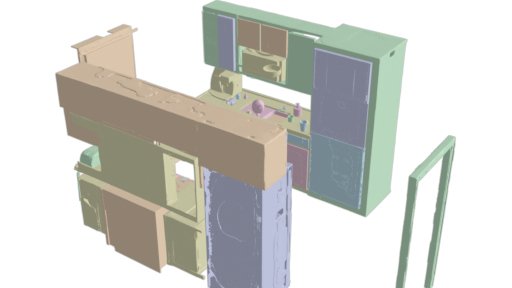} &
        \includegraphics[width=0.2\linewidth]{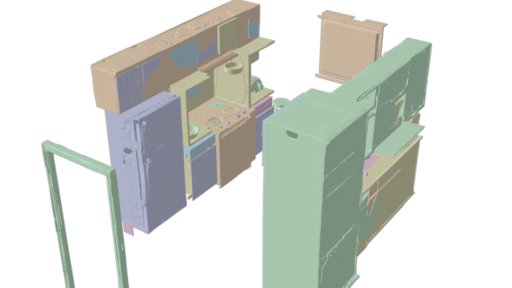} \\[1pt]

        \includegraphics[width=0.2\linewidth]{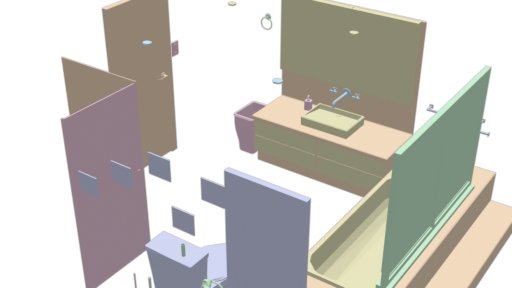} &
        \includegraphics[width=0.2\linewidth]{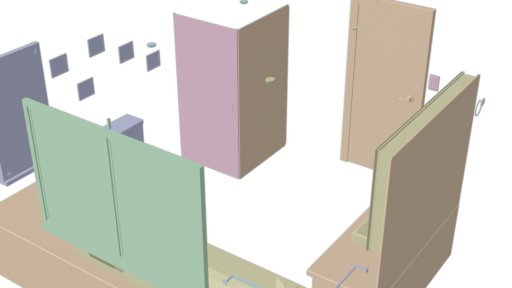} &
        \includegraphics[width=0.2\linewidth]{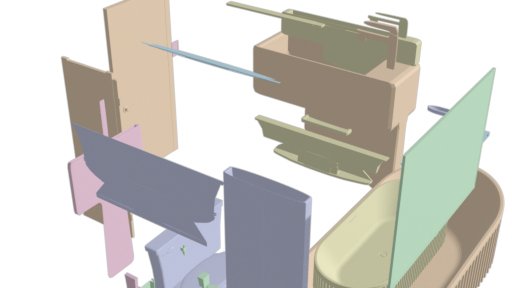} &
        \includegraphics[width=0.2\linewidth]{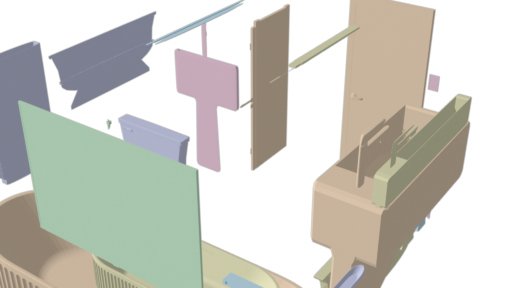} &
        \includegraphics[width=0.2\linewidth]{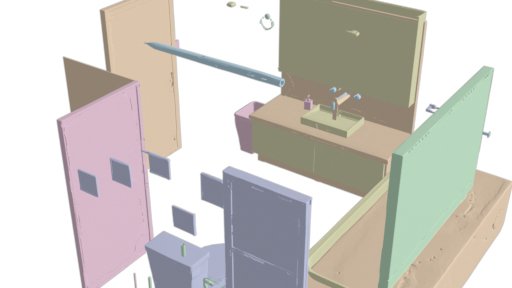} &
        \includegraphics[width=0.2\linewidth]{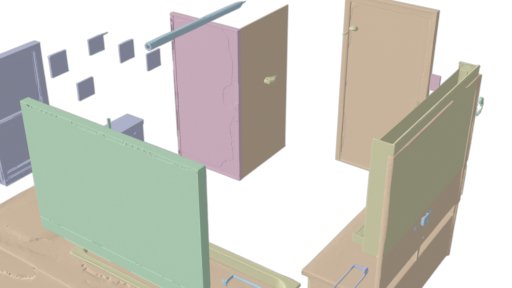} \\[1pt]
        GT View 1 & GT View 2 & LiteReality View 1 & LiteReality View 2 & Ours View 1 & Ours View 2
    \end{tabular}%
    }
    \captionof{figure}{\textbf{Geometry comparison with LiteReality under GT perception and pose.} \model{} can reconstruct the object geometry more accurately while requiring much less runtime.}
    \label{fig:litereality_floorplan12_411}
\end{table*}

\WrapNeedspace{81pt}
\paragraph{Comparison against composed pipeline}
{
\begin{wraptable}{r}{0.44\linewidth}
\vspace{-1em}
\small
\centering
\resizebox{\linewidth}{!}{
\begin{tabular}{lcccc}
\toprule
\textbf{Method} & \textbf{mAP}$\uparrow$ & \textbf{mIoU}$\uparrow$ & \textbf{CD} \revten{(cm)} $\downarrow$ & \textbf{PSNR} \revten{(dB)} $\uparrow$ \\
\midrule
Composed & 0.36 & 0.28 & 5.97 & 15.24 \\
Ours & \textbf{0.50} & \textbf{0.39} & \textbf{3.21} & \textbf{15.38} \\
\bottomrule
\end{tabular}
}
\caption{\textbf{End-to-end comparison} on 30 Imaginarium scenes.}
\label{tab:composed_cmp}
\vspace{-1em}
\end{wraptable}
We also construct a direct modular baseline that combines Boxer \cite{boxer} for 3D detection, SAM2 \cite{ravi2025sam} for image segmentation, and TRELLIS.2~\rebut{\cite{xiang2026native}} for object generation. We evaluate both pipelines end-to-end on the same 30 Imaginarium scenes. \rev{Both pipelines follow the evaluation protocols of Tabs.~\ref{tab:perception} and~\ref{tab:main_cmp}, including matched-object geometry and full-image novel-view rendering.} \rev{\model improves all reported quality metrics (Tab.~\ref{tab:composed_cmp}) and achieves $6.52\times$ faster network inference per object}, and \model{} also enjoys a faster mesh post-processing speed thanks to the implemented parallelism. \rev{This is a system-level comparison: differences in perception, conditioning, and training prevent attributing the gains to HC-VAE alone.} Visualizations can be found in Fig. \ref{fig:composed_manual8_final}. \par
}

\begin{table*}[h]
    \centering
    \setlength{\tabcolsep}{1pt}
    \renewcommand{\arraystretch}{0.5}
    \resizebox{\textwidth}{!}{%
    \begin{tabular}{cccccc}
        \includegraphics[width=0.164\linewidth]{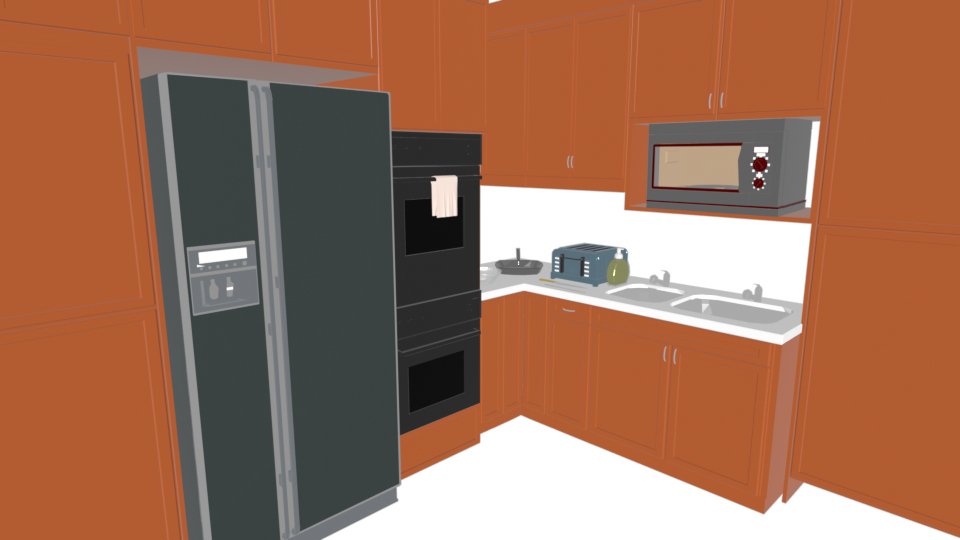} &
        \includegraphics[width=0.164\linewidth]{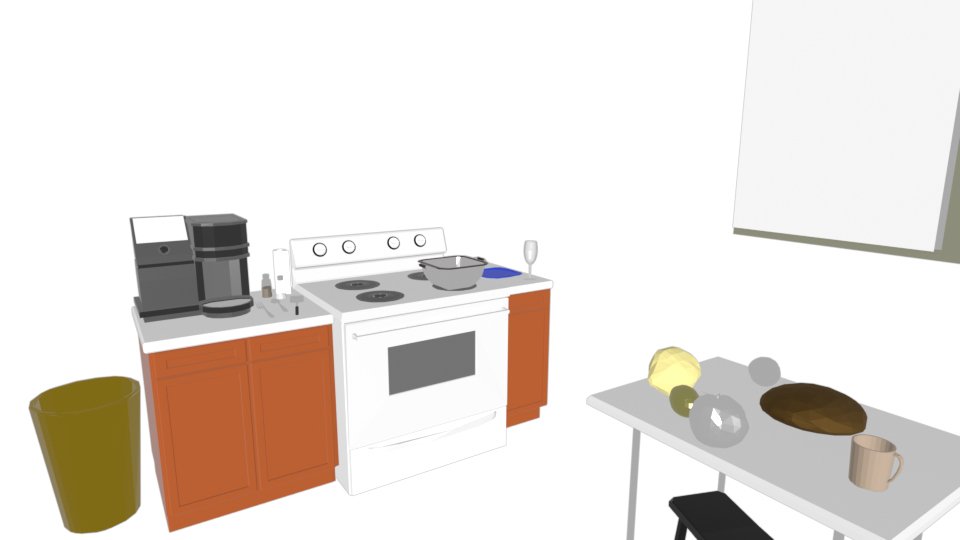} &
        \includegraphics[width=0.164\linewidth]{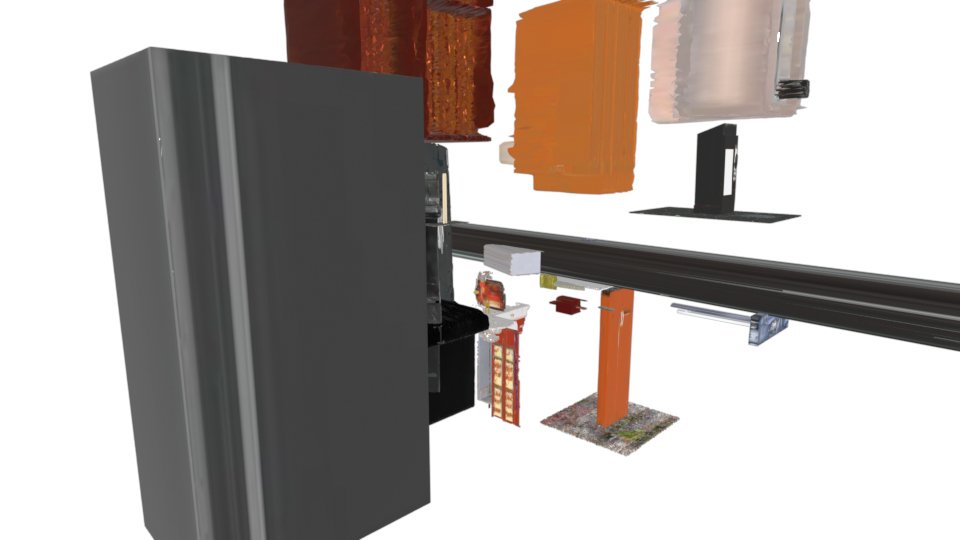} &
        \includegraphics[width=0.164\linewidth]{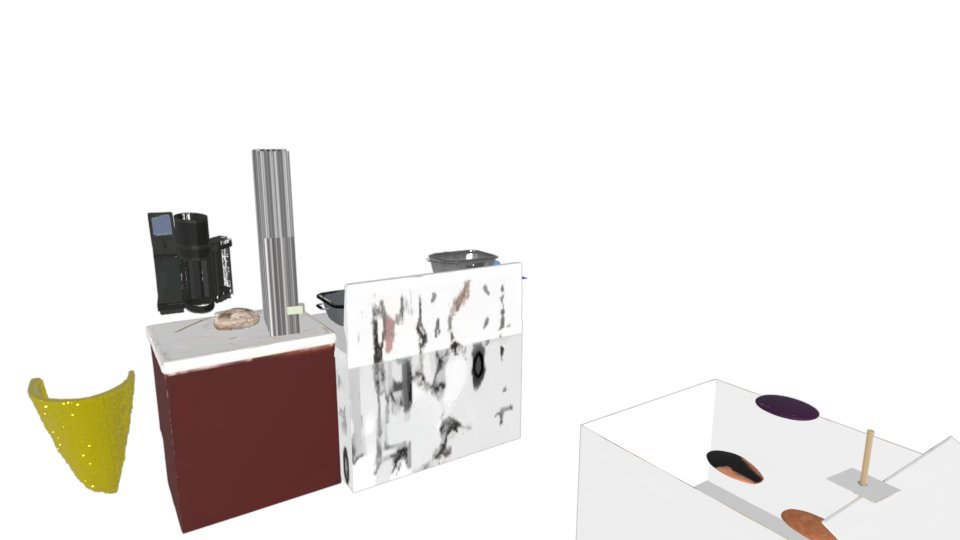} &
        \includegraphics[width=0.164\linewidth]{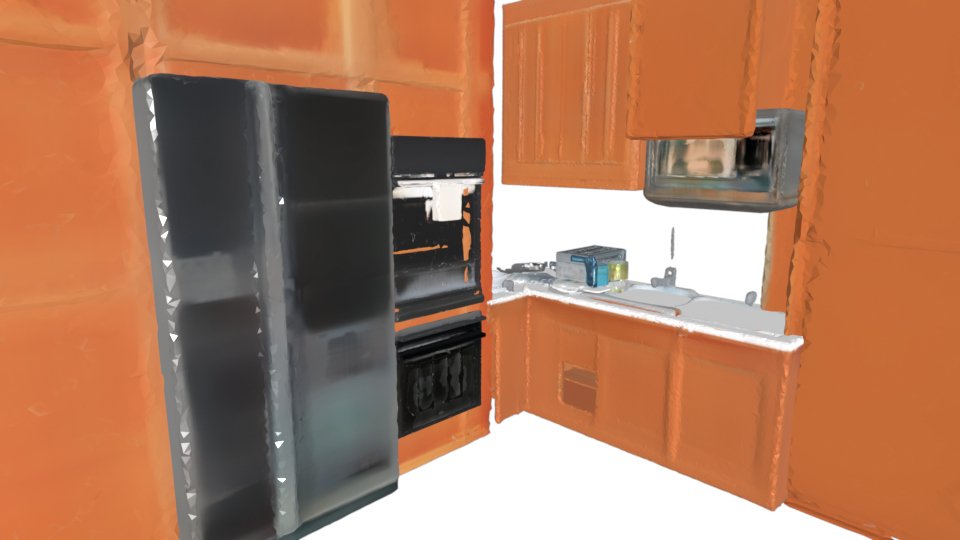} &
        \includegraphics[width=0.164\linewidth]{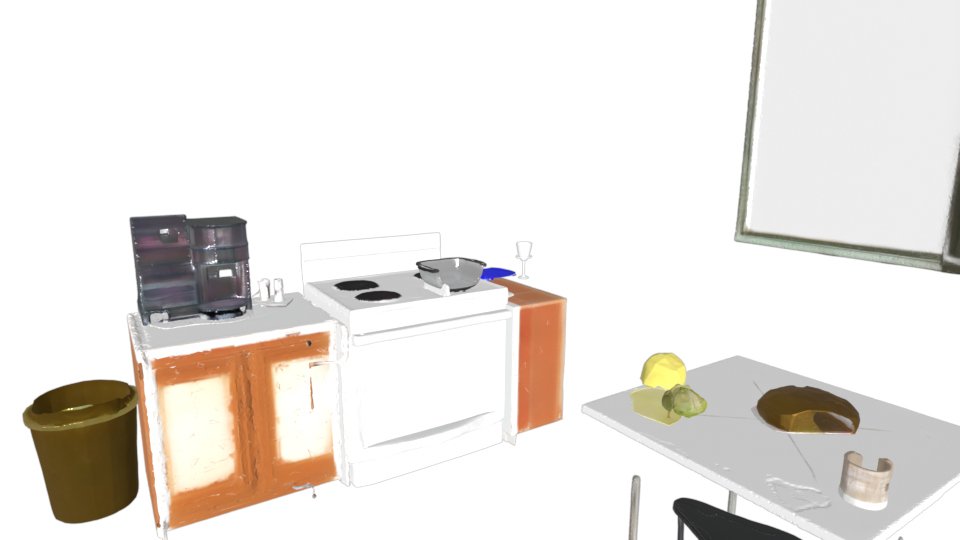} \\[1pt]

        \includegraphics[width=0.164\linewidth]{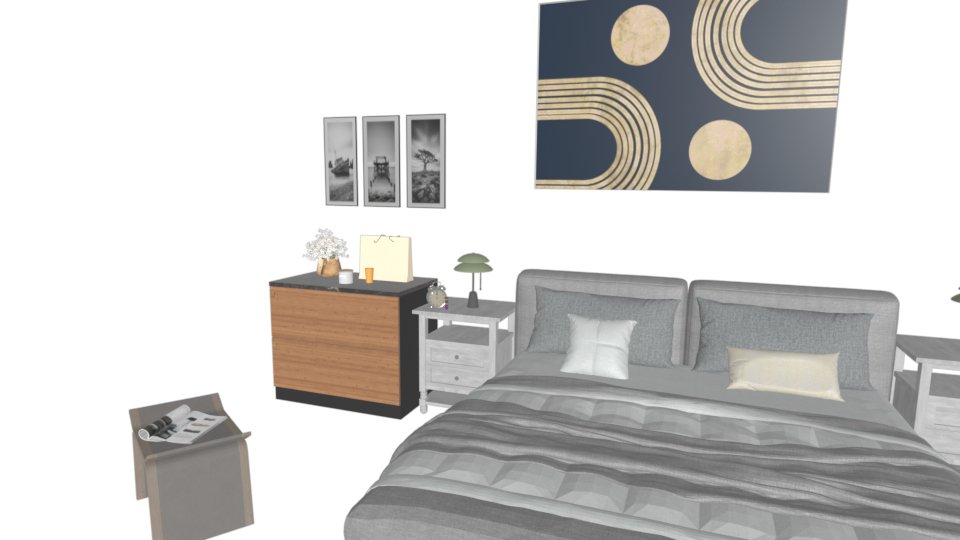} &
        \includegraphics[width=0.164\linewidth]{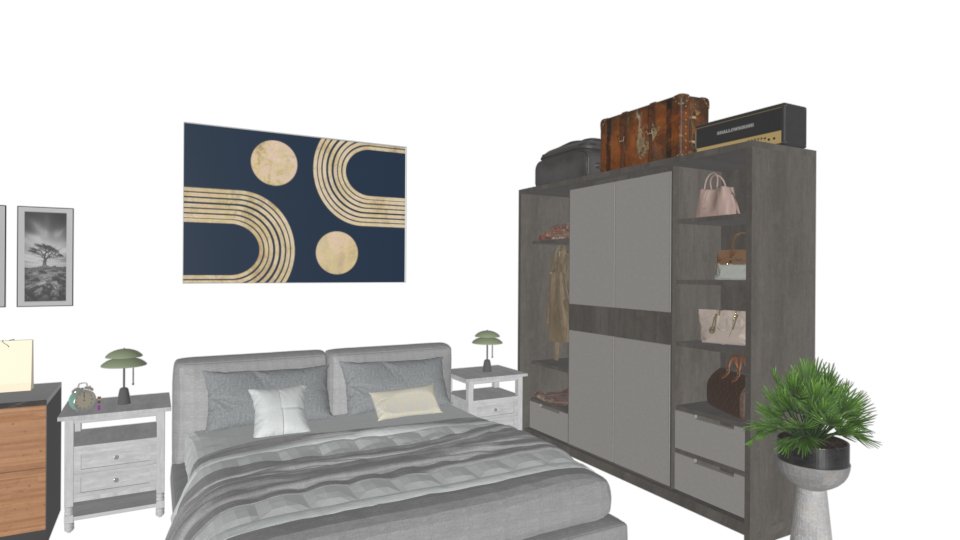} &
        \includegraphics[width=0.164\linewidth]{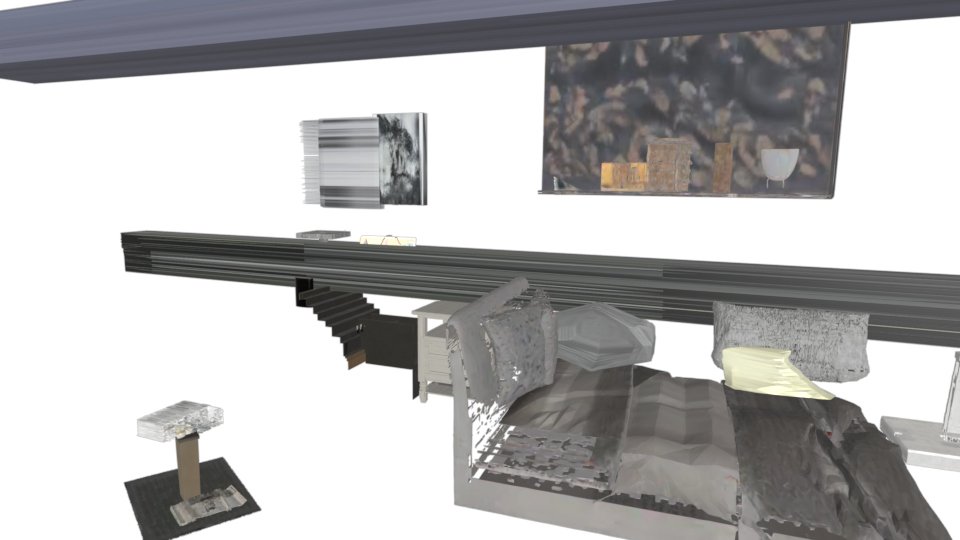} &
        \includegraphics[width=0.164\linewidth]{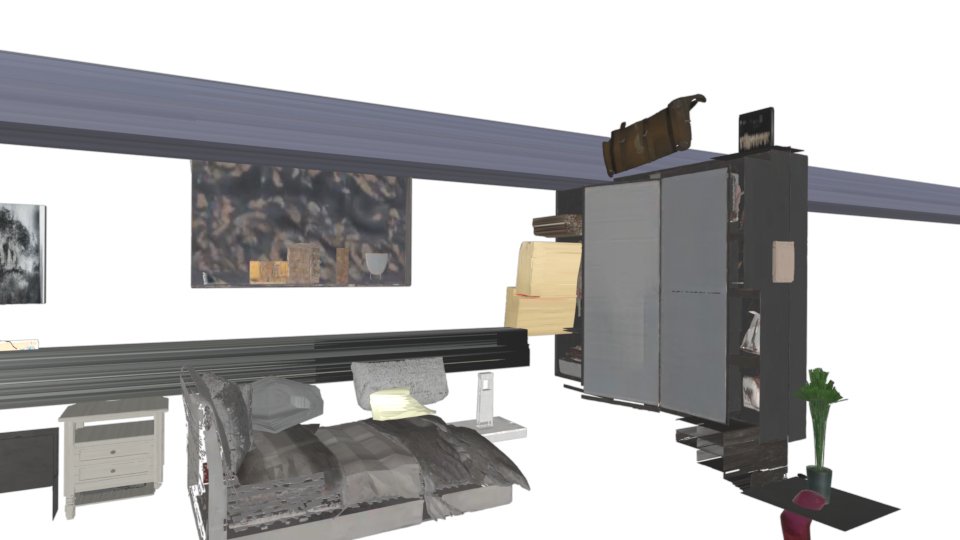} &
        \includegraphics[width=0.164\linewidth]{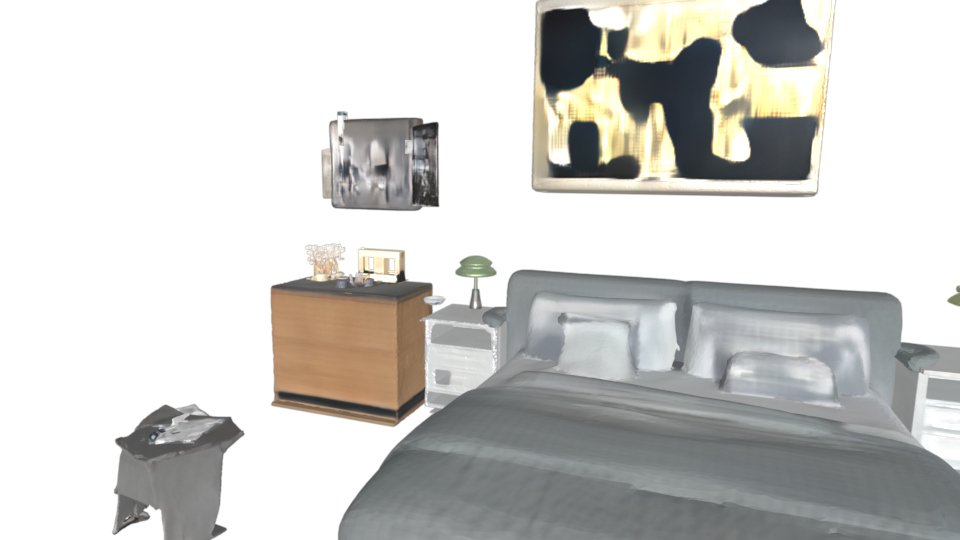} &
        \includegraphics[width=0.164\linewidth]{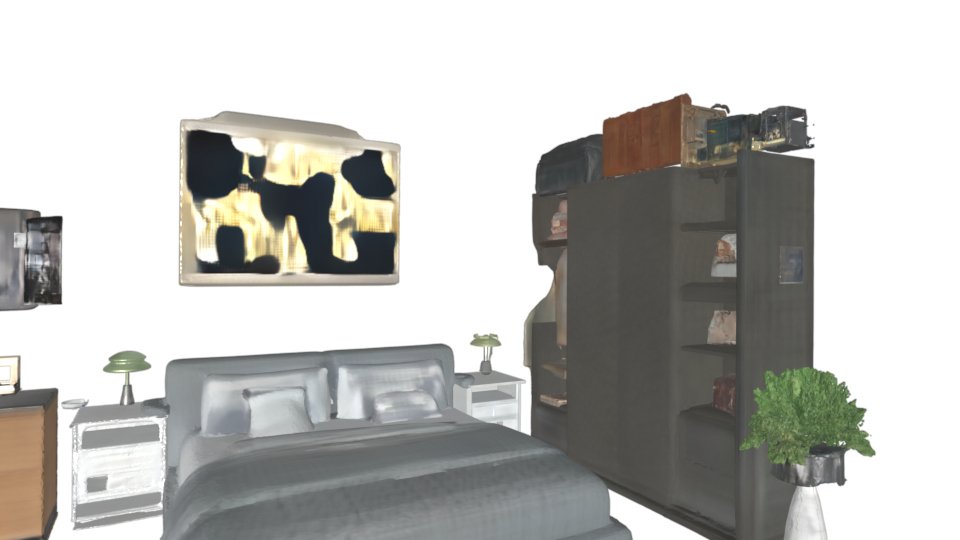} \\[1pt]
        
        GT View 1 & GT View 2 & Composed Pipe. View 1 & Composed Pipe. View 2 & \revten{Ours} View 1 & \revten{Ours} View 2
    \end{tabular}%
    }
    \captionof{figure}{\textbf{End-to-end qualitative comparison.} Both reconstruction methods use inferred perception, and background geometry is omitted for all columns. \rev{\model{} improves the reported system-level results while reducing runtime.}}
    \label{fig:composed_manual8_final}
\end{table*}

\paragraph{\rebut{Comparison against the Agentic Approach}}
\rebut{
General-purpose agents can also reconstruct a scene by iterative tool use. 
We set up Claude Opus 5.5 as a per-scene reconstruction agent with the same posed RGB-D input as \model{} and no external learned models. 
For each scene, the agent keeps iterating the reconstruction results to optimize the result. 
It back-projects the depth, fits planes and boxes, builds each object from primitive shapes in code, renders the result over the input frames, and fixes mismatches. 
On 4 test scenes, evaluated with the protocol of Tab.~\ref{tab:perception}, \model{} reconstructs each scene in 16 to 21\,s for about \$0.01 of GPU time and also predicts 3D instance masks. 
As a per-scene optimization method, the agent spends \textbf{13.9 to 24.4\,min} and \textbf{\$3.9 to \$9.2} per scene to reach higher accuracy (Tab.~\ref{tab:agent_cmp}). 
Like HoloScene (Tab.~\ref{tab:holoscene_cmp}), it trades runtime for accuracy, while \model{} makes reconstruction cheap enough to run on many scenes. 
Fig.~\ref{fig:agent_cmp} shows qualitative results.
}

\paragraph{\rebut{Combination with the Agentic Approach}}
\rebut{
The two approaches are also complementary. 
We initialize the same agent with the per-object meshes of \model{} instead of an empty scene, keeping its rules, tools and time limit unchanged. 
In Tab.~\ref{tab:agent_cmp}, we show that this gives a good result fast: \model{} provides a complete scene after 16 to 21\,s, and at 7\,min the combined scene reaches 0.642 mAP with 90 matched objects, against 0.353 and 47 for the agent alone, as shown in Fig.~\ref{fig:agent_cmp} as well. 
The per-object geometry at that point still favors the agent alone, whose fewer objects are already refined. 
Given enough time, the combination also raises the ceiling, ending with the best mAP, CD, F1 and NC of all methods, with mIoU on par. 
The agent rebuilds box-like furniture from the measured depth and keeps the meshes of \model{} for organic shapes such as beds, armchairs and plants, which are hard to model in code.
}

\begin{table}[ht]
\centering
\small
\resizebox{\linewidth}{!}{%
\begin{tabular}{lccccccccc}
\toprule
\textbf{Method} & \textbf{Time / scene} & \textbf{Cost / scene} & mAP $\uparrow$ & mIoU $\uparrow$ & CD (cm) $\downarrow$ & F1 $\uparrow$ & NC $\uparrow$ & Matched objects $\uparrow$ \\
\midrule
Opus 5.5 agent (per-scene opt.) & 13.9 to 24.4 min & \$3.9 to \$9.2 & 0.805 & 0.445 & 1.96 & 0.557 & 0.799 & 119 \\
\model{} (feed-forward) & \textbf{16 to 21 s} & \textbf{About \$0.01} & 0.627 & 0.468 & 4.31 & 0.353 & 0.750 & 89 \\
\model{} + Opus 5.5 agent (combined) & 17.5 to 28.0 min & \$5.7 to \$10.7 & 0.828 & 0.435 & 1.74 & 0.597 & 0.803 & 117 \\
\midrule
Opus 5.5 agent (7 min) & 7 min & \$1.9 to \$2.7 & 0.353 & 0.208 & 2.59 & 0.431 & 0.804 & 47 \\
\model{} + Opus 5.5 agent (7 min) & 7 min & \$1.5 to \$1.8 & 0.642 & 0.337 & 3.01 & 0.381 & 0.759 & 90 \\
\bottomrule
\end{tabular}}
\caption{
\textbf{Quantitative comparison against an agentic baseline.} 
\model{} runs in seconds instead of minutes per scene, while the per-scene agent trades runtime for accuracy.
The combined method starts the agent from the output of \model{}, and its time and cost include those of \model{}.
CD, F1 and NC are computed over \textbf{the matched objects only}. 
The matched objects are summed over the 4 scenes (136 ground-truth objects in total).
The mIoU of \model{} is computed from predicted masks, and that of the other methods from predicted boxes.
}
\label{tab:agent_cmp}
\end{table}

\begin{table*}[ht]
    \centering
    \setlength{\tabcolsep}{1pt}
    \renewcommand{\arraystretch}{0.5}
    \resizebox{\textwidth}{!}{%
    \begin{tabular}{cccccc}
        \includegraphics[width=0.3\linewidth]{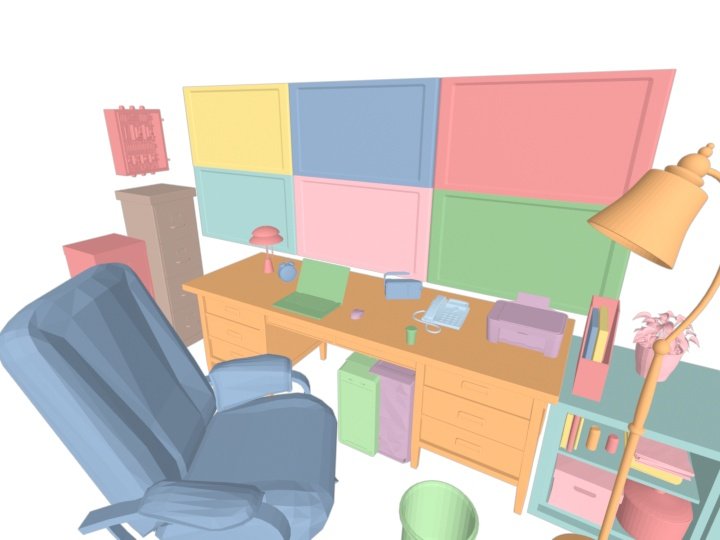} &
        \includegraphics[width=0.3\linewidth]{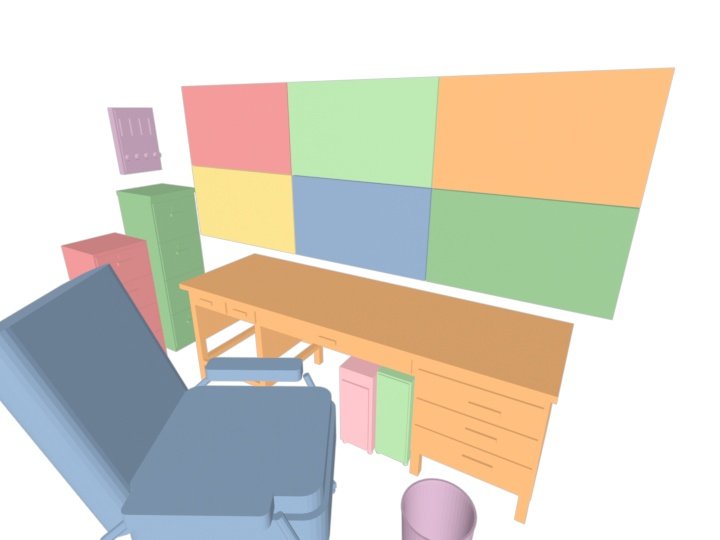} &
        \includegraphics[width=0.3\linewidth]{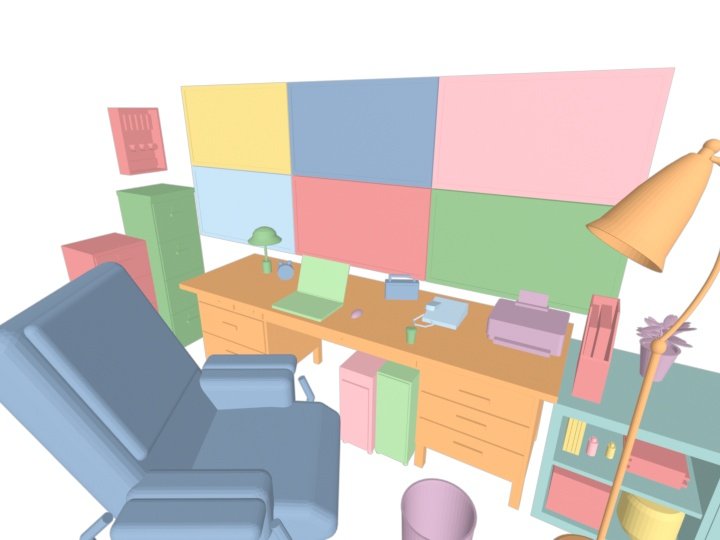} &
        \includegraphics[width=0.3\linewidth]{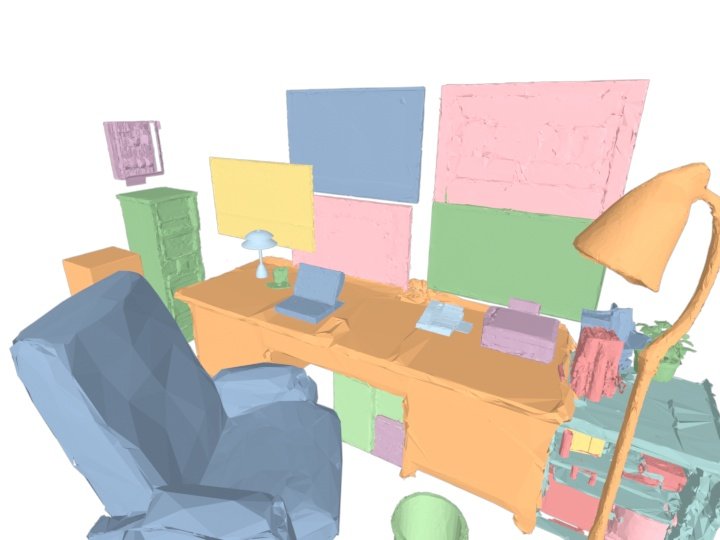} &
        \includegraphics[width=0.3\linewidth]{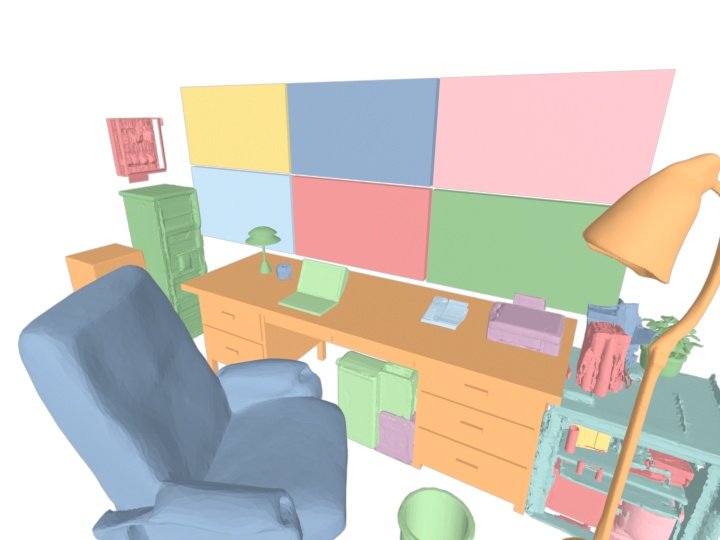} &
        \includegraphics[width=0.3\linewidth]{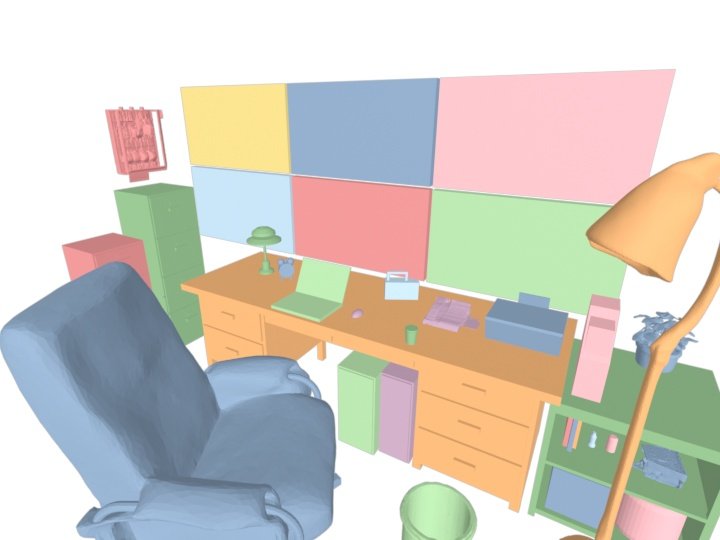} \\[1pt]
        \includegraphics[width=0.3\linewidth]{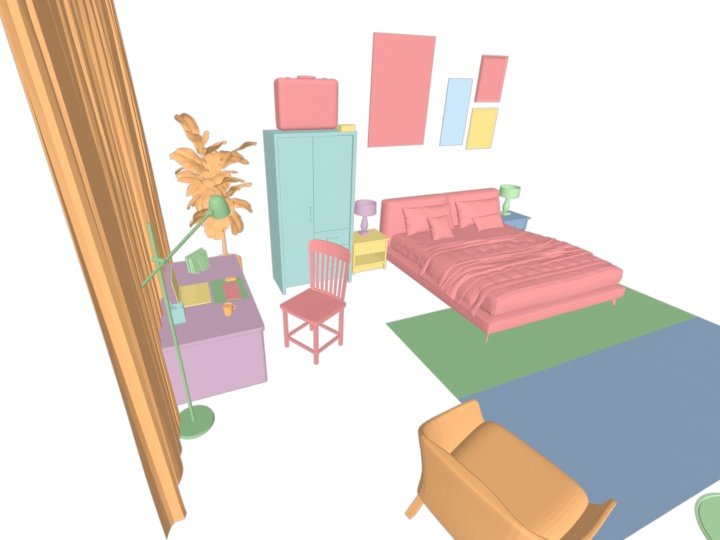} &
        \includegraphics[width=0.3\linewidth]{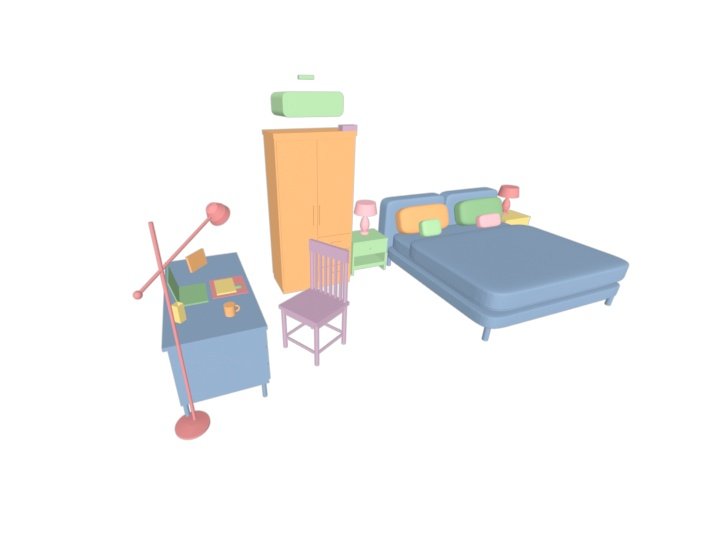} &
        \includegraphics[width=0.3\linewidth]{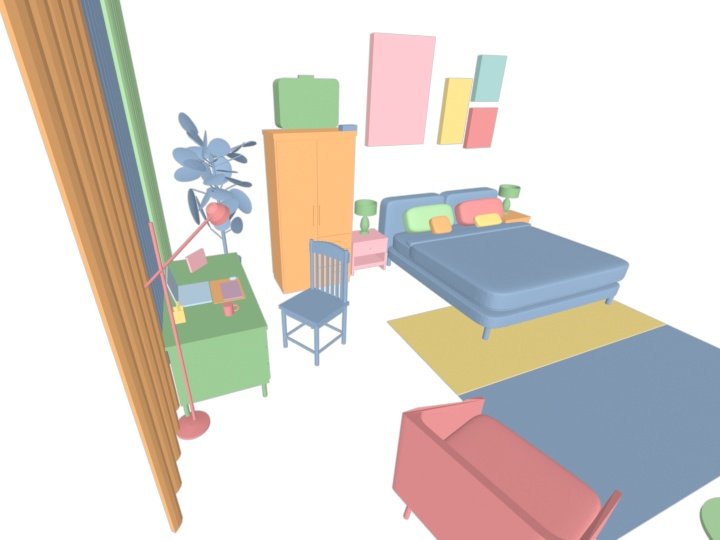} &
        \includegraphics[width=0.3\linewidth]{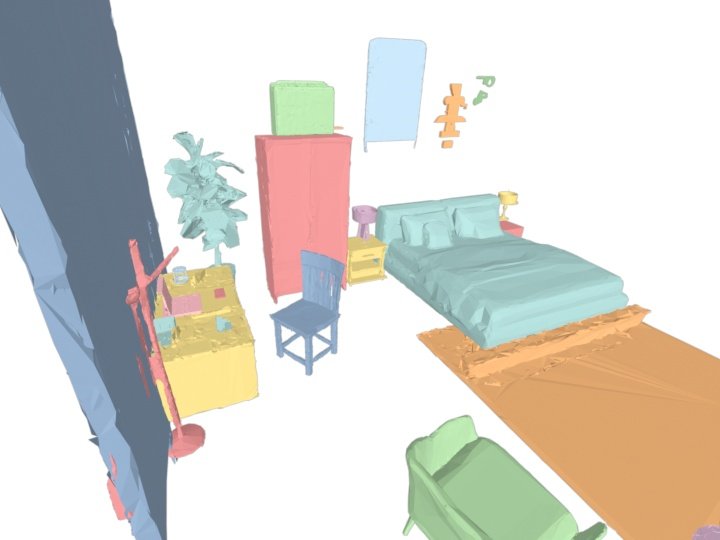} &
        \includegraphics[width=0.3\linewidth]{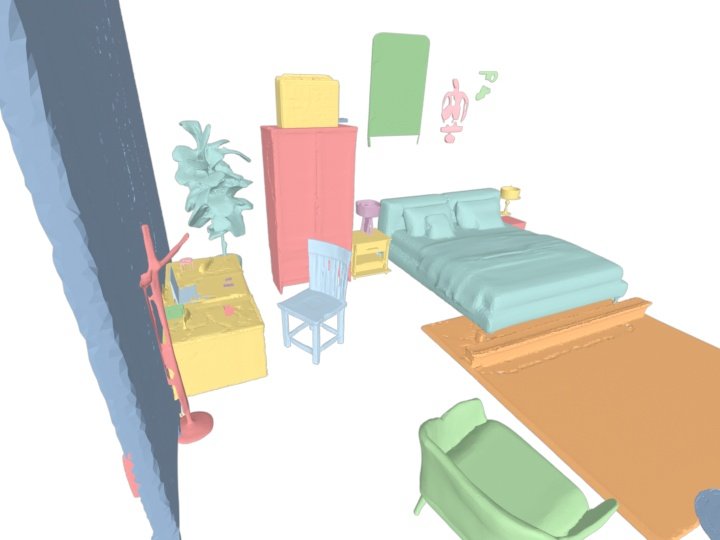} &
        \includegraphics[width=0.3\linewidth]{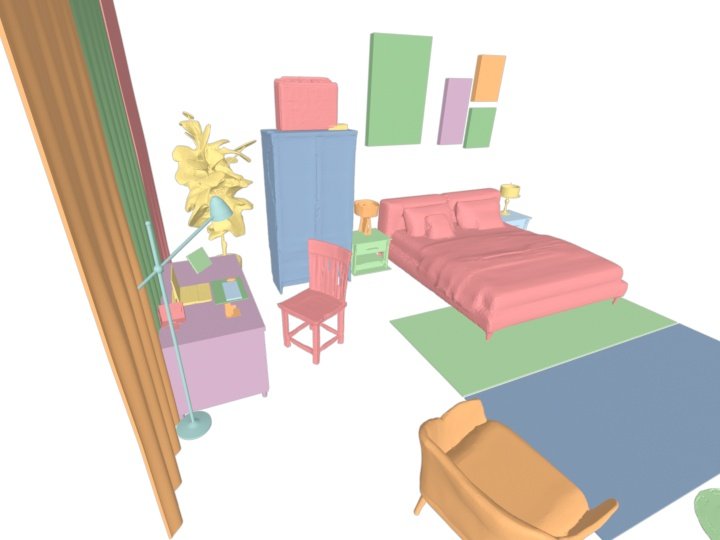} \\[1pt]
        GT & Agent (7 min) & Agent (final) & \model{} & \model{} + Agent (7 min) & \model{} + Agent (final)
    \end{tabular}%
    }
    \captionof{figure}{
        \textbf{Qualitative comparison of the Opus 5.5 Agent with and without \model{}.}
        At 7\,min of total time per scene, the agent alone has built only part of the scene (e.g., no curtains, rug, armchair or wall art in the second row), whereas starting from \model{} gives a complete scene, some of whose objects are still unrefined meshes of \model{}.
        At the end, the agent keeps the meshes of \model{} for organic shapes such as the office chair and the bed.
        We show the geometry with one color per object instance.
        Background is omitted for all columns.
    }
    \label{fig:agent_cmp}
\end{table*}

\begin{figure}[htb]
  \centering
  \begin{minipage}{0.70\columnwidth}
    \centering
    \resizebox{\linewidth}{!}{%
    \begin{tabular}{ll|cc|ccc|ccc}
    \toprule
    \multirow{2}{*}{Method} & \multirow{2}{*}{Dataset}
      & \multicolumn{2}{c|}{Representation}
      & \multicolumn{3}{c|}{Geometry}
      & \multicolumn{3}{c}{Rendering} \\
    \cmidrule(lr){3-4}\cmidrule(lr){5-7}\cmidrule(lr){8-10}
     & & Res. & Feat. Dim. & CD \revten{(cm)} & F1 & NC & PSNR \revten{(dB)} & SSIM & LPIPS \\
    \midrule
    SC-VAE only \rebut{\cite{xiang2026native}} & \multirow{4}{*}{Toys4K\cite{toys4k}} & 32 & 32 & 0.261 & 0.997&0.965&26.801&0.955&0.056 \\
    SC-VAE + HC-VAE (Ours)      &                         & 8 & 64  & 0.269&0.991&0.943&26.635&0.947&0.065\\
    \rebut{SC-VAE + HC-VAE} & & \rebut{8} & \rebut{32} & \rebut{0.282} & \rebut{0.983} & \rebut{0.927} & \rebut{26.184} & \rebut{0.932} & \rebut{0.081} \\
    \rebut{SC-VAE + HC-VAE} & & \rebut{8} & \rebut{128} & \rebut{0.264} & \rebut{0.994} & \rebut{0.956} & \rebut{26.748} & \rebut{0.952} & \rebut{0.060} \\
    
    \midrule
    
    SC-VAE only \rebut{\cite{xiang2026native}} & \multirow{4}{*}{Imaginarium\cite{imaginarium}} & 32 & 32 & 0.407&0.919&0.957&21.966&0.818&0.270 \\
    SC-VAE + HC-VAE (Ours)      &                         & 8 &  64 & 0.413&0.914&0.946&21.665&0.792&0.305 \\
    \rebut{SC-VAE + HC-VAE} & & \rebut{8} & \rebut{32} & \rebut{0.431} & \rebut{0.902} & \rebut{0.931} & \rebut{21.263} & \rebut{0.768} & \rebut{0.342} \\
    \rebut{SC-VAE + HC-VAE} & & \rebut{8} & \rebut{128} & \rebut{0.409} & \rebut{0.917} & \rebut{0.952} & \rebut{21.854} & \rebut{0.809} & \rebut{0.284} \\
    \bottomrule
    \end{tabular}%
    }
  \end{minipage}
  \hfill
  \begin{minipage}{0.24\columnwidth}
    \centering
    \resizebox{1.0\linewidth}{!}{%
    \begin{tabular}{ccc}
    \adjincludegraphics[trim=0.15\width 0.15\height 0.15\width 0.15\height, clip, width=0.3\linewidth]{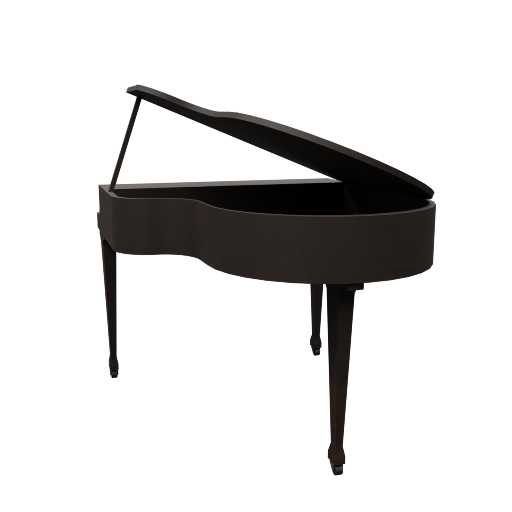} &  
    \adjincludegraphics[trim=0.15\width 0.15\height 0.15\width 0.15\height, clip, width=0.3\linewidth]{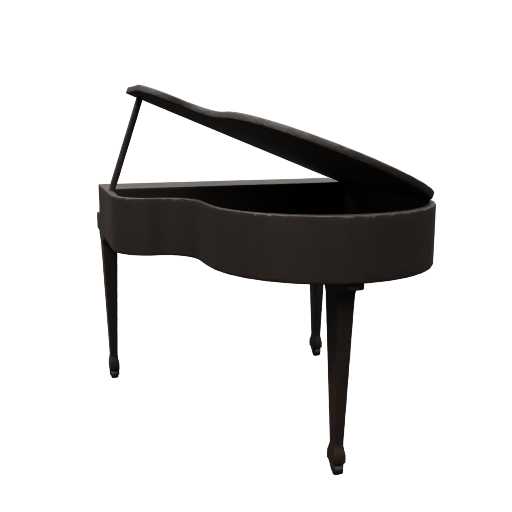} &  
    \adjincludegraphics[trim=0.15\width 0.15\height 0.15\width 0.15\height, clip, width=0.3\linewidth]{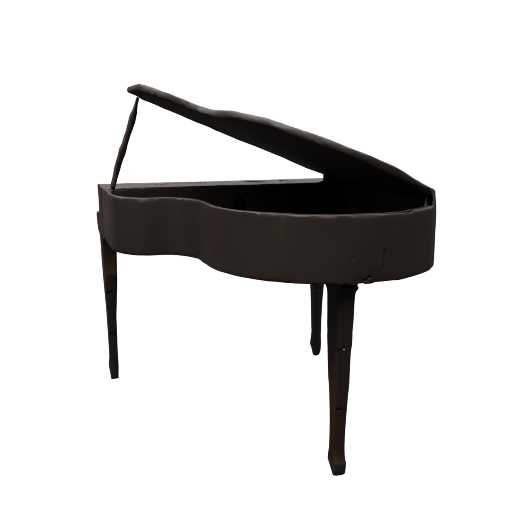} \\
    \adjincludegraphics[trim=0.15\width 0.15\height 0.15\width 0.15\height, clip, width=0.3\linewidth]{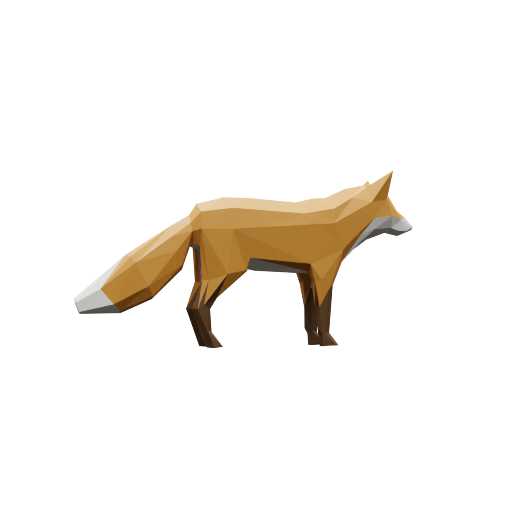} &  
    \adjincludegraphics[trim=0.15\width 0.15\height 0.15\width 0.15\height, clip, width=0.3\linewidth]{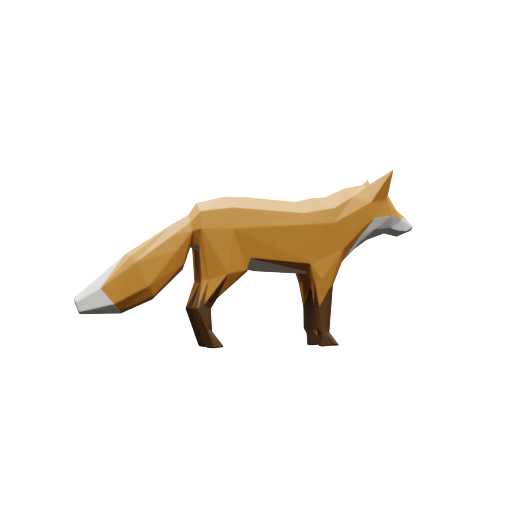} &  
    \adjincludegraphics[trim=0.15\width 0.15\height 0.15\width 0.15\height, clip, width=0.3\linewidth]{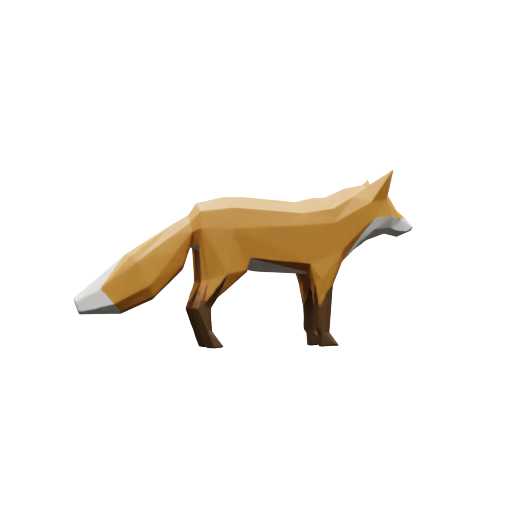} \\
    GT & SC-VAE only \rebut{\cite{xiang2026native}} & Ours \\
    \end{tabular}%
    }
  \end{minipage}
  \captionof{table}{\textbf{Qualitative and Quantitative results on VAE design comparison.} We evaluate the geometry and rendering metrics in Toys4k \cite{toys4k} Object Dataset and Imaginarium \cite{imaginarium} Scene Dataset. Results show that our added HC-VAE above SC-VAE further compresses the latent space and still maintains the high-quality reconstruction. \rebut{Rows with 32 and 128 feature channels vary the HC-VAE channel width.}}
  \label{tab:vae_recon}
\end{figure}

\paragraph{VAE Reconstruction Comparisons} 
We evaluate the reconstruction quality of our added HC-VAE above the original SC-VAE in \rebut{\cite{xiang2026native}} in the Toys4K object \cite{toys4k} dataset as well as the Imaginarium \cite{imaginarium} scene dataset. 
As shown in Tab. \ref{tab:vae_recon}, though we further compress the object latent by $32 \times$, our reconstruction still maintains high quality in terms of geometry and rendering metrics, which enables our model to perform batchified and accelerated inference \rev{with a small change in autoencoding fidelity. This comparison measures encoding--decoding distortion, not the effect of latent resolution on flow generation.} 
\rebut{
As an ablation study, we also vary the HC-VAE channel width, and find that 64 channels gives the best balance. 
Its performance stays close to 128 channels ($16\times$ compression) at half the latent size, while 32 channels ($64\times$) widens the gap to SC-VAE by 1.7 to 4 times on every metric. 
For inference, the width of latent does not set the speed because the flow transformer takes $8^3=512$ tokens per object for every width, and the width only changes its input and output layers.
However, for the flow matching model training, the wider the channel of latent, the harder the training of the flow matching model because it needs the model to learn a mapping in a higher dimensional space.
So choosing latent width with 64 channels is a good trade-off among inference speed, reconstruction quality, and training budget.
}

\Needspace{4\baselineskip}
\paragraph{\rev{Flow generation under equal training compute.}}
\rev{We train flow models directly on uncompressed SC-VAE latents using the same training data and augmentations and the same GPU-hour budget as the HC-VAE flow models. The matched budget covers flow training only, excluding HC-VAE training. Both variants use the same test samples and evaluation protocols as Tab.~\ref{tab:main_cmp}, with GT perception on all three datasets. The uncompressed models have not yet converged at this budget, so the comparison measures quality under equal training compute rather than fully converged performance.}

\begin{table}[ht]
\centering
\small
\setlength{\tabcolsep}{5pt}
\begin{tabular}{llrrrrrr}
\toprule
Dataset & Flow representation & CD \revten{(cm)} $\downarrow$ & F1 $\uparrow$ & NC $\uparrow$ & PSNR \revten{(dB)} $\uparrow$ & SSIM $\uparrow$ & LPIPS $\downarrow$ \\
\midrule
\multirow{2}{*}{ShapeR} & SC-VAE latents & 2.29 & 0.44 & 0.66 & -- & -- & -- \\
& HC-VAE latents & \textbf{1.64} & \textbf{0.52} & \textbf{0.73} & -- & -- & -- \\
\midrule
\multirow{2}{*}{iTHOR} & SC-VAE latents & 3.17 & 0.48 & 0.67 & 17.34 & 0.86 & 0.25 \\
& HC-VAE latents & \textbf{1.38} & \textbf{0.71} & \textbf{0.81} & \textbf{23.85} & \textbf{0.92} & \textbf{0.13} \\
\midrule
\multirow{2}{*}{Imaginarium} & SC-VAE latents & 1.84 & 0.53 & 0.70 & 17.39 & 0.85 & 0.22 \\
& HC-VAE latents & \textbf{1.08} & \textbf{0.68} & \textbf{0.82} & \textbf{20.23} & \textbf{0.89} & \textbf{0.14} \\
\bottomrule
\end{tabular}
\caption{\rev{\textbf{Flow generation with matched training compute.} Both variants use equal GPU-hours for flow training and GT perception. CD is in centimeters. HC-VAE results are from Tab.~\ref{tab:main_cmp}; the uncompressed SC-VAE flow models have not converged within this budget.}}
\label{tab:flow_compute_ablation}
\end{table}

\rev{HC-VAE improves every reported geometry and rendering metric under the matched budget (Tab.~\ref{tab:flow_compute_ablation}). For 12-step classifier-free-guided flow sampling and VAE decoding, the uncompressed models require approximately 4\,s per object, compared with 0.392\,s for HC-VAE (0.238\,s for shape and 0.154\,s for material in Tab.~\ref{tab:runtime_breakdown}), a roughly $10\times$ reduction under the same timing protocol. These times exclude perception and asset export. Compact latents also support concurrent generation of scene objects, making both compute-efficient training and fast batched inference practical for scene reconstruction.}

\Needspace{3\baselineskip}
\paragraph{\rev{Discussion of HC-VAE.}}
\rev{HC-VAE provides a compact representation for efficient batched generation. Its contribution is supported by complementary evidence at the representation, execution, and system levels. First, the $32\times$ reduction in nominal feature capacity retains high autoencoding fidelity (Tab.~\ref{tab:vae_recon}), showing that substantial compression is compatible with detailed geometry and appearance. Second, this representation supports joint generation of multiple objects, and our batched execution produces matching outputs to sequential execution with more than $10\times$ speedup in the tested configuration. Third, the complete system combines fast inference with competitive reconstruction quality (Tabs.~\ref{tab:main_cmp} and~\ref{tab:composed_cmp}). Together, these results establish the practical value of compact latents within our batched reconstruction system. The matched-budget flow ablation (Tab.~\ref{tab:flow_compute_ablation}) adds direct evidence of improved generated outputs at equal flow-training compute. Full-system speedups also depend on perception, batching, and export.}

\WrapNeedspace{166pt}
\paragraph{Runtime Analysis}
{
\begin{wraptable}{r}{0.6\linewidth}
\vspace{-1em}
\small
\centering
\resizebox{\linewidth}{!}{
\begin{tabular}{lclc}
    \toprule
    \textbf{Geometry pathway} & \textbf{s/obj.}
    & \textbf{Texture additions} & \textbf{s/obj.} \\
    \midrule
    Perception       & 0.209 & Texture inference \& dec. & 0.154 \\
    Shape inf. \& geo. dec.  & 0.238 & UV generation w. xatlas   & 1.799 \\
    Topology \& remeshing    & 1.396 & Baking materials        & 0.986 \\
    \textbf{Geometry total} & \textbf{1.844}
    & \textbf{Texture total} & \textbf{4.783} \\
    \midrule
    \multicolumn{3}{l}{\textbf{Network inference total}} & \textbf{0.601} \\
    \midrule
     \multicolumn{3}{l}{\textbf{Post-process total}} & \textbf{4.181} \\
    \bottomrule
\end{tabular}
}
\captionof{table}{\textbf{Per-object runtime breakdown.} \rev{FIRE3D stages start from posed RGB-D; Pi3X is timed separately. Texture total includes the geometry pathway, DINOv3/AnyUp features, and mesh/texture export; Tab.~\ref{tab:main_cmp} reports only network inference.}}
\label{tab:runtime_breakdown}
\vspace{-1em}
\end{wraptable}
\rev{Runtime averages use each comparison's own evaluation subset and processing stages: per scene in Tabs.~\ref{tab:perception} and~\ref{tab:simrecon_cmp}, and per object in Tabs.~\ref{tab:main_cmp},~\ref{tab:litereality_cmp}, and~\ref{tab:runtime_breakdown}. All methods in Tab.~\ref{tab:main_cmp} are timed for network inference only, excluding input preprocessing and asset export; external instance parsing (EFM3D for ShapeR Infer) and SAM3D frame selection are not timed.} We profile several representative scenes across various datasets with results in Tab.~\ref{tab:runtime_breakdown}. 
For geometry-only inference, it requires 1.844s per object, including network prediction and mesh post-processing.
Texture inference and decoding, UV generation, and texture baking increase the end-to-end texture total to 4.783s per object. 
\rev{We measure 1.05\,s for Pi3X network inference on 60 frames. Adding this preprocessing gives a total runtime of 58.4\,s \revten{($12\times4.783+1.05$)} for twelve textured instances, including the background. Direct end-to-end timing with batch size 12 also completes within one minute in this setting, consistent with the per-object average. The geometry-only pathway supports thirty instances within a minute.}\par
}

\paragraph{\rebut{Fine Details under HC-VAE Compression}}
\rebut{
We also study how the HC-VAE compression affects fine details of encoded meshes.
We evaluate this with 100 objects each from Toys4K and Imaginarium.
As Tab.~\ref{tab:hcvae_fine_detail} shows, HC-VAE can preserve geometry well at the scales used in our evaluation.
The F1 with HC-VAE is 0.999 at the 1\,cm threshold of our main metrics and stays above 0.94 at 0.2\,cm. 
Differences appear only at 0.1\,cm and below, mainly in high-curvature regions. 
On Toys4K, NC changes from 0.953 to 0.866 in the 10\% highest-curvature regions and from 0.987 to 0.958 elsewhere. 
In the hardest cases, thin structures such as foliage and sharp edges are slightly smoothed, while the overall shape is kept. 
Fig.~\ref{fig:hcvae_detail} shows qualitative results.
}

\begin{table}[ht]
\centering
\small
\begin{tabular}{lcccc}
\toprule
& \multicolumn{2}{c}{\textbf{Toys4K}} & \multicolumn{2}{c}{\textbf{Imaginarium}} \\
\cmidrule(lr){2-3}\cmidrule(lr){4-5}
\textbf{Metric} & SC-VAE & + HC-VAE & SC-VAE & + HC-VAE \\
\midrule
F1@0.05\,cm $\uparrow$ & 0.964 & 0.747 & 0.941 & 0.796 \\
F1@0.1\,cm $\uparrow$ & 0.985 & 0.869 & 0.986 & 0.906 \\
F1@0.2\,cm $\uparrow$ & 0.996 & 0.943 & 0.999 & 0.964 \\
F1@0.5\,cm $\uparrow$ & 1.000 & 0.990 & 1.000 & 0.994 \\
F1@1\,cm $\uparrow$ & 1.000 & 0.999 & 1.000 & 0.999 \\
99th-pct. GT-to-pred. distance (cm) $\downarrow$ & 0.071 & 0.335 & 0.068 & 0.267 \\
NC, top 10\% curvature $\uparrow$ & 0.953 & 0.866 & 0.961 & 0.907 \\
NC, other regions $\uparrow$ & 0.987 & 0.958 & 0.988 & 0.968 \\
\bottomrule
\end{tabular}
\caption{\textbf{Quantitative results on fine details under HC-VAE compression.} F1 stays above 0.94 at 0.2\,cm and reaches at least 0.99 at 0.5\,cm. Differences appear only at finer scales, mainly in high-curvature regions.}
\label{tab:hcvae_fine_detail}
\end{table}

\begin{table*}[ht]
    \centering
    \setlength{\tabcolsep}{1pt}
    \renewcommand{\arraystretch}{0.5}
    \resizebox{\textwidth}{!}{%
    \begin{tabular}{cccccccc}
        \includegraphics[width=0.122\linewidth]{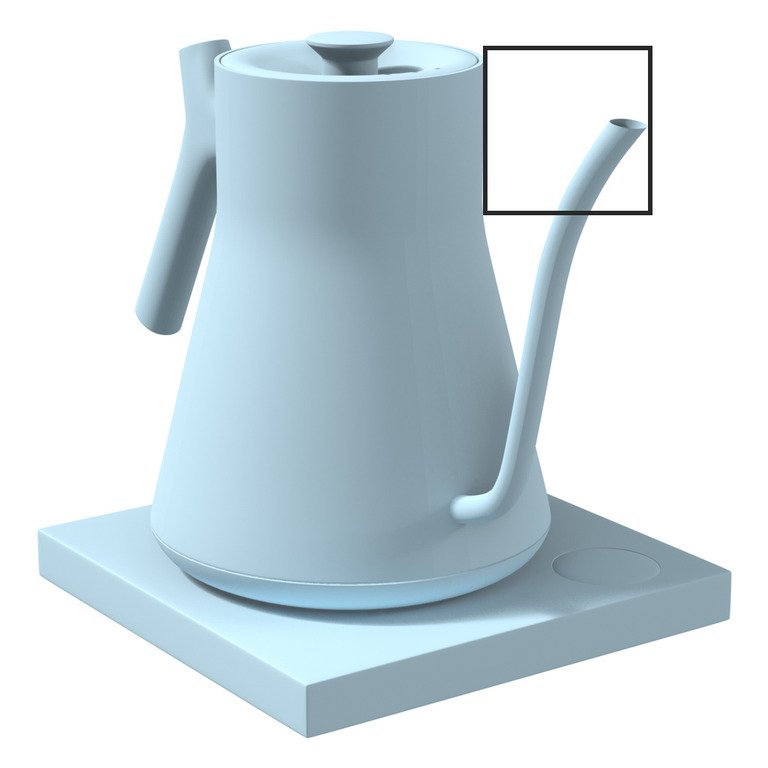} &
        \includegraphics[width=0.122\linewidth]{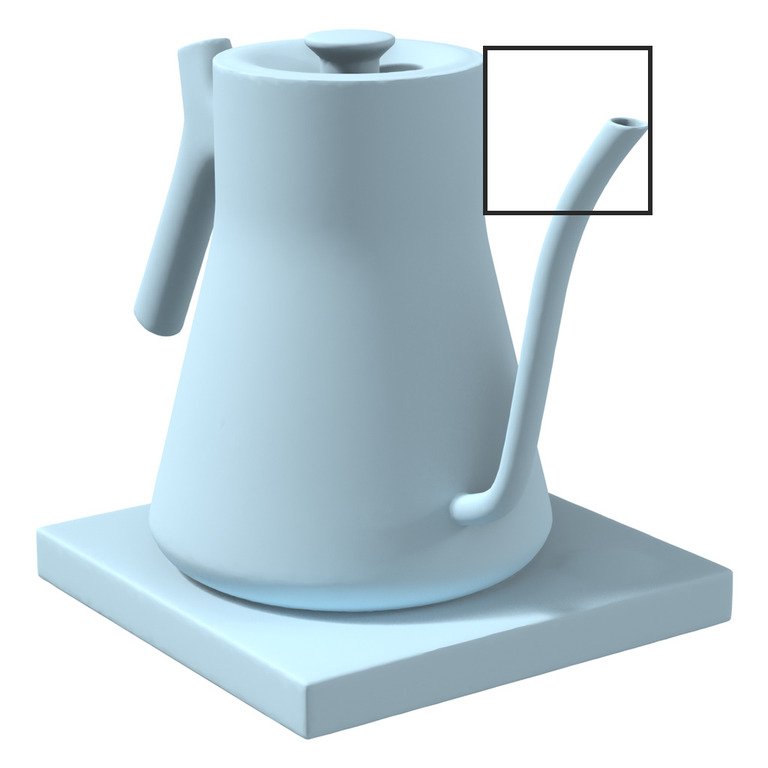} &
        \includegraphics[width=0.122\linewidth]{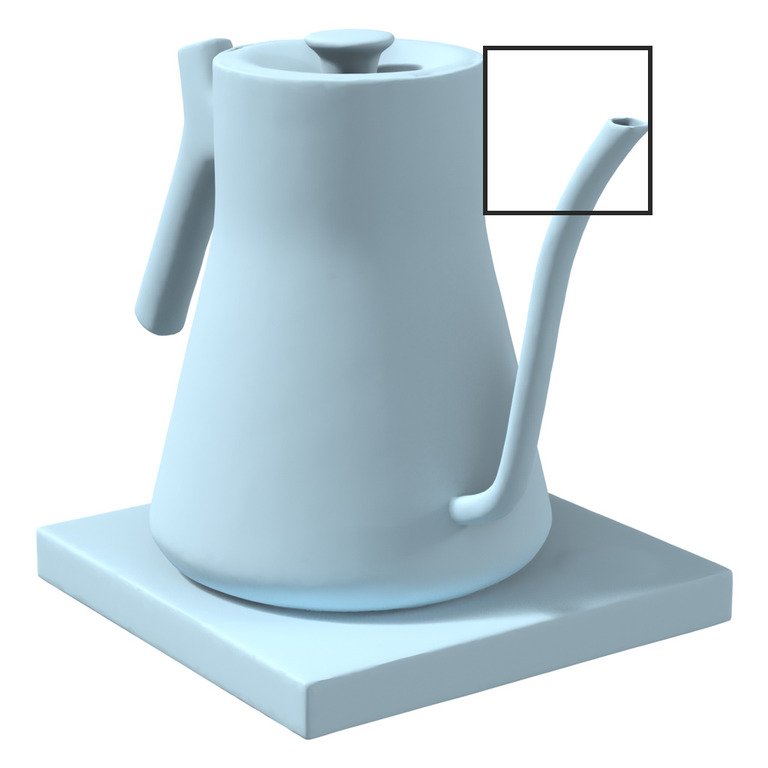} &
        \includegraphics[width=0.122\linewidth]{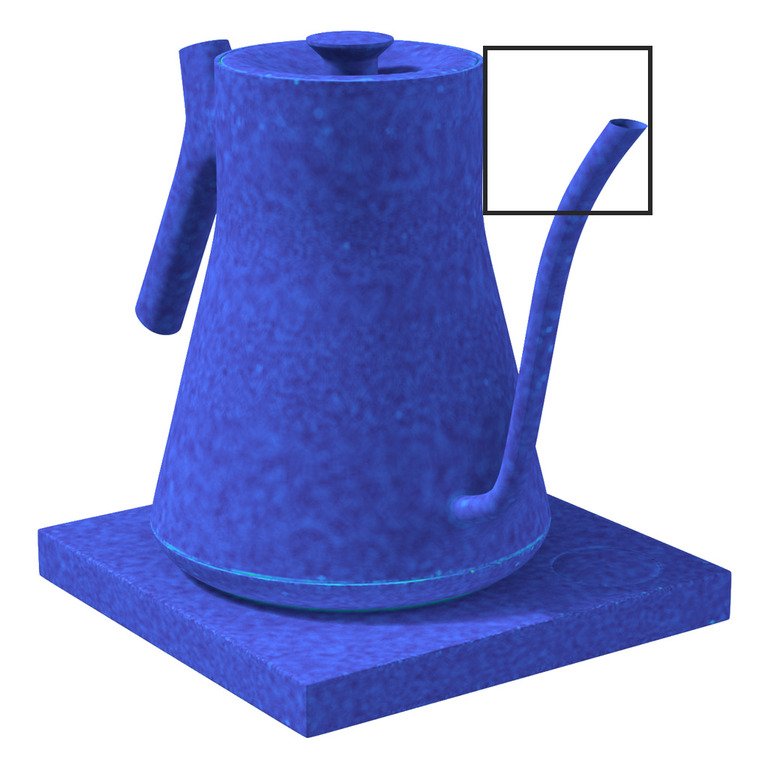} &
        \includegraphics[width=0.122\linewidth]{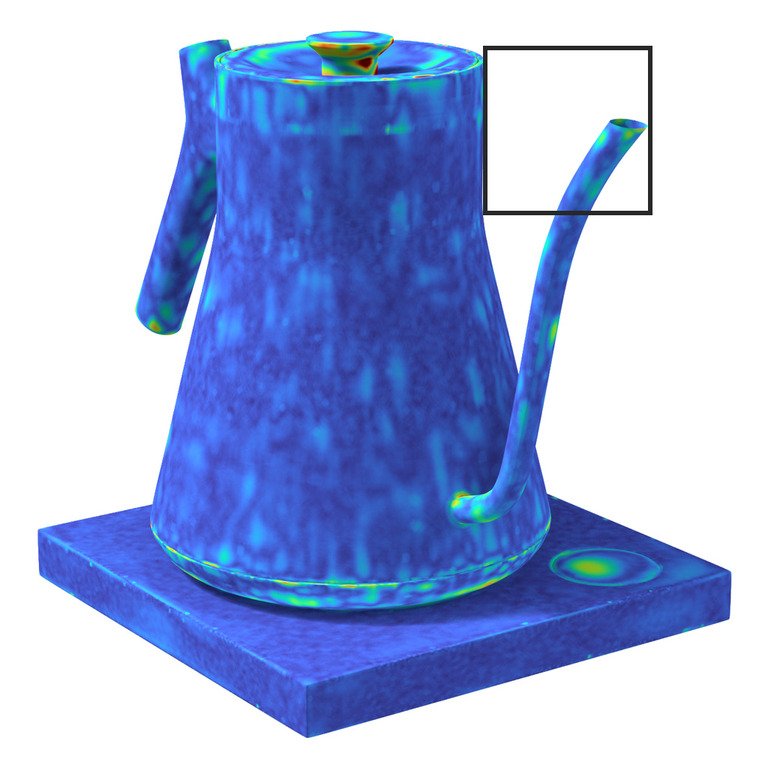} &
        \includegraphics[width=0.122\linewidth]{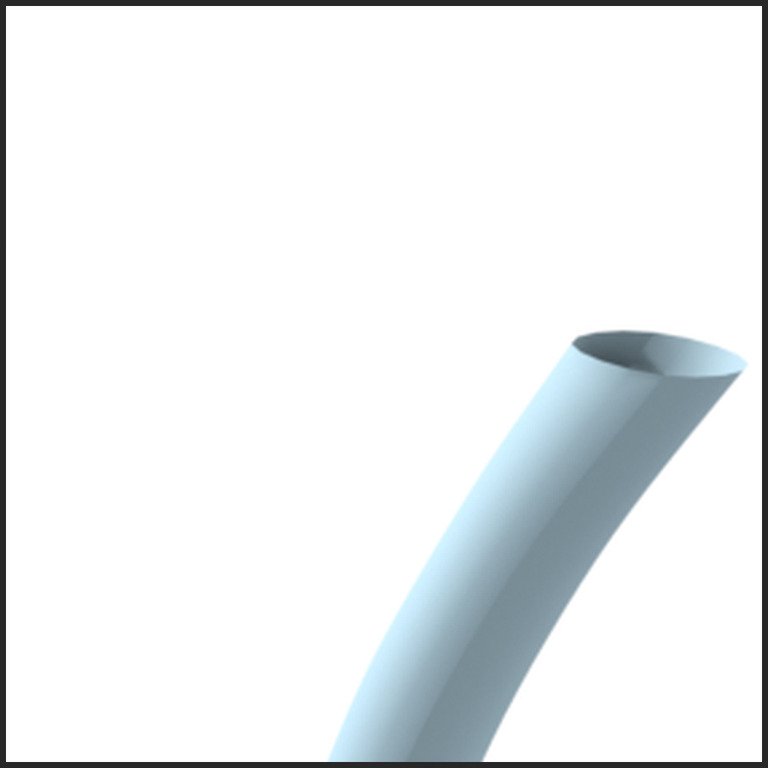} &
        \includegraphics[width=0.122\linewidth]{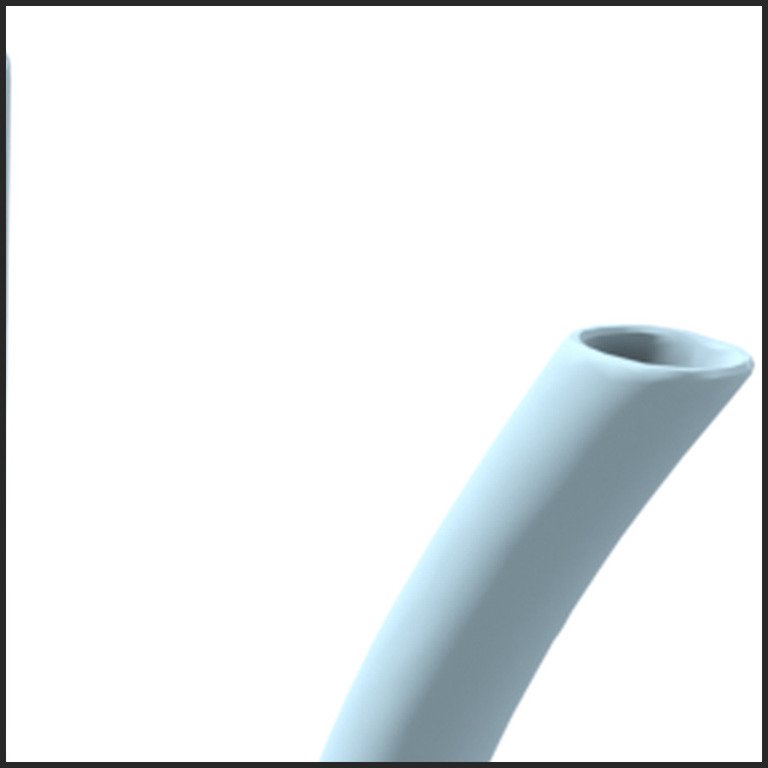} &
        \includegraphics[width=0.122\linewidth]{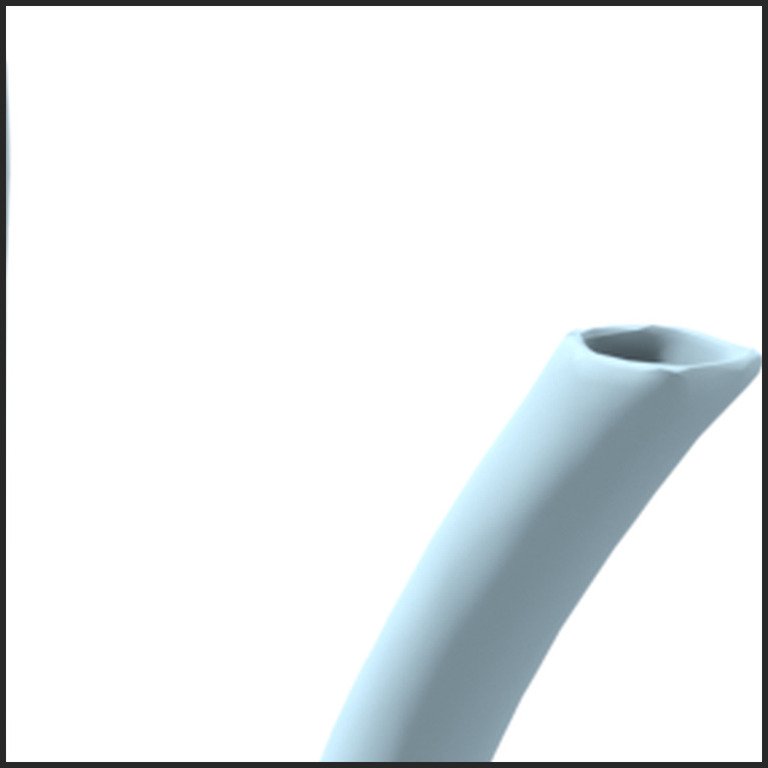} \\[1pt]
        \includegraphics[width=0.122\linewidth]{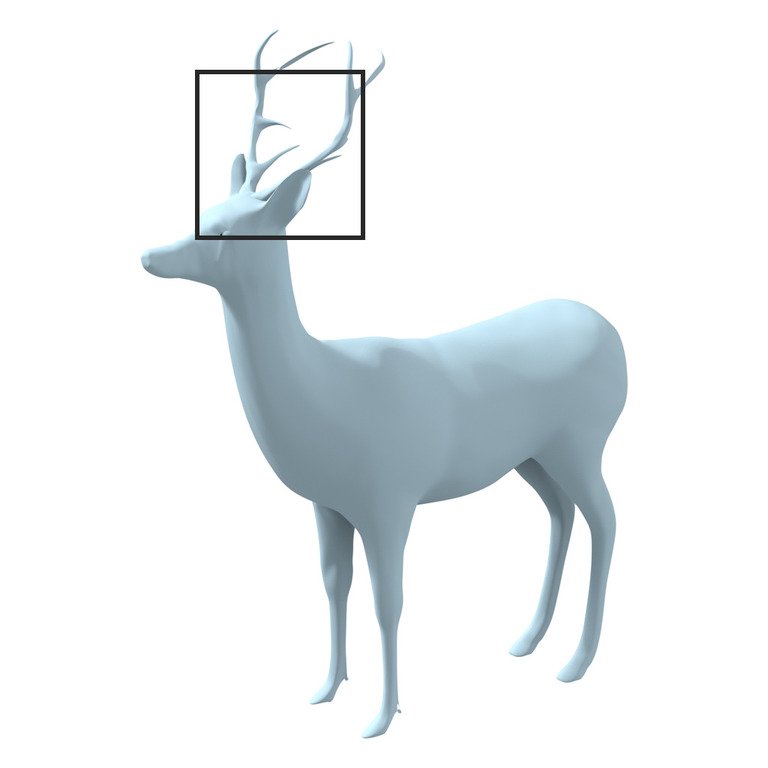} &
        \includegraphics[width=0.122\linewidth]{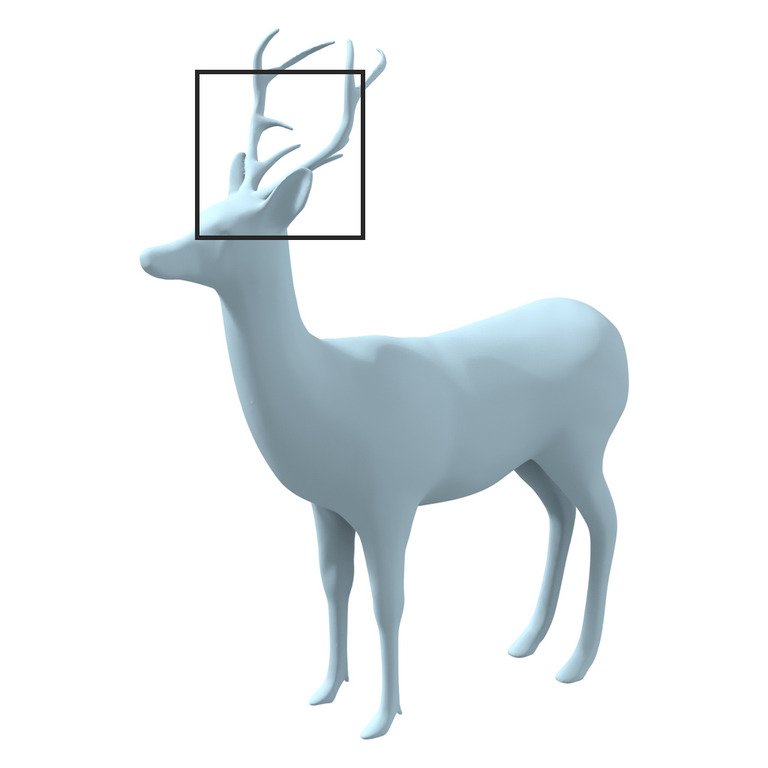} &
        \includegraphics[width=0.122\linewidth]{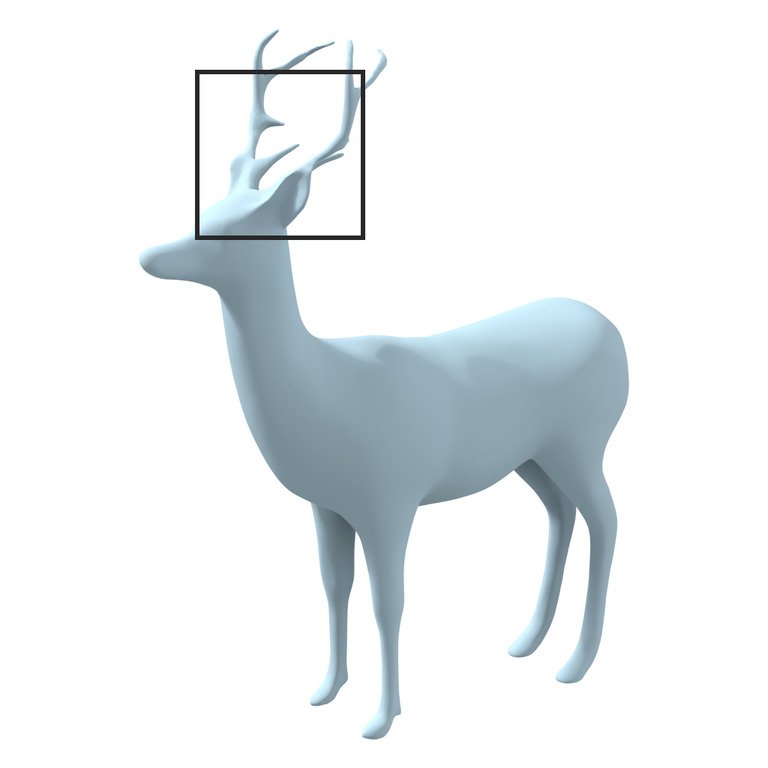} &
        \includegraphics[width=0.122\linewidth]{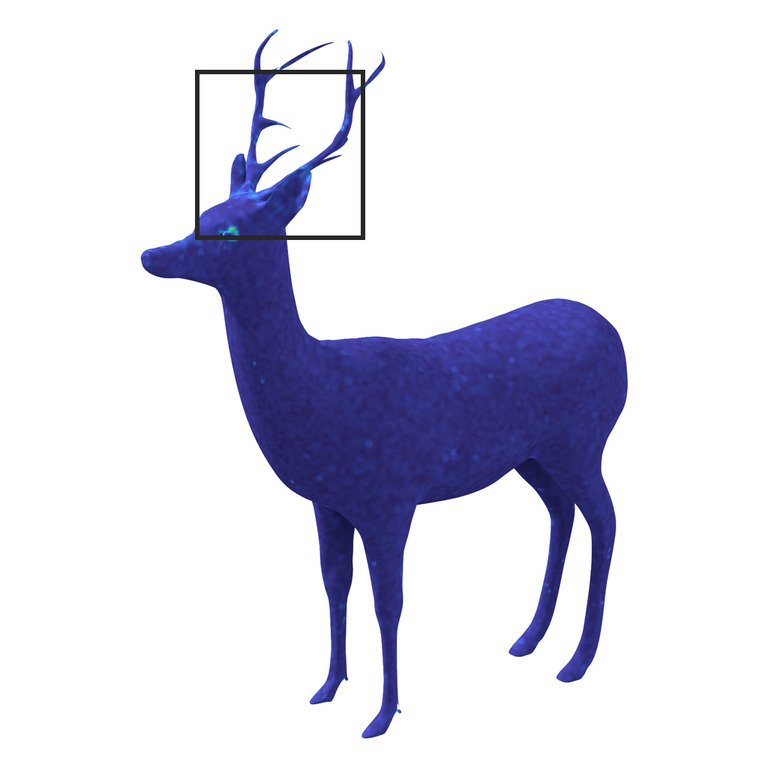} &
        \includegraphics[width=0.122\linewidth]{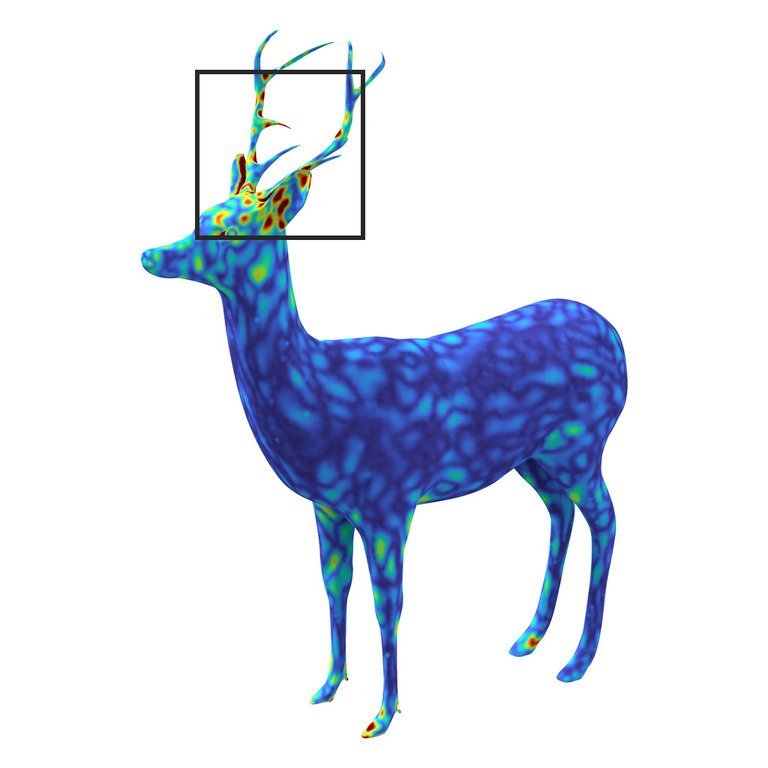} &
        \includegraphics[width=0.122\linewidth]{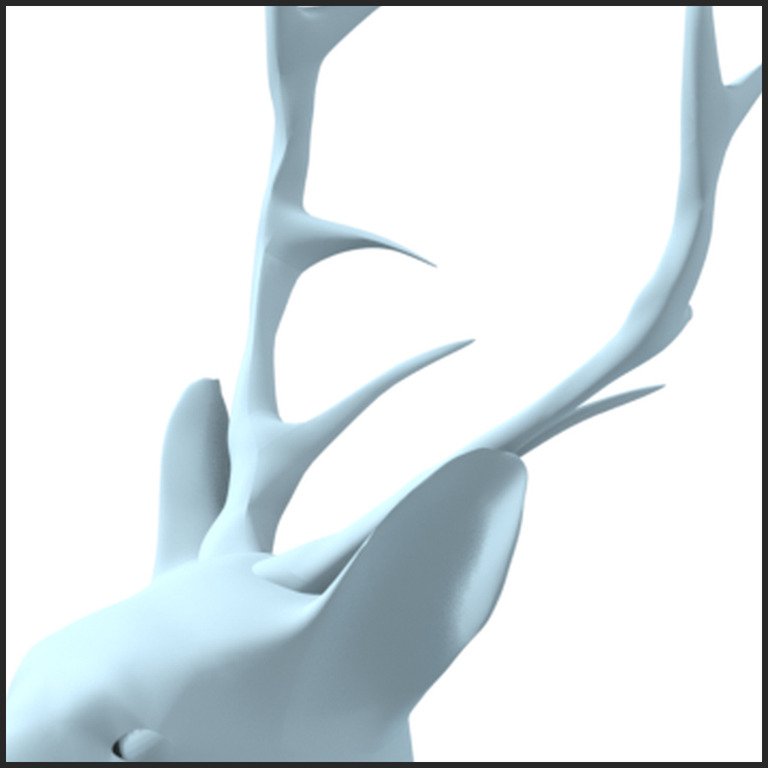} &
        \includegraphics[width=0.122\linewidth]{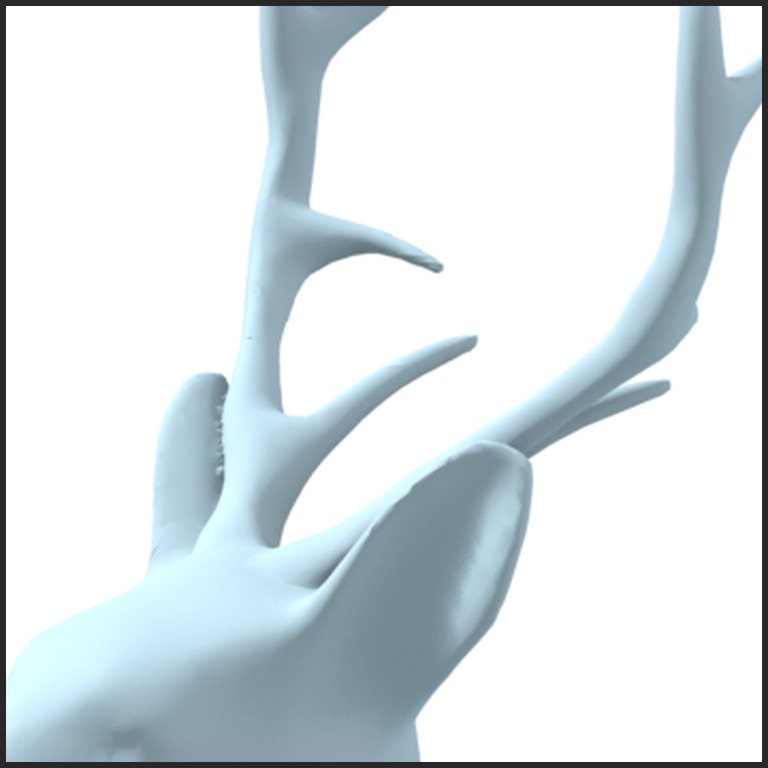} &
        \includegraphics[width=0.122\linewidth]{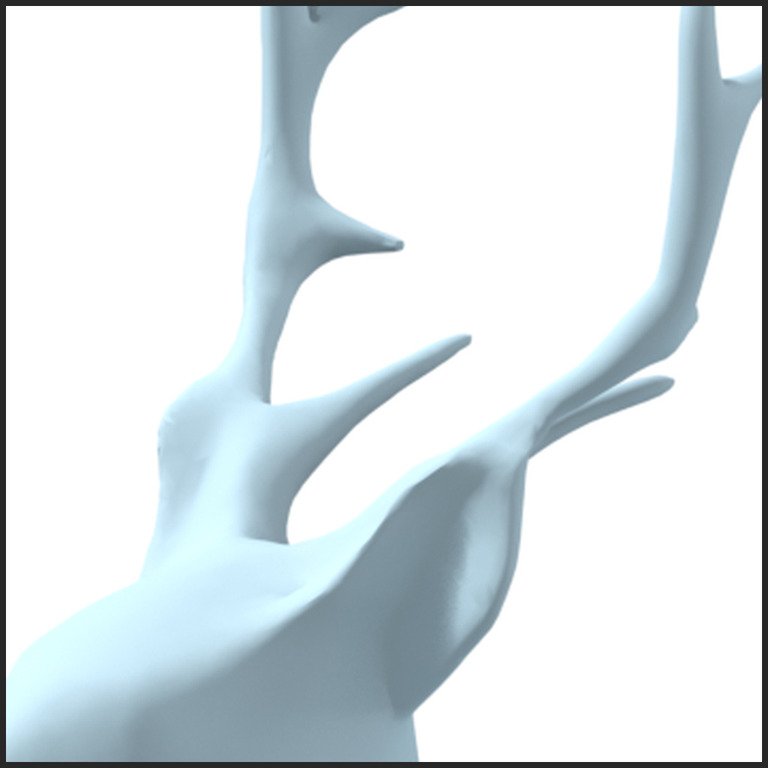} \\[1pt]
        \includegraphics[width=0.122\linewidth]{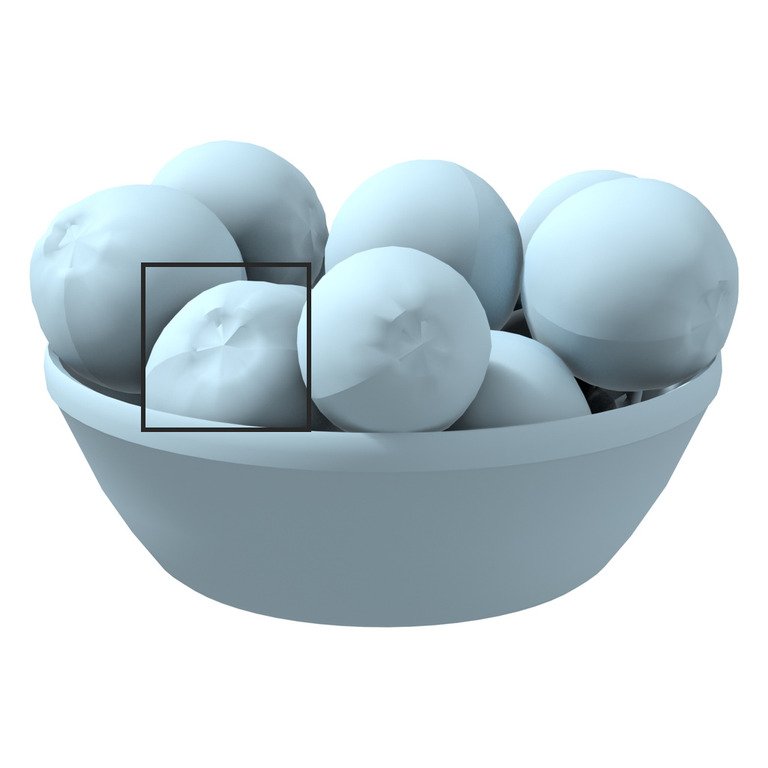} &
        \includegraphics[width=0.122\linewidth]{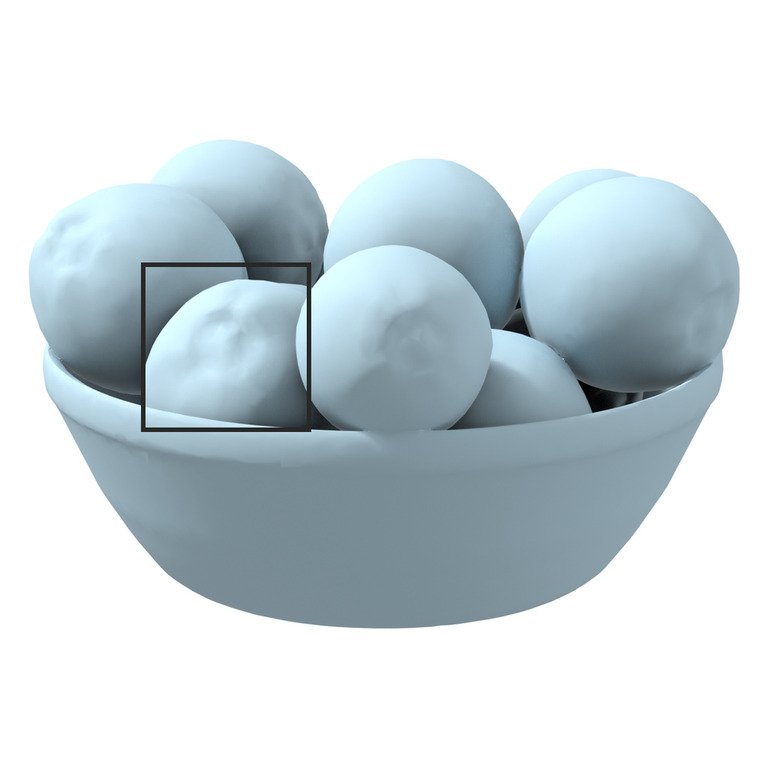} &
        \includegraphics[width=0.122\linewidth]{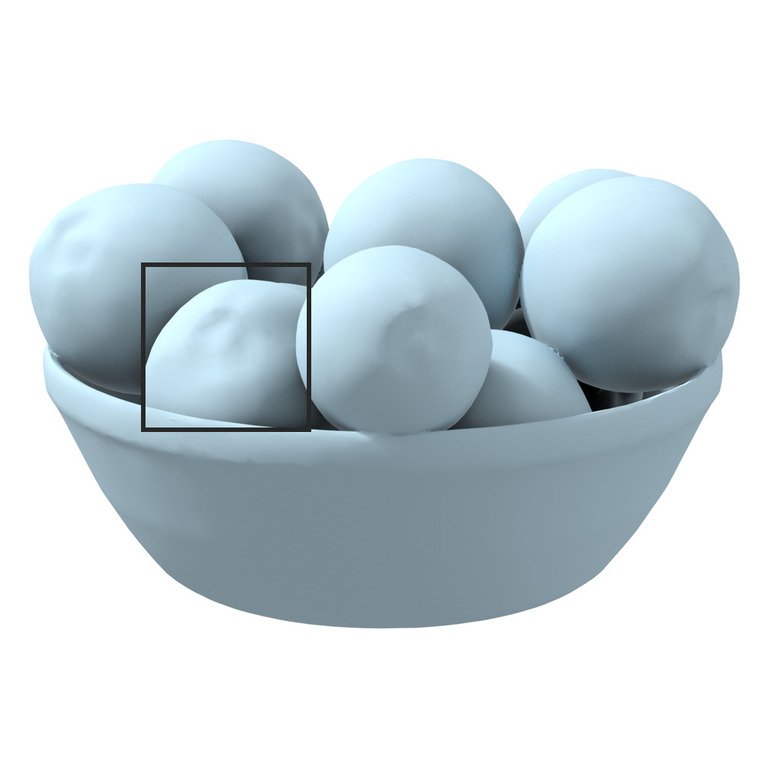} &
        \includegraphics[width=0.122\linewidth]{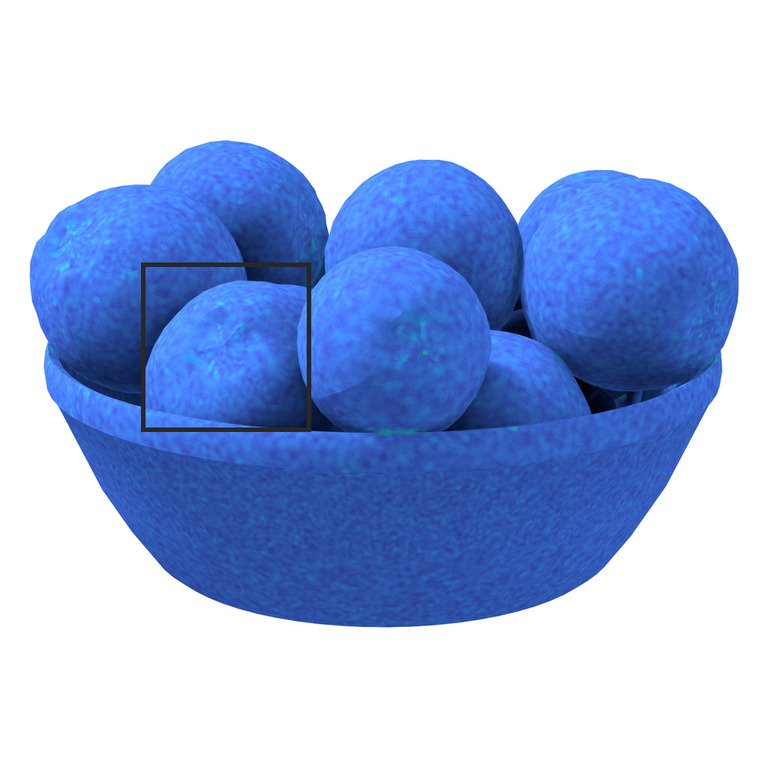} &
        \includegraphics[width=0.122\linewidth]{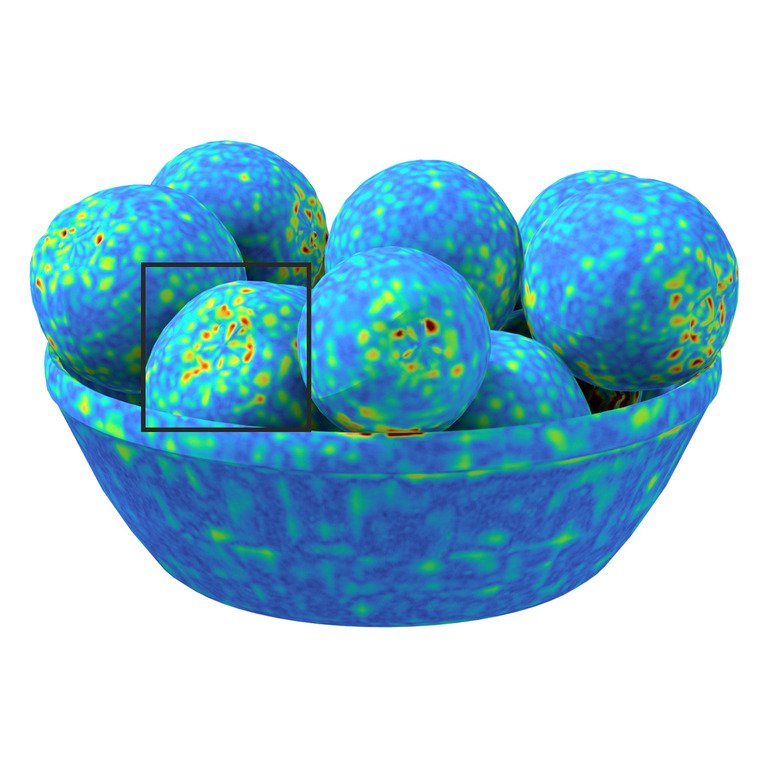} &
        \includegraphics[width=0.122\linewidth]{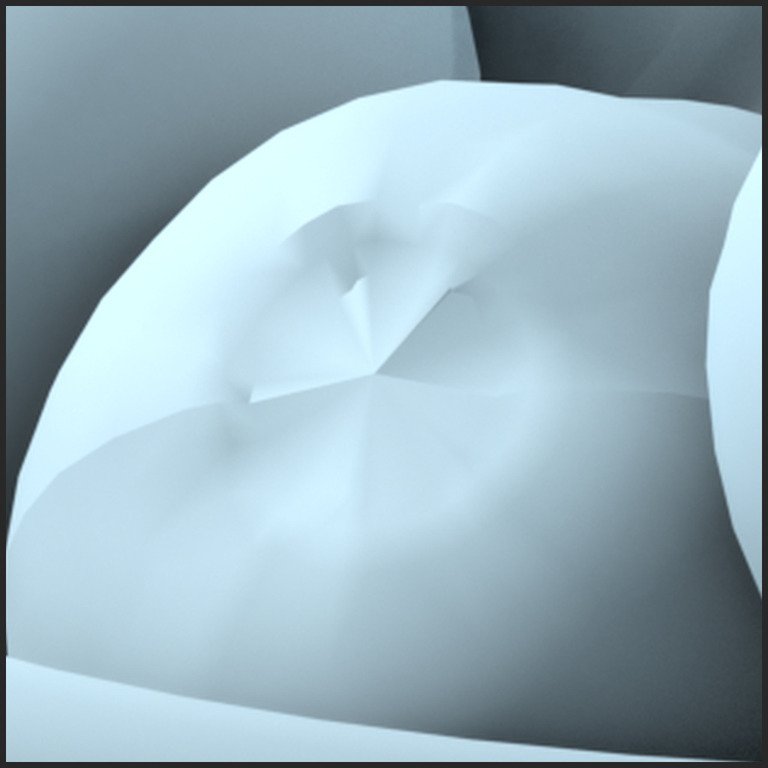} &
        \includegraphics[width=0.122\linewidth]{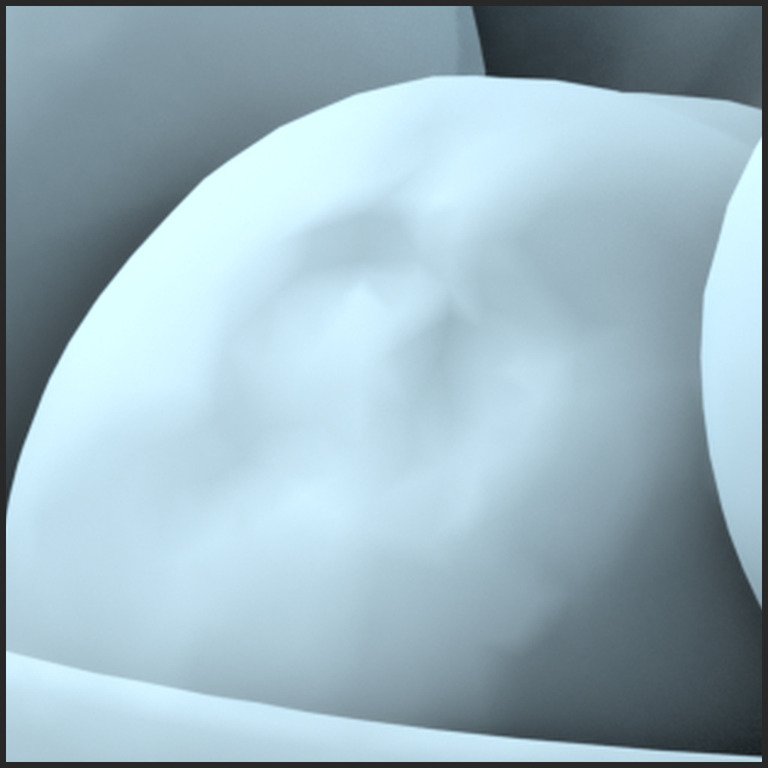} &
        \includegraphics[width=0.122\linewidth]{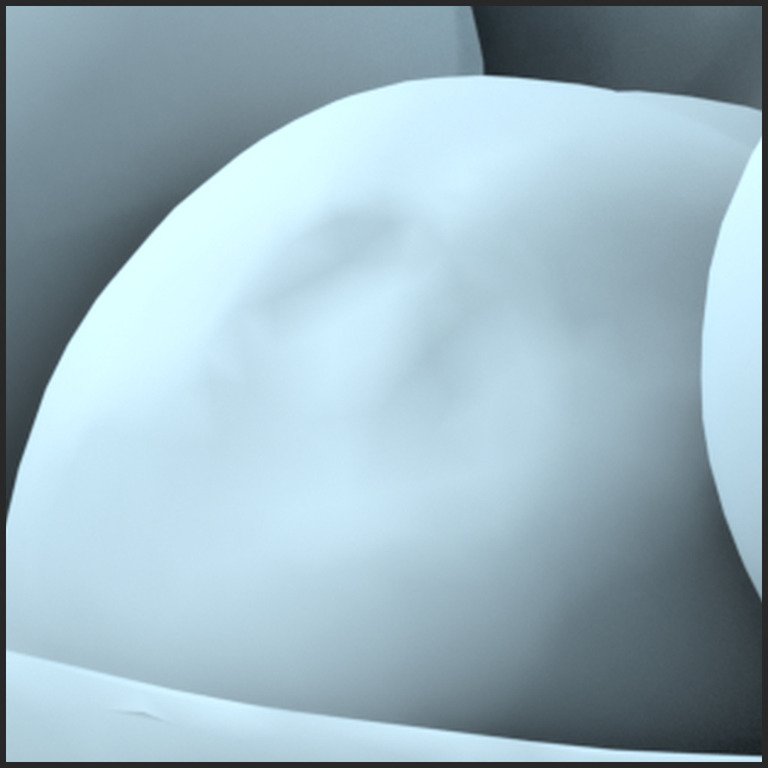} \\[1pt]
        \footnotesize GT & \footnotesize SC-VAE & \footnotesize + HC-VAE & \footnotesize \makecell{SC-VAE\\error} & \footnotesize \makecell{+ HC-VAE\\error} & \footnotesize GT & \footnotesize SC-VAE & \footnotesize + HC-VAE \\[2pt]
        & & & \multicolumn{2}{c}{\includegraphics[width=0.2\linewidth]{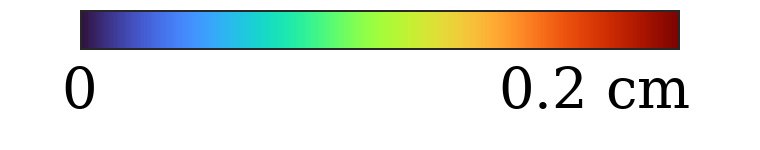}} & & & \\
    \end{tabular}%
    }
    \captionof{figure}{\textbf{Qualitative results on fine details under HC-VAE compression.} HC-VAE keeps the overall shape and thin parts such as the spout and the antlers. Error maps show the distance from the GT surface to each decoder output, and the right three columns enlarge the boxed regions.}
    \label{fig:hcvae_detail}
\end{table*}

\paragraph{\rebut{Limitations and} Failure Modes}\label{sec:limitations}
\model{} can produce errors if the perception doesn't detect objects correctly, which leads to missed objects. 
\rebut{Small objects are the main source of missed detections.} 
Also, point-based conditioning has limitations in perfect object shape and texture reconstructions. 
\rebut{Some finer details are possibly slightly smoothed by HC-VAE. 
Texture realism can be further improved for visual policy learning.} 
See the figure visualizations for details. 
\rebut{
A controlled study is necessary to separate the perception architecture and training data. 
End-to-end training of the perception and generation modules is another promising direction, which we leave for future work.
} 

\section{Implementation Details}

\subsection{Data Preparation Details}
\paragraph{Data curation: }
We build training data from diverse indoor scene datasets and render RGB-D observations from SAGE-10k~\cite{sage}, InternScenes~\cite{internscenes}, ProcTHOR~\cite{ai2thor}, MansionWorld~\cite{mansion}, and SceneSmith~\cite{scenesmith}, totaling \rev{80k scenes and 138,202 rendered videos} with randomized camera intrinsics. We further augment the diversity and realism with Flux.2~\cite{flux2}, which produces \rev{79,672 additional appearance-augmented videos, for 217,874 sequences sharing the underlying scenes}. Additionally, we use an extra 500k objects from four object datasets, including the 3D-Future dataset \cite{3dfuture}, the ABO dataset \cite{abo}, the HSSD dataset \cite{hssd}, and the Objaverse dataset \cite{objaverse}, in the flow matching reconstruction model training to further enhance its capability.

\paragraph{Data rendering: }
With the collected data of abundant indoor scenes, we leverage Blender to render RGB-D videos together with camera intrinsics and poses inside the rooms.
We design a heuristic algorithm to automatically generate a camera trajectory inside the room, and use Blender EEVEE and CYCLES renderers to render the videos with added lights.
The image resolution is fixed to 512x512, and camera intrinsics are chosen randomly with a FOV from 40 degrees to 90 degrees. {Each frame therefore provides RGB, metric depth, camera intrinsics, and a camera-to-world pose for constructing the posed RGB-D observation.}
Through this process, we totally rendered 138202 videos.
For datasets with diverse objects and layouts such as SAGE-10k \cite{sage}, \rev{SceneSmith \cite{scenesmith}}, and InternScenes \cite{internscenes}, we render 3 videos per scene. 
For other datasets, MansionWorld and ProcTHOR \cite{mansion, ai2thor}, we only render 2 and 1 video per room.

\subsection{Data Augmentation Details}
During training, we apply data augmentations to the scene renderings to boost the generalizability of the trained models. 
\paragraph{Scene rotation}
We apply random rotations of 90, 180, and 270 degrees to the whole scene along the z-axis. This can help the model learn the orientations of objects during the training of the perception model.
\paragraph{Noise-based augmentation}
In realistic capturing from the real world, the {camera poses and depth estimates obtained through geometric preprocessing} \cite{schoenberger2016sfm,vggt,wang2026pi}  {are} not perfectly accurate. {In our practical configurations, we use COLMAP poses with \rev{Pi3X} depth, or \rev{Pi3X} for both poses and depth.} However, in our synthetic rendering, the attained camera poses and depth rendering are too perfect. \rev{This motivates exposing the model to imperfect geometric inputs during training.}
To mitigate this, \rev{we apply Gaussian noise to every training sample, with standard deviations of $0.02$\,m for camera translation and depth, and $0.01^{\circ}$ for camera rotation}, to mimic the {noise in practically estimated posed RGB-D observations}.

\paragraph{Flux-based augmentation}
\begin{table*}[ht]
    \centering

    \setlength{\tabcolsep}{1pt}
    \renewcommand{\arraystretch}{0.5}

    \resizebox{\textwidth}{!}{%
    \begin{tabular}{cccccc}

        \includegraphics[width=0.2\linewidth]{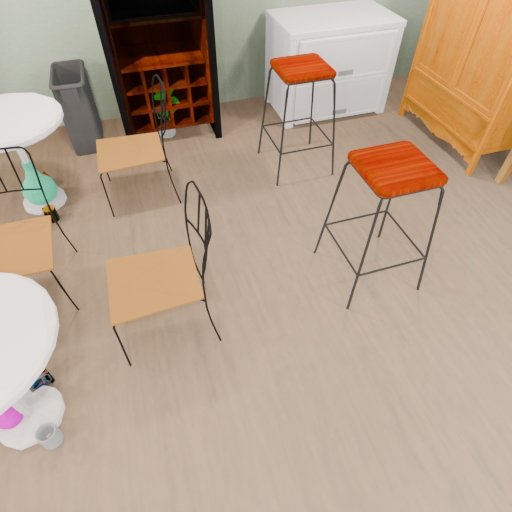} &
        \includegraphics[width=0.2\linewidth]{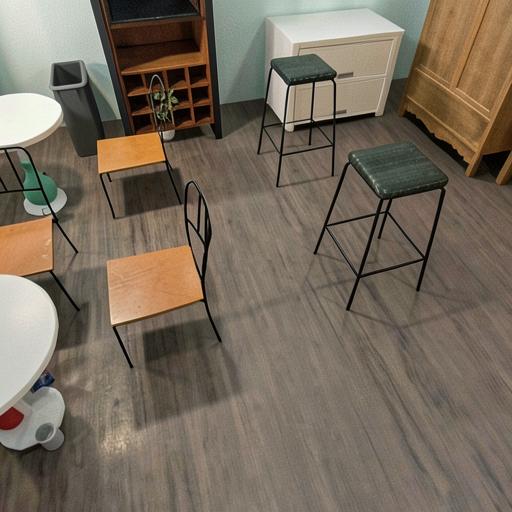} &
        \includegraphics[width=0.2\linewidth]{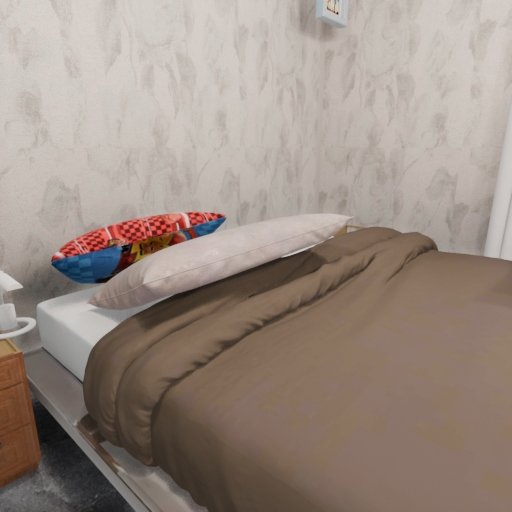} &
        \includegraphics[width=0.2\linewidth]{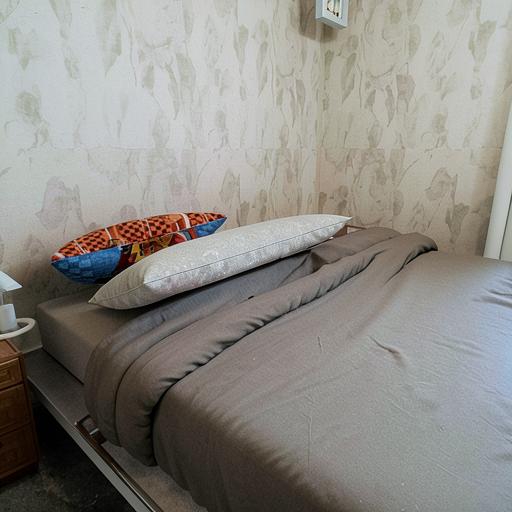} &
        \includegraphics[width=0.2\linewidth]{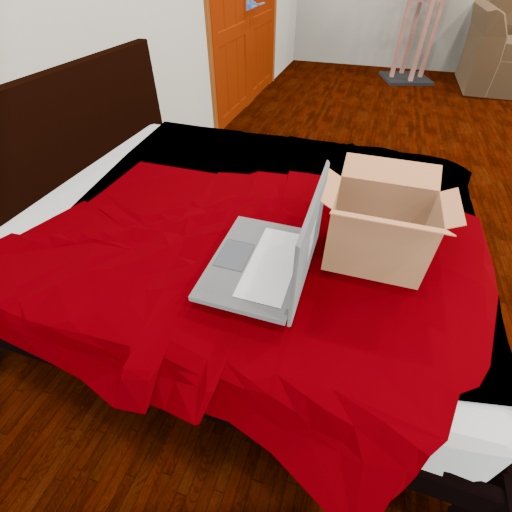} &
        \includegraphics[width=0.2\linewidth]{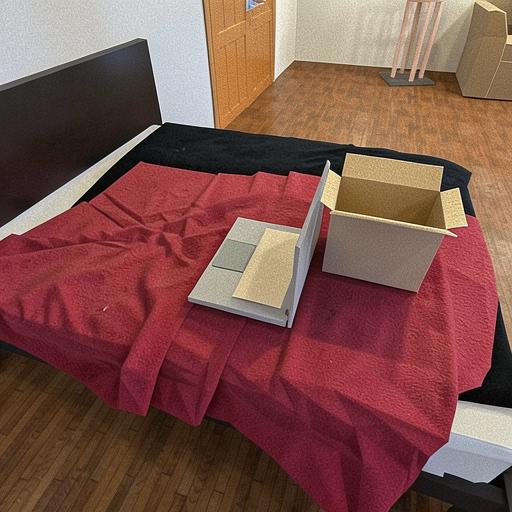} \\
       
        Synthetic & Realistic &
        Synthetic & Realistic & 
        Synthetic & Realistic  \\

    \end{tabular}
    }

    \captionof{figure}{
    \textbf{Visualization of comparisons between synthetic rendering and generated realistic pairs.} 
    Here we showcase the comparison of the original synthetic rendering and the Flux.2 generated realistic images.
    }
    \label{fig:flux}
\end{table*}
To mitigate the gap between synthetic rendering and realistic capturing, we also use the Flux.2 \cite{flux2} to synthesize photorealistic images from synthetic renders, and attain 79672 videos from this process.
The visualizations of comparisons between synthetic and realistic generation can be found in Fig. \ref{fig:flux}. \rebut{To check that the edits stay aligned with the labels, we measure 3,000 training frames from 100 scenes across the five scene datasets. For 97.0\% of instance boundary pixels, the edited frame has an image edge within 2\,px, close to 97.6\% for the original renders, and the median optical flow between each render and its edit is 0.40\,px. Since frames are edited separately, appearance varies across frames, while depth, poses, and instance labels stay consistent because they come from the renders. Fig.~\ref{fig:flux_align} shows qualitative results.}

\begin{table*}[ht]
    \centering
    \setlength{\tabcolsep}{1pt}
    \renewcommand{\arraystretch}{0.5}
    \resizebox{\textwidth}{!}{%
    \begin{tabular}{cccccc}
        \includegraphics[width=0.164\linewidth]{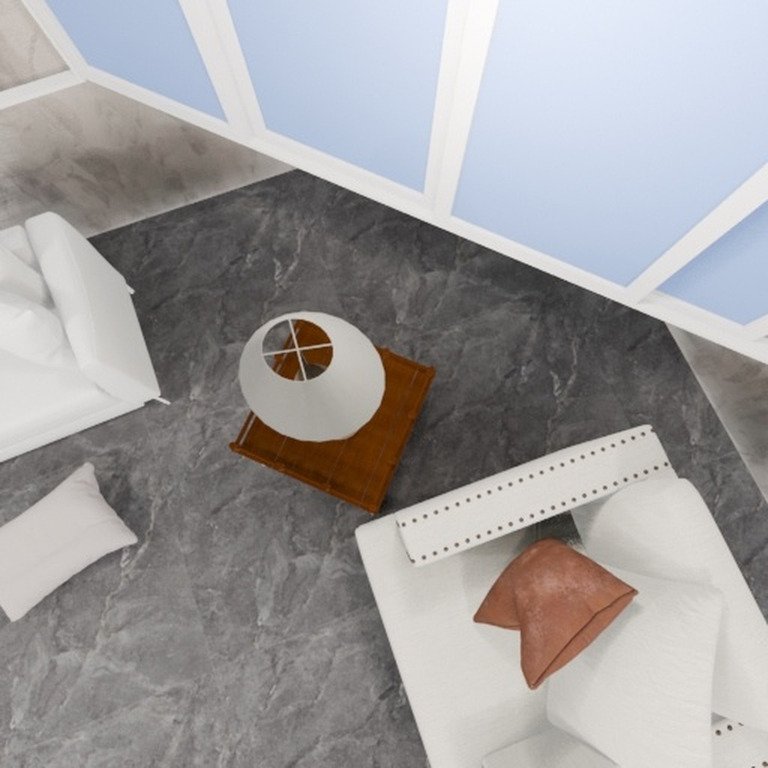} &
        \includegraphics[width=0.164\linewidth]{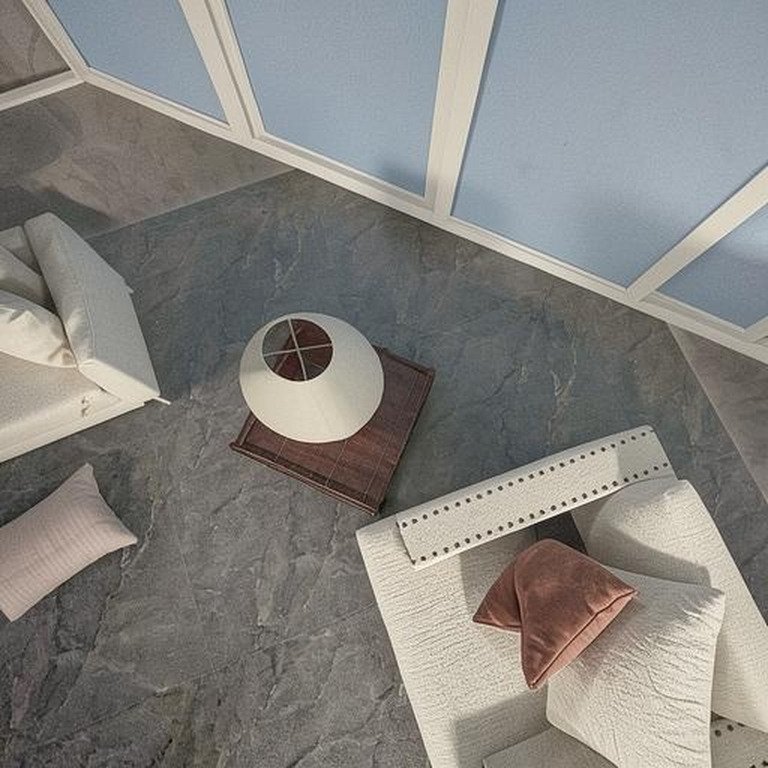} &
        \includegraphics[width=0.164\linewidth]{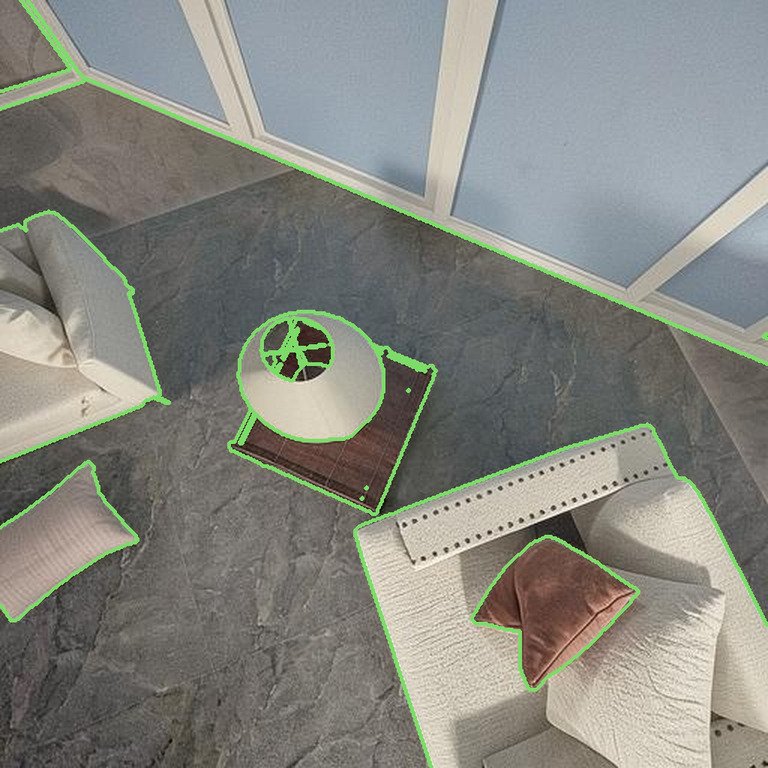} &
        \includegraphics[width=0.164\linewidth]{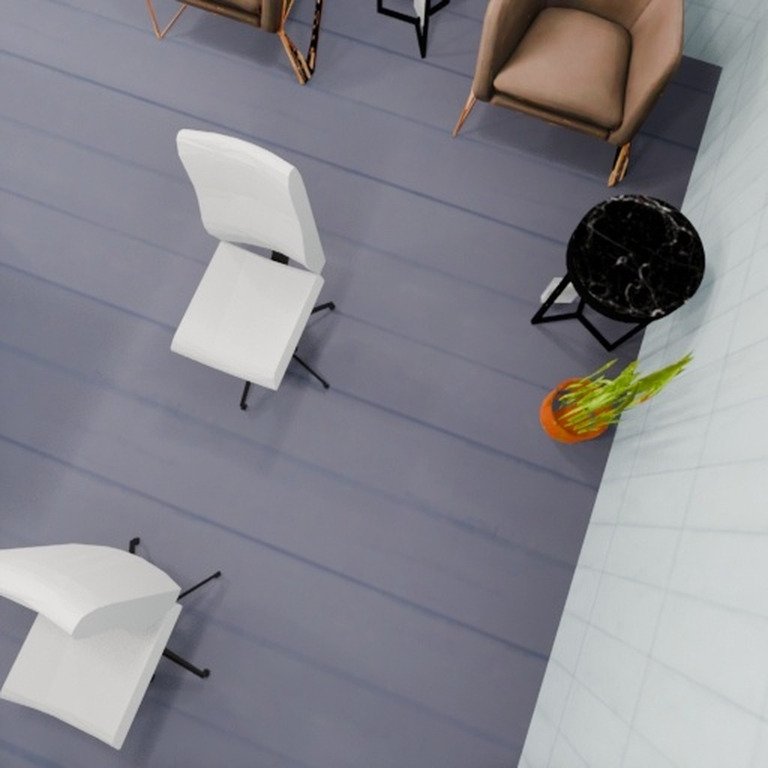} &
        \includegraphics[width=0.164\linewidth]{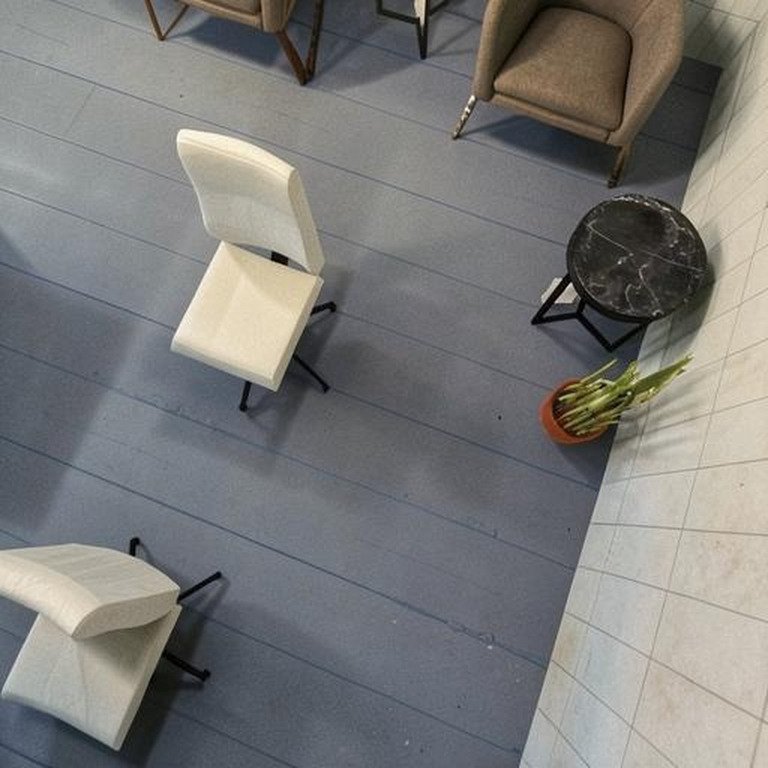} &
        \includegraphics[width=0.164\linewidth]{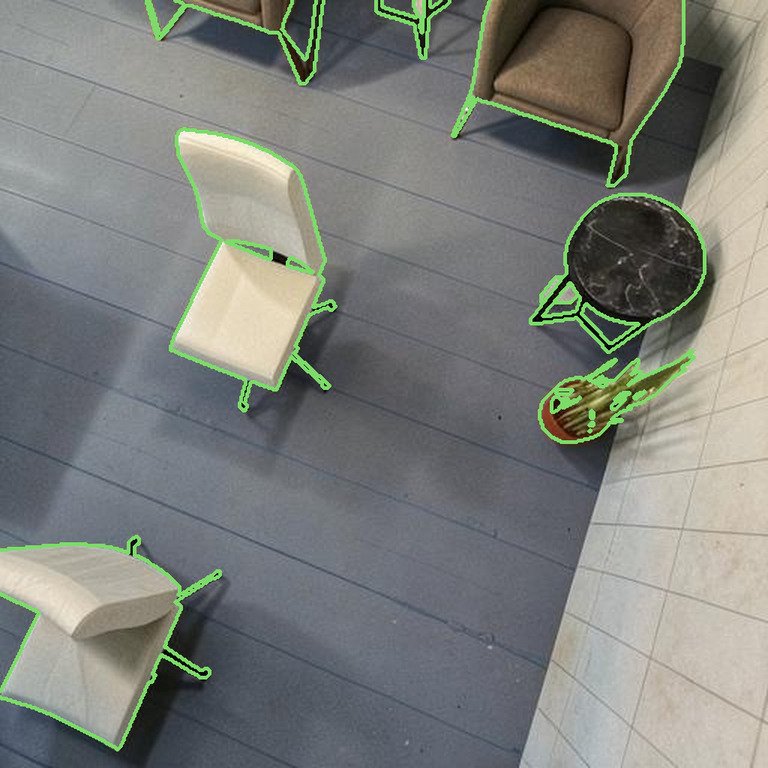} \\[1pt]
        \includegraphics[width=0.164\linewidth]{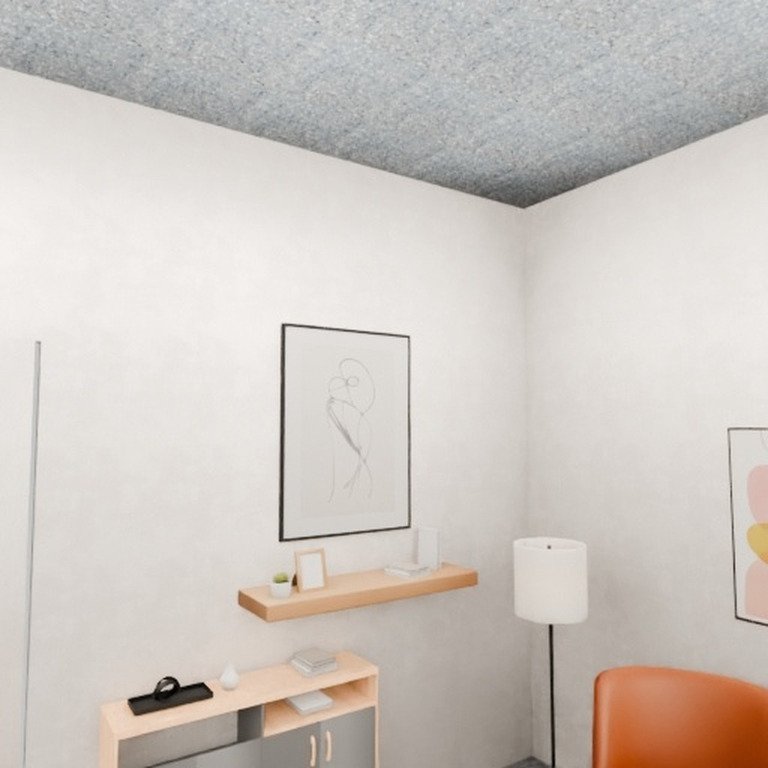} &
        \includegraphics[width=0.164\linewidth]{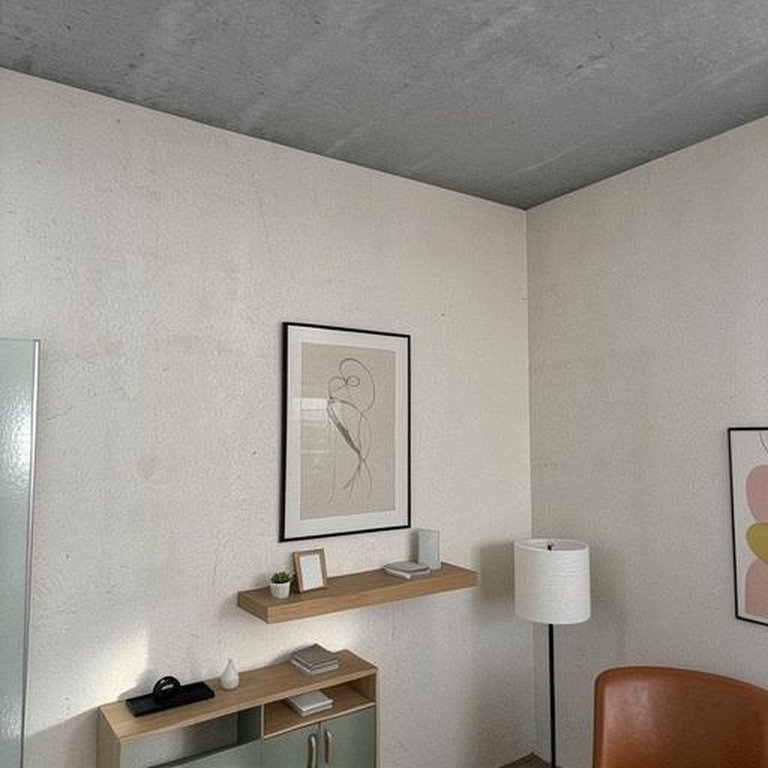} &
        \includegraphics[width=0.164\linewidth]{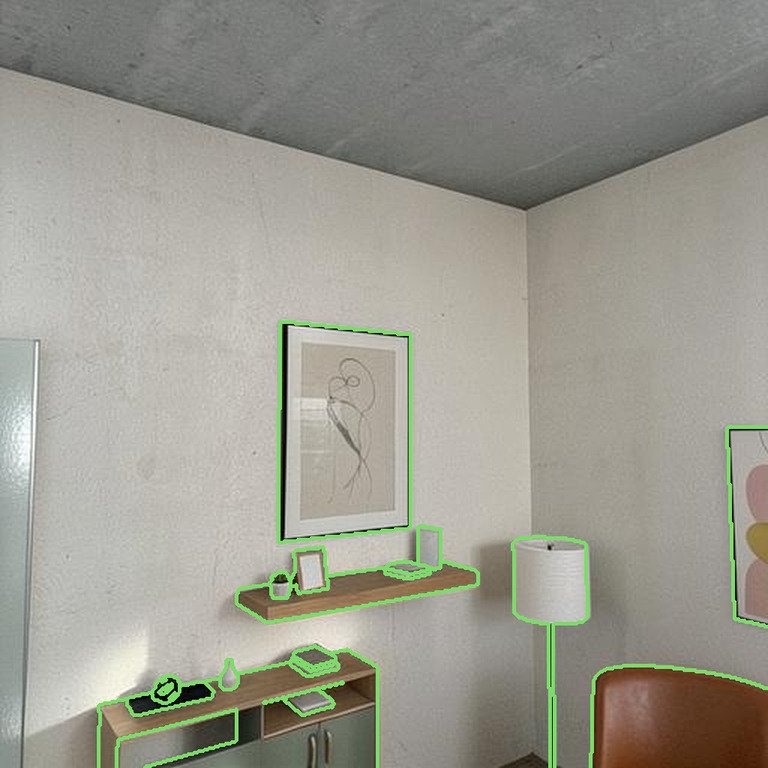} &
        \includegraphics[width=0.164\linewidth]{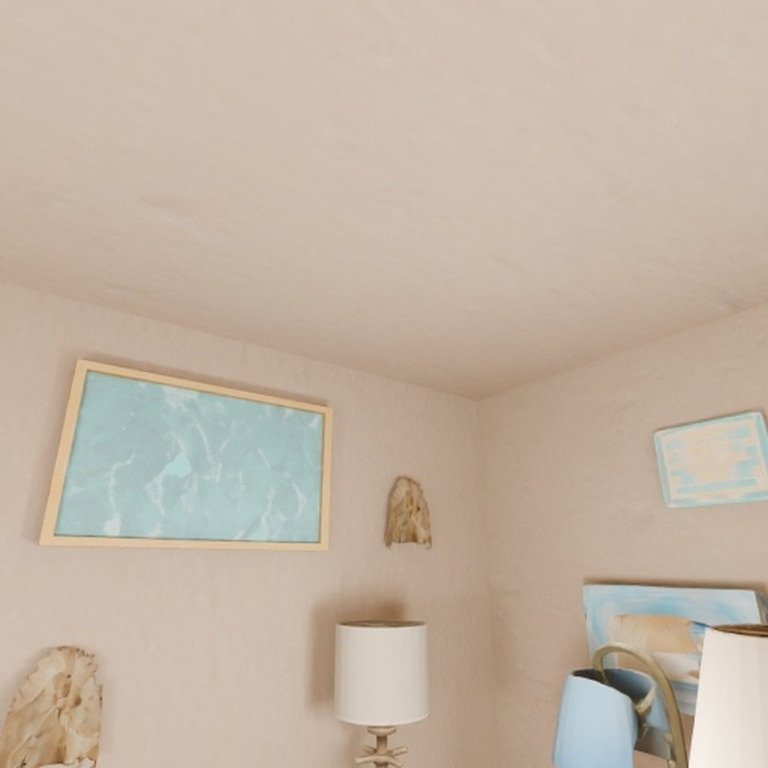} &
        \includegraphics[width=0.164\linewidth]{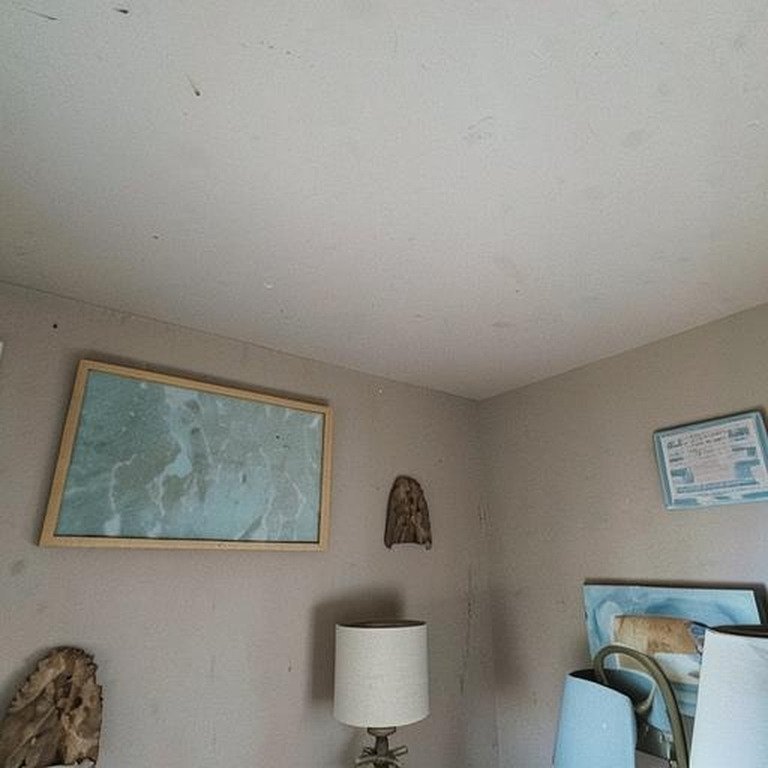} &
        \includegraphics[width=0.164\linewidth]{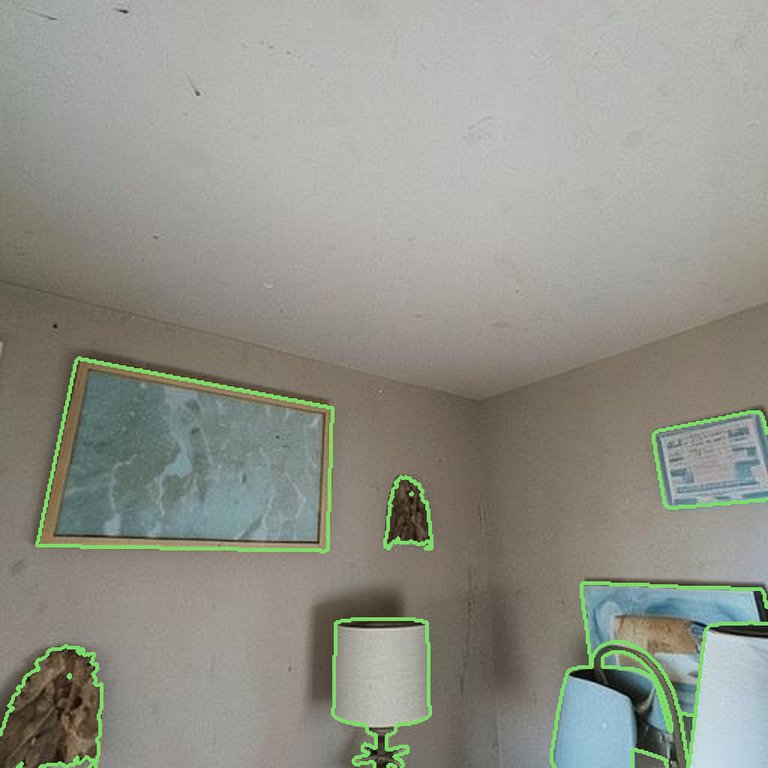} \\[1pt]
        Render & Flux.2 edit & Edit + labels & Render & Flux.2 edit & Edit + labels
    \end{tabular}%
    }
    \captionof{figure}{\textbf{Visualization of Flux.2-edited frames with instance labels.} The edits stay aligned with the labels. Instance boundaries of the original renders are drawn in green on the edited frames.}
    \label{fig:flux_align}
\end{table*}

\subsection{Object Compact Latent Representation Details}
\subsubsection{Hierarchical VAEs design}
To enable efficient multi-object generation, the proposed Hierarchical Compression VAE (HC-VAE) encodes the sparse feature tensor above the Sparse Compression VAE (SC-VAE) in \rebut{\cite{xiang2026native}}.
The HC-VAE \revten{consists of a lightweight sparse 3D convolutional encoder and decoder}, built with FlexGEMM \rebut{\cite{xiang2026native}} to further compress the sparse latent into an even more compact one. \rebut{HC-VAE is trained on SC-VAE latents of the same assets used by SC-VAE, from Objaverse-XL~\cite{deitke2023objaverse}, ABO~\cite{abo}, 3D-FUTURE~\cite{3dfuture}, and HSSD~\cite{hssd}, so it needs no extra data.}
In the following, we will describe the detailed HC-VAE structure as well. 
We also show the detailed network architecture in Tab. \ref{tab:hc_vae_arch} and Tab. \ref{tab:hc_vae_details}.

\paragraph{HC-VAE for shape latents.}
The shape HC-VAE is implemented as a sparse 3D U-Net-style variational autoencoder. The encoder receives 32-channel sparse shape features and progressively increases the feature width \rev{from 160 to 640 and 1280 channels}, using residual ConvNeXt-style 3D convolutional blocks at each resolution. Two stride-2 residual downsampling stages reduce the sparse spatial resolution, after which the representation is projected to a 64-channel latent code. The decoder mirrors this hierarchy with 3D residual upsampling stages, reducing the feature width \rev{from 1280 to 640 and 160 channels} before reconstructing the 32-channel shape feature field. This branch also predicts subdivision signals for refining the sparse structure.

\paragraph{HC-VAE for material latents.}
The material HC-VAE uses the same sparse 3D U-Net-style encoder--decoder design as the shape HC-VAE, but is trained to reconstruct physically based rendering attributes rather than geometry latents. Its encoder maps 32-channel sparse material
features through three feature stages with \rev{widths 160, 640, and 1280}, and compresses them into a 64-channel latent representation. The decoder applies the symmetric sequence of residual 3D convolution and upsampling blocks to recover 32-channel material features. In contrast to the shape branch, this model does not predict subdivision, since material attributes are decoded on the given sparse support.

\begin{table}[h]
\centering

\resizebox{\linewidth}{!}{
\begin{tabular}{lccc}
\toprule
Model & Architecture & Channel schedule & Latent channels \\
\midrule
HC-VAE, shape
& Sparse 3D U-Net VAE
& \rev{$(160,640,1280) \rightarrow (1280,640,160)$}
& 64 \\
HC-VAE, material
& Sparse 3D U-Net VAE
& \rev{$(160,640,1280) \rightarrow (1280,640,160)$}
& 64 \\
\bottomrule
\end{tabular}
}
\vspace{2px}
\caption{Architecture summary of HC-VAE.}
\label{tab:hc_vae_arch}
\end{table}

\begin{table}[h]
\centering

\resizebox{\linewidth}{!}{
\begin{tabular}{lcccccc}
\toprule
Model & In ch. & Out ch. & Blocks & Sampling blocks & Loss & Subdivision \\
\midrule
HC-VAE, shape
& 32 & 32 & $(4,6,8)/(8,6,4)$
& stride-2 residual conv. / residual upconv.
& \rev{L1+KL+subdiv.} & yes \\
HC-VAE, material
& 32 & 32 & $(4,6,8)/(8,6,4)$
& stride-2 residual conv. / residual upconv.
& \rev{L1+KL} & no \\
\bottomrule
\end{tabular}
}
\vspace{2px}
\caption{Detailed configuration of the latent and structure autoencoders.}
\label{tab:hc_vae_details}
\end{table}

\subsubsection{Hierarchical Batchified Object Decoding}

\paragraph{Batchified flow-matching inference.}
During inference, objects in the same scene are decoded in chunks of size $B$ rather than one at a time. {The background instance is included in the same packed batch and decoded with the same flow models.} For each chunk, the object points, point features, instance indices, and object-to-canonical transforms are concatenated into a single batched input. The instance index identifies which object each point belongs to, while the transform normalizes the object into its canonical frame.
The flow-matching model therefore denoises multiple object latents in one forward pass, with object-specific conditioning preserved by the packed instance labels and per-object transforms.

\paragraph{Sparse coordinate packing.}
For feature and material generation, the decoded sparse coordinates of all objects in a chunk are packed into one sparse tensor. The first coordinate dimension stores the local object index within the chunk, and the remaining three dimensions store the voxel coordinate. This produces a standard batched sparse representation of the form $(b,x,y,z)$, where $b \in \{0,\ldots,B-1\}$. The corresponding shape or material latent features are concatenated in the same order, allowing the sparse convolutional decoders to process all objects in the chunk jointly while keeping their sparse supports disjoint.

\paragraph{Batchified mesh post-processing.}
After latent decoding, the resulting per-object meshes generally have different numbers of vertices and faces. We batch them by padding each mesh to the maximum vertex and face count within the chunk and storing binary vertex and face masks. The padded tensors are then processed together on the GPU for hole filling, narrow-band remeshing, and mesh simplification. The masks are used to recover the valid vertices and faces for each object after post-processing, after which each mesh is transformed back to world coordinates and exported independently.

\paragraph{Batchified texture realization.}
For textured export, the same chunked mesh representation is used for UV unwrapping and rasterization. UV-space texels are rasterized for all objects in the chunk, and valid texel positions are projected back to the corresponding 3D surface. These 3D sample positions are used to query the decoded Material voxel field with trilinear interpolation, producing texture maps and material masks for each object. This keeps expensive UV, rasterization, and material sampling steps batched while preserving separate outputs for each object.

\subsection{Architecture Details}

\subsubsection{Instance-aware 3D scene perception model}

\paragraph{Point-level feature extraction.}
The perception model first extracts dense 2D visual descriptors using a DINOv3 ViT-L/16 backbone and lifts them to a 3D point representation. Each point is represented by a 1024-dimensional feature vector. A point-based U-Net then processes the lifted point cloud with hidden width 1024 and two 1024-channel convolutional layers. \rev{Encoded scene context is reduced to at most 30k tokens by farthest-point sampling; sparse input and segmentation features are not capped at 30k. Fourier features use sampling rate 60.} The main architecture hyper-parameters are summarized in
Tab.~\ref{tab:perception_arch_hparams}.

\paragraph{Instance-aware scene decoding.}
The scene decoder is a transformer-based instance prediction module with model dimension 1024, 16 attention heads, and a 4096-dimensional feed-forward network. It uses six transformer encoder layers for global scene reasoning, six additional encoder layers for segmentation feature refinement, and six decoder layers for object-level prediction. The decoder predicts object validity, translation, rotation, scale, and point-to-instance segmentation. \rev{The decoder has $K_Q=384$ candidate slots (including background), with nine tokens each. Translation/scale use 1024 bins over $[0,24]$ in model coordinates; Euler ranges are $[-\pi,\pi]$, $[-\pi/2,\pi/2]$, and $[-\pi,\pi]$. Segmentation features have dimension 1024.}

\subsubsection{Point cloud conditioned generative reconstruction model}

\paragraph{Conditional flow models.}
The reconstruction stage uses three point-cloud-conditioned flow-matching models for sparse structure, shape features, and material features. All three models share the same conditioning design: object-level point features are encoded with Fourier positional features and projected to a 1024-dimensional context. \rev{Each flow backbone has width 1536, 30 transformer-style blocks, and 12 attention heads.} The flow architecture details are listed in
Tab.~\ref{tab:flow_arch_hparams}.

\paragraph{Latent parameterization.}
The sparse-structure flow predicts 8-channel latents on a $2^3$ grid, which are decoded into object occupancy coordinates. The shape flow predicts 64-channel latents on the generated sparse support, and the material flow predicts 64-channel material latents conditioned on the shape support. \rev{Both flows operate at sparse resolution 8. Material input concatenates 64-channel shape and material latents (128 channels), with 64 output channels. The model resolution setting is 1024; we decode at resolution 512.} 

\subsection{Training Details}
\setlength{\intextsep}{4pt plus 1pt minus 1pt}

\begin{table}[h]
\centering
\resizebox{\linewidth}{!}{
\begin{tabular}{lcccccc}
\toprule
Component & Width & Layers & Heads & FFN dim. & Feature dim. & Precision \\
\midrule
Point U-Net & 1024 & 2 conv. layers & -- & -- & 1024 & \rev{float32} \\
Scene encoder & 1024 & 6 & 16 & 4096 & 1024 & \rev{float32} \\
Segmentation encoder & 1024 & 6 & 16 & 4096 & 1024 & \rev{float32} \\
Object decoder & 1024 & 6 & 16 & 4096 & 1024 & \rev{float32} \\
\bottomrule
\end{tabular}
}
\vspace{2px}
\caption{Architecture hyper-parameters of the instance-aware perception model.}
\label{tab:perception_arch_hparams}
\end{table}

\begin{table}[h]
\centering
\resizebox{\linewidth}{!}{
\begin{tabular}{lccccccc}
\toprule
Model & Latent res. & In ch. & Out ch. & Width & Cond. dim. & Blocks & Pos. enc. \\
\midrule
Sparse structure flow & $2^3$ & 8 & 8 & \rev{1536} & 1024 & 30 & RoPE \\
Shape feature flow & sparse $8$ & \rev{64} & \rev{64} & \rev{1536} & 1024 & 30 & RoPE \\
Material feature flow & sparse $8$ & \rev{128} & \rev{64} & \rev{1536} & 1024 & 30 & RoPE \\
\bottomrule
\end{tabular}
}
\vspace{2px}
\caption{Architecture hyper-parameters of the point-cloud-conditioned flow models.}
\label{tab:flow_arch_hparams}
\end{table}

\begin{table}[h]
\centering
\resizebox{\linewidth}{!}{
\begin{tabular}{lcccccccc}
\toprule
Model group & Steps & LR & WD & Betas & Precision & Grad. clip & EMA & Save interval \\
\midrule
Perception & \rev{400k} & \rev{$10^{-5}$} & 0.01 & $(0.9,0.95)$
& \rev{float32} & 0.1 & no & \rev{5000} \\
Structure / shape / material flows & \rev{200k} & $10^{-4}$ & 0.01 & $(0.9,0.95)$
& bfloat16 & 1.0 & 0.9999 & 5000 \\
\bottomrule
\end{tabular}
}
\vspace{2px}
\caption{\rev{Training hyper-parameters.}}
\label{tab:training_hparams}
\end{table}

\paragraph{Perception model training.}
\rev{We train the perception model from scratch for 400k steps using AdamW
at learning rate $10^{-5}$}, weight decay $0.01$, betas $(0.9,0.95)$, and
$\epsilon=10^{-8}$. \rev{We use no warmup, float32 precision,}
gradient clipping at norm $0.1$, and no EMA. The loss combines translation,
scale, rotation, validity, and segmentation terms; the matching cost weights
are $0.1$ for classification, $10.0$ for translation, $10.0$ for scale, and
$1.0$ for rotation. \rev{Checkpoints are saved every 5000 steps; validation/inference runs every 1000 steps. We evaluate the final 400k-step checkpoint.} The training hyper-parameters are summarized in
Tab.~\ref{tab:training_hparams}.

\paragraph{Generative reconstruction model training.}
\rev{We train the flow models from scratch for 200k steps with}
AdamW, learning rate $10^{-4}$, weight decay $0.01$, betas $(0.9,0.95)$, and
$\epsilon=10^{-8}$. Training uses bfloat16 mixed precision, gradient clipping
at norm $1.0$ with a 95th-percentile clipping statistic, and EMA with decay
$0.9999$. Classifier-free conditioning dropout is applied with probability
$p_{\mathrm{uncond}}=0.1$. \rev{Checkpoints and inference samples are
saved every 5000 steps}, as shown in Tab.~\ref{tab:training_hparams}.
\rev{We accumulate gradients over 2/1/2 steps for structure/shape/material and use non-EMA flow weights at inference.}

\end{document}